\documentclass{article}
\usepackage{arxiv}
\pdfoutput=1

\usepackage[utf8]{inputenc} 
\usepackage[T1]{fontenc}    
\usepackage{hyperref}       
\usepackage{url}            
\usepackage{booktabs}       
\usepackage{amsfonts}       
\usepackage{nicefrac}       
\usepackage{microtype}      
\usepackage{lipsum}
\usepackage{fancyhdr}       
\usepackage{graphicx}       
\usepackage[table]{xcolor}
\graphicspath{{figures/}}     

\usepackage{multirow}
\usepackage{float}
\usepackage[round, numbers, sort]{natbib}

\title{STREAMLINE: An Automated Machine Learning Pipeline for Biomedicine Applied to Examine the Utility of Photography-Based Phenotypes for OSA Prediction Across International Sleep Centers
}

\begin{document}
\maketitle

\vspace{-2.0cm}

\textbf{Ryan J. Urbanowicz\textsuperscript{1,4*}, Harsh Bandhey\textsuperscript{1}, Brendan T. Keenan\textsuperscript{2}, Greg Maislin\textsuperscript{2}, Sy Hwang\textsuperscript{3}, Danielle L. Mowery\textsuperscript{4}, Shannon M. Lynch\textsuperscript{5}, Diego R. Mazzotti\textsuperscript{6,7}, Fang Han\textsuperscript{8}, Qing Yun Li\textsuperscript{9}, Thomas Penzel\textsuperscript{10}, Sergio Tufik\textsuperscript{11}, Lia Bittencourt\textsuperscript{11}, Thorarinn Gislason\textsuperscript{12,13}, Philip de Chazal\textsuperscript{14}, Bhajan Singh\textsuperscript{15,16}, Nigel McArdle\textsuperscript{15,16}, Ning-Hung Chen\textsuperscript{17}, Allan Pack\textsuperscript{2}, Richard J. Schwab\textsuperscript{2}, Peter A. Cistulli\textsuperscript{18}, Ulysses J. Magalang\textsuperscript{19}}

\textsuperscript{1}Department of Computational Biomedicine, Cedars Sinai Medical Center, Los Angeles, CA \\
\textsuperscript{2}Division of Sleep Medicine, Department of Medicine, University of Pennsylvania, Philadelphia, PA \\
\textsuperscript{3}Institute for Biomedical Informatics, University of Pennsylvania, Philadelphia, PA \\
\textsuperscript{4}Department of Biostatistics Epidemiology and Informatics, University of Pennsylvania, Philadelphia, PA \\
\textsuperscript{5}Cancer Prevention and Control Program, Fox Chase Cancer Center, Philadelphia, PA \\
\textsuperscript{6}Division of Medical Informatics, Department of Internal Medicine, University of Kansas, Lawrence, KS \\
\textsuperscript{7}Division of Pulmonary, Critical Care, and Sleep Medicine, Department of Internal Medicine, University of Kansas, Lawrence, KS \\
\textsuperscript{8}Division of Sleep Medicine, Peking University People’s Hospital, Beijing, China \\
\textsuperscript{9}Department of Respiratory and Critical Care Medicine, Ruijin Hospital, Shanghai Jiao Tong University School of Medicine, Shanghai, China \\
\textsuperscript{10}Interdisciplinary Center of Sleep Medicine, Charite University Hospital, Berlin, Germany \\
\textsuperscript{11}Department of Psychobiology, Federal University of Sao Paulo, Sao Paulo, Brazil \\
\textsuperscript{12}Department of Sleep Medicine, Landspitali University Hospital, Reykjavik, Iceland \\
\textsuperscript{13}University of Iceland, Faculty of Medicine, Reykjavik, Iceland \\
\textsuperscript{14}Charles Perkins Center, Faculty of Engineering, University of Sydney, Sydney, Australia \\
\textsuperscript{15}Department of Pulmonary Physiology and Sleep Medicine, Sir Charles Gairdner Hospital, Nedlands, Australia \\
\textsuperscript{16}University of Western Australia, Crawley, Australia \\
\textsuperscript{17}Department of Pulmonary and Critical Care Medicine, Sleep Center, Linkou Chang Gung Memorial Hospital, Taoyuan, Taiwan \\
\textsuperscript{18}Charles Perkins Centre, Faculty of Medicine and Health, University of Sydney and Royal North Shore Hospital, Sydney, Australia \\
\textsuperscript{19}Division of Pulmonary, Critical Care, and Sleep Medicine, Department of Medicine, Ohio State University Wexner Medical Center, Columbus, OH \\

*Corresponding author: R.J.U. (ryan.urbanowicz@cshs.org) \\ 700 N. San Vicente Blvd., Pacific Design Center Suite G541E, Los Angeles, CA 90069

\vspace{1.0cm}
\pagebreak
\begin{abstract}
\textbf{Objective: }
While machine learning (ML) includes a valuable array of tools for analyzing biomedical data with multivariate and complex underlying associations, significant time and expertise is required to assemble effective, rigorous, comparable, reproducible, and unbiased pipelines. Automated ML (AutoML) tools seek to facilitate ML application by automating a subset of analysis pipeline elements. In this study we develop and validate a Simple, Transparent, End-to-end Automated Machine Learning Pipeline (STREAMLINE) and apply it to investigate the added utility of photography-based phenotypes for predicting obstructive sleep apnea (OSA); a common and underdiagnosed condition associated with a variety of health, economic, and safety consequences.

\textbf{Methods:}
STREAMLINE is designed to tackle biomedical binary classification tasks while (1) avoiding common mistakes, (2) accommodating complex associations and common data challenges, and (3) allowing scalability, reproducibility, and model interpretation. It automates the majority of established, generalizable, and reliably automatable aspects of an ML analysis pipeline while incorporating cutting edge algorithms and providing opportunities for human-in-the-loop customization. We present a broadly refactored and extended release of STREAMLINE, validating and benchmarking performance across simulated and real-world datasets. Then we applied STREAMLINE to evaluate the utility of demographics (DEM), self-reported comorbidities (DX), symptoms (SYM), and photography-based craniofacial (CF) and intraoral (IO) anatomy measures in predicting ‘any OSA’ or ‘moderate/severe OSA’ using 3,111 participants from Sleep Apnea Global Interdisciplinary Consortium (SAGIC).

\textbf{Results:}
Benchmarking analyses validated the efficacy of STREAMLINE across data simulations with increasingly complex patterns of association including epistatic interactions and genetic heterogeneity. OSA analyses identified a significant increase in ROC-AUC when adding CF to DEM+DX+SYM to predict ‘moderate/severe’ OSA. Additionally, a consistent but non-significant increase in PRC-AUC was observed with the addition of each subsequent feature set to predict ‘any OSA’, with CF and IO yielding minimal improvements. 

\textbf{Conclusion:}
STREAMLINE is an effective, rigorous, transparent, and easy-to-use AutoML approach to a comparative ML analysis that adheres to best practices in data science. Application of STREAMLINE to OSA data suggests that CF features provide additional value in predicting moderate/severe OSA, but neither CF nor IO features meaningfully improved the prediction of ‘any OSA’ beyond established demographics, comorbidity and symptom characteristics. 
\end{abstract}


\section{Introduction}
Machine learning (ML) encompasses a variety of computational methodologies for tackling optimization tasks, including predictive modeling \cite{jovel_introduction_2021,badillo_introduction_2020}. Interest in, and application of, ML has continued to expand across many domains including biomedical research; a field synonymous with complex, noisy, heterogeneous, and often large-scale data, as well as the need for interpretable/explainable solutions, statistical rigor, and reproducibility \cite{mirza_machine_2019,peek_technical_2014}. However, assembly of an ML analysis pipeline can be time-intensive, require significant programming and data science expertise, and hold countless opportunities for mistakes \cite{whalen_navigating_2022,garreta_scikit-learn_2017,greener_guide_2022,chicco_ten_2017,seghier_ten_2022,luo_guidelines_2016,heil_reproducibility_2021,riley_three_2019,smialowski_pitfalls_2010,ucar_effect_2020}. Packages such as \textit{pandas}, \textit{scikit-learn}, and \textit{scipy} make it easier to implement elements of a pipeline, but the challenge of how to correctly and effectively put these elements together into a complete analysis pipeline remains. As implied by the “no free lunch” theorem \cite{wolpert_no_1997}, no single ML algorithm can be expected to perform optimally for all problems. This expectation could similarly be extended to the entire ML analysis pipeline, such that for any new problem, users will not know which combination of ML algorithms, algorithm hyperparameter settings, or possible data preparation steps (e.g. cleaning and feature selection, engineering, or transformation) will yield an optimized model. 

 Automated machine learning (AutoML) represents a newer subfield of artificial intelligence and ML research focused on the development of tools or code libraries that facilitate and/or optimize the design, implementation, or execution of ML analyses in a manner that “relaxes the need for a user in the loop” \cite{escalante_automated_2021,waring_automated_2020}. A comparative survey of the scope and automation capabilities of 24 open-source AutoML implementations highlights many of their differences (see S.1, and Tables S1, S2). High-level differences include: (1) \textit{applicable data types}; they all handle structured (e.g. tabular) data, but a minority can use images, text, time-series, or multi-modal data (e.g. Auto-Keras \cite{jin_autokeras_2023}, FEDOT \cite{nikitin_automated_2022}, and Auto-Gluon \cite{erickson_autogluon-tabular_2020} can handle all these types); (2) \textit{applicable targets}; they all handle binary classification, most also handle multi-class and regression, and a small minority handle multi-task (e.g. Ludwig \cite{molino_ludwig_2019}), multi-label (Auto-Sklearn \cite{feurer_efficient_2015}), clustering and anomaly detection (e.g. PYCARET \cite{noauthor_welcome_2023}); (3) \textit{advanced utilization of a meta-learner} to either recommend analysis configurations (ALIRO \cite{la_cava_evaluating_2021} and AutoPyTorch \cite{zimmer_auto-pytorch_2021}) or automatically discover ML pipeline configurations through search (i.e. TPOT \cite{olson_evaluation_2016,olson_tpot_2016,le_scaling_2020}, FEDOT \cite{nikitin_automated_2022}, Auto-Sklearn \cite{feurer_efficient_2015}, GAMA \cite{gijsbers_gama_2021}, RECIPE \cite{de_sa_recipe_2017}, and ML-Plan \cite{mohr_ml-plan_2018}); (4) \textit{accessibility and ease of use}; some AutoMLs are designed to be broadly accessible requiring little to no coding experience to implement and run (e.g. ALIRO \cite{la_cava_evaluating_2021}, MLme \cite{akshay_machine_2023}, MLIJAR-supervised \cite{noauthor_mljar_2023}, H20-3 \cite{H2OAutoML20}, STREAMLINE \cite{urbanowicz_streamline_2023}, and Auto-WEKA \cite{kotthoff_auto-weka_2017}), while others are designed primarily as a code library to facilitate building a customizable pipeline with automated elements (e.g. LAMA \cite{vakhrushev_lightautoml_2021}, FLAML \cite{wang_flaml_2021}, Hyperopt-sklearn \cite{hutter_automated_2019}, TransmorgrifAI \cite{noauthor_transmogrifai_2023}, MLBox \cite{vasile_mlbox_2018}, Xcessiv \cite{nakano_xcessiv_2023}); (5) \textit{output focus}; the aims of AutoML vary, with focus on either a single best optimized model/pipeline (e.g. Auto-Sklearn \cite{feurer_efficient_2015}), a leaderboard of top models/pipelines (e.g. TPOT \cite{le_scaling_2020}), or a direct comparison of model performance across algorithms (e.g. STREAMLINE \cite{urbanowicz_streamline_2023}, MLme \cite{akshay_machine_2023}, and PYCARET \cite{noauthor_welcome_2023}); (6) \textit{inclusion and automation of different possible elements of a complete end-to-end ML pipeline}; with algorithm selection and hyperparameter optimization being most common; and (7) \textit{transparency in the documentation}, i.e. to what degree are the available elements, options, and automations defined and validated.

 Previously, we implemented and benchmarked a prototype version of a “Simple, Transparent, End-to-end AutoML analysis pipeline” (STREAMLINE) \cite{urbanowicz_streamline_2023} specifically designed for the needs and challenges of structured biomedical data analysis with binary outcomes. This implementation was also applied to investigate a variety of biomedical problems/datasets, including Alzheimer’s disease \cite{wang_exploring_2023,tong_comparing_2023}, oncology patient readmission \cite{hwang_toward_2023}, acute respiratory failure \cite{kohn_data-driven_2023}, and pancreatic cancer \cite{urbanowicz_rigorous_2020} with feedback from these collaborations informing the present implementation. STREAMLINE differs from other surveyed AutoML in that it (1) fully automates all generalizable ML pipeline elements that can be reliably automated without specific problem domain knowledge, (2) follows a carefully designed, transparent, and standardized pipeline workflow, rather than conducting a meta-algorithm search of pipeline configurations (often computationally expensive) or focusing on extensive user-driven pipeline customization (for ease of use and to avoid pitfalls, mistakes, and bias), (3) rigorously applies and directly compares a variety of established and cutting-edge ML modeling algorithms, (4) captures and facilitates interpretation of complex associations (e.g. feature interactions and heterogeneous associations), and (5) automatically generates and organizes figures, visualizations, results, and analysis summaries for each phase of the pipeline. STREAMLINE is among the minority of AutoMLs that include exploratory data analysis (EDA), basic data cleaning, or outputting separately trained models for each cross-validation (CV) partition to directly assess the potential impact of sample bias in data partitioning. 

 To the best of our knowledge, STREAMLINE is unique in its ability to automatically (1) analyze and compare multiple datasets (e.g. to examine the value of specific feature subsets, as in the present study), (2) engineer binary ‘missingness features’ to determine the predictive value of data entries that are ‘missing not at random’ (MNAR) \cite{emmanuel_survey_2021}, (3) utilize cutting-edge bioinformatics feature selection methodologies (i.e. MultiSURF \cite{urbanowicz_benchmarking_2018} and collective feature selection \cite{verma_collective_2018}), and modeling (i.e. ExSTraCS \cite{urbanowicz_exstracs_2015}, a rule-based algorithm designed specifically to address the challenges of detecting and characterizing epistasis \cite{mckinney_machine_2006} and heterogeneous associations \cite{woodward_genetic_2022} in biomedical data), (4) conduct statistical significance comparisons (between algorithms and datasets), (5) collectively compare and contrast feature importance (FI) estimates across modelling algorithms, and (6) generate a comprehensive sharable summary report. With respect to overall AutoML design and goals, STREAMLINE currently is most closely related to MLIJAR-supervised \cite{noauthor_mljar_2023} and MLme \cite{akshay_machine_2023} AutoML tools. 

Herein, we introduce a broadly refactored and extended release of STREAMLINE (v.0.3.4) as a comprehensive automated machine learning analysis methodology, including new capabilities targeting ease-of-use, parallelizability, automated data preparation, and customizability. We validated and benchmarked its performance on a diverse set of real and simulated datasets; some previously analyzed in v.0.2.5 \cite{urbanowicz_streamline_2023}. We then applied STREAMLINE to train, evaluate, and compare ML models for the prediction of Obstructive Sleep Apnea (OSA) using 3,111 participants from the Sleep Apnea Global Interdisciplinary Consortium (SAGIC) \cite{magalang_agreement_2013,magalang_agreement_2016,keenan_recognizable_2018,rizzatti_defining_2020,sutherland_facial_2023}. OSA is a common condition characterized by repetitive episodes of complete or partial collapse of the upper airway with a respective decrease in blood oxygen saturation or cortical arousal from sleep \cite{lyons_global_2020,gottlieb_diagnosis_2020}. OSA is estimated to affect one billion people worldwide and is associated with a variety of adverse health \cite{gami_day-night_2005,yaggi_obstructive_2005}, economic \cite{alghanim_economic_2008,mulgrew_impact_2007}, and safety consequences \cite{george_reduction_2001}. However, OSA remains underdiagnosed and untreated even within primary care settings \cite{arnardottir_obstructive_2016,arsic_assessing_2022,netzer_prevalence_2003}, leading to an estimated additional \$3.4 billion/year in medical costs \cite{kapur_medical_1999}. Hence, there is a need to identify reliable risk factors and biomarkers predictive of OSA that are ideally low cost and low burden to collect. Leveraging STREAMLINE, this study evaluates the utility of progressively adding distinct feature categories (i.e. demographics, self-reported comorbidities, symptoms \cite{maislin_survey_1995}, and photography-based measures of craniofacial and intraoral anatomy \cite{rizzatti_defining_2020,sutherland_facial_2023,schwab_digital_2017,sutherland_facial_2014,sutherland_craniofacial_2016,lee_craniofacial_2009,lee_prediction_2009}) in the prediction of OSA, defined by an apnea-hypopnea index (AHI) $\geq$ 15 events/hour (moderate/severe OSA) or AHI $\geq$ 5 events/hour (any OSA).

\section{Materials \& Methods}
\label{sec:materials and methods}

\subsection{STREAMLINE Automated Machine Learning Pipeline}
\begin{figure}
\centering
\includegraphics[width=\textwidth]{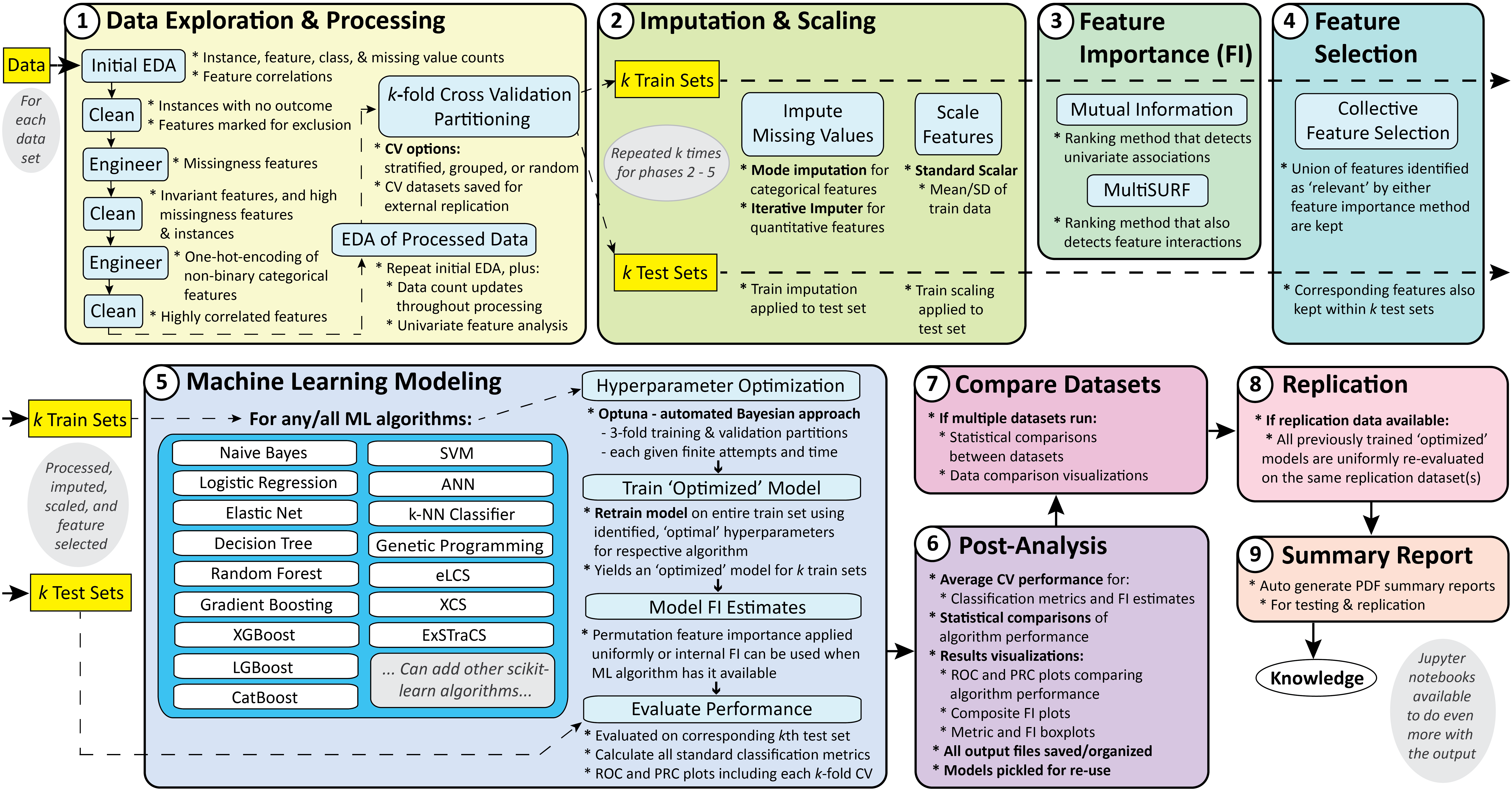}
\caption{STREAMLINE Automated Machine Learning Pipeline Schematic. Includes 9 parallelizable phases, run consecutively, with phase 7 only run in the presence of multiple input datasets and phase 8 only run when external hold-out data (replication data) is available. Results across all phases are visualized, summarized, and organized as output.}
\label{fig:fig1}
\end{figure}

The ML analyses presented in this paper were conducted with STREAMLINE (v0.3.4), a significantly refactored and extended release based on v0.2.5 \cite{urbanowicz_streamline_2023}. Currently, STREAMLINE is limited to structured, binary classification data analyses. As summarized in Figure \ref{fig:fig1}, STREAMLINE runs a carefully designed ML analysis pipeline across 9 consecutive phases following best practices in ML and data science and addressing common challenges in biomedical data (i.e. missing, imbalanced, large-scale, noisy, complex, and mixed type data, with a need for interpretability, reproducibility, and analytical rigor). 

In summary, STREAMLINE fully automates the following pipeline elements: (1) initial and post-processed EDAs, (2) basic data cleaning (i.e. numerical encoding, removal of instances with no outcome or high missingness, and removal of features with invariant values, high missingness, and high correlation with another feature), (3) basic feature engineering (i.e. adding binary ‘missingness features’, and one-hot-encoding of categorical features), (4) k-fold cross-validation partitioning with models trained and output for each partition, (5) missing value imputation, (6) feature scaling, (7) filter-based FI estimation using mutual information (MI) \cite{ross_mutual_2014} for univariate associations and MultiSURF \cite{urbanowicz_benchmarking_2018} for multivariate interactions, (8) collective feature selection \cite{verma_collective_2018} utilizing MI and MultiSURF FI estimates, (9) ML modeling across 16 scikit-learn compatible algorithms, including two prototype rule-based ML algorithm implementations (i.e. eLCS \cite{urbanowicz2017introduction} and XCS \cite{butz_xcs_2006}; not used in this study) incorporating Optuna-driven hyperparameter optimization \cite{akiba_optuna_2019}, final model training, permutation-based model feature importance estimation \cite{breiman_random_2001}, and performance evaluation using 16 standard classification metrics (see S.2.1.5), (10) generating and organizing all models, publication-ready plots, evaluation results, statistical significance comparisons across ML algorithms and analyzed datasets, and a formatted PDF summary report documenting run parameters and key results, and (11) application and re-evaluation of all trained models on further hold-out replication data, including separate EDA, cleaning, feature engineering, imputation, scaling, and feature selection to prepare the replication data in a manner consistent with the respective training partitions. A detailed explanation of STREAMLINE elements and key run parameters across all phases is given in S.2.1.

In addition to refactoring, and in contrast with (v0.2.5), STREAMLINE (v0.3.4) introduced the following automated capabilities: (1) numerical encoding of text-valued features, (2) enforced handling of categorical features including user-specification of feature types (which cannot always be reliably automated) and one-hot-encoding of categorical features for modeling, (3) engineering of ‘missingness features’ to consider missingness as a potentially informative feature, (4) cleaning (i.e. removal) of invariant features, (5) cleaning of instances and features with high missingness, (6) cleaning of features with high correlation (e.g., one of each pair of any perfectly correlated features is removed), (7) a secondary round of EDA documenting step-wise data modifications to ensure data processing quality, and (8) ML modeling with elastic net \cite{noauthor_sklearnlinear_modelelasticnet_nodate}. This new release can also now: (1) be parallelized on 7 different types of high-performance computing (HPC) clusters, (2) run all phases from a single command using a configuration file, (3) be run by those with no coding experience using an improved Google Colab run mode, and (4) easily add additional scikit-learn compatible ML modeling algorithms.

While STREAMLINE is configured to run using default parameters, users can optionally customize these parameters to improve performance, reduce runtime, or better scale the pipeline to large datasets. In this study, we largely adopted default STREAMLINE parameters (see S.2.4) with some problem-specific exceptions. Run parameters for each analysis are documented in the respective PDF reports (supplementary materials). 

STREAMLINE offers run modes suitable for non-programmers and data science experts alike, including: (1) Google Colab notebook, (2) Jupyter notebook, and (3) command line interface; either locally or with a dask-compatible HPC Cluster. Analyses in this study were performed on the Cedars Sinai HPC (1,012 CPU cores, 11.4 TB of RAM, using Slurm). STREAMLINE is open-source and available at: \href{https://github.com/UrbsLab/STREAMLINE}{https://github.com/UrbsLab/STREAMLINE}.

\subsection{Data and Analyses}
We first validated and benchmarked STREAMLINE’s performance on real and simulated datasets. We then applied STREAMLINE to train, evaluate, and compare ML models for the prediction of Obstructive Sleep Apnea (OSA) encoded as a binary outcome \cite{magalang_agreement_2013,magalang_agreement_2016,keenan_recognizable_2018,rizzatti_defining_2020} using two clinically-relevant thresholds: AHI $geq$ 15 events/hour (moderate/severe OSA) and AHI $geq$ 5 events/hour (any OSA).

\subsubsection{Benchmark Datasets}
STREAMLINE benchmarking was conducted over four separate STREAMLINE ‘experiments’ (i.e. individual runs of the AutoML pipeline). The first analyzed the hepatocellular carcinoma (HCC) survival dataset from the UCI repository \cite{miriam_santos_hcc_2015} in comparison to a modified ‘custom’ HCC dataset that removed covariate features and added simulated features and instances to explicitly verify all data processing elements of STREAMLINE (see S.2.2.1). The second analyzed and compared 6 different simulated genomic datasets each with a unique underlying pattern of association (i.e. univariate, additive, heterogeneous, 2-way epistasis, 2-way epistasis with heterogeneity, and 3-way epistasis). Each dataset was simulated by GAMETES \cite{urbanowicz_gametes_2012} to include 100 features, 1600 instances and a noisy signal (see S.2.2.2). The third analyzed and compared 6 different x-bit multiplexer datasets (i.e. 6, 11, 20, 37, 70, and 135-bit) often used as complex simulated ML benchmarks \cite{urbanowicz_benchmarking_2018,urbanowicz_exstracs_2015,urbanowicz2017introduction,butz_xcs_2006}. Multiplexer datasets represent clean problems (i.e. no noise) that involve underlying feature interactions and heterogeneous associations that increase in complexity with increasing ‘x’ (see S.2.2.3). The fourth analyzed and compared 4 different ‘XOR’ datasets (i.e. 2, 3, 4, and 5-way) applied in previous benchmarking of the MultiSURF feature selection algorithm \cite{urbanowicz_benchmarking_2018}. XOR datasets represent clean problems with increasing orders of pure/strict-epistatic interactions predicting a binary outcome (see S.2.2.4). The HCC, GAMETES, and multiplexer datasets were previously applied to benchmark STREAMLINE v.0.2.5 \cite{urbanowicz_streamline_2023}, while the XOR datasets are newly examined here.

\subsubsection{Obstructive Sleep Apnea Dataset}
These data include a total of 3,111 participants from the Sleep Apnea Global Interdisciplinary Consortium (SAGIC) with available data on severity of obstructive sleep apnea (OSA) based on the AHI (see Table \ref{tab1}) \cite{magalang_agreement_2013,magalang_agreement_2016,keenan_recognizable_2018,rizzatti_defining_2020,sutherland_facial_2023}. SAGIC is an international study of participants recruited from sleep centers across the world, including sites in the United States (The Ohio State University and University of Pennsylvania), China (Peking University People’s Hospital and Peking University International Hospital, Beijing), Germany (Charité University Hospital, Berlin), Brazil (Medicado Instituto do Sono, São Paulo), Iceland (Landspitali – The National University Hospital of Iceland, Reykjavik), Australia (Royal North Shore Hospital, Sydney and Sir Charles Gairdner Hospital, Perth), and Taiwan (Chang Gung Memorial Hospital, Taipei). The main goal of SAGIC is to establish a large international cohort with in-depth OSA-related phenotyping in which to understand overall and racial/ethnic differences in OSA causes, presentations, and consequences.

\begin{table}
\centering
\begin{tabular}{l|c|c|c|c|c}
\cellcolor{gray!20} & \multicolumn{2}{|c|}{\cellcolor{gray!20}\textbf{AHI$\geq$15 (Moderate/Severe OSA)}} & \multicolumn{2}{|c|}{\cellcolor{gray!20}\textbf{AHI$\geq$5 (Any OSA)}} & \cellcolor{gray!20}\multirow{2}{*}{} \\
\cellcolor{gray!20}{\textbf{Site}} & \cellcolor{gray!20}\textbf{< 15} & \cellcolor{gray!20}\textbf{$\geq$ 15} & \cellcolor{gray!20}\textbf{< 5} & \cellcolor{gray!20}\textbf{$\geq$ 5} & \cellcolor{gray!20}\textbf{Total Samples}  \\
 \hline
\textbf{Germany (Berlin)} & 134 (78.36\%) & 37 (21.64\%) & 92 (53.80\%) & 79 (46.20\%) & \textbf{171} \\
\hline
\textbf{Brazil} & 572 (69.59\%) & 250 (30.41\%) & 375 (45.62\%) & 447 (54.38\%) & \textbf{822} \\
\hline
\textbf{Iceland} & 75 (33.19\%) & 151 (66.81\%) & 4 (1.77\%) & 222 (98.23\%) & \textbf{226} \\
\hline
\textbf{USA } \textbf{(Ohio State)} & 76 (42.94\%) & 101 (57.06\%) & 35 (19.77\%) & 142 (80.23\%) & \textbf{177} \\
\hline
\textbf{USA} \textbf{(UPenn)} & 102 (51.00\%) & 98 (49.00\%) & 36 (18.00\%) & 164 (82.00\%) & \textbf{200} \\
\hline
\textbf{Australia} \textbf{(Perth)} & 72 (62.61\%) & 43 (37.39\%) & 46 (40.00\%) & 69 (60.00\%) & \textbf{115} \\
\hline
\textbf{Australia} \textbf{(Sydney)} & 162 (60.22\%) & 107 (39.78\%) & 93 (34.57\%) & 176 (65.43\%) & \textbf{269} \\
\hline
\textbf{Taiwan} & 19 (12.03\%) & 139 (87.97\%) & 5 (3.16\%) & 153 (96.84\%) & \textbf{158} \\
\hline
\textbf{China} \textbf{(Beijing)} & 489 (50.26\%) & 484 (49.74\%) & 283 (29.09\%) & 690 (70.91\%) & \textbf{973} \\
\hline
\cellcolor{gray!20}\textbf{Totals} & \cellcolor{gray!20}\textbf{1701} & \cellcolor{gray!20}\textbf{1410} & \cellcolor{gray!20}\textbf{969} & \cellcolor{gray!20}\textbf{2142} & \cellcolor{gray!20}\textbf{3111} \\
\end{tabular}
\caption{{Summary of SAGIC OSA data across participating sites.} Includes sample counts and percent ‘no OSA’ vs. ‘OSA’ for AHI$\geq$15 (moderate/severe OSA) and AHI$\geq$5 (any OSA) outcome encodings. }
\label{tab1}
\end{table}

This SAGIC data includes the following feature sets: (1) demographics (DEM), including age, sex, race/ethnicity, and body mass index (BMI); (2) self-reported comorbidities (DX), including hypertension, coronary artery disease, heart failure, stroke, atrial fibrillation, high cholesterol, and diabetes; (3) symptoms (SYM) from the Multivariable Apnea Prediction Index (MAP), including frequency of loud snoring, snorting or gasping, and breathing stops or witnessed apneas, and MAP index [68]; (4) craniofacial measures from photographs (CF) \cite{sutherland_facial_2023,sutherland_facial_2014,sutherland_craniofacial_2016,lee_craniofacial_2009,lee_prediction_2009} (i.e. 46 quantitative measures capturing craniofacial widths, heights, and angles); and (5) intraoral measures from photographs (IO) \cite{rizzatti_defining_2020,sutherland_facial_2023,schwab_digital_2017} (i.e. 12 quantitative measures of mouth and tongue size and two Mallampati score classifications I-IV \cite{mallampati_clinical_1983}). Section S.2.3 and Tables S3 and S4 provide data summary statistics and details of these features. Further details on study methods, including sleep study scoring, photography-based phenotyping, and patient questionnaires have been previously published \cite{magalang_agreement_2013,magalang_agreement_2016,keenan_recognizable_2018,rizzatti_defining_2020,laharnar_simple_2021,qin_heart_2021,holfinger_diagnostic_2022,sutherland_global_2019}.

Prior to applying STREAMLINE, the complete SAGIC dataset had been pre-processed by numerically encoding SYM features as ordinal and one-hot-encoding race/ethnicity and Mallampati classifications. Next, 20 dataset variations were created, each with a unique combination of feature set(s) and binary OSA outcome encoding. Considered feature sets included each set on its own, as well as the sequential addition of features sets regarded as more challenging/time-consuming/expensive to collect (DEM<DX<SYM<CF<IO), as well as specifically examining DEM+DX+SYM+IO in contrast with DEM+DX+SYM+CF to understand the relative importance of different anatomical measures. Two binary OSA outcomes were evaluated: “moderate/severe OSA” (AHI$\geq$15 events/hour) and “any OSA” (AHI$\geq$5 events/hour). Table S5 summarizes these 20 datasets (identified as A-T). 

Each of the 20 datasets were split into respective ‘development’ (70\%) and ‘replication’ (30\%) sets using stratified partitioning (see S.2.3.1). Each development set was used by STREAMLINE to train and test CV models in phases 1-7, and the replication set was used in phase 8 to evaluate all models on the same hold-out evaluation data to further evaluate model generalizability and select a best performer.

\section{Results}
\label{sec:results}
\subsection{Benchmark Analyses Yielded Consistent Findings}

Detailed model evaluation and feature importance results of STREAMLINE benchmarking analyses across HCC, GAMETES, multiplexer, and XOR datasets are given in S.3.1, S.3.2, S.3.3, and S.3.4, respectively. In summary, the custom HCC dataset validated the efficacy of STREAMLINE’s automated data processing, and ML modeling performance across all datasets was consistent with previous findings including ‘ground-truth’ expectations for model feature importance in simulated GAMETES, multiplexer, and XOR datasets.

For XOR datasets, we noted that STREAMLINE’s default settings for phase 4 collective feature selection (i.e. removing features that score $\leq$ 0 by both FI algorithms) incorrectly removed predictive features involved in pure 4 or 5-way epistatic interactions prior to modeling. This was in line with previous MultiSURF benchmarking observations noting that predictive features involved in high-order interactions (i.e. >3-way) yielded highly negative rather than highly positive feature importance scores with respect to non-predictive features [47]. However, performance was restored on the 4-way XOR by turning off feature selection; allowing all features to be present for modeling. In line with previous findings, the biomedically-tuned rule-based algorithm (ExSTraCS) dramatically outperformed all other algorithms on the 4-way XOR, as well other benchmarks with a high degree of underlying complexity, i.e. the GAMETES dataset ‘E’ (simulating both epistasis and heterogeneity) and the 70-bit multiplexer dataset. 

\subsection{Obstructive Sleep Apnea}

A detailed examination of STREAMLINE results across all 20 OSA datasets including EDA, data processing, feature importance, model testing performance, and model replication performance is given in S.3.5. Key results are presented below.

\subsubsection{Exploratory Data Analysis, Processing, and Feature Importance Estimation}

The next two subsections focus exclusively on development partitions of the 20 OSA datasets (70\% of data used for training and testing within STREAMLINE phases 1-7). Prior to data processing, AHI$\geq$15 datasets had similarly balanced class counts (no OSA = 1098, OSA = 1080), while AHI$\geq$5 datasets did not (no OSA = 606, OSA = 1573); see Figure S14. Across all 20 datasets, processing (via automated data cleaning and feature engineering steps) removed instances with > 50\% missing values in certain datasets. The largest removal occurred datasets with DX features alone (384 and 381 instances removed for datasets B and L, respectively). None of the datasets had features removed due to > 50\% missingness, feature value invariance, or perfect feature correlation (see Figure S16), nor instances removed due to missing outcome. Following processing, all AHI$\geq$15 datasets remained largely balanced, and all AHI$\geq$5 datasets remained imbalanced (see Tables S11 and S12). Consequently, we primarily focus on receiver operating characteristic area under the curve (ROC-AUC) to evaluate balanced AHI$\geq$15 datasets and precision-recall curve area under the curve (PRC-AUC) to evaluate imbalanced AHI$\geq$5 datasets.

Prior to 10-fold CV partitioning, univariate analyses were conducted across all 85 features included in datasets J and T (examining OSA as AHI$\geq$15 and AHI$\geq$5, respectively). The 10 most significant features for AHI$\geq$15 include BMI (from DEM feature set) and 9 CF features; with face width as most significant (p=1.046x10\textsuperscript{-51}) (see Table S13). The 10 most significant features for AHI$\geq$5 include BMI, 8 CF features, and loud snoring (from SYM feature set); with BMI as most significant (p=1.060x10\textsuperscript{-42}) (see Table S14).

Following CV partitioning, STREAMLINE estimated FI independently within each training set using Mutual Information and MultiSURF with respective median FI scores across CV partitions being examined (see Figures S19 and S20). The 10 top scoring features for AHI$\geq$15 with Mutual Information included; BMI, 7 CF features (mandibular width scoring highest), MAP index and loud snoring (from SYM feature set). With MultiSURF, the top 10 included BMI and male sex (from DEM), 6 CF features (mandibular width scoring highest), and loud snoring and breathing stops or choking/struggling for breath (from SYM).

The 10 top scoring features for AHI$\geq$5 with Mutual Information included; BMI, 6 CF features (mandibular width scoring highest), and MAP index, loud snoring, and breathing stops or choking/struggling for breath (from SYM). With MultiSURF, the top 10 included BMI, age, male (from DEM), 6 CF features (cervicomental angle scoring highest), and loud snoring (from SYM).

\subsubsection{ ML Modeling Testing Evaluations Across 20 OSA Datasets}
\begin{figure}
    \centering
    \includegraphics[width=\textwidth]{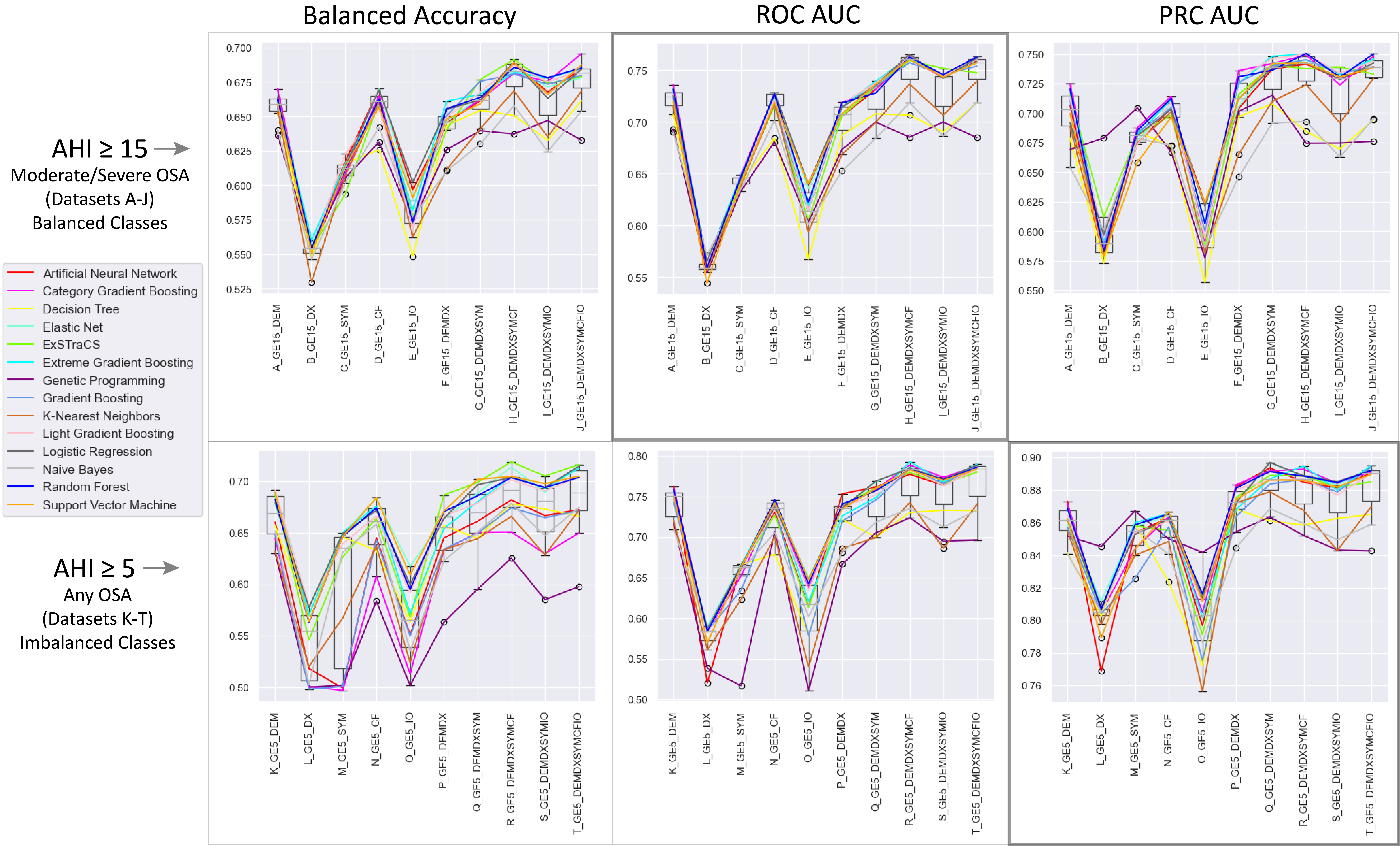}
    \caption{\textbf{STREAMLINE OSA Model Testing Evaluation Summary. }These 6 subplots report key evaluation metrics in datasets where OSA outcome is defined by either AHI $\geq$ 15 or AHI $\geq$ 5. Algorithm hyperparameters were selected to optimize balanced accuracy, and we focus on ROC-AUC or PRC-AUC to compare overall algorithm and dataset performance based on whether the data was balanced or not (see the two darkly outlined subplots). In each subplot the y-axis is the metric value (to improve visibility; the y-axis range is not fixed) and the x-axis is the target dataset (with dataset names specifying included feature sets, i.e. DEM, DX, SYM, CF, and IO). Each of the 60 individual boxplots give the spread of target metric performance (averaged over 10-fold CV models) across the 14 ML algorithms. Mean CV testing performance of a given algorithm is identified by the colored lines in the key.  }
    \label{fig:fig2}
\end{figure}

Figure \ref{fig:fig2} gives a global summary of model testing performance (with 14 ML algorithms) across the 20 OSA datasets. Pairwise performance metric comparisons (using Wilcoxon Signed-Rank Test) between datasets having the same underlying features (e.g. A vs. K), indicate that balanced accuracy and ROC-AUC were consistently (but not always significantly) higher for models predicting AHI$\geq$5 (see Table S15). The AHI$\geq$5 datasets consistently yielded significantly higher PRC-AUC (p<0.05) than AHI$\geq$15 datasets (see S.3.5.5 for details). 

Pairwise performance metric comparisons between datasets having only a single feature set (i.e. A-E and K-O) identified DEM and CF features as being significantly more informative of either OSA outcome than DX, SYM, or IO based on balanced accuracy and ROC-AUC. For PRC-AUC in AHI$\geq$15, DEM was significantly more informative than DX or IO and non-significantly better than SYM or CF. CF was significantly more informative than IO, and non-significantly better than SYM or DX. For PRC-AUC in AHI$\geq$5, DEM and CF were both significantly more informative than DX or IO, and non-significantly better than SYM. SYM was consistently (but not always significantly) higher than DX or IO for AHI$\geq$15 and AHI$\geq$5 and all three metrics. DEM and CF yielded similar performance for the three metrics and both outcome encodings. IO performed significantly better than DX for AHI$\geq$15 and AHI$\geq$5, with balanced accuracy and ROC-AUC, but yielded a significantly lower PRC-AUC for AHI$\geq$15 and a similar PRC-AUC for AHI$\geq$5. To generalize, DEM and CF were similarly most informative on their own, followed by SYM, IO, and lastly DX. Details of these comparisons are given in Tables S16 and S17.

\begin{figure}
  \centering
  \includegraphics[width=\textwidth]{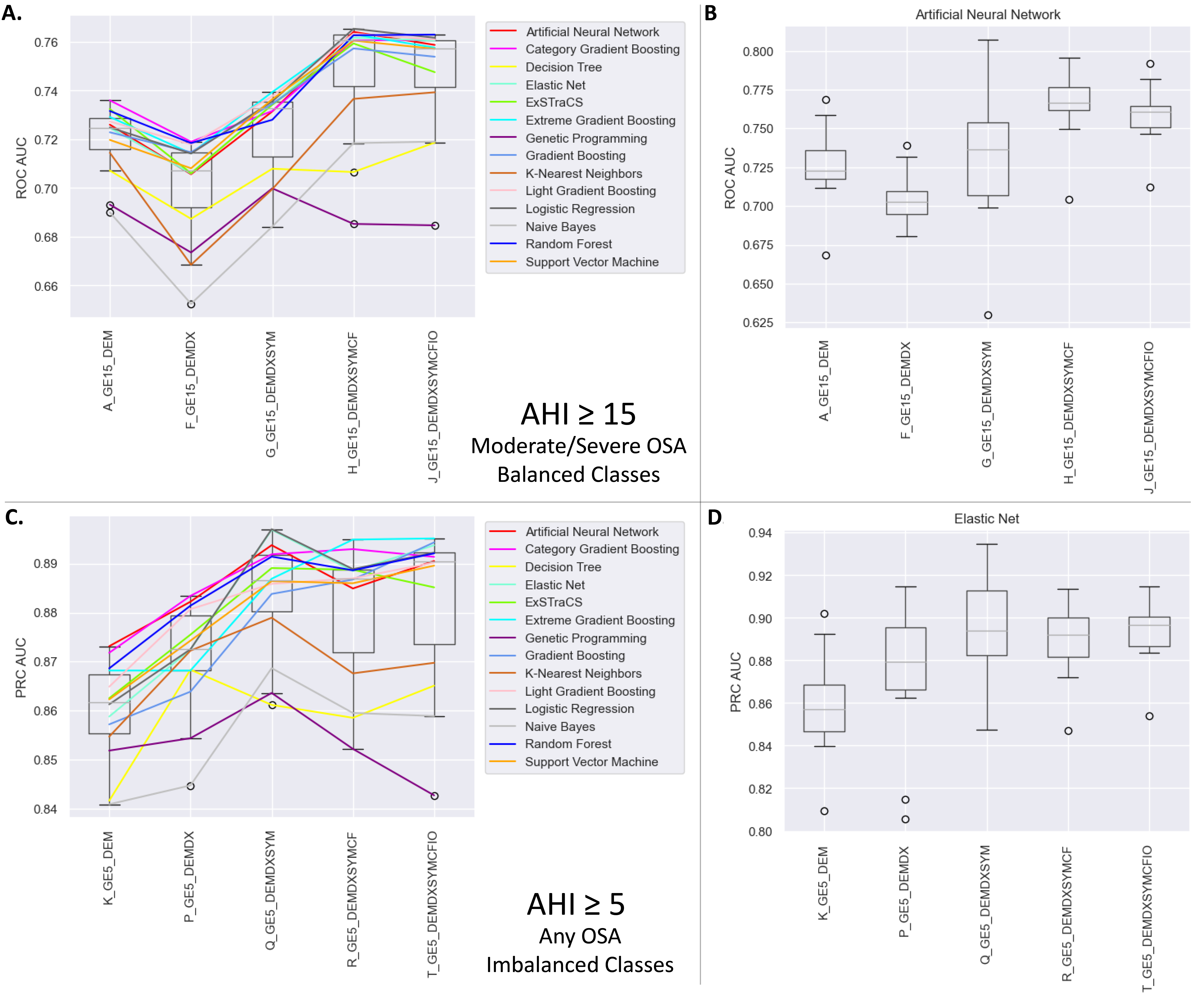}
  \caption{\textbf{Testing Evaluations for the Progressive Addition of OSA Feature Sets. }Sub-plots A and C represent a subset of the comparisons found in Figure \ref{fig:fig2}, with individual boxplots capturing the distribution of mean (10-fold CV) model metric performance across 14 algorithms. Differently, sub-plots B and D focus on the algorithm with the highest overall median metric performance for either AHI$\geq$15 or AHI$\geq$5 (as identified in Tables S18 and S19, respectively). For these, individual boxplots capture the distributions of metric performance across 10-fold CV models trained by that algorithm. (A) Algorithm comparisons on datasets A, F, G, H, and J (AHI$\geq$15) with ROC-AUC. (B) 10-fold CV comparisons for artificial neural network on datasets A, F, G, H, and J (AHI$\geq$15) with ROC-AUC. (C) Algorithm comparisons on datasets K, P, Q, R, and T (AHI$\geq$5) with PRC-AUC. (D) 10-fold CV comparisons for elastic net on datasets K, P, Q, R, and T (AHI$\geq$5) with PRC-AUC. }
  \label{fig:fig3}
\end{figure}

Next, we examined pairwise performance metric comparisons between datasets progressively adding feature sets (i.e. DEM, +DX, +SYM, +CF, +IO). Figure \ref{fig:fig3} focuses on these specific comparisons (narrowing the results from Figure \ref{fig:fig2}), using ROC-AUC for AHI$\geq$15 and PRC-AUC for AHI$\geq$5. For AHI$\geq$15 datasets, the only step-wise feature set addition that led to a significant increase in ROC-AUC was CF added to DEM+DX+SYM (dataset G vs. H; p=0.0137). Significant ROC-AUC increases were also observed between (1) A vs. either H or J, (2) F vs. either H or J, and (3) G vs. J (see Table S18). The highest overall median ROC-AUC was obtained on dataset H (DEM+DX+SYM+CF) using the artificial neural network (ANN) algorithm. Figure 3B reveals a similar pattern of ROC-AUC performance across AHI$\geq$15 datasets when exclusively focusing on the 10 ANN models trained for each CV partition.

\begin{figure}
  \centering
  \includegraphics[width=\textwidth]{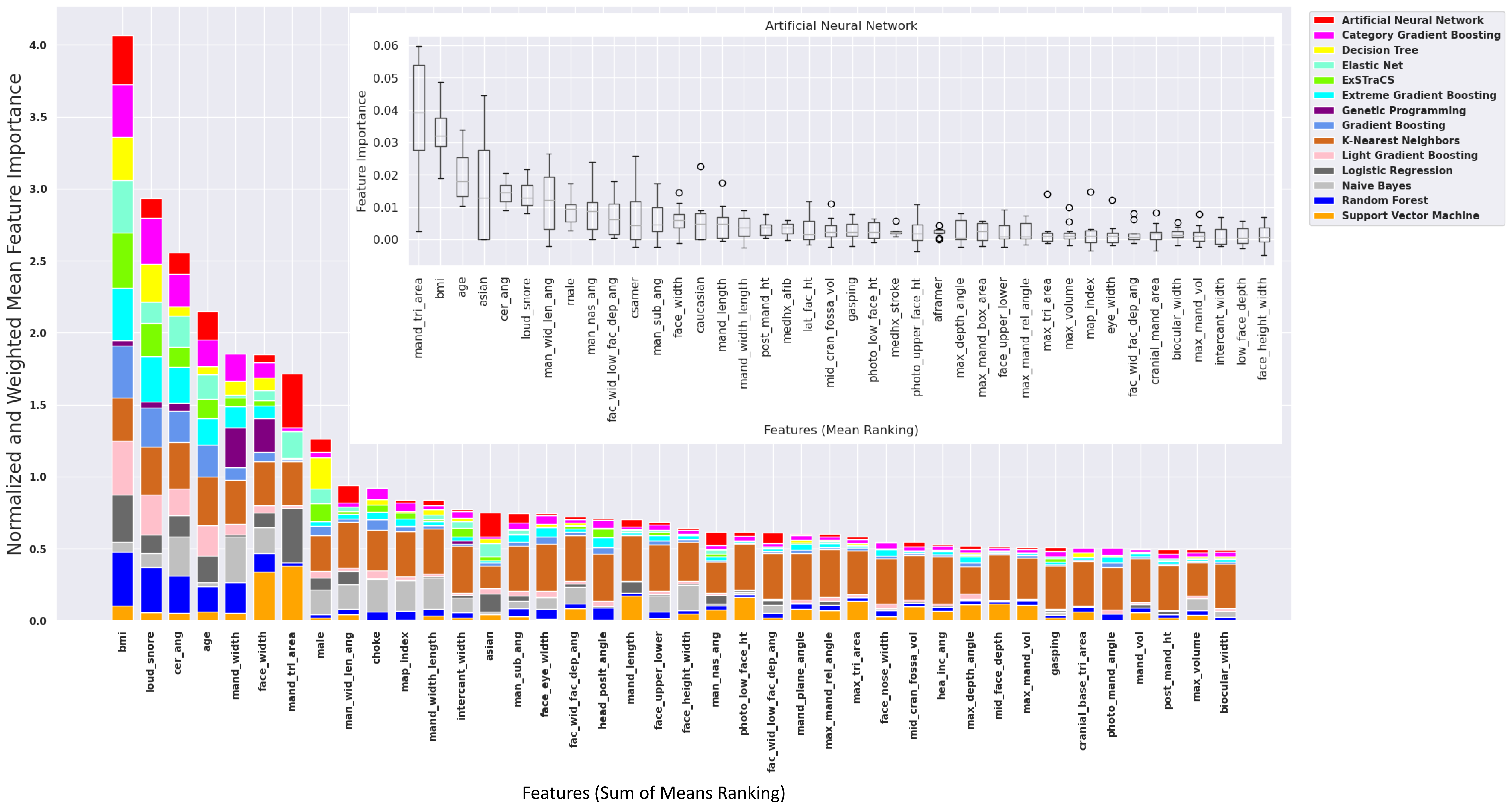}
  \caption{\textbf{Dataset H (AHI$\geq$15 and DEM+DX+SYM+CF) Model FI Estimates. }This composite feature importance plot illustrates normalized and (balanced accuracy)-weighted mean FI scores summed across all algorithms. Only the top 40 features are displayed. Embedded in the main plot is a boxplot of ANN FI estimates for the same dataset, showing the distribution of model FI estimates across all 10-fold CV partition models.}
  \label{fig:fig4}
\end{figure}

Focusing further on dataset H (which includes DEM+DX+SYM+CF), Figure \ref{fig:fig4} presents model FI estimates across all algorithms and a more detailed look at FI estimates for the ANN algorithm. Across all algorithms, the top 10 mean FI scores included DEM features (BMI, age, and male), SYM features (loud snoring and choking/struggling for breath), and CF features (cervicomental angle, mandibular width, face width, mandibular triangular area, and mandibular width to length angle). For the top performing ANN algorithm, the top 10 mean FI scores included DEM features (BMI, age, Asian race, and male), SYM feature (loud snoring) and CF features (mandibular triangular area, cervicomental angle, mandibular width to length angle, mandibular-nasion angle, and face width to lower face depth angle). 

Shifting to AHI$\geq$5 (Figure 3C), no step-wise feature set additions led to a significant increase in PRC-AUC. However, we observe consistently increasing median best-algorithm performance with the addition of each subsequent feature set, with CF and IO yielding minimal improvements. The only significant PRC-AUC increases were observed between datasets K and R (i.e. DEM vs. DEM+DX+SYM+CF) and K and T (i.e. DEM vs. DEM+DX+SYM+CF+IO) (see Table S19). The highest overall median PRC-AUC across both AHI$\geq$5 and AHI$\geq$15 datasets was obtained on dataset T (DEM+DX+SYM+CF+IO) using elastic net. Figure 3D reveals a similar pattern of PRC-AUC performance across AHI$\geq$5 datasets when exclusively focusing on the 10 elastic net models trained for each CV partition. In comparing all algorithm PRC-AUCs on dataset T, elastic net performed significantly better than decision tree, genetic programming, K-nearest neighbors, and Naïve Bayes algorithms, and similarly to the rest (see Table S20). A detailed evaluation of dataset T across 14 ML algorithms using 16 classification metrics is given in Figures S22-S24 and Tables S21-S22.

\begin{figure}
  \centering
  \includegraphics[width=\textwidth]{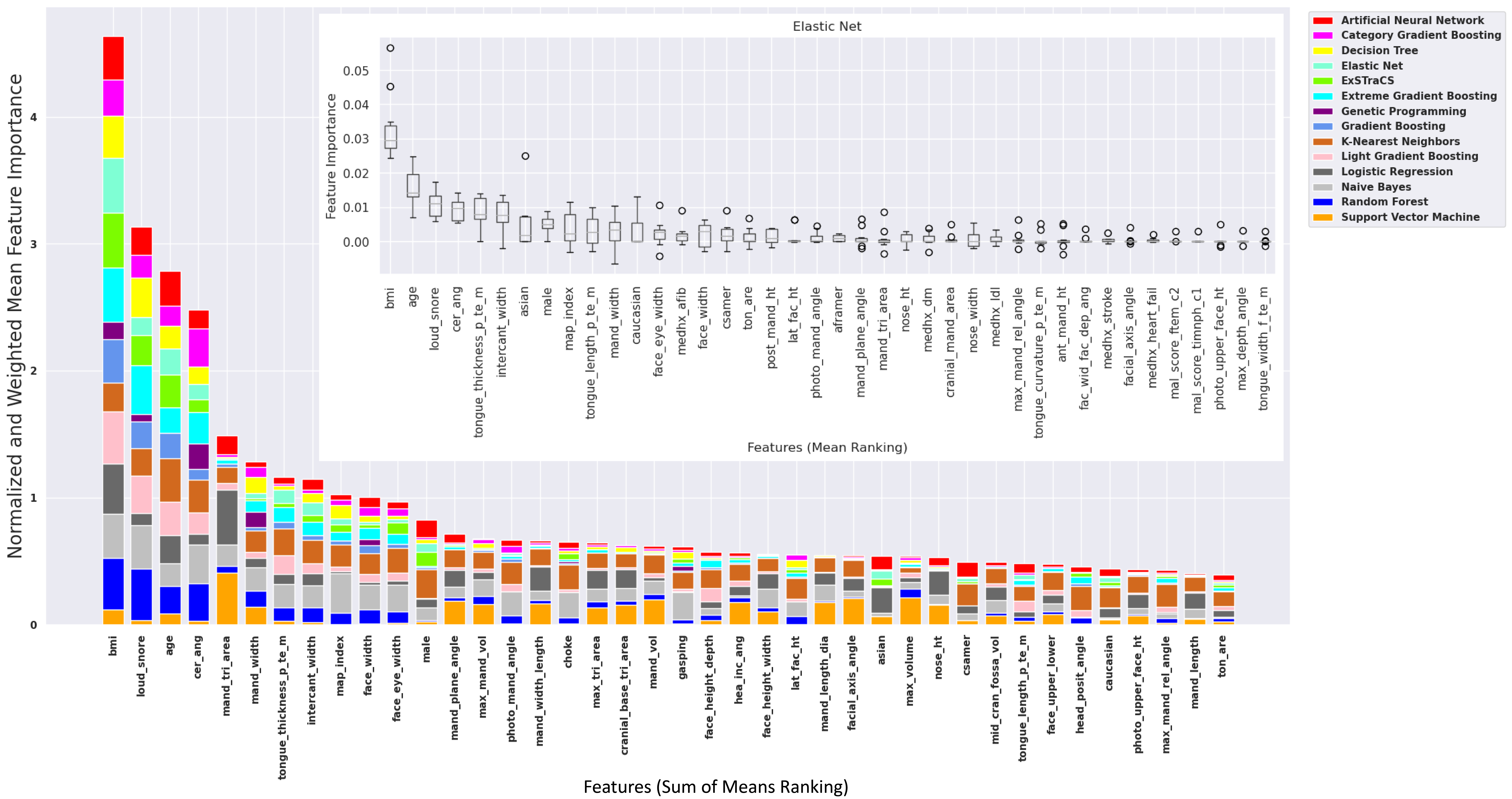}
  \caption{\textbf{Dataset T (AHI$\geq$5 and DEM+DX+SYM+CF+IO) Model FI Estimates. }This composite feature importance plot illustrates normalized and (balanced accuracy)-weighted mean FI scores summed across all algorithms. Only the top 40 features are displayed. Embedded in the main plot is a boxplot of logistic regression FI estimates for the same dataset, showing the distribution of model FI estimates across all 10-fold CV partition models.}
  \label{fig:fig5}
\end{figure}

Focusing further on dataset T (which includes all feature sets), Figure \ref{fig:fig5} presents model FI estimates across all algorithms and a more detailed look at FI estimates for the elastic net algorithm. Across all algorithms, the top 10 mean FI scores included DEM features (BMI and age), SYM features (loud snoring and MAP index), CF features (cervicomental angle, mandibular triangular area, mandibular width, intercanthal width, and face width) and one IO feature (tongue thickness). For the top performing elastic net algorithm, the top 10 mean FI scores included DEM features (BMI, age, Asian race, and male), SYM features (loud snoring and MAP index), CF features (cervicomental angle, and intercanthal width) and IO features (tongue thickness and tongue length). 

\subsubsection{ML Modeling Replication Evaluations}
Thus far, all OSA model evaluations have been conducted on respective hold-out testing datasets (different for each CV partition). Next, STREAMLINE was applied to re-evaluate all ML models (i.e. 10 individual CV models trained by each of 14 ML algorithms) on the same replication dataset (processed differently/appropriately for each respective CV training partition; see S.2.1.8). 

\begin{figure}
  \centering
  \includegraphics[width=\textwidth]{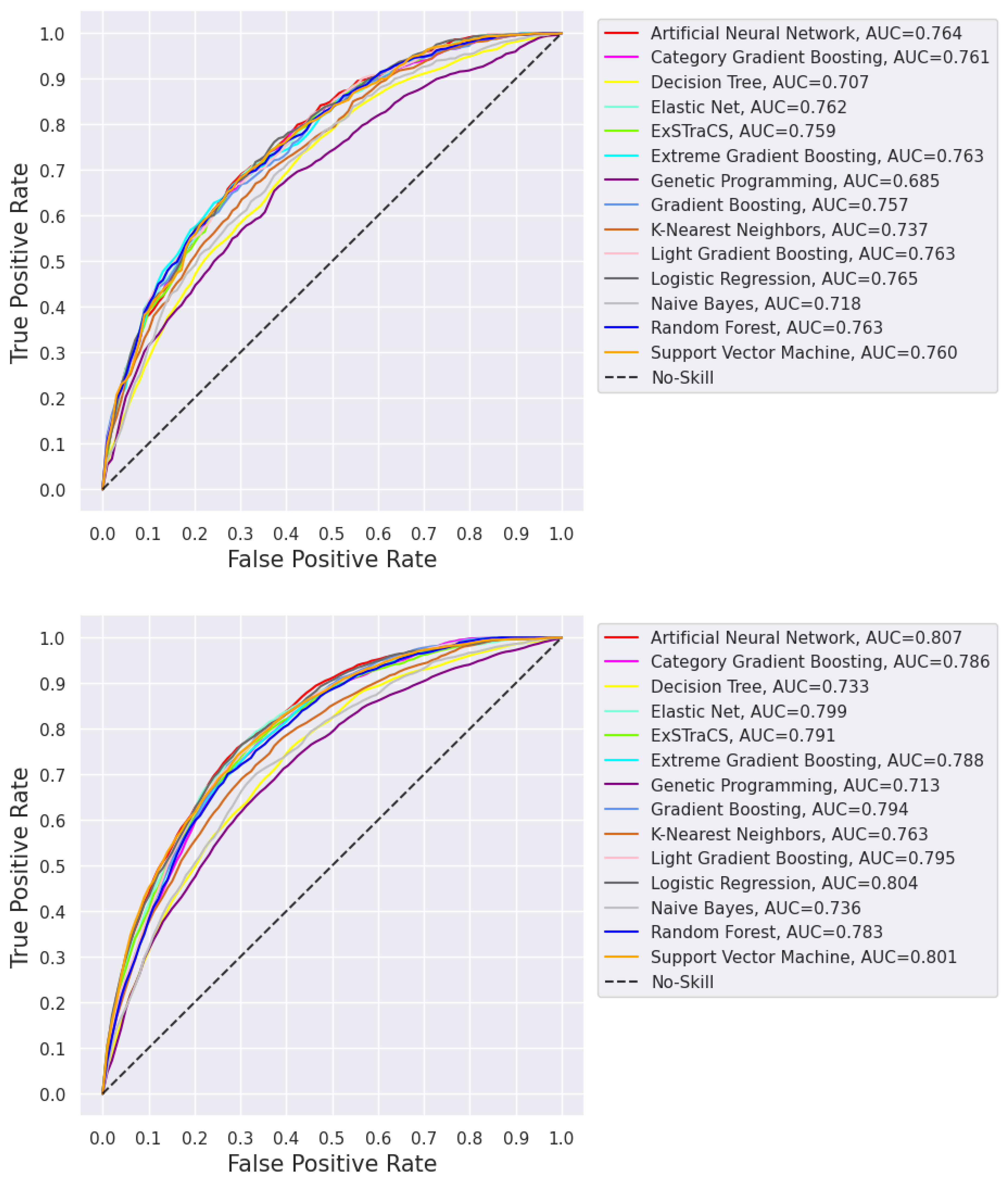}
  \caption{\textbf{ROC Plots for Dataset H Contrasting Testing vs Replication Algorithm Performance. }Dataset H encodes OSA as AHI$\geq$15 and includes feature sets DEM+DX+SYM+CF. Each algorithm line represents an average ROC across 10-fold CV models. (Top) Testing data evaluation; each CV model is evaluated on a unique hold-out testing partition. (Bottom) Replication data evaluation; each CV model is evaluated on the same hold-out replication dataset processed consistent with the model’s corresponding training data.}
  \label{fig:fig6}
\end{figure}

\begin{figure}
  \centering
  \includegraphics[width=\textwidth]{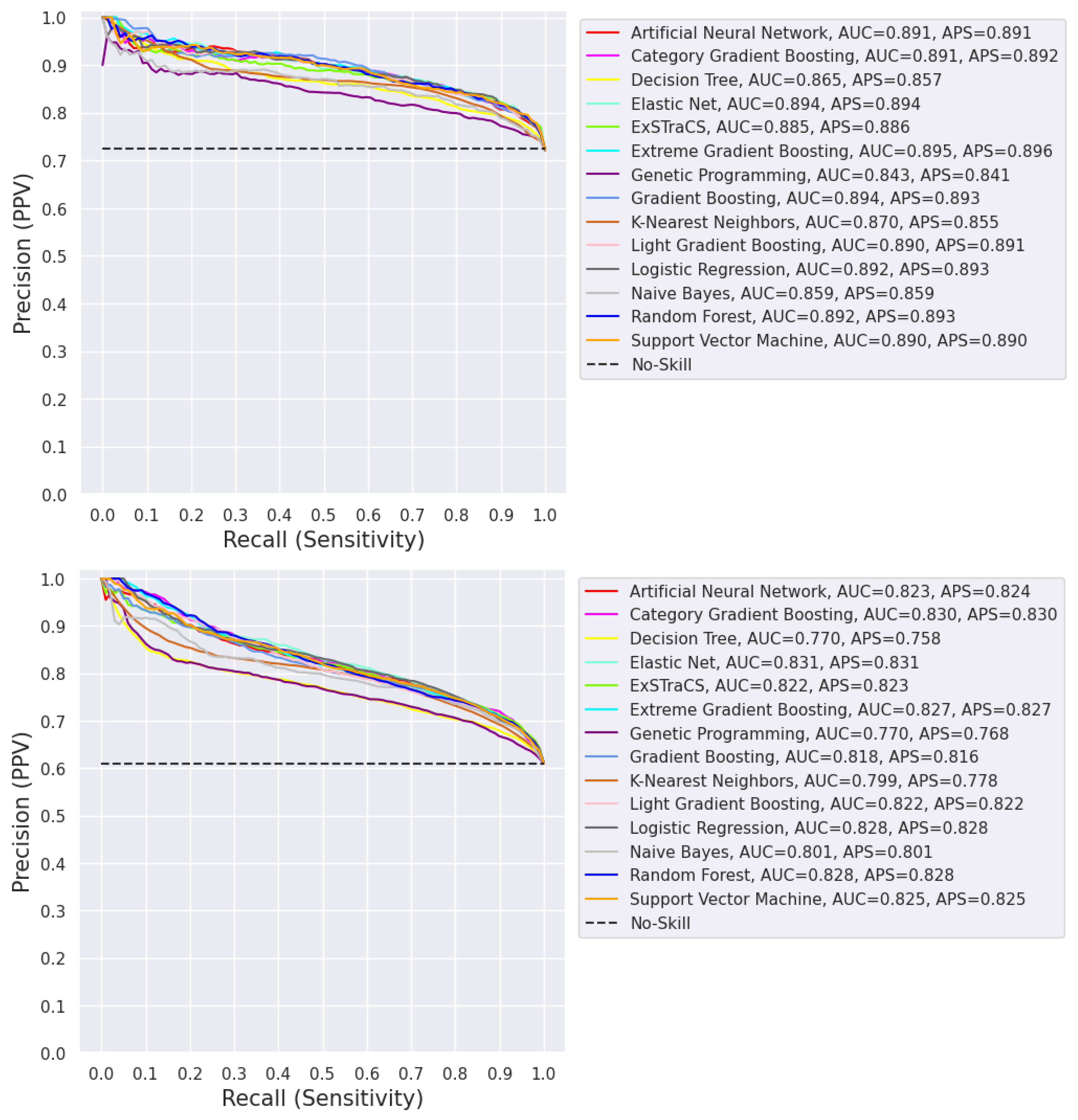}
  \caption{\textbf{PRC Plots for Dataset T Contrasting Testing vs Replication Algorithm Performance. }Dataset T encodes OSA as AHI$\geq$5 and includes feature sets DEM+DX+SYM+CF+IO. Each algorithm line represents an average PRC across 10-fold CV models. (Top) Testing data evaluation; each CV model is evaluated on a unique hold-out testing partition. (Bottom) Replication data evaluation; each CV model is evaluated on the same hold-out replication dataset processed consistent with the model’s corresponding training data.}
  \label{fig:fig7}
\end{figure}

For each of the 20 OSA datasets we compare the models that previously yielded the best median testing performance metric (either ROC-AUC for AHI$\geq$15 or PRC-AUC for AHI$\geq$5) to the performance of the same models on the replication data (see Table S23). For AHI$\geq$15, the best replication performance was again observed for dataset H with ANN models, having a significantly increased ROC-AUC of 0.807 (p=0.0002) compared to testing evaluation. Figure \ref{fig:fig6} gives average ROC plots comparing all algorithms on testing vs. replication data for dataset H. For all other AHI$\geq$15 datasets, we observed a mix of significant and non-significant increases and decreases in ROC-AUC replication performance. For AHI$\geq$5, the best replication performance was again observed for dataset T with elastic net models, although there was a significantly decreased PRC-AUC of 0.830 (p=0.0002) compared to the testing evaluation. Figure \ref{fig:fig7} gives average PRC plots comparing all algorithms on testing vs. replication data for dataset T. For all other AHI$\geq$5 datasets we similarly observed significant decreases in PRC-AUC replication performance compared to the testing evaluation.

Additionally, for each of the 20 OSA datasets we compare the models that previously yielded the best median testing performance metric (either ROC-AUC for AHI$\geq$15 or PRC-AUC for AHI$\geq$5) to the same (or different) models that yielded the best median performance on the replication data (see Table S24). For AHI$\geq$15, the best replication ROC-AUC was again observed for dataset H with ANN models and all other AHI$\geq$15 datasets yielded a consistently (but not always significantly) higher median ROC-AUC on replication data compared with testing evaluations. For AHI$\geq$5, the best median replication performance was again observed for dataset T with elastic net models, and all other AHI$\geq$5 datasets yielded significantly lower median PRC-AUC on replication data compared to testing evaluations.

Table S24 also identifies the best performing individual model on the replication data across all 20 OSA datasets. For AHI$\geq$15, a top ROC-AUC of 0.813 was obtained by CV model ‘1’ using the ANN algorithm on dataset H (DEM+DX+SYM+CF). For AHI$\geq$5, a top PRC-AUC of 0.847 was obtained by CV model ‘8’ using the category gradient boosting algorithm on dataset R (DEM+DX+SYM+CF). Optuna-optimized hyperparameter settings for the two models are specified in S.3.5.6.1

\section{Discussion}
In the subsections below we discuss (1) benchmarking of STREAMLINE, (2) applying STREAMLINE to OSA prediction including clinical relevance and potential limitations, (3) potential uses of STREAMLINE, and (4) future directions for STREAMLINE development. 

\subsection{Benchmarking}
Our primary goal in benchmarking analyses was to transparently present the design, capabilities, and efficacy of STREAMLINE as an automated ML analysis pipeline. The benchmarking analyses conducted in this study (sections S.3.1-S.3.4) successfully validated STREAMLINE efficacy, contrasted performance of v.0.3.4 with the v0.2.5, and compared the performance of individual ML modeling algorithms across simple to extremely complex dataset scenarios. These analyses also identified future directions for expanding and improving the STREAMLINE framework as discussed in section 4.4.

We acknowledge that there are countless ways to assemble a valid ML analysis pipeline and therefore do not claim that STREAMLINE offers an ‘ideal’ analytical configuration for every problem/dataset. STREAMLINE uniquely brings together reliable, well-established, but often overlooked or misused approaches to data preparation and modeling as a wholistic methodology for machine learning analyses in biomedicine while uniquely incorporating cutting-edge machine learning methods well-suited to the challenges and complexities of biomedical data (e.g. MultiSURF \cite{urbanowicz_benchmarking_2018}, collective feature selection \cite{verma_collective_2018}, and ExSTraCS \cite{urbanowicz_exstracs_2015}). While we have highlighted the rigor and novelty of STREAMLINE in contrast with other AutoML tools (see sections 1 and S.1), this study has not made direct performance comparisons to other AutoML tools. Benchmarking between different AutoML tools will be a focus of future work as it presents a variety of major challenges regarding AutoML comparability, computational expense, and the choice of benchmarking datasets and benchmarking study design (as detailed in S.4.1). 

In brief, ‘how’ to benchmark effectively remains an open topic of debate even in the comparison of individual modeling algorithms \cite{liao_are_2021}. These challenges are only exaggerated in seeking to compare the performance of different AutoML tools which becomes a meta-optimization problem. Ultimately, we expect convincing AutoML comparison to necessarily examine predictive performance across multiple metrics using a wide variety of simulated and real-world datasets with different underlying association complexities, data dimensions, signal/noise ratio, feature types, class balance/imbalance, feature value frequencies and distributions, and ratio of informative to non-informative features. AutoML stochasticity should also be taken into account in performance comparisons by conducting multiple runs of each tool on each benchmark dataset. Sufficiently rigorous AutoML benchmarking is thus expected to be extremely computationally expensive. Beyond metric performance, AutoML comparisons should also examine computational complexity/run time, model/pipeline interpretability and transparency, what pipeline elements are automated, and ease of use.

Furthermore, convincing AutoML benchmarking will have to establish how to fairly compare AutoML tools that adopt different approaches to data partitioning (impacting the quantity and consistency of which instances are used for training vs. hold-out evaluation) as well as how to fairly choose different AutoML run parameters and computational constraints in such comparisons. While some attempts have been made to compare the performance of a subset of AutoML tools \cite{nikitin_automated_2022,gijsbers_gama_2021,H2OAutoML20,zoller_benchmark_2021,balaji_benchmarking_2018,ferreira_comparison_2021,gijsbers_amlb_2022}, no study has yet addressed all of the above issues. 

In lieu of reliable cross-AutoML benchmarking, we recommend users select an AutoML tool that can handle the demands of the given task (e.g. data type, intended model use), and which is transparent enough to understand exactly how the data is being processed, how the model is trained vs. evaluated, and whether best practices are being properly adopted.  

\subsection{OSA Feature Set Utility and ML Model Performance}
STREAMLINE phases 1-7 focus on the development data, which was further partitioned into training and testing sets for evaluation. This analysis was used to compare model performance between algorithms and datasets having different OSA feature subsets and outcomes. As highlighted in section 3.2.1, AHI$\geq$5 datasets had a relatively large degree of class imbalance (majority of instances labeled as positive for ‘any OSA’), leading us to primarily evaluate these datasets with PRC-AUC. In contrast, AHI$\geq$15 datasets (moderate/severe OSA), were balanced and focused on ROC-AUC. Across all feature sets, AHI$\geq$5 datasets yielded higher model performance than AHI$\geq$15 datasets (see Table S15). While a higher PRC-AUC would be expected by default in AHI$\geq$5 datasets given the imbalanced classes, this performance difference was similarly observed for both balanced accuracy and ROC-AUC, indicating that the ML algorithms were more effective at predicting ‘any OSA’. Comparisons between datasets having only a single feature set (i.e. A-E and K-O) provided insight to their standalone utility. This highlighted the similarly high predictive value of DEM and CF feature sets, and the lower predictive value of available DX features. 

Comparisons between datasets with stepwise addition of feature sets, ordered by general availability/accessibility (i.e. DEM, +DX, +SYM, +CF, +IO), indicated that CF added significant value for predicting ‘moderate/severe’ OSA, but both CF and IO added minimal value for predicting ‘any OSA’. Thus, facial photographic measurements appear to not significantly improve the prediction of OSA status when subjects with mild OSA (defined as AHI of 5-15/hour) are included. However, craniofacial features from digital photographs did improve the prediction of moderate-to-severe OSA. As a key purpose of predicting OSA risk is to prioritize diagnosis and treatment of those individuals with more significant disease (e.g., those with moderate/severe OSA), these results suggest CF features may have clinical value. Of note, it is also possible that the predictive value of CF and IO digital photographs may be underestimated by the specific set of quantitative measures that were extracted for analysis, and that ML algorithms able to directly train on the images (e.g. convolutional neural networks) could potentially derive greater predictive value, albeit with limited interpretability.
 
Interestingly, the addition of DX features to DEM led to a non-significant reduction in ROC-AUC but a non-significant improvement in PRC-AUC, indicating that DX features improved OSA prediction at the expense of more false positives (highlighted in Figures \ref{fig:fig2} and \ref{fig:fig3}). The same trend was observed when adding IO features to DEM+DX+SYM+CF. Ultimately, in this testing evaluation, the highest median ROC-AUC for moderate/severe OSA was obtained on dataset H (DEM+DX+SYM+CF) using the ANN algorithm and the highest median PRC-AUC for predicting any OSA was obtained on dataset T (DEM+DX+SYM+CF+IO) using the elastic net algorithm. Future work should more closely examine OSA prediction in the top performing feature sets alone (i.e. DEM+SYM+CF), as well as evaluate how the predictive value of the CF and IO features may differ based on certain DEM traits such as race/ethnicity (as suggested by a recent SAGIC publication \cite{sutherland_facial_2023}). A potential clinical limitation of this study is that subjects in the SAGIC dataset had already been referred for a sleep test on suspicion of OSA (explaining the imbalance in AHI$\geq$5). Thus, future work should ideally re-evaluate model performance for screening for OSA in general population samples.  

Figures S25-S27 compare feature ranking consistency among the top 10 features within univariate analyses, pre-modeling FI estimates, model FI estimates across all algorithms, and model FI estimates for the top median performing algorithm. BMI was the only feature consistently identified for both AHI$\geq$15 and AHI$\geq$5, in-line with the well-established relationship between OSA and obesity \cite{young_occurrence_1993,peppard_increased_2013}. For AHI$\geq$15, cervicomental angle was also consistently top-ranked, followed by loud snoring, face width, mandibular width, and mandibular triangular area, each in the top 10 in 4/5 rankings. For AHI$\geq$5, loud snoring was also consistently top-ranked, followed by mandibular width, cervicomental angle, and face width, each in the top 10 in 4/5 rankings. For model FI estimates, only DEM, SYM, CF and IO features ranked in the top 10 scorings.

STREAMLINE phase 8 focused on the replication data, which is used to further assess model generalizability and globally compare model performance to pick a “best” model. For AHI$\geq$15 and AHI$\geq$5, the best median performance was obtained on datasets H and T, respectively, consistent with testing evaluations. Notably, as summarized by Table S24, AHI$\geq$15 datasets yielded larger ROC-AUCs on replication data compared to testing data, while AHI$\geq$5 replication datasets yielded smaller PRC-AUCs, reflecting a general increase in true negative rate and reduction in true positive rate. Ideally, when using STREAMLINE to select a best performing model, investigators would conduct additional replication analyses across multiple sites or subpopulations and select either (1) the model that performs best across these sites or (2) the best performing model for a given site/sub-population. However, this is beyond the scope of the current OSA study. 

Lastly, across all benchmarking and OSA datasets, we observed that the ‘best’ performing algorithm often varies depending on the dataset, metric, and whether the evaluation is on testing or replication data. This variability is due to many factors including algorithm and pipeline stochasticity, class balance, feature types, and underlying sample bias in data partitioning. This highlights the importance of considering a variety of ML modeling algorithms and metrics which can be facilitated by AutoML tools in tackling each new data analysis.

\subsection{STREAMLINE Uses}
STREAMLINE was primarily designed for biomedical researchers to easily run a carefully designed and rigorous ML analysis across one or more target datasets, comparing performance and assess feature importance across multiple algorithms with testing data evaluations, as well as pick a top performing model with the replication evaluations. This is in-line with the objectives of the OSA analyses above. However, STREAMLINE is also designed to be used as a rigorous framework for benchmarking and comparing newly developed (scikit-learn compatible) ML modeling algorithms to the other established algorithms (currently included). This can be achieved with minimal coding to add other algorithms to its repertoire. Furthermore, we expect STREAMLINE will be useful as a standard of comparison (i.e. a positive control) with which to evaluate and benchmark other emerging AutoML tools, in particular those that seek to automatically discover and optimize pipeline assembly as a whole (e.g. TPOT \cite{le_scaling_2020}, FEDOT \cite{nikitin_automated_2022}, Auto-Sklearn \cite{feurer_efficient_2015}, GAMA \cite{gijsbers_gama_2021}, RECIPE \cite{de_sa_recipe_2017}, and ML-Plan \cite{mohr_ml-plan_2018}).

\subsection{STREAMLINE Future Development}
While STREAMLINE (v.0.3.4) was successfully validated, and performed well on the analyses conducted herein, it is actively being improved/extended (see also S.4.3). Currently, the biggest limitation being addressed is extension to data with multi-class or quantitative (i.e. regression) outcomes. Other limitations highlighted by this study that we aim to address include: (1) improve the collective feature selection strategy (phase 4) to account for the possibility of pure, higher-order feature interactions as identified in the XOR benchmarking analysis (see S.3.4); (2) consider alternative approaches to comparing PRC-AUC between datasets with differences in class balance; (3) directly incorporate model fairness evaluations and fairness-inclusive optimization into the framework to help address underlying biases in the data \cite{caton2020fairness}; and (4) add a new phase to STREAMLINE following modeling (phase 5) that automatically trains voting and stacking ensemble models using phase 5 baseline models and comparing ensembles to other baseline models in post analysis (phase 6). We welcome feedback on how to improve/extend STREAMLINE to address a broader range of analytical needs and challenges.

\section{Conclusion}
This work introduced a broadly refactored and extended release of STREAMLINE (v.0.3.4), an end-to-end AutoML analysis pipeline designed to address the needs of biomedical structured data analysis. While currently limited to data analysis tasks with binary outcome, STREAMLINE automation makes it easy for users with and without programming experience to understand, process, partition, and prepare their dataset(s), as well as train, evaluate, and replicate models using a diversity of established and cutting-edge ML algorithms in a carefully structured, rigorous, and transparent manner. Simulated and real-world benchmarking analyses validated STREAMLINE efficacy and highlighted the value of using a diversity of ML algorithms and evaluation metrics, including ExSTraCS, a rule-based ML algorithm, that outperformed other algorithms on problems with a high degree of underlying complexity (e.g. epistatic interactions and heterogeneous associations). 

Application of STREAMLINE AutoML to evaluate the utility of available feature sets in predicting moderate/severe OSA (AHI$\geq$15) or any OSA (AHI$\geq$5) supported the following: (1) ML models performed significantly and substantially better in predicting any OSA vs. moderate/severe OSA, (2) the predictive value of individual features sets can be ranked as DEM>CF>SYM>IO>DX, (3) CF features provide significant added value in predicting moderate/severe OSA, and (4) neither CF or IO features significantly improved the prediction of any OSA. Overall, these findings suggest that in combination with DEM, DX, and SYM features, photography-based anatomy measures provided conditional added value in the prediction of OSA outcomes, with the strongest benefit coming from adding CF features in the prediction of moderate/severe OSA.

\section*{Acknowledgments}
The authors acknowledge the helpful comments and suggestions by Dr. Jason H. Moore, Dr. Li Shen, and Pedro Ribeiro in the development and testing of STREAMLINE and Dr. Kate Sutherland for her involvement in SAGIC data collection. The study was supported in part by the following NIH grants: R01s AI173095, AG066833, and P01 HL160471.

\section*{Funding}
The study was supported in part by the following NIH grants: R01s AI173095, AG066833, and P01 HL160471.

\section*{Data Availability Statement}
Benchmarking datasets used in this study are all available publicly or upon request. The OSA data analyzed here have been supplied by the Sleep Apnea Global Interdisciplinary Consortium (SAGIC) and are not publicly available but may be made available upon request and approval by the SAGIC investigators. Queries or clarifications regarding the results can be made to the corresponding author. The authors are open to communicating and/or collaborating with other researchers interested in this topic. 

\section*{CrediT Authorship Contribution Statement}
\textbf{Ryan J. Urbanowicz:} Conceptualization; Data curation; Formal Analysis; Methodology; Project administration; Software; Supervision; Visualization; Validation; Writing – Original Draft; Writing – Review \& Editing
\textbf{Harsh Bandhey:} Formal Analysis; Methodology; Software; Visualization; Writing – Review \& Editing
\textbf{Brendan T. Keenan:} Data curation; Writing – Original Draft; Writing – Review \& Editing
\textbf{Greg Maislin:} Data curation; Writing – Review \& Editing
\textbf{Sy Hwang:} Visualization; Writing – Review \& Editing
\textbf{Danielle L. Mowery:} Methodology; Writing – Review \& Editing
\textbf{Shannon M. Lynch:} Methodology; Writing – Review \& Editing
\textbf{Diego R. Mazzotti:} Methodology; Writing – Review \& Editing
\textbf{Fang Han:} Resources; Writing – Review \& Editing
\textbf{Qing Yun Li:} Resources; Writing – Review \& Editing
\textbf{Thomas Penzel:} Resources; Writing – Review \& Editing
\textbf{Sergio Tufik:} Resources; Writing – Review \& Editing
\textbf{Lia Bittencourt:} Resources; Writing – Review \& Editing
\textbf{Thorarinn Gislason:} Resources; Writing – Review \& Editing
\textbf{Philip de Chazal:} Resources; Writing – Review \& Editing
\textbf{Bhajan Sing:} Resources; Writing – Review \& Editing
\textbf{Nigel McArdle:} Resources; Writing – Review \& Editing
\textbf{Ning-Hung Chen:} Resources; Writing – Review \& Editing
\textbf{Allan Pack:} Conceptualization; Project administration; Resources; Supervision; Writing – Review \& Editing
\textbf{Richard J. Schwab:} Resources; Writing – Review \& Editing
\textbf{Peter A. Cistulli:} Resources; Writing – Review \& Editing
\textbf{Ulysses J. Magalang:} Conceptualization; Resources; Supervision; Writing – Review \& Editing

\section*{Declaration of Competing Interest}
The authors declare the following financial interests/personal relationships not directly related to this present study but which may be considered as potential competing interests from the past 3 years: Thomas Penzel reports consultation and speaker fees from Bayer, Bioprojet, Cerebra, Idorsia, Jazz Pharma, Sleepimage, Löwenstein Medical, Philips, and the National Sleep Foundation, and grant funding from Cidelec, Löwenstein Medical, and Novarti. Bhajan Singh reports honoraria from SomnoMed Australia. Richard J. Schwab reports grant funding from ResMed, Inspire, and CryOSA, royalties from UpToDate and Merck Manual, research consulting for Eli Lilly, and is on the Medical Advisory Board for eXciteOSA. Peter A. Cistulli has an appointment to an endowed academic Chair at the University of Sydney that was established from ResMed funding, has received research support from ResMed and SomnoMed, and is a consultant to ResMed, SomnoMed, Signifier Medical Technologies, Bayer, and Sunrise Medical.

\section*{Appendix A. Supplementary Materials}
The following are the supplementary materials to this article:
\begin{itemize}
    \item Supporting Information Document
    \item hcc$\_$exp$\_$STREAMLINE$\_$Report.pdf
    \item hcc$\_$exp$\_$STREAMLINE$\_$Replication$\_$Report.pdf
    \item gametes$\_$exp$\_$STREAMLINE$\_$Report.pdf
    \item multiplexer$\_$exp$\_$STREAMLINE$\_$Report.pdf
    \item xor$\_$exp$\_$STREAMLINE$\_$Report.pdf
    \item xor$\_$nofs$\_$exp$\_$STREAMLINE$\_$Report.pdf
    \item sagic$\_$exp$\_$STREAMLINE$\_$Report.pdf
\end{itemize}

\bibliographystyle{unsrtnat}  
\bibliography{main}

\begin{thebibliography}{96}
\providecommand{\natexlab}[1]{#1}
\providecommand{\url}[1]{\texttt{#1}}
\expandafter\ifx\csname urlstyle\endcsname\relax
  \providecommand{\doi}[1]{doi: #1}\else
  \providecommand{\doi}{doi: \begingroup \urlstyle{rm}\Url}\fi

\bibitem[Jovel and Greiner(2021)]{jovel_introduction_2021}
Juan Jovel and Russell Greiner.
\newblock An {Introduction} to {Machine} {Learning} {Approaches} for {Biomedical} {Research}.
\newblock \emph{Frontiers in Medicine}, 8, 2021.
\newblock ISSN 2296-858X.
\newblock URL \url{https://www.frontiersin.org/articles/10.3389/fmed.2021.771607}.

\bibitem[Badillo et~al.(2020)Badillo, Banfai, Birzele, Davydov, Hutchinson, Kam-Thong, Siebourg-Polster, Steiert, and Zhang]{badillo_introduction_2020}
Solveig Badillo, Balazs Banfai, Fabian Birzele, Iakov~I. Davydov, Lucy Hutchinson, Tony Kam-Thong, Juliane Siebourg-Polster, Bernhard Steiert, and Jitao~David Zhang.
\newblock An {Introduction} to {Machine} {Learning}.
\newblock \emph{Clinical Pharmacology \& Therapeutics}, 107\penalty0 (4):\penalty0 871--885, 2020.
\newblock ISSN 1532-6535.
\newblock \doi{10.1002/cpt.1796}.
\newblock URL \url{https://onlinelibrary.wiley.com/doi/abs/10.1002/cpt.1796}.
\newblock \_eprint: https://onlinelibrary.wiley.com/doi/pdf/10.1002/cpt.1796.

\bibitem[Mirza et~al.(2019)Mirza, Wang, Wang, Choi, Chung, and Ping]{mirza_machine_2019}
Bilal Mirza, Wei Wang, Jie Wang, Howard Choi, Neo~Christopher Chung, and Peipei Ping.
\newblock Machine {Learning} and {Integrative} {Analysis} of {Biomedical} {Big} {Data}.
\newblock \emph{Genes}, 10\penalty0 (2):\penalty0 87, February 2019.
\newblock ISSN 2073-4425.
\newblock \doi{10.3390/genes10020087}.
\newblock URL \url{https://www.mdpi.com/2073-4425/10/2/87}.
\newblock Number: 2 Publisher: Multidisciplinary Digital Publishing Institute.

\bibitem[Peek et~al.(2014)Peek, Holmes, and Sun]{peek_technical_2014}
N.~Peek, J.~H. Holmes, and J.~Sun.
\newblock Technical {Challenges} for {Big} {Data} in {Biomedicine} and {Health}: {Data} {Sources}, {Infrastructure}, and {Analytics}.
\newblock \emph{Yearbook of Medical Informatics}, 23\penalty0 (1):\penalty0 42--47, 2014.
\newblock ISSN 0943-4747, 2364-0502.
\newblock \doi{10.15265/IY-2014-0018}.
\newblock URL \url{http://www.thieme-connect.de/DOI/DOI?10.15265/IY-2014-0018}.
\newblock Publisher: Georg Thieme Verlag KG.

\bibitem[Whalen et~al.(2022)Whalen, Schreiber, Noble, and Pollard]{whalen_navigating_2022}
Sean Whalen, Jacob Schreiber, William~S. Noble, and Katherine~S. Pollard.
\newblock Navigating the pitfalls of applying machine learning in genomics.
\newblock \emph{Nature Reviews Genetics}, 23\penalty0 (3):\penalty0 169--181, March 2022.
\newblock ISSN 1471-0064.
\newblock \doi{10.1038/s41576-021-00434-9}.
\newblock URL \url{https://www.nature.com/articles/s41576-021-00434-9}.
\newblock Number: 3 Publisher: Nature Publishing Group.

\bibitem[Garreta et~al.(2017)Garreta, Moncecchi, Hauck, and Hackeling]{garreta_scikit-learn_2017}
Raul Garreta, Guillermo Moncecchi, Trent Hauck, and Gavin Hackeling.
\newblock \emph{scikit-learn : {Machine} {Learning} {Simplified}: {Implement} scikit-learn into every step of the data science pipeline}.
\newblock Packt Publishing Ltd, November 2017.
\newblock ISBN 978-1-78883-152-9.
\newblock Google-Books-ID: sEFPDwAAQBAJ.

\bibitem[Greener et~al.(2022)Greener, Kandathil, Moffat, and Jones]{greener_guide_2022}
Joe~G. Greener, Shaun~M. Kandathil, Lewis Moffat, and David~T. Jones.
\newblock A guide to machine learning for biologists.
\newblock \emph{Nature Reviews Molecular Cell Biology}, 23\penalty0 (1):\penalty0 40--55, January 2022.
\newblock ISSN 1471-0080.
\newblock \doi{10.1038/s41580-021-00407-0}.
\newblock URL \url{https://www.nature.com/articles/s41580-021-00407-0}.
\newblock Number: 1 Publisher: Nature Publishing Group.

\bibitem[Chicco(2017)]{chicco_ten_2017}
Davide Chicco.
\newblock Ten quick tips for machine learning in computational biology.
\newblock \emph{BioData Mining}, 10\penalty0 (1):\penalty0 35, December 2017.
\newblock ISSN 1756-0381.
\newblock \doi{10.1186/s13040-017-0155-3}.
\newblock URL \url{https://doi.org/10.1186/s13040-017-0155-3}.

\bibitem[Seghier(2022)]{seghier_ten_2022}
Mohamed~L. Seghier.
\newblock Ten simple rules for reporting machine learning methods implementation and evaluation on biomedical data.
\newblock \emph{International Journal of Imaging Systems and Technology}, 32\penalty0 (1):\penalty0 5--11, 2022.
\newblock ISSN 1098-1098.
\newblock \doi{10.1002/ima.22674}.
\newblock URL \url{https://onlinelibrary.wiley.com/doi/abs/10.1002/ima.22674}.
\newblock \_eprint: https://onlinelibrary.wiley.com/doi/pdf/10.1002/ima.22674.

\bibitem[Luo et~al.(2016)Luo, Phung, Tran, Gupta, Rana, Karmakar, Shilton, Yearwood, Dimitrova, Ho, Venkatesh, and Berk]{luo_guidelines_2016}
Wei Luo, Dinh Phung, Truyen Tran, Sunil Gupta, Santu Rana, Chandan Karmakar, Alistair Shilton, John Yearwood, Nevenka Dimitrova, Tu~Bao Ho, Svetha Venkatesh, and Michael Berk.
\newblock Guidelines for {Developing} and {Reporting} {Machine} {Learning} {Predictive} {Models} in {Biomedical} {Research}: {A} {Multidisciplinary} {View}.
\newblock \emph{Journal of Medical Internet Research}, 18\penalty0 (12):\penalty0 e5870, December 2016.
\newblock \doi{10.2196/jmir.5870}.
\newblock URL \url{https://www.jmir.org/2016/12/e323}.
\newblock Company: Journal of Medical Internet Research Distributor: Journal of Medical Internet Research Institution: Journal of Medical Internet Research Label: Journal of Medical Internet Research Publisher: JMIR Publications Inc., Toronto, Canada.

\bibitem[Heil et~al.(2021)Heil, Hoffman, Markowetz, Lee, Greene, and Hicks]{heil_reproducibility_2021}
Benjamin~J. Heil, Michael~M. Hoffman, Florian Markowetz, Su-In Lee, Casey~S. Greene, and Stephanie~C. Hicks.
\newblock Reproducibility standards for machine learning in the life sciences.
\newblock \emph{Nature Methods}, 18\penalty0 (10):\penalty0 1132--1135, October 2021.
\newblock ISSN 1548-7105.
\newblock \doi{10.1038/s41592-021-01256-7}.
\newblock URL \url{https://www.nature.com/articles/s41592-021-01256-7}.
\newblock Number: 10 Publisher: Nature Publishing Group.

\bibitem[Riley(2019)]{riley_three_2019}
Patrick Riley.
\newblock Three pitfalls to avoid in machine learning.
\newblock \emph{Nature}, 572\penalty0 (7767):\penalty0 27--29, August 2019.
\newblock \doi{10.1038/d41586-019-02307-y}.
\newblock URL \url{https://www.nature.com/articles/d41586-019-02307-y}.
\newblock Bandiera\_abtest: a Cg\_type: Comment Number: 7767 Publisher: Nature Publishing Group Subject\_term: Mathematics and computing, Software, Research data.

\bibitem[Smialowski et~al.(2010)Smialowski, Frishman, and Kramer]{smialowski_pitfalls_2010}
Pawel Smialowski, Dmitrij Frishman, and Stefan Kramer.
\newblock Pitfalls of supervised feature selection.
\newblock \emph{Bioinformatics}, 26\penalty0 (3):\penalty0 440--443, February 2010.
\newblock ISSN 1367-4803.
\newblock \doi{10.1093/bioinformatics/btp621}.
\newblock URL \url{https://doi.org/10.1093/bioinformatics/btp621}.

\bibitem[Uçar et~al.(2020)Uçar, Nour, Sindi, and Polat]{ucar_effect_2020}
Muhammed~Kürşad Uçar, Majid Nour, Hatem Sindi, and Kemal Polat.
\newblock The {Effect} of {Training} and {Testing} {Process} on {Machine} {Learning} in {Biomedical} {Datasets}.
\newblock \emph{Mathematical Problems in Engineering}, 2020:\penalty0 e2836236, May 2020.
\newblock ISSN 1024-123X.
\newblock \doi{10.1155/2020/2836236}.
\newblock URL \url{https://www.hindawi.com/journals/mpe/2020/2836236/}.
\newblock Publisher: Hindawi.

\bibitem[Wolpert and Macready(1997)]{wolpert_no_1997}
D.H. Wolpert and W.G. Macready.
\newblock No free lunch theorems for optimization.
\newblock \emph{IEEE Transactions on Evolutionary Computation}, 1\penalty0 (1):\penalty0 67--82, April 1997.
\newblock ISSN 1941-0026.
\newblock \doi{10.1109/4235.585893}.
\newblock Conference Name: IEEE Transactions on Evolutionary Computation.

\bibitem[Escalante(2021)]{escalante_automated_2021}
Hugo~Jair Escalante.
\newblock Automated {Machine} {Learning}—{A} {Brief} {Review} at the {End} of the {Early} {Years}.
\newblock In Nelishia Pillay and Rong Qu, editors, \emph{Automated {Design} of {Machine} {Learning} and {Search} {Algorithms}}, Natural {Computing} {Series}, pages 11--28. Springer International Publishing, Cham, 2021.
\newblock ISBN 978-3-030-72069-8.
\newblock \doi{10.1007/978-3-030-72069-8_2}.
\newblock URL \url{https://doi.org/10.1007/978-3-030-72069-8_2}.

\bibitem[Waring et~al.(2020)Waring, Lindvall, and Umeton]{waring_automated_2020}
Jonathan Waring, Charlotta Lindvall, and Renato Umeton.
\newblock Automated machine learning: {Review} of the state-of-the-art and opportunities for healthcare.
\newblock \emph{Artificial Intelligence in Medicine}, 104:\penalty0 101822, April 2020.
\newblock ISSN 0933-3657.
\newblock \doi{10.1016/j.artmed.2020.101822}.
\newblock URL \url{https://www.sciencedirect.com/science/article/pii/S0933365719310437}.

\bibitem[Jin et~al.(2023)Jin, Chollet, Song, and Hu]{jin_autokeras_2023}
Haifeng Jin, François Chollet, Qingquan Song, and Xia Hu.
\newblock {AutoKeras}: {An} {AutoML} {Library} for {Deep} {Learning}.
\newblock \emph{Journal of Machine Learning Research}, 24\penalty0 (6):\penalty0 1--6, 2023.
\newblock ISSN 1533-7928.
\newblock URL \url{http://jmlr.org/papers/v24/20-1355.html}.

\bibitem[Nikitin et~al.(2022)Nikitin, Vychuzhanin, Sarafanov, Polonskaia, Revin, Barabanova, Maximov, Kalyuzhnaya, and Boukhanovsky]{nikitin_automated_2022}
Nikolay~O. Nikitin, Pavel Vychuzhanin, Mikhail Sarafanov, Iana~S. Polonskaia, Ilia Revin, Irina~V. Barabanova, Gleb Maximov, Anna~V. Kalyuzhnaya, and Alexander Boukhanovsky.
\newblock Automated evolutionary approach for the design of composite machine learning pipelines.
\newblock \emph{Future Generation Computer Systems}, 127:\penalty0 109--125, February 2022.
\newblock ISSN 0167-739X.
\newblock \doi{10.1016/j.future.2021.08.022}.
\newblock URL \url{https://www.sciencedirect.com/science/article/pii/S0167739X21003307}.

\bibitem[Erickson et~al.(2020)Erickson, Mueller, Shirkov, Zhang, Larroy, Li, and Smola]{erickson_autogluon-tabular_2020}
Nick Erickson, Jonas Mueller, Alexander Shirkov, Hang Zhang, Pedro Larroy, Mu~Li, and Alexander Smola.
\newblock {AutoGluon}-{Tabular}: {Robust} and {Accurate} {AutoML} for {Structured} {Data}, March 2020.
\newblock URL \url{http://arxiv.org/abs/2003.06505}.
\newblock arXiv:2003.06505 [cs, stat].

\bibitem[Molino et~al.(2019)Molino, Dudin, and Miryala]{molino_ludwig_2019}
Piero Molino, Yaroslav Dudin, and Sai~Sumanth Miryala.
\newblock Ludwig: a type-based declarative deep learning toolbox, September 2019.
\newblock URL \url{http://arxiv.org/abs/1909.07930}.
\newblock arXiv:1909.07930 [cs, stat].

\bibitem[Feurer et~al.(2015)Feurer, Klein, Eggensperger, Springenberg, Blum, and Hutter]{feurer_efficient_2015}
Matthias Feurer, Aaron Klein, Katharina Eggensperger, Jost Springenberg, Manuel Blum, and Frank Hutter.
\newblock Efficient and {Robust} {Automated} {Machine} {Learning}.
\newblock In \emph{Advances in {Neural} {Information} {Processing} {Systems}}, volume~28. Curran Associates, Inc., 2015.
\newblock URL \url{https://proceedings.neurips.cc/paper_files/paper/2015/hash/11d0e6287202fced83f79975ec59a3a6-Abstract.html}.

\bibitem[noa(2023{\natexlab{a}})]{noauthor_welcome_2023}
Welcome to {PyCaret}, September 2023{\natexlab{a}}.
\newblock URL \url{https://github.com/pycaret/pycaret}.
\newblock original-date: 2019-11-23T18:40:48Z.

\bibitem[La~Cava et~al.(2021)La~Cava, Williams, Fu, Vitale, Srivatsan, and Moore]{la_cava_evaluating_2021}
William La~Cava, Heather Williams, Weixuan Fu, Steve Vitale, Durga Srivatsan, and Jason~H Moore.
\newblock Evaluating recommender systems for {AI}-driven biomedical informatics.
\newblock \emph{Bioinformatics}, 37\penalty0 (2):\penalty0 250--256, April 2021.
\newblock ISSN 1367-4803.
\newblock \doi{10.1093/bioinformatics/btaa698}.
\newblock URL \url{https://doi.org/10.1093/bioinformatics/btaa698}.

\bibitem[Zimmer et~al.(2021)Zimmer, Lindauer, and Hutter]{zimmer_auto-pytorch_2021}
Lucas Zimmer, Marius Lindauer, and Frank Hutter.
\newblock Auto-{PyTorch} {Tabular}: {Multi}-{Fidelity} {MetaLearning} for {Efficient} and {Robust} {AutoDL}, April 2021.
\newblock URL \url{http://arxiv.org/abs/2006.13799}.
\newblock arXiv:2006.13799 [cs, stat].

\bibitem[Olson et~al.(2016)Olson, Bartley, Urbanowicz, and Moore]{olson_evaluation_2016}
Randal~S. Olson, Nathan Bartley, Ryan~J. Urbanowicz, and Jason~H. Moore.
\newblock Evaluation of a {Tree}-based {Pipeline} {Optimization} {Tool} for {Automating} {Data} {Science}.
\newblock In \emph{Proceedings of the {Genetic} and {Evolutionary} {Computation} {Conference} 2016}, {GECCO} '16, pages 485--492, New York, NY, USA, July 2016. Association for Computing Machinery.
\newblock ISBN 978-1-4503-4206-3.
\newblock \doi{10.1145/2908812.2908918}.
\newblock URL \url{https://dl.acm.org/doi/10.1145/2908812.2908918}.

\bibitem[Olson and Moore(2016)]{olson_tpot_2016}
Randal~S. Olson and Jason~H. Moore.
\newblock {TPOT}: {A} {Tree}-based {Pipeline} {Optimization} {Tool} for {Automating} {Machine} {Learning}.
\newblock In \emph{Proceedings of the {Workshop} on {Automatic} {Machine} {Learning}}, pages 66--74. PMLR, December 2016.
\newblock URL \url{https://proceedings.mlr.press/v64/olson_tpot_2016.html}.
\newblock ISSN: 1938-7228.

\bibitem[Le et~al.(2020)Le, Fu, and Moore]{le_scaling_2020}
Trang~T Le, Weixuan Fu, and Jason~H Moore.
\newblock Scaling tree-based automated machine learning to biomedical big data with a feature set selector.
\newblock \emph{Bioinformatics}, 36\penalty0 (1):\penalty0 250--256, January 2020.
\newblock ISSN 1367-4803.
\newblock \doi{10.1093/bioinformatics/btz470}.
\newblock URL \url{https://doi.org/10.1093/bioinformatics/btz470}.

\bibitem[Gijsbers and Vanschoren(2021)]{gijsbers_gama_2021}
Pieter Gijsbers and Joaquin Vanschoren.
\newblock {GAMA}: {A} {General} {Automated} {Machine} {Learning} {Assistant}.
\newblock In Yuxiao Dong, Georgiana Ifrim, Dunja Mladenić, Craig Saunders, and Sofie Van~Hoecke, editors, \emph{Machine {Learning} and {Knowledge} {Discovery} in {Databases}. {Applied} {Data} {Science} and {Demo} {Track}}, Lecture {Notes} in {Computer} {Science}, pages 560--564, Cham, 2021. Springer International Publishing.
\newblock ISBN 978-3-030-67670-4.
\newblock \doi{10.1007/978-3-030-67670-4_39}.

\bibitem[de~Sá et~al.(2017)de~Sá, Pinto, Oliveira, and Pappa]{de_sa_recipe_2017}
Alex G.~C. de~Sá, Walter José G.~S. Pinto, Luiz Otavio V.~B. Oliveira, and Gisele~L. Pappa.
\newblock {RECIPE}: {A} {Grammar}-{Based} {Framework} for {Automatically} {Evolving} {Classification} {Pipelines}.
\newblock In James McDermott, Mauro Castelli, Lukas Sekanina, Evert Haasdijk, and Pablo García-Sánchez, editors, \emph{Genetic {Programming}}, Lecture {Notes} in {Computer} {Science}, pages 246--261, Cham, 2017. Springer International Publishing.
\newblock ISBN 978-3-319-55696-3.
\newblock \doi{10.1007/978-3-319-55696-3_16}.

\bibitem[Mohr et~al.(2018)Mohr, Wever, and Hüllermeier]{mohr_ml-plan_2018}
Felix Mohr, Marcel Wever, and Eyke Hüllermeier.
\newblock {ML}-{Plan}: {Automated} machine learning via hierarchical planning.
\newblock \emph{Machine Learning}, 107\penalty0 (8):\penalty0 1495--1515, September 2018.
\newblock ISSN 1573-0565.
\newblock \doi{10.1007/s10994-018-5735-z}.
\newblock URL \url{https://doi.org/10.1007/s10994-018-5735-z}.

\bibitem[Akshay et~al.(2023)Akshay, Katoch, Shekarchizadeh, Abedi, Sharma, Burkhard, Adam, Monastyrskaya, and Gheinani]{akshay_machine_2023}
Akshay Akshay, Mitali Katoch, Navid Shekarchizadeh, Masoud Abedi, Ankush Sharma, Fiona~C. Burkhard, Rosalyn~M. Adam, Katia Monastyrskaya, and Ali~Hashemi Gheinani.
\newblock Machine {Learning} {Made} {Easy} ({MLme}): {A} {Comprehensive} {Toolkit} for {Machine} {Learning}-{Driven} {Data} {Analysis}, July 2023.
\newblock URL \url{https://www.biorxiv.org/content/10.1101/2023.07.04.546825v1}.
\newblock Pages: 2023.07.04.546825 Section: New Results.

\bibitem[noa(2023{\natexlab{b}})]{noauthor_mljar_2023}
{MLJAR} {Automated} {Machine} {Learning} for {Humans}, September 2023{\natexlab{b}}.
\newblock URL \url{https://github.com/mljar/mljar-supervised}.
\newblock original-date: 2018-11-05T12:58:04Z.

\bibitem[LeDell and Poirier(2020)]{H2OAutoML20}
Erin LeDell and Sebastien Poirier.
\newblock {H2O} {A}uto{ML}: Scalable automatic machine learning.
\newblock \emph{7th ICML Workshop on Automated Machine Learning (AutoML)}, July 2020.
\newblock URL \url{https://www.automl.org/wp-content/uploads/2020/07/AutoML_2020_paper_61.pdf}.

\bibitem[Urbanowicz et~al.(2023)Urbanowicz, Zhang, Cui, and Suri]{urbanowicz_streamline_2023}
Ryan Urbanowicz, Robert Zhang, Yuhan Cui, and Pranshu Suri.
\newblock {STREAMLINE}: {A} {Simple}, {Transparent}, {End}-{To}-{End} {Automated} {Machine} {Learning} {Pipeline} {Facilitating} {Data} {Analysis} and {Algorithm} {Comparison}.
\newblock In Leonardo Trujillo, Stephan~M. Winkler, Sara Silva, and Wolfgang Banzhaf, editors, \emph{Genetic {Programming} {Theory} and {Practice} {XIX}}, Genetic and {Evolutionary} {Computation}, pages 201--231. Springer Nature, Singapore, 2023.
\newblock ISBN 978-981-19846-0-0.
\newblock \doi{10.1007/978-981-19-8460-0_9}.
\newblock URL \url{https://doi.org/10.1007/978-981-19-8460-0_9}.

\bibitem[Kotthoff et~al.(2017)Kotthoff, Thornton, Hoos, Hutter, and Leyton-Brown]{kotthoff_auto-weka_2017}
Lars Kotthoff, Chris Thornton, Holger~H. Hoos, Frank Hutter, and Kevin Leyton-Brown.
\newblock Auto-{WEKA} 2.0: {Automatic} model selection and hyperparameter optimization in {WEKA}.
\newblock \emph{Journal of Machine Learning Research}, 18\penalty0 (25):\penalty0 1--5, 2017.
\newblock ISSN 1533-7928.
\newblock URL \url{http://jmlr.org/papers/v18/16-261.html}.

\bibitem[Vakhrushev et~al.(2021)Vakhrushev, Ryzhkov, Savchenko, Simakov, Damdinov, and Tuzhilin]{vakhrushev_lightautoml_2021}
Anton Vakhrushev, Alexander Ryzhkov, Maxim Savchenko, Dmitry Simakov, Rinchin Damdinov, and Alexander Tuzhilin.
\newblock {LightAutoML}: {AutoML} {Solution} for a {Large} {Financial} {Services} {Ecosystem}, September 2021.
\newblock URL \url{https://arxiv.org/abs/2109.01528v2}.

\bibitem[Wang et~al.(2021)Wang, Wu, Weimer, and Zhu]{wang_flaml_2021}
Chi Wang, Qingyun Wu, Markus Weimer, and Eric Zhu.
\newblock {FLAML}: {A} {Fast} and {Lightweight} {AutoML} {Library}, 2021.
\newblock URL \url{https://github.com/microsoft/FLAML}.
\newblock original-date: 2020-08-20T20:46:11Z.

\bibitem[Hutter et~al.(2019)Hutter, Kotthoff, and Vanschoren]{hutter_automated_2019}
Frank Hutter, Lars Kotthoff, and Joaquin Vanschoren, editors.
\newblock \emph{Automated {Machine} {Learning}: {Methods}, {Systems}, {Challenges}}.
\newblock The {Springer} {Series} on {Challenges} in {Machine} {Learning}. Springer International Publishing, Cham, 2019.
\newblock ISBN 978-3-030-05317-8 978-3-030-05318-5.
\newblock \doi{10.1007/978-3-030-05318-5}.
\newblock URL \url{http://link.springer.com/10.1007/978-3-030-05318-5}.

\bibitem[noa(2023{\natexlab{c}})]{noauthor_transmogrifai_2023}
{TransmogrifAI}, September 2023{\natexlab{c}}.
\newblock URL \url{https://github.com/salesforce/TransmogrifAI}.
\newblock original-date: 2017-11-02T16:15:15Z.

\bibitem[Vasile et~al.(2018)Vasile, Pop, Niţă, and Cristea]{vasile_mlbox_2018}
Mihaela-Andreea Vasile, Florin Pop, Mihaela-Cătălina Niţă, and Valentin Cristea.
\newblock {MLBox}: {Machine} learning box for asymptotic scheduling.
\newblock \emph{Information Sciences}, 433-434:\penalty0 401--416, April 2018.
\newblock ISSN 0020-0255.
\newblock \doi{10.1016/j.ins.2017.01.005}.
\newblock URL \url{https://www.sciencedirect.com/science/article/pii/S0020025517300099}.

\bibitem[Nakano(2023)]{nakano_xcessiv_2023}
Reiichiro Nakano.
\newblock Xcessiv, August 2023.
\newblock URL \url{https://github.com/reiinakano/xcessiv}.
\newblock original-date: 2017-03-07T18:18:25Z.

\bibitem[Wang et~al.(2023)Wang, Feng, Tong, Bao, Ritchie, Saykin, Moore, Urbanowicz, and Shen]{wang_exploring_2023}
Xinkai Wang, Yanbo Feng, Boning Tong, Jingxuan Bao, Marylyn~D. Ritchie, Andrew~J. Saykin, Jason~H. Moore, Ryan Urbanowicz, and Li~Shen.
\newblock Exploring {Automated} {Machine} {Learning} for {Cognitive} {Outcome} {Prediction} from {Multimodal} {Brain} {Imaging} using {STREAMLINE}.
\newblock \emph{AMIA Summits on Translational Science Proceedings}, 2023:\penalty0 544--553, June 2023.
\newblock ISSN 2153-4063.
\newblock URL \url{https://www.ncbi.nlm.nih.gov/pmc/articles/PMC10283099/}.

\bibitem[Tong et~al.(2023)Tong, Risacher, Bao, Feng, Wang, Ritchie, Moore, Urbanowicz, Saykin, and Shen]{tong_comparing_2023}
Boning Tong, Shannon~L. Risacher, Jingxuan Bao, Yanbo Feng, Xinkai Wang, Marylyn~D. Ritchie, Jason~H. Moore, Ryan Urbanowicz, Andrew~J. Saykin, and Li~Shen.
\newblock Comparing {Amyloid} {Imaging} {Normalization} {Strategies} for {Alzheimer}’s {Disease} {Classification} using an {Automated} {Machine} {Learning} {Pipeline}.
\newblock \emph{AMIA Summits on Translational Science Proceedings}, 2023:\penalty0 525--533, June 2023.
\newblock ISSN 2153-4063.
\newblock URL \url{https://www.ncbi.nlm.nih.gov/pmc/articles/PMC10283108/}.

\bibitem[Hwang et~al.(2023)Hwang, Urbanowicz, Lynch, Vernon, Bresz, Giraldo, Kennedy, Leabhart, Bleacher, Ripchinski, Mowery, and Oyer]{hwang_toward_2023}
Sy~Hwang, Ryan Urbanowicz, Selah Lynch, Tawnya Vernon, Kellie Bresz, Carolina Giraldo, Erin Kennedy, Max Leabhart, Troy Bleacher, Michael~R. Ripchinski, Danielle~L. Mowery, and Randall~A. Oyer.
\newblock Toward {Predicting} 30-{Day} {Readmission} {Among} {Oncology} {Patients}: {Identifying} {Timely} and {Actionable} {Risk} {Factors}.
\newblock \emph{JCO Clinical Cancer Informatics}, \penalty0 (7):\penalty0 e2200097, September 2023.
\newblock \doi{10.1200/CCI.22.00097}.
\newblock URL \url{https://ascopubs.org/doi/abs/10.1200/CCI.22.00097}.
\newblock Publisher: Wolters Kluwer.

\bibitem[Kohn et~al.(2023)Kohn, Harhay, Weissman, Urbanowicz, Wang, Anesi, Scott, Bayes, Greysen, Halpern, and Kerlin]{kohn_data-driven_2023}
Rachel Kohn, Michael~O. Harhay, Gary~E. Weissman, Ryan Urbanowicz, Wei Wang, George~L. Anesi, Stefania Scott, Brian Bayes, S.~Ryan Greysen, Scott~D. Halpern, and Meeta~Prasad Kerlin.
\newblock A {Data}-{Driven} {Analysis} of {Ward} {Capacity} {Strain} {Metrics} {That} {Predict} {Clinical} {Outcomes} {Among} {Survivors} of {Acute} {Respiratory} {Failure}.
\newblock \emph{Journal of Medical Systems}, 47\penalty0 (1):\penalty0 83, August 2023.
\newblock ISSN 1573-689X.
\newblock \doi{10.1007/s10916-023-01978-5}.
\newblock URL \url{https://doi.org/10.1007/s10916-023-01978-5}.

\bibitem[Urbanowicz et~al.(2020)Urbanowicz, Suri, Cui, Moore, Ruth, Stolzenberg-Solomon, and Lynch]{urbanowicz_rigorous_2020}
Ryan~J. Urbanowicz, Pranshu Suri, Yuhan Cui, Jason~H. Moore, Karen Ruth, Rachael Stolzenberg-Solomon, and Shannon~M. Lynch.
\newblock A {Rigorous} {Machine} {Learning} {Analysis} {Pipeline} for {Biomedical} {Binary} {Classification}: {Application} in {Pancreatic} {Cancer} {Nested} {Case}-control {Studies} with {Implications} for {Bias} {Assessments}, September 2020.
\newblock URL \url{http://arxiv.org/abs/2008.12829}.
\newblock arXiv:2008.12829 [cs, stat].

\bibitem[Emmanuel et~al.(2021)Emmanuel, Maupong, Mpoeleng, Semong, Mphago, and Tabona]{emmanuel_survey_2021}
Tlamelo Emmanuel, Thabiso Maupong, Dimane Mpoeleng, Thabo Semong, Banyatsang Mphago, and Oteng Tabona.
\newblock A survey on missing data in machine learning.
\newblock \emph{Journal of Big Data}, 8\penalty0 (1):\penalty0 140, October 2021.
\newblock ISSN 2196-1115.
\newblock \doi{10.1186/s40537-021-00516-9}.
\newblock URL \url{https://doi.org/10.1186/s40537-021-00516-9}.

\bibitem[Urbanowicz et~al.(2018)Urbanowicz, Olson, Schmitt, Meeker, and Moore]{urbanowicz_benchmarking_2018}
Ryan~J. Urbanowicz, Randal~S. Olson, Peter Schmitt, Melissa Meeker, and Jason~H. Moore.
\newblock Benchmarking relief-based feature selection methods for bioinformatics data mining.
\newblock \emph{Journal of Biomedical Informatics}, 85:\penalty0 168--188, September 2018.
\newblock ISSN 1532-0464.
\newblock \doi{10.1016/j.jbi.2018.07.015}.
\newblock URL \url{https://www.sciencedirect.com/science/article/pii/S1532046418301412}.

\bibitem[Verma et~al.(2018)Verma, Lucas, Zhang, Veturi, Dudek, Li, Li, Urbanowicz, Moore, Kim, and Ritchie]{verma_collective_2018}
Shefali~S. Verma, Anastasia Lucas, Xinyuan Zhang, Yogasudha Veturi, Scott Dudek, Binglan Li, Ruowang Li, Ryan Urbanowicz, Jason~H. Moore, Dokyoon Kim, and Marylyn~D. Ritchie.
\newblock Collective feature selection to identify crucial epistatic variants.
\newblock \emph{BioData Mining}, 11\penalty0 (1):\penalty0 5, April 2018.
\newblock ISSN 1756-0381.
\newblock \doi{10.1186/s13040-018-0168-6}.
\newblock URL \url{https://doi.org/10.1186/s13040-018-0168-6}.

\bibitem[Urbanowicz and Moore(2015)]{urbanowicz_exstracs_2015}
Ryan~J. Urbanowicz and Jason~H. Moore.
\newblock {ExSTraCS} 2.0: description and evaluation of a scalable learning classifier system.
\newblock \emph{Evolutionary Intelligence}, 8\penalty0 (2):\penalty0 89--116, September 2015.
\newblock ISSN 1864-5917.
\newblock \doi{10.1007/s12065-015-0128-8}.
\newblock URL \url{https://doi.org/10.1007/s12065-015-0128-8}.

\bibitem[McKinney et~al.(2006)McKinney, Reif, Ritchie, and Moore]{mckinney_machine_2006}
Brett~A. McKinney, David~M. Reif, Marylyn~D. Ritchie, and Jason~H. Moore.
\newblock Machine {Learning} for {Detecting} {Gene}-{Gene} {Interactions}.
\newblock \emph{Applied Bioinformatics}, 5\penalty0 (2):\penalty0 77--88, June 2006.
\newblock ISSN 1175-5636.
\newblock \doi{10.2165/00822942-200605020-00002}.
\newblock URL \url{https://doi.org/10.2165/00822942-200605020-00002}.

\bibitem[Woodward et~al.(2022)Woodward, Urbanowicz, Naj, and Moore]{woodward_genetic_2022}
Alexa~A. Woodward, Ryan~J. Urbanowicz, Adam~C. Naj, and Jason~H. Moore.
\newblock Genetic heterogeneity: {Challenges}, impacts, and methods through an associative lens.
\newblock \emph{Genetic Epidemiology}, 46\penalty0 (8):\penalty0 555--571, 2022.
\newblock ISSN 1098-2272.
\newblock \doi{10.1002/gepi.22497}.
\newblock URL \url{https://onlinelibrary.wiley.com/doi/abs/10.1002/gepi.22497}.
\newblock \_eprint: https://onlinelibrary.wiley.com/doi/pdf/10.1002/gepi.22497.

\bibitem[Magalang et~al.(2013)Magalang, Chen, Cistulli, Fedson, Gíslason, Hillman, Penzel, Tamisier, Tufik, Phillips, Pack, and {SAGIC Investigators}]{magalang_agreement_2013}
Ulysses~J. Magalang, Ning-Hung Chen, Peter~A. Cistulli, Annette~C. Fedson, Thorarinn Gíslason, David Hillman, Thomas Penzel, Renaud Tamisier, Sergio Tufik, Gary Phillips, Allan~I. Pack, and {SAGIC Investigators}.
\newblock Agreement in the scoring of respiratory events and sleep among international sleep centers.
\newblock \emph{Sleep}, 36\penalty0 (4):\penalty0 591--596, April 2013.
\newblock ISSN 1550-9109.
\newblock \doi{10.5665/sleep.2552}.

\bibitem[Magalang et~al.(2016)Magalang, Arnardottir, Chen, Cistulli, Gíslason, Lim, Penzel, Schwab, Tufik, Pack, and {SAGIC Investigators}]{magalang_agreement_2016}
Ulysses~J. Magalang, Erna~S. Arnardottir, Ning-Hung Chen, Peter~A. Cistulli, Thorarinn Gíslason, Diane Lim, Thomas Penzel, Richard Schwab, Sergio Tufik, Allan~I. Pack, and {SAGIC Investigators}.
\newblock Agreement in the {Scoring} of {Respiratory} {Events} {Among} {International} {Sleep} {Centers} for {Home} {Sleep} {Testing}.
\newblock \emph{Journal of clinical sleep medicine: JCSM: official publication of the American Academy of Sleep Medicine}, 12\penalty0 (1):\penalty0 71--77, January 2016.
\newblock ISSN 1550-9397.
\newblock \doi{10.5664/jcsm.5398}.

\bibitem[Keenan et~al.(2018)Keenan, Kim, Singh, Bittencourt, Chen, Cistulli, Magalang, McArdle, Mindel, Benediktsdottir, Arnardottir, Prochnow, Penzel, Sanner, Schwab, Shin, Sutherland, Tufik, Maislin, Gislason, and Pack]{keenan_recognizable_2018}
Brendan~T. Keenan, Jinyoung Kim, Bhajan Singh, Lia Bittencourt, Ning-Hung Chen, Peter~A. Cistulli, Ulysses~J. Magalang, Nigel McArdle, Jesse~W. Mindel, Bryndis Benediktsdottir, Erna~Sif Arnardottir, Lisa~Kristin Prochnow, Thomas Penzel, Bernd Sanner, Richard~J. Schwab, Chol Shin, Kate Sutherland, Sergio Tufik, Greg Maislin, Thorarinn Gislason, and Allan~I. Pack.
\newblock Recognizable clinical subtypes of obstructive sleep apnea across international sleep centers: a cluster analysis.
\newblock \emph{Sleep}, 41\penalty0 (3):\penalty0 zsx214, March 2018.
\newblock ISSN 1550-9109.
\newblock \doi{10.1093/sleep/zsx214}.

\bibitem[Rizzatti et~al.(2020)Rizzatti, Mazzotti, Mindel, Maislin, Keenan, Bittencourt, Chen, Cistulli, McArdle, Pack, Singh, Sutherland, Benediktsdottir, Fietze, Gislason, Lim, Penzel, Sanner, Han, Li, Schwab, Tufik, Pack, and Magalang]{rizzatti_defining_2020}
Fabiola~G. Rizzatti, Diego~R. Mazzotti, Jesse Mindel, Greg Maislin, Brendan~T. Keenan, Lia Bittencourt, Ning-Hung Chen, Peter~A. Cistulli, Nigel McArdle, Frances~M. Pack, Bhajan Singh, Kate Sutherland, Bryndis Benediktsdottir, Ingo Fietze, Thorarinn Gislason, Diane~C. Lim, Thomas Penzel, Bernd Sanner, Fang Han, Qing~Yun Li, Richard Schwab, Sergio Tufik, Allan~I. Pack, and Ulysses~J. Magalang.
\newblock Defining {Extreme} {Phenotypes} of {OSA} {Across} {International} {Sleep} {Centers}.
\newblock \emph{Chest}, 158\penalty0 (3):\penalty0 1187--1197, September 2020.
\newblock ISSN 1931-3543.
\newblock \doi{10.1016/j.chest.2020.03.055}.

\bibitem[Sutherland et~al.(2023)Sutherland, Kim, Veatch, Keenan, Bittencourt, Chen, Gislason, Han, Jafari, Li, Lim, Maislin, Magalang, Mazzotti, McArdle, Mindel, Pack, Penzel, Singh, Wiemken, Xu, Sun, Tufik, Schwab, and Cistulli]{sutherland_facial_2023}
Kate Sutherland, Soriul Kim, Olivia~J. Veatch, Brendan~T. Keenan, Lia Bittencourt, Ning-Hung Chen, Thorarinn Gislason, Fang Han, Niusha Jafari, Qing~Yun Li, Diane~C. Lim, Greg Maislin, Ulysses Magalang, Diego~R. Mazzotti, Nigel McArdle, Jesse Mindel, Allan~I. Pack, Thomas Penzel, Bhajan Singh, Andrew Wiemken, Liyue Xu, Yun Sun, Sergio Tufik, Richard~J. Schwab, and Peter~A. Cistulli.
\newblock Facial and {Intraoral} {Photographic} {Traits} {Related} to {Sleep} {Apnea} in a {Clinical} {Sample} with {Genetic} {Ancestry} {Analysis}.
\newblock \emph{Annals of the American Thoracic Society}, 20\penalty0 (6):\penalty0 880--890, June 2023.
\newblock ISSN 2325-6621.
\newblock \doi{10.1513/AnnalsATS.202207-577OC}.

\bibitem[Lyons et~al.(2020)Lyons, Bhatt, Pack, and Magalang]{lyons_global_2020}
M.~Melanie Lyons, Nitin~Y. Bhatt, Allan~I. Pack, and Ulysses~J. Magalang.
\newblock Global burden of sleep-disordered breathing and its implications.
\newblock \emph{Respirology}, 25\penalty0 (7):\penalty0 690--702, 2020.
\newblock ISSN 1440-1843.
\newblock \doi{10.1111/resp.13838}.
\newblock URL \url{https://onlinelibrary.wiley.com/doi/abs/10.1111/resp.13838}.
\newblock \_eprint: https://onlinelibrary.wiley.com/doi/pdf/10.1111/resp.13838.

\bibitem[Gottlieb and Punjabi(2020)]{gottlieb_diagnosis_2020}
Daniel~J. Gottlieb and Naresh~M. Punjabi.
\newblock Diagnosis and {Management} of {Obstructive} {Sleep} {Apnea}: {A} {Review}.
\newblock \emph{JAMA}, 323\penalty0 (14):\penalty0 1389--1400, April 2020.
\newblock ISSN 0098-7484.
\newblock \doi{10.1001/jama.2020.3514}.
\newblock URL \url{https://doi.org/10.1001/jama.2020.3514}.

\bibitem[Gami et~al.(2005)Gami, Howard, Olson, and Somers]{gami_day-night_2005}
Apoor~S. Gami, Daniel~E. Howard, Eric~J. Olson, and Virend~K. Somers.
\newblock Day-night pattern of sudden death in obstructive sleep apnea.
\newblock \emph{The New England Journal of Medicine}, 352\penalty0 (12):\penalty0 1206--1214, March 2005.
\newblock ISSN 1533-4406.
\newblock \doi{10.1056/NEJMoa041832}.

\bibitem[Yaggi et~al.(2005)Yaggi, Concato, Kernan, Lichtman, Brass, and Mohsenin]{yaggi_obstructive_2005}
H.~Klar Yaggi, John Concato, Walter~N. Kernan, Judith~H. Lichtman, Lawrence~M. Brass, and Vahid Mohsenin.
\newblock Obstructive sleep apnea as a risk factor for stroke and death.
\newblock \emph{The New England Journal of Medicine}, 353\penalty0 (19):\penalty0 2034--2041, November 2005.
\newblock ISSN 1533-4406.
\newblock \doi{10.1056/NEJMoa043104}.

\bibitem[AlGhanim et~al.(2008)AlGhanim, Comondore, Fleetham, Marra, and Ayas]{alghanim_economic_2008}
Nayef AlGhanim, Vikram~R. Comondore, John Fleetham, Carlo~A. Marra, and Najib~T. Ayas.
\newblock The economic impact of obstructive sleep apnea.
\newblock \emph{Lung}, 186\penalty0 (1):\penalty0 7--12, 2008.
\newblock ISSN 0341-2040.
\newblock \doi{10.1007/s00408-007-9055-5}.

\bibitem[Mulgrew et~al.(2007)Mulgrew, Ryan, Fleetham, Cheema, Fox, Koehoorn, Fitzgerald, Marra, and Ayas]{mulgrew_impact_2007}
A.~T. Mulgrew, C.~F. Ryan, J.~A. Fleetham, R.~Cheema, N.~Fox, M.~Koehoorn, J.~M. Fitzgerald, C.~Marra, and N.~T. Ayas.
\newblock The impact of obstructive sleep apnea and daytime sleepiness on work limitation.
\newblock \emph{Sleep Medicine}, 9\penalty0 (1):\penalty0 42--53, December 2007.
\newblock ISSN 1389-9457.
\newblock \doi{10.1016/j.sleep.2007.01.009}.

\bibitem[George(2001)]{george_reduction_2001}
C.~F. George.
\newblock Reduction in motor vehicle collisions following treatment of sleep apnoea with nasal {CPAP}.
\newblock \emph{Thorax}, 56\penalty0 (7):\penalty0 508--512, July 2001.
\newblock ISSN 0040-6376.
\newblock \doi{10.1136/thorax.56.7.508}.

\bibitem[Arnardottir et~al.(2016)Arnardottir, Bjornsdottir, Olafsdottir, Benediktsdottir, and Gislason]{arnardottir_obstructive_2016}
Erna~S. Arnardottir, Erla Bjornsdottir, Kristin~A. Olafsdottir, Bryndis Benediktsdottir, and Thorarinn Gislason.
\newblock Obstructive sleep apnoea in the general population: highly prevalent but minimal symptoms.
\newblock \emph{The European Respiratory Journal}, 47\penalty0 (1):\penalty0 194--202, January 2016.
\newblock ISSN 1399-3003.
\newblock \doi{10.1183/13993003.01148-2015}.

\bibitem[Arsic et~al.(2022)Arsic, Zebic, Sajid, Bhave, Passalacqua, White-Perkins, Lamerato, Rees, and Budzynska]{arsic_assessing_2022}
Benjamin Arsic, Kristina Zebic, Aamna Sajid, Neha Bhave, Karla~D. Passalacqua, Denise White-Perkins, Lois Lamerato, Della Rees, and Katarzyna Budzynska.
\newblock Assessing the {Adequacy} of {Obstructive} {Sleep} {Apnea} {Diagnosis} for {High}-{Risk} {Patients} in {Primary} {Care}.
\newblock \emph{Journal of the American Board of Family Medicine: JABFM}, 35\penalty0 (2):\penalty0 320--328, 2022.
\newblock ISSN 1558-7118.
\newblock \doi{10.3122/jabfm.2022.02.210296}.

\bibitem[Netzer et~al.(2003)Netzer, Hoegel, Loube, Netzer, Hay, Alvarez-Sala, Strohl, and {Sleep in Primary Care International Study Group}]{netzer_prevalence_2003}
Nikolaus~C. Netzer, Josef~J. Hoegel, Daniel Loube, Cordula~M. Netzer, Birgit Hay, Rudolfo Alvarez-Sala, Kingman~P. Strohl, and {Sleep in Primary Care International Study Group}.
\newblock Prevalence of symptoms and risk of sleep apnea in primary care.
\newblock \emph{Chest}, 124\penalty0 (4):\penalty0 1406--1414, October 2003.
\newblock ISSN 0012-3692.
\newblock \doi{10.1378/chest.124.4.1406}.

\bibitem[Kapur et~al.(1999)Kapur, Blough, Sandblom, Hert, de~Maine, Sullivan, and Psaty]{kapur_medical_1999}
V.~Kapur, D.~K. Blough, R.~E. Sandblom, R.~Hert, J.~B. de~Maine, S.~D. Sullivan, and B.~M. Psaty.
\newblock The medical cost of undiagnosed sleep apnea.
\newblock \emph{Sleep}, 22\penalty0 (6):\penalty0 749--755, September 1999.
\newblock ISSN 0161-8105.
\newblock \doi{10.1093/sleep/22.6.749}.

\bibitem[Maislin et~al.(1995)Maislin, Pack, Kribbs, Smith, Schwartz, Kline, Schwab, and Dinges]{maislin_survey_1995}
G.~Maislin, A.~I. Pack, N.~B. Kribbs, P.~L. Smith, A.~R. Schwartz, L.~R. Kline, R.~J. Schwab, and D.~F. Dinges.
\newblock A survey screen for prediction of apnea.
\newblock \emph{Sleep}, 18\penalty0 (3):\penalty0 158--166, April 1995.
\newblock ISSN 0161-8105.
\newblock \doi{10.1093/sleep/18.3.158}.

\bibitem[Schwab et~al.(2017)Schwab, Leinwand, Bearn, Maislin, Rao, Nagaraja, Wang, and Keenan]{schwab_digital_2017}
Richard~J. Schwab, Sarah~E. Leinwand, Cary~B. Bearn, Greg Maislin, Ramya~Bhat Rao, Adithya Nagaraja, Stephen Wang, and Brendan~T. Keenan.
\newblock Digital {Morphometrics}: {A} {New} {Upper} {Airway} {Phenotyping} {Paradigm} in {OSA}.
\newblock \emph{Chest}, 152\penalty0 (2):\penalty0 330--342, August 2017.
\newblock ISSN 1931-3543.
\newblock \doi{10.1016/j.chest.2017.05.005}.

\bibitem[Sutherland et~al.(2014)Sutherland, Schwab, Maislin, Lee, Benedikstdsottir, Pack, Gislason, Juliusson, and Cistulli]{sutherland_facial_2014}
Kate Sutherland, Richard~J. Schwab, Greg Maislin, Richard W.~W. Lee, Bryndis Benedikstdsottir, Allan~I. Pack, Thorarinn Gislason, Sigurdur Juliusson, and Peter~A. Cistulli.
\newblock Facial phenotyping by quantitative photography reflects craniofacial morphology measured on magnetic resonance imaging in {Icelandic} sleep apnea patients.
\newblock \emph{Sleep}, 37\penalty0 (5):\penalty0 959--968, May 2014.
\newblock ISSN 1550-9109.
\newblock \doi{10.5665/sleep.3670}.

\bibitem[Sutherland et~al.(2016)Sutherland, Lee, Petocz, Chan, Ng, Hui, and Cistulli]{sutherland_craniofacial_2016}
Kate Sutherland, Richard W.~W. Lee, Peter Petocz, Tat~On Chan, Susanna Ng, David~S. Hui, and Peter~A. Cistulli.
\newblock Craniofacial phenotyping for prediction of obstructive sleep apnoea in a {Chinese} population.
\newblock \emph{Respirology (Carlton, Vic.)}, 21\penalty0 (6):\penalty0 1118--1125, August 2016.
\newblock ISSN 1440-1843.
\newblock \doi{10.1111/resp.12792}.

\bibitem[Lee et~al.(2009{\natexlab{a}})Lee, Chan, Grunstein, and Cistulli]{lee_craniofacial_2009}
Richard W.~W. Lee, Andrew S.~L. Chan, Ronald~R. Grunstein, and Peter~A. Cistulli.
\newblock Craniofacial phenotyping in obstructive sleep apnea--a novel quantitative photographic approach.
\newblock \emph{Sleep}, 32\penalty0 (1):\penalty0 37--45, January 2009{\natexlab{a}}.
\newblock ISSN 0161-8105.

\bibitem[Lee et~al.(2009{\natexlab{b}})Lee, Petocz, Prvan, Chan, Grunstein, and Cistulli]{lee_prediction_2009}
Richard W.~W. Lee, Peter Petocz, Tania Prvan, Andrew S.~L. Chan, Ronald~R. Grunstein, and Peter~A. Cistulli.
\newblock Prediction of obstructive sleep apnea with craniofacial photographic analysis.
\newblock \emph{Sleep}, 32\penalty0 (1):\penalty0 46--52, January 2009{\natexlab{b}}.
\newblock ISSN 0161-8105.

\bibitem[Ross(2014)]{ross_mutual_2014}
Brian~C. Ross.
\newblock Mutual {Information} between {Discrete} and {Continuous} {Data} {Sets}.
\newblock \emph{PLOS ONE}, 9\penalty0 (2):\penalty0 e87357, February 2014.
\newblock ISSN 1932-6203.
\newblock \doi{10.1371/journal.pone.0087357}.
\newblock URL \url{https://journals.plos.org/plosone/article?id=10.1371/journal.pone.0087357}.
\newblock Publisher: Public Library of Science.

\bibitem[Urbanowicz and Browne(2017)]{urbanowicz2017introduction}
Ryan~J Urbanowicz and Will~N Browne.
\newblock \emph{Introduction to learning classifier systems}.
\newblock Springer, 2017.

\bibitem[Butz(2006)]{butz_xcs_2006}
Martin~V. Butz.
\newblock The {XCS} {Classifier} {System}.
\newblock In Martin~V. Butz, editor, \emph{Rule-{Based} {Evolutionary} {Online} {Learning} {Systems}: {A} {Principled} {Approach} to {LCS} {Analysis} and {Design}}, Studies in {Fuzziness} and {Soft} {Computing}, pages 51--64. Springer, Berlin, Heidelberg, 2006.
\newblock ISBN 978-3-540-31231-4.
\newblock \doi{10.1007/3-540-31231-5_4}.
\newblock URL \url{https://doi.org/10.1007/3-540-31231-5_4}.

\bibitem[Akiba et~al.(2019)Akiba, Sano, Yanase, Ohta, and Koyama]{akiba_optuna_2019}
Takuya Akiba, Shotaro Sano, Toshihiko Yanase, Takeru Ohta, and Masanori Koyama.
\newblock Optuna: {A} {Next}-generation {Hyperparameter} {Optimization} {Framework}.
\newblock In \emph{Proceedings of the 25th {ACM} {SIGKDD} {International} {Conference} on {Knowledge} {Discovery} \& {Data} {Mining}}, {KDD} '19, pages 2623--2631, New York, NY, USA, July 2019. Association for Computing Machinery.
\newblock ISBN 978-1-4503-6201-6.
\newblock \doi{10.1145/3292500.3330701}.
\newblock URL \url{https://doi.org/10.1145/3292500.3330701}.

\bibitem[Breiman(2001)]{breiman_random_2001}
Leo Breiman.
\newblock Random {Forests}.
\newblock \emph{Machine Learning}, 45\penalty0 (1):\penalty0 5--32, October 2001.
\newblock ISSN 1573-0565.
\newblock \doi{10.1023/A:1010933404324}.
\newblock URL \url{https://doi.org/10.1023/A:1010933404324}.

\bibitem[noa()]{noauthor_sklearnlinear_modelelasticnet_nodate}
sklearn.linear\_model.{ElasticNet}.
\newblock URL \url{https://scikit-learn/stable/modules/generated/sklearn.linear_model.ElasticNet.html}.

\bibitem[Miriam~Santos(2015)]{miriam_santos_hcc_2015}
Pedro~Abreu Miriam~Santos.
\newblock {HCC} {Survival}, 2015.
\newblock URL \url{https://archive.ics.uci.edu/dataset/423}.

\bibitem[Urbanowicz et~al.(2012)Urbanowicz, Kiralis, Sinnott-Armstrong, Heberling, Fisher, and Moore]{urbanowicz_gametes_2012}
Ryan~J. Urbanowicz, Jeff Kiralis, Nicholas~A. Sinnott-Armstrong, Tamra Heberling, Jonathan~M. Fisher, and Jason~H. Moore.
\newblock {GAMETES}: a fast, direct algorithm for generating pure, strict, epistatic models with random architectures.
\newblock \emph{BioData Mining}, 5\penalty0 (1):\penalty0 16, October 2012.
\newblock ISSN 1756-0381.
\newblock \doi{10.1186/1756-0381-5-16}.
\newblock URL \url{https://doi.org/10.1186/1756-0381-5-16}.

\bibitem[Mallampati(1983)]{mallampati_clinical_1983}
S.~Rao Mallampati.
\newblock Clinical sign to predict difficult tracheal intubation (hypothesis).
\newblock \emph{Canadian Anaesthetists’ Society Journal}, 30\penalty0 (3):\penalty0 316--317, May 1983.
\newblock ISSN 1496-8975.
\newblock \doi{10.1007/BF03013818}.
\newblock URL \url{https://doi.org/10.1007/BF03013818}.

\bibitem[Laharnar et~al.(2021)Laharnar, Herberger, Prochnow, Chen, Cistulli, Pack, Schwab, Keenan, Mazzotti, Fietze, and Penzel]{laharnar_simple_2021}
Naima Laharnar, Sebastian Herberger, Lisa-Kristin Prochnow, Ning-Hung Chen, Peter~A. Cistulli, Allan~I. Pack, Richard Schwab, Brendan~T. Keenan, Diego~R. Mazzotti, Ingo Fietze, and Thomas Penzel.
\newblock Simple and {Unbiased} {OSA} {Prescreening}: {Introduction} of a {New} {Morphologic} {OSA} {Prediction} {Score}.
\newblock \emph{Nature and Science of Sleep}, 13:\penalty0 2039--2049, 2021.
\newblock ISSN 1179-1608.
\newblock \doi{10.2147/NSS.S333471}.

\bibitem[Qin et~al.(2021)Qin, Keenan, Mazzotti, Vaquerizo-Villar, Kraemer, Wessel, Tufik, Bittencourt, Cistulli, de~Chazal, Sutherland, Singh, Pack, Chen, Fietze, Gislason, Holfinger, Magalang, and Penzel]{qin_heart_2021}
Hua Qin, Brendan~T. Keenan, Diego~R. Mazzotti, Fernando Vaquerizo-Villar, Jan~F. Kraemer, Niels Wessel, Sergio Tufik, Lia Bittencourt, Peter~A. Cistulli, Philip de~Chazal, Kate Sutherland, Bhajan Singh, Allan~I. Pack, Ning-Hung Chen, Ingo Fietze, Thorarinn Gislason, Steven Holfinger, Ulysses~J. Magalang, and Thomas Penzel.
\newblock Heart rate variability during wakefulness as a marker of obstructive sleep apnea severity.
\newblock \emph{Sleep}, 44\penalty0 (5):\penalty0 zsab018, May 2021.
\newblock ISSN 1550-9109.
\newblock \doi{10.1093/sleep/zsab018}.

\bibitem[Holfinger et~al.(2022)Holfinger, Lyons, Keenan, Mazzotti, Mindel, Maislin, Cistulli, Sutherland, McArdle, Singh, Chen, Gislason, Penzel, Han, Li, Schwab, Pack, and Magalang]{holfinger_diagnostic_2022}
Steven~J. Holfinger, M.~Melanie Lyons, Brendan~T. Keenan, Diego~R. Mazzotti, Jesse Mindel, Greg Maislin, Peter~A. Cistulli, Kate Sutherland, Nigel McArdle, Bhajan Singh, Ning-Hung Chen, Thorarinn Gislason, Thomas Penzel, Fang Han, Qing~Yun Li, Richard Schwab, Allan~I. Pack, and Ulysses~J. Magalang.
\newblock Diagnostic {Performance} of {Machine} {Learning}-{Derived} {OSA} {Prediction} {Tools} in {Large} {Clinical} and {Community}-{Based} {Samples}.
\newblock \emph{Chest}, 161\penalty0 (3):\penalty0 807--817, March 2022.
\newblock ISSN 1931-3543.
\newblock \doi{10.1016/j.chest.2021.10.023}.

\bibitem[Sutherland et~al.(2019)Sutherland, Keenan, Bittencourt, Chen, Gislason, Leinwand, Magalang, Maislin, Mazzotti, McArdle, Mindel, Pack, Penzel, Singh, Tufik, Schwab, Cistulli, and {SAGIC Investigators}]{sutherland_global_2019}
Kate Sutherland, Brendan~T. Keenan, Lia Bittencourt, Ning-Hung Chen, Thorarinn Gislason, Sarah Leinwand, Ulysses~J. Magalang, Greg Maislin, Diego~R. Mazzotti, Nigel McArdle, Jesse Mindel, Allan~I. Pack, Thomas Penzel, Bhajan Singh, Sergio Tufik, Richard~J. Schwab, Peter~A. Cistulli, and {SAGIC Investigators}.
\newblock A {Global} {Comparison} of {Anatomic} {Risk} {Factors} and {Their} {Relationship} to {Obstructive} {Sleep} {Apnea} {Severity} in {Clinical} {Samples}.
\newblock \emph{Journal of clinical sleep medicine: JCSM: official publication of the American Academy of Sleep Medicine}, 15\penalty0 (4):\penalty0 629--639, April 2019.
\newblock ISSN 1550-9397.
\newblock \doi{10.5664/jcsm.7730}.

\bibitem[Liao et~al.(2021)Liao, Taori, Raji, and Schmidt]{liao_are_2021}
Thomas Liao, Rohan Taori, Inioluwa~Deborah Raji, and Ludwig Schmidt.
\newblock Are {We} {Learning} {Yet}? {A} {Meta} {Review} of {Evaluation} {Failures} {Across} {Machine} {Learning}.
\newblock August 2021.
\newblock URL \url{https://openreview.net/forum?id=mPducS1MsEK}.

\bibitem[Zöller and Huber(2021)]{zoller_benchmark_2021}
Marc-André Zöller and Marco~F. Huber.
\newblock Benchmark and {Survey} of {Automated} {Machine} {Learning} {Frameworks}.
\newblock \emph{Journal of Artificial Intelligence Research}, 70:\penalty0 409--472, January 2021.
\newblock ISSN 1076-9757.
\newblock \doi{10.1613/jair.1.11854}.
\newblock URL \url{https://www.jair.org/index.php/jair/article/view/11854}.

\bibitem[Balaji and Allen(2018)]{balaji_benchmarking_2018}
Adithya Balaji and Alexander Allen.
\newblock Benchmarking {Automatic} {Machine} {Learning} {Frameworks}, August 2018.
\newblock URL \url{http://arxiv.org/abs/1808.06492}.
\newblock arXiv:1808.06492 [cs, stat].

\bibitem[Ferreira et~al.(2021)Ferreira, Pilastri, Martins, Pires, and Cortez]{ferreira_comparison_2021}
Luís Ferreira, André Pilastri, Carlos~Manuel Martins, Pedro~Miguel Pires, and Paulo Cortez.
\newblock A {Comparison} of {AutoML} {Tools} for {Machine} {Learning}, {Deep} {Learning} and {XGBoost}.
\newblock In \emph{2021 {International} {Joint} {Conference} on {Neural} {Networks} ({IJCNN})}, pages 1--8, July 2021.
\newblock \doi{10.1109/IJCNN52387.2021.9534091}.
\newblock URL \url{https://ieeexplore.ieee.org/abstract/document/9534091}.
\newblock ISSN: 2161-4407.

\bibitem[Gijsbers et~al.(2022)Gijsbers, Bueno, Coors, LeDell, Poirier, Thomas, Bischl, and Vanschoren]{gijsbers_amlb_2022}
Pieter Gijsbers, Marcos L.~P. Bueno, Stefan Coors, Erin LeDell, Sébastien Poirier, Janek Thomas, Bernd Bischl, and Joaquin Vanschoren.
\newblock {AMLB}: an {AutoML} {Benchmark}, July 2022.
\newblock URL \url{http://arxiv.org/abs/2207.12560}.
\newblock arXiv:2207.12560 [cs, stat].

\bibitem[Young et~al.(1993)Young, Palta, Dempsey, Skatrud, Weber, and Badr]{young_occurrence_1993}
T.~Young, M.~Palta, J.~Dempsey, J.~Skatrud, S.~Weber, and S.~Badr.
\newblock The occurrence of sleep-disordered breathing among middle-aged adults.
\newblock \emph{The New England Journal of Medicine}, 328\penalty0 (17):\penalty0 1230--1235, April 1993.
\newblock ISSN 0028-4793.
\newblock \doi{10.1056/NEJM199304293281704}.

\bibitem[Peppard et~al.(2013)Peppard, Young, Barnet, Palta, Hagen, and Hla]{peppard_increased_2013}
Paul~E. Peppard, Terry Young, Jodi~H. Barnet, Mari Palta, Erika~W. Hagen, and Khin~Mae Hla.
\newblock Increased prevalence of sleep-disordered breathing in adults.
\newblock \emph{American Journal of Epidemiology}, 177\penalty0 (9):\penalty0 1006--1014, May 2013.
\newblock ISSN 1476-6256.
\newblock \doi{10.1093/aje/kws342}.

\bibitem[Caton and Haas(2020)]{caton2020fairness}
Simon Caton and Christian Haas.
\newblock Fairness in machine learning: A survey.
\newblock \emph{ACM Computing Surveys}, 2020.

\end{thebibliography}


\end{document}



\newcommand{\SubItem}[1]{
    {\setlength\itemindent{15pt} \item[*] #1}
}
\begin{center}

\large

\textbf{STREAMLINE: An Automated Machine Learning Pipeline for Biomedicine Applied to Examine the Utility of Photography-Based Phenotypes for OSA Prediction Across International Sleep Centers}

\vspace{2cm}

\huge
    
\textbf{Supporting Information}

\vspace{2cm}

\footnotesize

\textbf{Ryan J. Urbanowicz\textsuperscript{1,4*}, Harsh Bandhey\textsuperscript{1}, Brendan T. Keenan\textsuperscript{2}, Greg Maislin\textsuperscript{2}, Sy Hwang\textsuperscript{3}, Danielle L. Mowery\textsuperscript{4}, Shannon M. Lynch\textsuperscript{5}, Diego R. Mazzotti\textsuperscript{6,7}, Fang Han\textsuperscript{8}, Qing Yun Li\textsuperscript{9}, Thomas Penzel\textsuperscript{10}, Sergio Tufik\textsuperscript{11}, Lia Bittencourt\textsuperscript{11}, Thorarinn Gislason\textsuperscript{12,13}, Philip de Chazal\textsuperscript{14}, Bhajan Singh\textsuperscript{15,16}, Nigel McArdle\textsuperscript{15,16}, Ning-Hung Chen\textsuperscript{17}, Allan Pack\textsuperscript{2}, Richard J. Schwab\textsuperscript{2}, Peter A. Cistulli\textsuperscript{18}, Ulysses J. Magalang\textsuperscript{19}}

\textsuperscript{1}Department of Computational Biomedicine, Cedars Sinai Medical Center, Los Angeles, CA \\
\textsuperscript{2}Division of Sleep Medicine, Department of Medicine, University of Pennsylvania, Philadelphia, PA \\
\textsuperscript{3}Institute for Biomedical Informatics, University of Pennsylvania, Philadelphia, PA \\
\textsuperscript{4}Department of Biostatistics Epidemiology and Informatics, University of Pennsylvania, Philadelphia, PA \\
\textsuperscript{5}Cancer Prevention and Control Program, Fox Chase Cancer Center, Philadelphia, PA \\
\textsuperscript{6}Division of Medical Informatics, Department of Internal Medicine, University of Kansas, Lawrence, KS \\
\textsuperscript{7}Division of Pulmonary, Critical Care, and Sleep Medicine, Department of Internal Medicine, University of Kansas, Lawrence, KS \\
\textsuperscript{8}Division of Sleep Medicine, Peking University People’s Hospital, Beijing, China \\
\textsuperscript{9}Department of Respiratory and Critical Care Medicine, Ruijin Hospital, Shanghai Jiao Tong University School of Medicine, Shanghai, China \\
\textsuperscript{10}Interdisciplinary Center of Sleep Medicine, Charite University Hospital, Berlin, Germany \\
\textsuperscript{11}Department of Psychobiology, Federal University of Sao Paulo, Sao Paulo, Brazil \\
\textsuperscript{12}Department of Sleep Medicine, Landspitali University Hospital, Reykjavik, Iceland \\
\textsuperscript{13}University of Iceland, Faculty of Medicine, Reykjavik, Iceland \\
\textsuperscript{14}Charles Perkins Center, Faculty of Engineering, University of Sydney, Sydney, Australia \\
\textsuperscript{15}Department of Pulmonary Physiology and Sleep Medicine, Sir Charles Gairdner Hospital, Nedlands, Australia \\
\textsuperscript{16}University of Western Australia, Crawley, Australia \\
\textsuperscript{17}Department of Pulmonary and Critical Care Medicine, Sleep Center, Linkou Chang Gung Memorial Hospital, Taoyuan, Taiwan \\
\textsuperscript{18}Charles Perkins Centre, Faculty of Medicine and Health, University of Sydney and Royal North Shore Hospital, Sydney, Australia \\
\textsuperscript{19}Division of Pulmonary, Critical Care, and Sleep Medicine, Department of Medicine, Ohio State University Wexner Medical Center, Columbus, OH \\

*Corresponding author: R.J.U. (ryan.urbanowicz@cshs.org) \\ 700 N. San Vicente Blvd., Pacific Design Center Suite G541E, Los Angeles, CA 90069

\normalsize

\end{center}

\pagebreak

\titlecontents{section}[0em]
{\vskip 0.5ex}%
{}
{}
{}%
\titlecontents{subsection}[2em]
{\vskip 0.5ex}%
{}
{}
{}%
\titlecontents{subsubsection}[4em]
{\vskip 0.5ex}%
{}
{}
{}%

\tableofcontents

\newpage

\addcontentsline{toc}{section}{S.1: AutoML Survey/Comparison}
\section*{S.1: AutoML Survey/Comparison}

We surveyed a total of 19 automated machine learning (AutoML) tools and libraries (including STREAMLINE). “Tools” represent a more self-contained (i.e. ready-to-run) AutoML approach with little to no pipeline implementation required, while “libraries” generally are designed to simplify the coding required to design and run a machine learning analysis pipeline (but can include automation of some aspects of the pipeline (e.g. algorithm selection and hyperparameter optimization being most common).

We limited our survey to include popular and relevant open-source (i.e non-enterprise) AutoML since they are free to access/use and their documentation and capabilities are largely transparent. We also limited our survey to AutoML designed (at minimum) to analyze structured data (i.e. tabular data) for binary classification tasks (but most could also do multi-class and regression tasks).

This survey focused on documentation provided within the respective GitHub repositories of each AutoML approach, including README’s, User’s Guides, and other documentation links within each repository. In each case we focused on the current release (as of September 2023) in the ‘main’ branch of each repository. \textbf{Table S1} summarized basic information on each of the 24 AutoML tools surveyed.

\textbf{Table S2} provides a high-level comparison of the scope and capabilities of each AutoML approach including (1) the data types they can work with, (2) the AutoML objective(s), e.g. classification, regression, etc., (3) aspects impacting ease-of-use, convenience, and parallizability for those with and without programming experience, and (4) what aspects of the ML analysis pipeline have some degree of automation. Some of these characterizations rely on our best subjective interpretation of the respective AutoML documentation, as to whether the AutoML approach automates (to at least some degree) the specified element (i.e. row in the table). This table does not capture many nuances of the different AutoML approaches, including; (1) the quality and efficiency of their implementation (2) the quality and degree of detail provided in the documentation, (3) other uncommon or more specific capabilities and (4) the degree to which each capability is satisfied, e.g. how many options are available?, how much user input is required for setup?, what assumptions are being made? To date, we have only directly used STREAMLINE, TPOT, and ALIRO, thus characterizations of other approaches exclusively rely on their documentation, therefore we apologize to developers if we have inadvertently mischaracterized any of these capabilities, and would be happy to receive feedback. Note that the characterization of STREAMLINE’s capabilities reflects those most recent release used in this current publication (0.3.4), rather than the original published version (0.2.5).

\addcontentsline{toc}{subsection}{Table S1: Summary of AutoML Tools and Libraries Surveyed. }
\begin{table}[H]
\centering
\scalebox{0.73}{
\begin{tabular}{| p{2.3cm} | p{1.3cm} | p{1.3cm} | p{1.3cm} | p{1.5cm} | l | l | l |}
\hline
 \rowcolor{light-gray}
\textbf{AutoML Tool/Library} & \textbf{Release Version} & \textbf{Initial Release} & \textbf{Latest Release} & \textbf{ML Package Reliance} & \textbf{Code Language} & \textbf{Code Link} & \textbf{Paper Citation} \\
\hline
STREAMLINE & 0.3.4 & 2022 & 2023 & sklearn & python & \href{https://github.com/UrbsLab/STREAMLINE}{https://github.com/UrbsLab/STREAMLINE} & \cite{urbanowicz_streamline_2023} \\
\hline
Auto-Keras & 1.1.0 & 2023 & 2023 & keras & python & \href{https://github.com/keras-team/autokeras}{https://github.com/keras-team/autokeras} &  \cite{jin_autokeras_2023} \\
\hline
H20-3 & NA & 2023 & 2023 &  & java & \href{https://github.com/h2oai/h2o-3}{https://github.com/h2oai/h2o-3} & NA \\
\hline
MLme & NA & 2023 & 2023 & sklearn & python & \href{https://github.com/FunctionalUrology/Mlme}{https://github.com/FunctionalUrology/Mlme} & \cite{akshay_machine_2023} \\
\hline
LAMA & 0.3.7.3 & 2021 & 2023 & sklearn & python & \href{https://github.com/sb-ai-lab/LightAutoML}{https://github.com/sb-ai-lab/LightAutoML} & \cite{vakhrushev_lightautoml_2021} \\
\hline
FEDOT & 0.7.2 & 2021 & 2023 & sklearn & python & \href{https://github.com/aimclub/FEDOT}{https://github.com/aimclub/FEDOT} & \cite{nikitin_automated_2022} \\
\hline
FLAML & 2.1.0 & 2020 & 2023 & Apache-Spark & python & \href{https://github.com/microsoft/FLAML}{https://github.com/microsoft/FLAML} & NA \\
\hline
ALIRO & 0.19 & 2020 & 2023 & sklearn & javascript, python & \href{https://github.com/EpistasisLab/Aliro}{https://github.com/EpistasisLab/Aliro} & \cite{la_cava_evaluating_2021} \\
\hline
PYCARET & 3.1.0 & 2020 & 2023 & sklearn & python & \href{https://github.com/pycaret/pycaret}{https://github.com/pycaret/pycaret} & NA \\
\hline
Auto-Gluon & 0.8.2 & 2020 & 2023 & sklearn & python & \href{https://github.com/autogluon/autogluon}{https://github.com/autogluon/autogluon} & \cite{erickson_autogluon-tabular_2020} \\
\hline
MLIJAR-supervised & 1.0.2 & 2019 & 2023 & sklearn & python & \href{https://github.com/mljar/mljar-supervised}{https://github.com/mljar/mljar-supervised} & NA \\
\hline
Ludwig & 0.8.3 & 2019 & 2023 &  & python & \href{https://github.com/ludwig-ai/ludwig/}{https://github.com/ludwig-ai/ludwig/} & \cite{molino_ludwig_2019} \\
\hline
TPOT & 0.12.1 & 2016 & 2023 & sklearn & python & \href{https://github.com/EpistasisLab/tpot}{https://github.com/EpistasisLab/tpot} & \cite{olson_evaluation_2016,olson_tpot_2016,le_scaling_2020} \\
\hline
Auto-Sklearn & 0.15.0 & 2015 & 2023 & sklearn & python & \href{https://github.com/automl/auto-sklearn}{https://github.com/automl/auto-sklearn} & \cite{feurer_efficient_2015} \\
\hline
Auto-PyTorch & 0.2.1 & 2021 & 2022 & Py-Torch & python & \href{https://github.com/automl/Auto-PyTorch}{https://github.com/automl/Auto-PyTorch} & \cite{zimmer_auto-pytorch_2021} \\
\hline
GAMA & 23.0.0 & 2020 & 2022 &  & python & \href{https://github.com/openml-labs/gama}{https://github.com/openml-labs/gama} & \cite{gijsbers_gama_2021} \\
\hline
Hyperopt-sklearn & NA & 2014 & 2022 & sklearn & python & \href{https://github.com/hyperopt/hyperopt-sklearn}{https://github.com/hyperopt/hyperopt-sklearn} & \cite{hutter_automated_2019} \\
\hline
Auto-WEKA & 2.6.4 & 2013 & 2022 & WEKA & java & \href{https://github.com/automl/autoweka}{https://github.com/automl/autoweka} & \cite{kotthoff_auto-weka_2017} \\
\hline
RECIPE & NA & 2017 & 2021 & sklearn & python, C, JavaScript & \href{https://github.com/laic-ufmg/Recipe}{https://github.com/laic-ufmg/Recipe} & \cite{de_sa_recipe_2017} \\
\hline
ML-Plan & 0.2.3 & 2018 & 2020 & sklearn, WEKA & java & \href{https://github.com/starlibs/AILibs}{https://github.com/starlibs/AILibs} & \cite{mohr_ml-plan_2018} \\
\hline
TransmogrifAI & 0.7.0 & 2018 & 2020 & Apache-Spark & Scala & \href{https://github.com/salesforce/TransmogrifAI}{https://github.com/salesforce/TransmogrifAI} & NA \\
\hline
MLBox & 0.8.1 & 2017 & 2019 & sklearn & python & \href{https://github.com/AxeldeRomblay/MLBox}{https://github.com/AxeldeRomblay/MLBox} & \cite{vasile_mlbox_2018} \\
\hline
Xcessiv & 0.5.1 & 2017 & 2017 & sklearn & javascript, python & \href{https://github.com/reiinakano/xcessiv}{https://github.com/reiinakano/xcessiv} & NA \\
\hline
Auto\_ML & 2.7.0 & 2016 & 2017 & sklearn & python & \href{https://github.com/ClimbsRocks/auto_ml}{https://github.com/ClimbsRocks/auto\_ml} & NA \\
\hline
\end{tabular}}
\caption*{\textbf{Table S1: Summary of AutoML Tools and Libraries Surveyed.} Key logistical information for popular and relevant AutoML tools and/or libraries. Not all methods included a publication or pre-print by the developers (but were still included). Methods are sorted in descending year based on latest release and initial release (with STREAMLINE at the top for convenience). Empty cells indicate that we were unsure or unable to confirm the respective value.}
\label{tab:1}
\end{table}
\pagebreak

Below we explain our definitions of each capability category (note that the documentation for different AutoML’s sometimes use different terms and term definitions):

\begin{itemize}  \itemsep0em 
\item \textbf{‘Data Types’}: Indicates whether respective data types can be utilized.  

\item\textbf{‘Target’}: Indicates what ML tasks can be automated. Note that anomaly detection (as an objective) is different than outlier detection (as a data cleaning element). Also note that Auto-sklearn appeared to be unique in its capability to handle multi-label tasks and PYCARET appeared to be unique in its capability to handle anomaly detection.

\item\textbf{‘Ease of Use’}: 

\begin{itemize}  \itemsep0em 
\item\textbf{‘Codeless implementation}: Identifies whether the user is required to utilize any level of coding to implement the pipeline itself, i.e. to set up the analysis pipeline, specify algorithms or hyperparameters, utilize (or not) different elements of an ML pipeline, and in what order.

\item\textbf{‘Codeless running}: Identifies whether has the capability to run the AutoML from a GUI, Web-interface, or a preconfigured notebook, at most having to adjust pipeline run parameters and file paths, and not requiring command line usage. I.e. is the AutoML able to be used by someone with no coding experience.

\item\textbf{‘GUI’}: Is there a downloadable or web-based graphical user interface? Note this does not necessarily mean that the AutoML can be used without coding (which is the case for Xcessiv).

\item\textbf{‘Generates Figures/Viz’}: Does the tool provide automation for the generation of results figures, model vizualizations, and or other comparisons? This can either be run automatically as part of the pipeline, or automated secondarily when asked for by the user.

\item\textbf{‘Outputs/organizes results’}: The approach automates or facilitates generating /downloading/ saving/ accessible-archiving of individual models, performance results, and visualizations. The focus here is allowing the user to easily examine and take a deeper dive into their results.

\item\textbf{‘Summary report’}: The approach automates or facilitates the generation and presentation of some type of global analysis summary, e.g. documenting run parameters, key results, key visualizations, etc.

\item\textbf{‘Recommender system’}: The approach includes some higher-level AI or decision making to suggest or decide what analysis configurations to try.

\item\textbf{‘Parallelizable’}: Is there a clear and documented approach in place to parallelize the analysis, either on a compute cluster, multi-core CPU, or GPU.
\end{itemize}

\item\textbf{‘Automated Pipeline Elements’}:

\begin{itemize}  \itemsep0em 
\item\textbf{‘Pipeline config. discovery’}: The approach provides some automation for determining/discovering what pipeline elements to include and in what order – beyond some pre-specified pipeline of elements.

\item\textbf{‘Feature Extraction}: Automates the extraction of structured features from unstructured data (e.g. text, images, time-series, etc) so that the structured features can be analyzed by algorithms designed to run on structured data only. 

\item\textbf{‘Exploratory Data Analysis’}: Automates multiple aspects of an exploratory data analysis (without requiring manual implementation) including but not limited to the data dimensions, data missingness, feature types, outcome balance or distribution, univariate analysis, feature correlations, clustering, etc.

\item\textbf{‘Categorical Encoding’}: Based on automatic detection or user specification of feature types (e.g. categorical vs. quanatiative), any categorical features are automatically appropriately encoded for the user (e.g. one-hot-encoding), so that they are treated appropriately in ML modeling.

\item\textbf{‘Numerical Encoding’}: Can the pipeline automatically work with data features that include text-values (e.g. Hair color – ‘brown’, ‘black’, ‘blond’, ‘red’), and ideally encode and treat ordinal vs. categorical features appropriately. Users may still need to specify a given feature’s type as this often requires domain knowledge.

\item\textbf{‘Basic Data Cleaning’}: Is there automation of some basic data cleaning elements (excluding missing value imputation), e.g. removing rows with missing outcome, removing redundant instances or features (i.e. perfect/high correlation), outlier detection/removal, etc.

\item\textbf{‘Feature Engineering’}: This term is often confusingly used synonymously with feature extraction, feature learning, and even feature transformation. In modern days its most accurately defined as utilizing domain knowledge to manually ‘engineer’ new features from the dataset (without using knowledge of the target outcome), e.g. taking drug start/stop dates and engineering a drug duration variable. Here we consider feature engineering in a slightly more flexible manner – that does not necessarily require specific problem domain knowledge (but is still distinct from feature extraction, feature, learning, and feature transformation). We define it here as the generation of new features from existing structured features (which may or may not then be removed from the dataset). Here we consider this to include elements of data encoding (including one-hot-encoding, where multiple new features replace an original feature), adding new features generated by dimensionality reduction (e.g. PCA), multivariate interaction encoding, or missing value encoding to explore feature missingness as a predictive variable. The key here is that new features are added to the dataset, and these features were generated without knowledge of the target outcome.

\item\textbf{‘Missing Val. Imputation’}: Automatically imputes or is capable of ignoring missing values in the dataset so that modeling can be applied. Could be simple mean/median/mode imputation or more sophisticated multi-variate imputation.

\item\textbf{‘Feature Transformation’}: One or more features in the dataset can be automatically transformed, e.g. normalization, scaling, or any other transformation that does not directly impact the dimensionality of the dataset (i.e. the number of features in the dataset)

\item\textbf{‘Feature Learning’}: Prior to ML modeling, the approach automatically learns new features that are some combination of existing features, where knowledge of the outcome is available. I.e. generating new informative features that can help simpler algorithms identify more complex relationships in the data, e.g. Multi-Factor Dimensionality reduction, or using earlier baseline models as inputs for downstream modeling (similar but not necessarily the same as ‘stacking’, a type of ensembling)

\item\textbf{‘Feature Selection’}: The AutoML automates ‘feature selection’ (the removal of non or less informative features) prior to modeling (i.e. we do not consider any feature selection internally conducted by a given ML modeling algorithm to be relevant here)

\item\textbf{‘Algorithm Selection’}: The AutoML automates the utilization and comparison of multiple ML algorithms for modeling. Different AutoML approaches were found to have access to a widely varying number of potential algorithms to apply (anywhere from 2, to all algorithms available in a given ML library, e.g. scikit-learn). Additionally while some AutoML’s allowed the user to manually add other algorithms (including STREAMLINE), we do not consider this as a criteria here.

\item\textbf{‘Deep Learning Included’}: The AutoML includes the availability of some form of deep learning, beyond a standard artificial neural network (ANN) that can be extended beyond a depth of three layers in a hyperparameter sweep. E.g. STREAMLINE can be adapted to look beyond 3 layers using scikit-learns ANN, however we do not consider it as including deep learning here.

\item\textbf{‘Hyperparam. Optimization’}: Does the AutoML allow automatic search, seeking to optimize one or more hyperparameters of the included ML algorithms? We consider neural architecture search to be included here. This is one of the most commonly automated elements of AutoML. 

\item\textbf{‘Models for each CV part.’}: This aspect is focused on comparing the models generated by different ML algorithms. Specifically are individual models, each trained on a respective CV training partition, saved/output/evaluated. Most AutoML tools use internal CV either for the hyperparmeter optimization, or to internally estimate model generalizability before either being evaluated on a hold out dataset (i.e. which could be termed evaluation, testing, or replication data). However this aspect asks if individual models are output by the autoML as part of a more global CV analysis, rather than a single model.

\item\textbf{‘Model Evaluation’}: Does the AutoML automate model evaluation (either on a testing partition, or globally held out data, e.g. replication data) across a variety of metrics, and easily present or summarize these results for the user? This differs from the ‘summary report’ in that it is specifically focused on summarizing and presenting model performance metrics either upon command request, within a GUI, or saved as output.

\item\textbf{‘Ensembling’}: Does the AutoML automate some form of ensembling (e.g. voting, bagging, stacking), apart from what could be considered ‘feature learning’. More specifically is the AutoML capable of generating an ensemble model using some or all other models trained by the different ML modeling algorithms. This could be across CV partitions or one model for each algorithm. Ideally this capability would also include comparing ensemble model performance to models generated by individual ML algorithms (but this was not always the case).

\item\textbf{‘Significance Comparisons’}: Does the AutoML automatically conduct statistical significance analysis comparing model and or dataset performance? STREAMLINE appears to be unique in this regard.

\item\textbf{‘Directly Compare Datasets’}: Does the AutoML support running multiple datasets through the pipeline at once and facilitate ML performance comparisons between those multiple datasets?

\item\textbf{‘Replication Analysis’}: Does the AutoML tool make it easy to re-evaluate a model on some additional holdout or new data (including target outcome information), beyond the primary hold out data. This often includes saving/archiving the trained model(s), and ideally should include applying the same data processing that was applied to the training data (e.g. imputation, feature transformations, feature engineering, feature selection, etc.) to ensure the model is applicable to that new data. This was difficult to clearly determine from the documentation of most other AutoMLs.

\item\textbf{‘Model Deploy (Predictions)’}: Does the AutoML make it easy to apply the model to making predictions on data that does not include target outcome information, e.g. making predictions for a competition, or in deploying the model to make real-world predictions (e.g. clinical decision making, or risk prediction). Ideally this would also include some form of explanation indicating the factors or reasoning from the model that led to a given prediction.
\end{itemize}
\end{itemize}

\addcontentsline{toc}{subsection}{Table S2: A High-Level Comparison of AutoML Tools and Libraries.}
\begin{figure}[H]
    \centering
    \includegraphics[width=\linewidth]{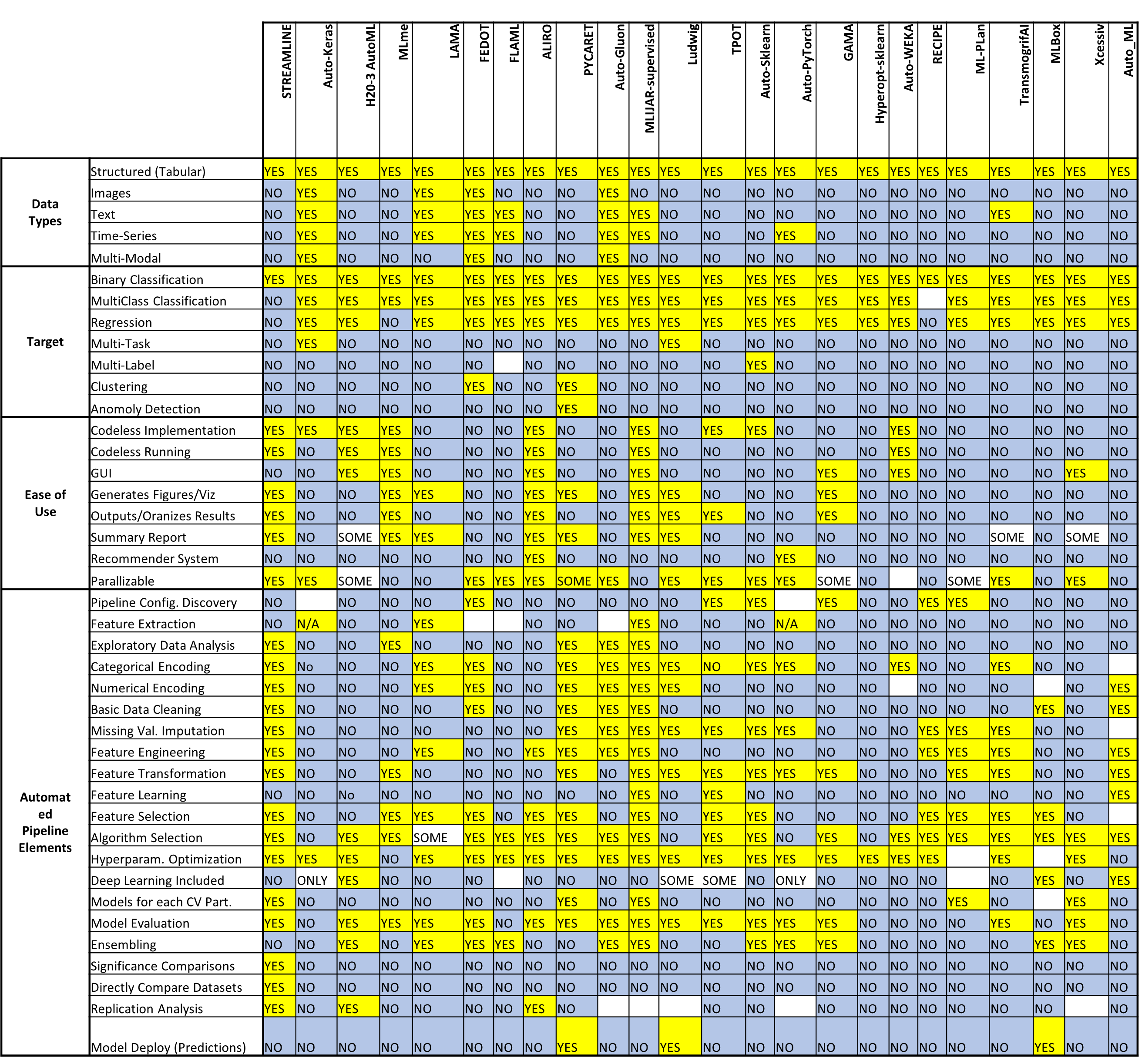}
    \caption*{\textbf{Table S2: A High-Level Comparison of AutoML Tools and Libraries.} 24 AutoML methods are compared with respect to various aspects of capabilities and scope of automation. ‘YES’ entries are highlighted in green, and ‘NO’ entries are highlighted in red. Ambiguous entries and those that we could not confidently determine from the documentation are left white and/or blank. Note, that it should not be concluded from this table that AutoMLs with more capabilities are necessarily better with respect to overall performance, ease of use, or adoption of data science best practices (in general or for specific applications).}
    \label{tab:tab2}
\end{figure}

\addcontentsline{toc}{section}{S.2 Materials \& Methods}
\section*{S.2 Materials \& Methods}

\addcontentsline{toc}{subsection}{S.2 STREAMLINE Detailed Pipeline Walkthrough}
\subsection*{S.2.1 STREAMLINE Detailed Pipeline Walkthrough}

In this section we explain: (1) what STREAMLINE does, (2) what happens in during each phase, (3) why it's designed the way it has been, (4) what user options are available to customize a run, and (5) what to expect when running a given phase. Phases 1-6 make up the core automated pipeline, with Phase 7 and beyond being run optionally based on user needs. Phases are organized to both encapsulate related pipeline elements, as well as to address practical computational needs. STREAMLINE includes reliable default run parameters so that it can easily be used 'as-is', but these parameters can be adjusted for further customization. We refer to a single run of the entire STREAMLINE pipeline as an 'experiment', with all outputs saved to a single 'experiment folder' for later examination and re-use.

 To avoid confusion on 'dataset' terminology we briefly review our definitions here:

1. \textbf{Target dataset} - A whole dataset (minus any instances the user may wish to hold out for replication) that has not yet undergone any other data partitioning and is intended to be used in the training and testing of models within STREAMLINE. Could also be referred to as the 'development dataset'.

2. \textbf{Training dataset} - A generally larger partition of the target dataset used in training a model

3. \textbf{Testing dataset }- A generally smaller partition of the target dataset used to evaluate the trained model

4. \textbf{Validation dataset} - The temporary, secondary hold-out partition of a given training dataset used for hyperparameter optimization. This is the product of using nested (aka double) cross-validation in STREAMLINE as a whole.

5. \textbf{Replication dataset} - Further data that is withheld from STREAMLINE phases 1-7 to (1) compare model evaluations on the same hold-out data and (2) verify the replicability and generalizability of model performance on data collected from other sites or sample populations. A replication dataset should have at least all of the features present in the target dataset which it seeks to replicate. 

 In the sections below, specific STREAMLINE run parameters are indicated by italics. Further details on STREAMLINE installation, run modes, run instructions, run parameters, output files, and additional code to do more with the output is given in the STREAMLINE documentation ( \href{https://urbslab.github.io/STREAMLINE/index.html}{https://urbslab.github.io/STREAMLINE/index.html}).

\addcontentsline{toc}{subsubsection}{S.2.1.1: Phase 1 - Data Exploration \& Processing}
\subsubsection*{S.2.1.1: Phase 1 - Data Exploration \& Processing}
This phase; (1) provides the user with key information about the target dataset(s) they wish to analyze, via an initial exploratory data analysis (EDA) (2) numerically encodes any text-based feature values in the data, (3) applies basic data cleaning and feature engineering to process the data, (4) informs the user how the data has been changed by the data processing, via a secondary, more in-depth EDA, and then (5) partitions the data using k-fold cross validation. 

\begin{itemize}
\itemsep 0em
\item Parallelizability: Runs once for each target dataset to be analyzed
\item Run Time: Typically fast, except when evaluating and visualizing feature correlation in datasets with a large number of features
\end{itemize}

\paragraph{Initial EDA}
Characterizes the orignal dataset as loaded by the user, including: data dimensions, feature type counts, missing value counts, class balance, other standard pandas data summaries (i.e. describe(), dtypes(), nunique()) and feature correlations (pearson). 

For precision, we strongly suggest users identify which features in their data should be treated as categorical vs. quantitative using the \textit{categorical\_feature\_path} and/or \textit{quantitative\_feature\_path} run parameters. However, if not specified by the user, STREAMLINE will attempt to automatically determine feature types relying on the \textit{categorical\_cutoff} parameter. Any features with fewer unique values than \textit{categorical\_cutoff} will be treated as categorical, and all others will be treated as quantitative. 

\begin{itemize}
\itemsep 0em
 \item Output: (1) CSV files for all above data characteristics, (2) bar plot of class balance, (3) histogram of missing values in data, (4) feature correlation heatmap
\end{itemize}

\paragraph{Numerical Encoding of Text-based Features}
Detects any features in the data with non-numeric values and applies scikit-learn’s ‘LabelEncoder’ to make them numeric as required by scikit-learn machine learning packages.

\paragraph{Basic Data Cleaning and Feature Engineering}
Applies the following steps to the target data, keeping track of changes to all data counts along the way:

1. Remove any instances that are missing an outcome label (as these cannot be used while conducting supervised learning)

2. Remove any features identified by the user with \textit{ignore\_features\_path }(a convenience for users that may wish to exclude one or more features from the analysis without changing the original dataset)

3. Engineer/add 'missingness' features. Any original feature with a missing value proportion greater than \textit{featureeng\_missingness} will have a new feature added to the dataset that encodes missingness with 0 = not missing and 1 = missing. This allows the user to examine whether missingness is not at random (MNAR) and is predictive of outcome itself.

4. Remove any features that have invariant values (i.e. they are always the same), or that have only one value in addition to missing values. Then remove any features with a missingness greater than \textit{cleaning\_missingness}. Afterwards, remove any instances in the data that may have a missingness greater than \textit{cleaning\_missingness}.

5. Engineer/add one-hot-encoding for any categorical features in the data. This ensures that all categorical features are treated as such throughout all aspects of the pipeline. For example, a single categorical feature with 3 possible states will be encoded as 3 separate binary-valued features indicating whether an instance has that feature's state or not. Feature names are automatically updated by STREAMLINE to reflect this change.

6. Remove highly correlated features based on \textit{correlation\_removal\_threshold}. Randomly removes one feature of a highly correlated feature pair (Pearson). While perfectly correlated features can be safely cleaned in this way, there is a chance of information loss when removing less correlated features.

\begin{itemize}
\itemsep 0em
 \item Output: CSV file summarizing changes to data counts during these cleaning and engineering steps.
\end{itemize}

\paragraph{Processed Data EDA}
Completes a more comprehensive EDA of the processed dataset including: everything examined in the initial EDA, as well as a univariate association analysis of all features using Chi-Square (for categorical features), or Mann-Whitney U-Test (for quantitative features). 

\begin{itemize}
\itemsep 0em
 \item Output: (1) CSV files for all above data characteristics, (2) bar plot of class balance, (3) histogram of missing values in data, (4) feature correlation heatmap, (5) a CSV file summarizing the univariate analyses including the test applied, test statistic, and p-value for each feature, (6) for any feature with a univariate analysis p-value less than \textit{sig\_cutoff} (i.e. significant association with outcome), a bar-plot will be generated if it is categorical, and a box-plot will be generated if it is quantitative.
\end{itemize}

\paragraph{k-fold Cross Validation (CV) Partitioning}

For k-fold CV, STREAMLINE uses 'Stratified' partitioning by default, which aims to maintain the same/similar class balance within the 'k' training and testing datasets. The value of 'k' can be adjusted with \textit{cv\_partitions}. However, using \textit{partition\_method}, users can also select 'Random' or 'Group' partitioning. 

 Of note, 'Group' partitioning requires the dataset to include a column identified by \textit{match\_label}. This column includes a group membership identifier for each instance which indicates that any instance with the same group ID should be kept within the same partition during cross validation. This was originally intended for running STREAMLINE on epidemiological data that had been matched for one or more covariates (e.g. age, sex, race) in order to adjust for their effects during modeling.

 This strategy of performing k-fold cross validation in order to train (and ultimately evaluate k models for each algorithm is different from most other AutoML tools, which typically only use k-fold CV internally as a way to try and optimize model generalizability during optimization prior to final hold out data evaluation. This approach has the advantage of yielding more than one model to later allow statistical comparisons between algorithms and datasets, but it has the disadvantage of having a smaller number of instances being used to optimize the model(s) during training/optimization. 

\begin{itemize}
\itemsep 0em
 \item Output: CSV files for each training and testing dataset generated following partitioning. Note, by default STREAMLINE will overwrite these files as the working datasets undergo imputation, scaling and feature selection in subsequent phases. However, the user can keep copies of these intermediary CV datasets for review using the parameter \textit{overwrite\_cv}.
\end{itemize}
 
\addcontentsline{toc}{subsubsection}{S.2.1.2: Phase 2 - Imputation and Scaling}
\subsubsection*{S.2.1.2: Phase 2 - Imputation and Scaling}
This phase conducts additional data preparation elements of the pipeline that occur after CV partitioning, i.e. missing value imputation and feature scaling. Both elements are 'trained' and applied separately to each individual training dataset. The respective testing datasets are not looked at when running imputation or feature scaling learning to avoid potential data leakage. However, the learned imputation and scaling patterns are applied in the same way to the testing data as they were in the training data. Both imputation and scaling can optionally be turned off using the parameters \textit{impute\_data }and \textit{scale\_data}, respectively for some specific use cases, however imputation must be on when missing data is present in order to run most scikit-learn modeling algorithms, and scaling should be on for certain modeling algorithms learn effectively (e.g. artificial neural networks), and for if the user wishes to infer feature importances directly from certain algorithm's internal estimators (e.g. logistic regression).

\begin{itemize}
\itemsep 0em
 \item Parallelizability: Runs 'k' times for each target dataset being analyzed (where k is number of CV partitions)
\item Run Time: Typically fast, with the exception of imputing larger datasets with many missing values
\end{itemize}

\paragraph{Imputation}
This phase first conducts imputation to replace any remaining missing values in the dataset with a 'value guess'. While missing value imputation could be reasonably viewed as data manufacturing, it is a common practice and viewed here as a necessary 'evil' in order to run scikit-learn modeling algorithms downstream (which mostly require complete datasets). Imputation is completed prior to scaling since it can influence the correct center and scale to be used.

 Missing value imputation seeks to make a reasonable, educated guess as to the value of a given missing data entry. By default, STREAMLINE uses 'mode imputation' for all categorical values, and multivariate imputation for all quantitative features. However, for larger datasets, multivariate imputation can be slow and require a lot of memory. Therefore, the user can deactivate multiple imputation with the \textit{multi\_impute }parameter, and STREAMLINE will use median imputation for quantitative features instead.

\paragraph{Scaling}
Second, this phase conducts feature scaling with scikit-learn’s ‘StandardScalar’ to transform features to have a mean at zero with unit variance. This is only necessary for certain modeling algorithms, but it should not hinder the performance of other algorithms. The primary drawback to scaling prior to modeling is that any future data applied to the model will need to be scaled in the same way prior to making predictions. Furthermore, for algorithms that have directly interpretable models (e.g. decision tree), the values specified by these models need to be un-scaled in order to understand the model in the context of the original data values. STREAMLINE includes a ‘Useful Notebook’ that can generate direct model visualizations for decision tree and genetic programming models. This code automatically un-scales the values specified in these models so they retain their interpretability.

\begin{itemize}
\itemsep 0em
 \item Output: (1) Learned imputation and scaling strategies for each training dataset are saved as pickled objects allowing any replication or other future data to be identically processed prior to running it through the model. (2) If \textit{overwrite\_cv} is False, new imputed and scaled copies of the training and testing datasets are saved as CSV output files, otherwise the old dataset files are overwritten with these new ones to save space.
\end{itemize}
 
\addcontentsline{toc}{subsubsection}{S.2.1.3: Phase 3 - Feature Importance Estimation}
\subsubsection*{S.2.1.3: Phase 3 - Feature Importance Estimation}
This phase applies feature importance estimation algorithms (i.e. Mutual information (MI) and MultiSURF \cite{urbanowicz_benchmarking_2018}, found in the ReBATE software package) used as filter-based feature selection algorithms. Both algorithms are run by default, however the user can deactivate either using \textit{do\_mutual\_info} or \textit{do\_multisurf}, respectively. MI scores features based on their univariate association with outcome, while MultiSURF scores features in a manner that is sensitive to both univariate and epistatic (i.e. multivariate feature interaction) associations.

 For datasets with a larger number of features (i.e. > 10,000) we recommend turning on the TuRF \cite{moore_tuning_2007} wrapper algorithm with \textit{use\_turf}, which has been shown to improve the sensitivity of MultiSURF to interactions, particularly in larger feature spaces. Users can increase the number of TuRF iterations (and performance) by decreasing \textit{turf\_pct} from 0.5, to approaching 0. However, this will significantly increase MultiSURF run time.

 Overall, this phase is important not only for subsequent feature selection, but as an opportunity to evaluate feature importance estimates prior to modeling outside of the initial univariate analyses (conducted on the entire dataset). Further, comparing feature rankings between MI, and MultiSURF can highlight features that may have little or no univariate effects, but that are involved in epistatic interactions that are predictive of outcome.

\begin{itemize}
\itemsep 0em
 \item Parallelizability: Runs 'k' times for each algorithm (MI and MultiSURF) and each target dataset being analyzed (where k is number of CV partitions)
\item Run Time: Typically reasonably fast, but takes more time to run MultiSURF, in particular as the number of training instances approaches the default \textit{instance\_subset} parameter of 2000 instances, or if this parameter set higher in larger datasets. This is because MultiSURF scales quadratically with the number of training instances.
\item Output: CSV files of feature importance scores for both algorithms and each CV partition ranked from largest to smallest scores.
\end{itemize}
 
\addcontentsline{toc}{subsubsection}{S.2.1.4: Phase 4 - Feature Selection}
\subsubsection*{S.2.1.4: Phase 4 - Feature Selection}
This phase uses the feature importance estimates learned in the prior phase to conduct feature selection using a 'collective' feature selection approach \cite{verma_collective_2018}. By default, STREAMLINE will remove any features from the training data that scored 0 or less by both feature importance algorithms (i.e. features deemed uninformative). Users can optionally ensure retention of all features prior to modeling by setting \textit{filter\_poor\_features} to False. Users can also specify a maximum number of features to retain in each training dataset using \textit{max\_features\_to\_keep} (which can help reduce overall pipeline runtime and make learning easier for modeling algorithms). If after removing 'uninformative features' there are still more features present than the user specified maximum, STREAMLINE will pick the unique top scoring features from one algorithm then the next until the maximum is reached and all other features are removed. Any features identified for removal from the training data are similarly removed from the testing data.

\begin{itemize}
\itemsep 0em
 \item Parallelizability: Runs 'k' times for each target dataset being analyzed (where k is number of CV partitions)
\item Run Time: Fast
\item Output: (1) CSV files summarizing feature selection for a target dataset (i.e. how many features were identified as informative or uninformative within each CV partition) and (2) a barplot of mean feature importance scores (across CV partitions). The user can specify the maximum number of top scoring features to be plotted using \textit{top\_fi\_features}.
\end{itemize}
 
\addcontentsline{toc}{subsubsection}{S.2.1.5: Phase 5 - Machine Learning (ML) Modeling}
\subsubsection*{S.2.1.5: Phase 5 - Machine Learning (ML) Modeling}
At the heart of STREAMLINE, this phase conducts (1) machine learning modeling using the training data, (2) model feature importance estimation (also with the training data), and (3) model evaluation on testing data. STREAMLINE uniquely includes 3 rule-based machine learning algorithms: ExSTraCS, XCS, and eLCS. These 'learning classifier systems' have been demonstrated to be able to detect complex associations while providing human interpretable models in the form of IF:THEN rule-sets. The ExSTraCS algorithm was developed by our research group to specifically handle the challenges of scalability, noise, and detection of epistasis and genetic heterogeneity in biomedical data mining. 

\begin{itemize}
\itemsep 0em
 \item Parallelizability: Runs 'k' times for each algorithm and each target dataset being analyzed (where k is number of CV partitions)
\item Run Time: Slowest phase, but can be sped up by reducing the set of ML methods selected to run, or deactivating ML methods that run slowly on large datasets
\end{itemize}

\paragraph{Model Selection}
The following 16 scikit-learn compatible ML modeling algorithms are currently included as options:

\begin{enumerate}
    \item Naive Bayes (NB) \cite{noauthor_sklearnnaive_bayesgaussiannb_nodate} \textit{(scikit-learn)}
    \item Logistic Regression (LR) \cite{noauthor_sklearnlinear_modellogisticregression_nodate} \textit{(scikit-learn)}
    \item Elastic Net (EN) \cite{noauthor_sklearnlinear_modelelasticnet_nodate} \textit{(scikit-learn)}
    \item Decision Tree (DT) \cite{noauthor_sklearntreedecisiontreeclassifier_nodate} \textit{(scikit-learn)}
    \item Random Forest (RF) \cite{noauthor_sklearnensemblerandomforestclassifier_nodate} \textit{(scikit-learn)}
    \item Gradient Boosting (GB) \cite{noauthor_sklearnensemblegradientboostingclassifier_nodate} \textit{(scikit-learn)}
    \item XGBoost (XGB) \cite{chen_xgboost_2016} \textit{(scikit-learn compatible)}
    \item LGBoost (LGB) \cite{ke_lightgbm_2017} \textit{(scikit-learn compatible)}
    \item CatBoost (CGB) \cite{dorogush_catboost_2018} \textit{(scikit-learn compatible)}
    \item Support Vector Machine (SVM) \cite{noauthor_sklearnsvmsvc_nodate} \textit{(scikit-learn)}
    \item Artificial Neural Network (ANN) \cite{noauthor_sklearnneural_networkmlpclassifier_nodate} \textit{(scikit-learn)}
    \item K-Nearest Neighbors (k-NN) \cite{noauthor_sklearnneighborskneighborsclassifier_nodate} \textit{(scikit-learn)}
    \item Genetic Programming (GP) \cite{stephens_trevorstephensgplearn_2023} \textit{(GPLearn – scikit-learn compatible)}
    \item Educational Learning Classifier System (eLCS) \cite{urbanowicz2017introduction} \textit{(scikit-learn compatible)}
    \item ‘X’ Classifier System (XCS) \cite{butz_xcs_2006} \textit{(scikit-learn compatible)}
    \item Extended Supervised Tracking and Classifying System (ExSTraCS) \cite{urbanowicz_exstracs_2015} \textit{(scikit-learn compatible)}
\end{enumerate}
 NB, LR, EN, DT, RF, GB, SVM, ANN, and k-NN are all part of the standard scikit-learn package. XGB, LGB, CGB, GP, eLCS, XCS, and ExSTraCS are all separately imported scikit-learn compatible packages.

 The first step is to decide which modeling algorithms to run. By default, STREAMLINE applies 14 of the 16 algorithms (excluding eLCS and XCS) it currently has built in. eLCS and XCS are currently experimental implementations of rule-based ML algorithms. Users can specify a specific subset of algorithms to run using \textit{algorithms}, or alternatively indicate a list of algorithms to exclude from all available algorithms using \textit{exclude}. STREAMLINE is also set up so that more advanced users can add other scikit-learn compatible modeling algorithms to run within the pipeline (as explained in ‘Adding New Modeling Algorithms’ in the documentation). This allows STREAMLINE to be used as a rigorous framework to easily benchmark new modeling algorithms in comparison to other established algorithms.

 Modeling algorithms vary in the implicit or explicit assumptions they make, the manner in which they learn, how they represent a solution (as a model), how well they handle different patterns of association in the data, how long they take to run, and how complex and/or interpretable the resulting model can be. To reduce runtime in datasets with a large number of training instances, users can specify a limited random number of training instances to use in training algorithms that run slowly in such datasets using \textit{training\_subsample}.

 In the STREAMLINE demo, we run only the fastest/simplest three algorithms (Naive Bayes, Logistic Regression, and Decision Tree), however these algorithms all have known limitations in their ability to detect complex associations in data. We encourage users to utilize a wider variety of the 14 established available algorithms, in their analyses, to give STREAMLINE the best opportunity to identify a best performing model for the given task (which is effectively impossible to predict for a given dataset ahead of time). 

 We recommend users utilize at least the following set of algorithms within STREAMLINE: Naive Bayes, Logistic Regression, Decision Tree, Random Forest, XGBoost, SVM, ANN, and ExSTraCS as we have found these to be a reliable set of algorithms with an array of complementary strengths and weaknesses on different problems. ExSTraCS is a rule-based machine learning algorithm in the sub-family of algorithms called learning classifier systems (LCS) that has been specifically adapted to the challenges of biomedical data analysis \cite{urbanowicz_exstracs_2015}.

\paragraph{Hyperparameter Optimization}
Most machine learning algorithms have a variety of hyperparameter options that influence how the algorithm runs and performs on a given dataset. In designing STREAMLINE we sought to identify the full set of important hyperparameters for each algorithm, along with a comprehensive range of possible settings and hard-coded these into the pipeline. STREAMLINE adopts the ‘Optuna’ package to conduct automated Bayesian optimization of hyperparameters for most algorithms by default. The evaluation metric 'balanced accuracy' is used to optimize hyperparameters as it takes class imbalance into account and places equal weight on the accurate prediction of both class outcomes. However, users can select an alternative metric with \textit{primary\_metric} and whether that metric needs to be maximized or minimized using the \textit{metric\_direction} parameter. To conduct the hyperparameter sweep, Optuna splits a given training dataset further, applying 3-fold cross validation to generate further internal training and validation partitions with which to evaluate different hyperparameter combinations.

 Users can also configure how Optuna operates in STREAMLINE with \textit{n\_trials }and\textit{ timeout} which controls the target number of hyperparameter value combination trials to conduct, as well as how much total time to try and complete these trials before picking the best hyperparameters found. To ensure reproducibility of the pipeline, note that \textit{timeout} should be set to ‘None’ (however this can take much longer to complete depending on other pipeline settings).

 Notable exceptions to most algorithms; Naive Bayes has no parameters to optimize, and rule-based (i.e. LCS) ML algorithms including ExSTraCS, eLCS, and XCS can be computationally expensive thus STREAMLINE is set to use their default hyperparameter settings without a sweep unless the user updates \textit{do\_lcs\_sweep} and \textit{lcs\_timeout}. While these LCS algorithms have many possible hyperparameters to manipulate they have largely stable performance when leaving most of their parameters to default values. Exceptions to this include the key LCS hyperparameters (1) number of learning iterations, (2) maximum rule-population size, and (3) accuracy pressure (nu), which can be manually controlled without a hyperparameter sweep by \textit{lcs\_iterations}, \textit{lcs\_N}, and \textit{lcs\_nu}, respectively.

\begin{itemize}
\itemsep 0em
 \item Output: CSV files specifying the optimized hyperparameter settings found by Optuna for each partition and algorithm combination.
\end{itemize}

\paragraph{Train 'Optimized' Model}
Having selected the best hyperparameter combination identified for a given training dataset and algorithm, STREAMLINE now retrains each model on the entire training dataset using those respective hyperparameter settings. This yields a total of 'k' potentially 'optimized' models for each algorithm. 

\begin{itemize}
\itemsep 0em
 \item Output: All trained models are pickled as python objects that can be loaded and applied later.
\end{itemize}

\paragraph{Model Feature Importance Estimation}
Next, STREAMINE estimates and summarizes model feature importance scores for every algorithm run. This is distinct from the initial feature importance estimation phase, in that these estimates are specific to a given model as a useful part of model interpretation/explanation. By default, STREAMLINE employes scikit-learn’s ‘permutation feature importance’ for estimating feature importances scores in the same uniform manner across all algorithms. Some ML algorithms that have a build in strategy to gather model feature importance estimates (i.e. LR,DT,RF,XGB,LGB,GB,eLCS,XCS,ExSTraCS). The user can instead use these estimates by setting \textit{use\_uniform\_fi} to ‘False’. This will direct STREAMLINE to report any available internal feature importance estimate for a given algorithm, while still utilizing permutation feature importance for algorithms with no such internal estimator. 

\begin{itemize}
\itemsep 0em
 \item Output: All feature importance scores are pickled as python objects for use in the next phase of the pipeline.
\end{itemize}

\paragraph{Evaluate Performance}
The last step in this phase is to evaluate all trained models using their respective testing datasets. A total of 16 standard classification metrics calculated for each model including: balanced accuracy, standard accuracy, F1 Score, sensitivity (recall), specificity, precision (positive predictive value), true positives, true negatives, false postitives, false negatives, negative predictive value, likeliehood ratio positive, likeliehood ratio negative, area under the ROC, area under the PRC, and average precision of PRC. 

\begin{itemize}
\itemsep 0em
 \item Output: All evaluation metrics are pickled as python objects for use in the next phase of the pipeline.
\end{itemize}
 
\addcontentsline{toc}{subsubsection}{S.2.1.6: Phase 6 - Post-Analysis}
\subsubsection*{S.2.1.6: Phase 6 - Post-Analysis}
This phase combines all modeling results to generate summary statistics files, generate results plots, and conduct non-parametric statistical significance analysis comparing ML performance across CV runs.

\begin{itemize}  \itemsep0em 
\item Output (Evaluation Metrics):

\begin{itemize}  \itemsep0em 
\item Testing data evaluation metrics for each CV partition - CSV file for each modeling algorithm 
\item Mean testing data evaluation metrics for each modeling algorithm - CSV file for mean, median, and standard deviation 
\item ROC and PRC curves for each CV partition in contrast with the average curve - ROC and PRC plot for each modeling algorithm
\item The generation of these plots can be turned off with \textit{exclude\_plots} 
\item Summary ROC and PRC plots - compares average ROC or PRC curves over all CV partitions for each modeling algorithm
\item Boxplots for each evaluation metric - comparing algorithm performance over all CV partitions

\begin{itemize}  \itemsep0em 
\item The generation of these plots can be turned off with \textit{exclude\_plots} 
\end{itemize}

\end{itemize}

\item Output (Model Feature Importance):

\begin{itemize}  \itemsep0em 
\item Model feature importance estimates for each CV partition - CSV file for each modeling algorithm
\item Boxplots comparing feature importance scores for each CV partition - CSV file for each modeling algorithm
\item Composite feature importance barplots illustrating mean feature importance scores across all algorithms (2 versions)

\begin{itemize}  \itemsep0em 

\item Feature importance scores are normalized prior to visualization

\item Feature importance scores are normalized and weighted by mean model performance metric (balanced accuracy by default)

\begin{itemize}  \itemsep0em 
\item The metric used to weight this plot can be changed with \textit{metric\_weight}
\end{itemize}

\end{itemize}

\item Number of top scoring features illustrated in feature importance plots is controlled by \textit{top\_model\_fi\_features}

\end{itemize}

 \item Output (Statistical Significance):

 \begin{itemize}  \itemsep0em 

\item Kruskal Wallis test results assessing whether there is a significant difference for each performance metric among all algorithms - CSV file

\begin{itemize}  \itemsep0em 
 
\item For any metric that yields a significant difference based on \textit{sig\_cutoff}, pairwise statistical tests between algorithms will be conducted using both the Mann Whitney U and the Wilcoxon Rank Test.
\end{itemize}

\item Pairwise statistical tests between algorithms using both the Mann Whitney U-test and the Wilcoxon Rank Test - CSV file for each statistic and significance test.
\item Parallelizability: Runs once for each target dataset being analyzed.
\item Run Time: Moderately fast - turning off some figure generation can make this phase faster

 \end{itemize}
 
\end{itemize}

\addcontentsline{toc}{subsubsection}{S.2.1.7: Phase 7 - Compare Datasets}
\subsubsection*{S.2.1.7: Phase 7 - Compare Datasets}
This phase should be run when STREAMLINE was applied to more than one target dataset during the 'experiment'. It applies further non-parametric statistical significance testing between datasets to identify if performance differences were observed among datasets comparing (1) the best performing algorithms or (2) on an algorithm-by-algorithm basis. It also generates plots to compare performance across datasets and examine algorithm performance consistency. 

\begin{itemize}  \itemsep0em 
\item Parallelizability: Runs once - not parallelizable
\item Run Time: Fast
\item Output (Comparing Best Performing Algorithms for each Metric):

\begin{itemize}  \itemsep0em 
\item Kruskal Wallis test results assessing whether there is a significant difference between median CV performance metric among all datasets focused on the best performing algorithm for each dataset - CSV file

\begin{itemize}  \itemsep0em 
\item For any metric that yields a significant difference based on \textit{sig\_cutoff} pairwise statistical tests between datasets will be conducted using both the Mann Whitney U-test and the Wilcoxon Rank Test 
\end{itemize}

\item Pairwise statistical tests between datasets focused on the best performing algorithm for each dataset using both the Mann Whitney U-test and the Wilcoxon Rank Test - CSV file for each significance test.

\end{itemize}

\item Output (Comparing Algorithms Independently):

\begin{itemize}  \itemsep0em 

\item Kruskal Wallis test results assessing whether there is a significant difference for each median CV performance metric among all datasets - CSV file for each algorithm

\begin{itemize}  \itemsep0em 

\item For any metric that yields a significant difference based on \textit{sig\_cutoff}, pairwise statistical tests between datasets will be conducted using both the Mann Whitney U-test and the Wilcoxon Rank Test 

\end{itemize}

\item Pairwise statistical tests between datasets using both the Mann Whitney U-test and the Wilcoxon Rank Test - CSV file for each significance test and algorithm
\end{itemize}
\end{itemize}

\addcontentsline{toc}{subsubsection}{S.2.1.8: Phase 8 - Replication}
\subsubsection*{S.2.1.8: Phase 8 - Replication}
This phase of STREAMLINE is only run when the user has further hold out data , i.e. one or more 'replication' datasets, which will be used to re-evaluate all models trained on a given target dataset. This means that this phase would need to be run once to evaluate the models of each original target dataset Notably, this phase would be the first time that all models are evaluated on the same set of data which is useful for more confidently picking a 'best' model and further evaluating model generalizability and it's ability to replicate performance on data collected at different times, sites, or populations.

To run this phase the user needs to specify the filepath to the target dataset to be replicated with \textit{dataset\_for\_rep} as well as the folderpath to the folder containing one or more replication datasets using \textit{rep\_data\_path}.

This phase begins by conducting an initial exploratory data analysis (EDA) on the new replication dataset(s), followed by uniquely processing versions of the replication dataset(s) to correspond with each original CV training dataset, yielding the same number of features (but not necessarily the same number of instances). This processing includes cleaning of instances in replication data that have no outcome or with a missingness over the user-defined threshold. Then to ensure that the same features exist in the replication data that existed in the respective model’s original training set, feature engineering creates the same one-hot-encoded categorical features, as well as the same ‘missingness features’ previously generated. If a new categorical feature values is observed (that didn’t exist in the original training data for one-hot-encoding) this new value is treated as ‘missing’ in replication data by assigning 0’s to all one hot encoded values for that feature. Similarly, with respect to feature cleaning, the same features removed in respective training data are removed from the corresponding replication dataset (i.e. features marked for exclusion or features removed due to high missingness, high feature correlation, or removed during feature selection. Then EDA is repeated to confirm processing of the replication dataset and generate a feature correlation heatmap, however univariate analyses are not repeated on the replication data.

Next all models previously trained for the target dataset are re-evaluated across all metrics using each replication dataset with results saved separately. Similarly, all model evaluation plots (with the exception of feature importance plots) are automatically generated. As before non-parametric statistical tests are applied to examine differences in algorithm performance. 

\begin{itemize}  \itemsep0em 
\item Parallelizability: Runs once for each replication dataset being analyzed for a given target dataset.
\item Run Time: Moderately fast
\item Output: Similar outputs to Phase 1 minus univariate analyses, and similar outputs to Phase 6 minus feature importance assessments.
\end{itemize}

\addcontentsline{toc}{subsubsection}{S.2.1.9: Phase 9 – Summary Report}
\subsubsection*{S.2.1.9: Phase 9 – Summary Report}
This final phase generates a pre-formatted PDF report summarizing (1) STREAMLINE run parameters, (2) key exploratory analysis for the processed target data, (3) key ML modeling results (including metrics and feature importances), (4) dataset comparisons (if relevant), (5) key statistical significance comparisons, and (6) runtime. STREAMLINE collects run-time information on each phase of the pipeline and for the training of each ML algorithm model.

Separate reports are generated representing the findings from running Phases 1-7, i.e. 'Testing Data Evaluation Report', as well as for Phase 8 if run on replication data, i.e. 'Replication Data Evaluation Report'. 

\begin{itemize}  \itemsep0em 
\item Parallelizability: Runs once - not parallelizable
\item Run Time: About a minute
\item Output: One or more PDF reports
\end{itemize}

\addcontentsline{toc}{subsection}{S.2.2: Benchmark Datasets}
\subsection*{S.2.2: Benchmark Datasets}
Much of the text describing the GAMETES and x-bit multiplexer datasets below directly quotes our previous STREAMLINE paper [1], which we are replicating in the benchmarking analyses below. Note that since the STREAMLINE codebase has changed, even though we are using the same random seed for these analyses, the CV partitions will not be the same as previous analyses, so the results are not expected to be exactly the same (i.e. we expect to see different performance for different algorithms and different CV partitions due to differences in data partitioning as well as stochastic differences in hyperparameter optimization sweep). However, we expect similar overall performance.

\addcontentsline{toc}{subsubsection}{S.2.2.1: HCC Benchmark Data}
\subsubsection*{S.2.2.1: HCC Benchmark Data}

\paragraph{Real-World HCC Dataset}
The first benchmarking dataset (hcc\_data.csv) is an example of a real-world biomedical classification task. This is a Hepatocellular Carcinoma (HCC) dataset taken from the UCI Machine Learning Repository. It includes 165 instances, 49 features, and a binary class label. It also includes a mix of categorical and quantitative features (however all categorical features are binary), about 10\% missing values, and class imbalance, i.e. 63 deceased (class = 1), and 102 survived (class 0).

\paragraph{Custom Simulated HCC Dataset}
The second benchmarking dataset (hcc\_data\_custom.csv) is similar to the first, but we have made a number of modifications to it in order to test the data cleaning and feature engineering functionalities of STREAMLINE.

Modifications include the following:

\begin{itemize}  \itemsep0em 
\item Removal of covariate features (i.e. ‘Gender’ and ‘Age at diagnosis’)
\item Addition of two simulated instances with a missing class label (to test basic data cleaning)
\item Addition of two simulated instances with a high proportion of missing values (to test instance missingness data cleaning)
\item Addition of three simulated, numerically encoded categorical features with 2, 3, or 4 unique values, respectively (to test one-hot-encoding)
\item Addition of three simulated, text-valued categorical features with 2, 3, or 4 unique values, respectively (to test one-hot-encoding of text-based features)
\item Addition of three simulated quantiative features with high missingness (to test feature missingness data cleaning)
\item Addition of three pairs of correlated quanatiative features (6 features added in total), with correlations of -1.0, 0.9, and 1.0, respectively (to test high correlation data cleaning)
\item Addition of three simulated features with (1) invariant feature values, (2) all missing values, and (3) a mix of invariant values and missing values.
\end{itemize}

These simulated features and instances have been clearly identified in the feature names and instances IDs of this dataset.

\paragraph{Custom Simulated HCC Replication Dataset}
The last demo dataset (hcc\_data\_custom\_rep.csv) was simulated as a mock replication dataset for hcc\_data\_custom.csv. To generate this dataset we first took hcc\_data\_custom.csv and for 30\% of instances randomly generated realistic looking new values for each feature and class outcome (effectively adding noise to this data). Furthermore, we simulated further instances that test the ability of STREAMLINE’s one-hot-encoding to ignore new (as-of-yet unseen) categorical features values during STREAMLINE’s replication phase. If this were to happen, the new value would be ignored (i.e. no new feature columns added).

Modifications included adding a simulated instance that includes a new (as-of-yet unseen) value for the following previously simulated features:

\begin{itemize}  \itemsep0em 
\item The binary text-valued categorical feature
\item The 3-value text-valued categorical feature
\item The binary numerically encoded categorical feature
\item The 3-value numerically encoded categorical feature
\item We also added a new, previously unseen feature value to each of the invariant feature columns.
 \end{itemize}

 The code to generate the additional features and instances within the custom hcc\_data\_custom.csv and hcc\_data\_-custom\_rep.csv can be found in the notebook at /data/Generate\_Expanded\_HCC\_dataset.ipynb.

\paragraph{HCC Analysis With STREAMLINE}
We applied STREAMLINE to analyze and compare both the HCC and custom HCC datasets. For the replication phase we evaluated the custom HCC dataset using it’s corresponding simulated custom replication dataset to validate the efficacy of replication data preparation and model evaluation. See the first page of ‘hcc\_exp\_STREAMLINE\_Report.pdf’ in the supplementary materials for all STREAMLINE run parameters used in this analysis.

 We repeated this analysis twice in parallel using a fixed number of Optuna trials and no time limit to confirm the replicability of STREAMLINE.

\addcontentsline{toc}{subsubsection}{S.2.2.2: GAMETES Benchmark Data}
\subsubsection*{S.2.2.2: GAMETES Benchmark Data}

We evaluated 6 genomics datasets simulated with GAMETES \cite{urbanowicz_gametes_2012} using a newer version of the code extended to simulated additive and heterogeneous associations (\href{https://github.com/UrbsLab/GAMETES/tree/v2.2} {https://github.com/UrbsLab/GAMETES/tree/v2.2}). 
Each dataset was modeled with a different underlying pattern of association to illustrate how STREAMLINE can be utilized to evaluate the strengths and weaknesses of different ML algorithms in different contexts. Each dataset simulated 100 single nucleotide polymorphisms (SNPs) as features and included 1600 instances using a minor allele frequency of 0.2 for relevant features and the ‘easiest’ model architecture generated for each configuration using the GAMETES simulation approach. All irrelevant features were randomly simulated with a minor allele frequency between 0.05 and 0.5. We detail dataset differences here (A) Univariate association between a single feature and outcome with 1 relevant and 99 irrelevant features and a heritability of 0.4, (B) Additive combination of 4 univariate associations with 4 relevant and 96 irrelevant features and heritability of 0.4 for each relevant feature, (C) Heterogeneous combination of 4 univariate associations, i.e. each univariate association is only predictive in a respective quarter of instances (with all else the same as dataset B), (D) Pure 2-way feature interaction with 2 relevant and 98 irrelevant features and heritability of 0.4, (E) Heterogeneous combination of 2 independent pure 2-way feature interactions with 4 relevant and 96 irrelevant features and heritability of 0.4 for each 2-way interaction (used in above sections for figure examples), and (F) Pure 3-way feature interaction with 3 relevant and 97 irrelevant features and heritability of 0.2 (i.e. the most difficult dataset).

\paragraph{GAMETES Analysis With STREAMLINE}
We applied STREAMLINE to analyze and compare all 6 GAMETES datasets, treating all features as quantitative unless it had only two possible values in the training (largely in-line with the original 0.2.5 STREAMLINE analysis which was not set up to apply one-hot-encoding). No replication data was applied in this analysis as the ground truth in these dataset simulations is known. See the first page of ‘gametes\_exp\_STREAMLINE\_Report.pdf’ in the supplementary materials for all STREAMLINE run parameters used in this analysis.

 When STREAMLINE conducts modeling in the datasets discussed so far, ExSTraCS, which was trained without a hyperparameter sweep, is set to run with hyperparameters (nu = 1, rule population size =2K, and training iterations = 200K), where the nu setting represents the emphasis on discovering and keeping rules with maximum accuracy. Previous work suggests that nu = 1 is more effective in noisy data, while a nu of 5 or 10 is more effective in clean data (i.e. no noise) \cite{urbanowicz_exstracs_2015}.
 
\addcontentsline{toc}{subsubsection}{S.2.2.3: Multiplexer Benchmark Data}
\subsubsection*{S.2.2.3: Multiplexer Benchmark Data}
We applied STREAMLINE to 6 different x-bit multiplexer (MUX) binary classification benchmark datasets (i.e. 6, 11, 20, 37, 70, and 135-bit) often utilized to evaluate ML algorithms such as RBML \cite{urbanowicz_benchmarking_2018,urbanowicz2017introduction,butz_xcs_2006,urbanowicz_exstracs_2015}. Like GAMETES dataset ‘E’, these MUX datasets involve both feature interactions and heterogeneous associations with outcome, but differently involve binary features and a clean association with outcome. The ‘x’-bit value denotes the number of relevant features underlying the association, and increasing values from 6 to 135 dramatically scale up the complexity of the underlying pattern of association. Specifically, solving the 6-bit problem involves modeling 4 independent 3-way interactions, while the 135-bit problem involves modeling

128 independent 8-way interactions. In these datasets, all features are relevant to solving the problem, where a subset of features serves as ‘address bits’ which point to one of the remaining ‘register bits’. The value of that corresponding register bit indicates the correct outcome (i.e. 0 or 1). No replication data was applied in this analysis as the ground truth in these dataset simulations is known.

 Previous work benchmarking MultiSURF on these datasets indicated that these address bits would yield the largest feature importance scores of all features in the dataset, as these features are important for the correct prediction of every instance in the dataset \cite{urbanowicz_benchmarking_2018}.

 In other previous work, the ExSTraCS algorithm was the first ML demonstrated to directly model the 135-bit problem successfully, in a dataset including 40K training instances, with hyperparameters (nu = 10, rule population size =10K, and training iterations = 1.5 million) \cite{urbanowicz_exstracs_2015}. ExSTraCS has the advantage over other ML algorithms of using the feature importance estimates from MultiSURF In this study, 6 to 135-bit datasets were generated with 500, 1000, 2000, 5000, 10000, and 20000 instances, respectively, with 90\% of each used by STREAMLINE for training (due to 10-fold CV). In previous work, it was demonstrated that larger numbers of instances are required to solve or nearly solve increasingly complex MUX problems. In contrast with modeling on the noisy datasets, for the MUX datasets, ExSTraCS was assigned the following hyperparameter settings: nu = 10, rule population size =5K, and training iterations = 500K. As such, we do not expect ExSTraCS to be able to solve the 135-bit problem in this analysis.

\addcontentsline{toc}{subsubsection}{S.2.2.4: XOR Benchmark Data}
\subsubsection*{S.2.2.4: XOR Benchmark Data}
We evaluated 4 simulated ‘XOR’ datasets with models manually generated and respective datasets simulated with GAMETES [39] using (https://github.com/UrbsLab/GAMETES/tree/v2.2). This benchmark was not previously applied to STREAMLINE v0.2.5. These datasets are designed to examine the capability of ML algorithms (and the STREAMLINE pipeline as a whole) to detect increasingly high dimensional pure n-way interactions (i.e. 2-way, 3-way, 4-way, and 5-way). Each dataset simulated 20 single nucleotide polymorphisms (SNPs) as features and included 1600 instances. Unlike the other GAMETES datasets (described above), but like the multiplexer datasets, these data are simulated to have a perfectly clean signal, such that it is possible to train models with 100\% testing accuracy. These are the same XOR datasets previously used to benchmark the ReBATE suite of Relief-based feature selection algorithms, including ‘MultiSURF \cite{urbanowicz_benchmarking_2018}. All features in this analysis were treated as quantitative unless it had only two possible values in the training (largely in-line with the original 0.2.5 STREAMLINE analysis which was not set up to apply one-hot-encoding). No replication data was applied in this analysis as the ground truth in these dataset simulations is known.

 In that previous work, we found that MultiSURF failed prioritize the correct predictive features in the case of 4-way and 5-way pure interactions. More specifically those feature actually scored the most ‘negative’ of all feature values. Traditionally, Relief-based algorithms are applied by prioritizing the most positively scoring features, and or removing features with a score <= 0, and this is how we have implemented feature selection with MulitSURF in STREAMLINE. However, based on the results in [20] as well as in the current study (see S.3.4), we hypothesize that Relief-based algorithms should prioritize features based on the absolute value of their scores in order to detect higher order interactions (i.e. 4-way interactions and above). Preliminary work supports this hypothesis, and future work will seek to improve STREAMLINE’s current feature selection strategy to better account for the possible presence of high order interactions (which currently most research and benchmarking of algorithms does not explicitly consider), by taking the absolute value of MultiSURF scoring into account. 
 
\newpage
\addcontentsline{toc}{subsection}{S.2.3: Obstructive Sleep Apnea Data}
\subsection*{S.2.3: Obstructive Sleep Apnea Data}
Here we provide a detailed summary of the SAGIC OSA dataset a total of 3111 instances, unique instance IDs, class/outcome (encoded by AHI threshold), and features including demographics (DEM), comorbidities (DX), symptoms (SYM), craniofacial measures from photos (CF), and intraoral measures from photos (IO). \textbf{Table S3} provides summary statistics for original SAGIC dataset. \textbf{Table S4} provides a key explaining all features, their group, type and any underlying encoding. A total of 85 features are available in datasets with all feature groups. The two Mallampati score classifications include “Frontal, Tongue Extended Maximally” (FTEM), and “Tongue in Mouth, No Phonation” (TIMNPH) \cite{schwab_digital_2017}.

\scriptsize
\begingroup
\setlength{\LTleft}{-20cm plus -1fill}
\setlength{\LTright}{\LTleft}
\begin{longtable}{| l | p{1cm} | p{1cm} | p{1cm} | p{1cm} | p{1.2cm} | p{1cm} | p{1cm} | p{1cm} | p{1cm} | p{1cm} |}
\hline
\cellcolor{gray!20}\textbf{Feature} & \cellcolor{gray!20}\textbf{Count} & \cellcolor{gray!20}\textbf{Missing Values} & \cellcolor{gray!20}\textbf{Unique Values} & \cellcolor{gray!20}\textbf{Mean} & \cellcolor{gray!20}\textbf{Standard Deviation} & \cellcolor{gray!20}\textbf{Min} & \cellcolor{gray!20}\textbf{25\%} & \cellcolor{gray!20}\textbf{Median} & \cellcolor{gray!20}\textbf{75\%} & \cellcolor{gray!20}\textbf{Max} \\
\hline
subnum & 3111 & 0 & 3111 & NA & NA & NA & NA & NA & NA & NA \\
\hline
class (AHI$\geq$15) & 3111 & 0 & 2 & 0.45323 & 0.497888 & 0 & 0 & 0 & 1 & 1 \\
\hline
class (AHI$\geq$5) & 3111 & 0 & 2 & 0.688525 & 0.463171 & 0 & 0 & 1 & 1 & 1 \\
\hline
age & 3111 & 0 & 73 & 49.97653 & 13.93849 & 17 & 39 & 50 & 60 & 89 \\
\hline
male & 3111 & 0 & 2 & 0.604629 & 0.489009 & 0 & 0 & 1 & 1 & 1 \\
\hline
bmi & 3111 & 0 & 2426 & 29.75954 & 7.049058 & 17.03923 & 25.01352 & 28.3737 & 32.92469 & 75.30469 \\
\hline
caucasian & 3104 & 7 & 2 & 0.308956 & 0.462137 & 0 & 0 & 0 & 1 & 1 \\
\hline
aframer & 3104 & 7 & 2 & 0.037371 & 0.1897 & 0 & 0 & 0 & 0 & 1 \\
\hline
asian & 3104 & 7 & 2 & 0.378222 & 0.485021 & 0 & 0 & 0 & 1 & 1 \\
\hline
csamer & 3104 & 7 & 2 & 0.209407 & 0.406951 & 0 & 0 & 0 & 0 & 1 \\
\hline
other & 3104 & 7 & 2 & 0.066044 & 0.248399 & 0 & 0 & 0 & 0 & 1 \\
\hline
medhx\_hbp & 2537 & 574 & 2 & 0.44659 & 0.497237 & 0 & 0 & 0 & 1 & 1 \\
\hline
medhx\_cad & 2517 & 594 & 2 & 0.097735 & 0.297016 & 0 & 0 & 0 & 0 & 1 \\
\hline
medhx\_heart\_fail & 2520 & 591 & 2 & 0.04127 & 0.198953 & 0 & 0 & 0 & 0 & 1 \\
\hline
medhx\_stroke & 2526 & 585 & 2 & 0.034046 & 0.181383 & 0 & 0 & 0 & 0 & 1 \\
\hline
medhx\_ldl & 2489 & 622 & 2 & 0.451185 & 0.497711 & 0 & 0 & 0 & 1 & 1 \\
\hline
medhx\_dm & 2525 & 586 & 2 & 0.127129 & 0.333183 & 0 & 0 & 0 & 0 & 1 \\
\hline
medhx\_afib & 2293 & 818 & 2 & 0.060183 & 0.237878 & 0 & 0 & 0 & 0 & 1 \\
\hline
loud\_snore & 2571 & 540 & 5 & 2.054842 & 1.729045 & 0 & 0 & 2 & 4 & 4 \\
\hline
gasping & 2554 & 557 & 5 & 1.287001 & 1.558197 & 0 & 0 & 0 & 3 & 4 \\
\hline
choke & 2554 & 557 & 5 & 1.100235 & 1.512753 & 0 & 0 & 0 & 2 & 4 \\
\hline
map\_index & 2582 & 529 & 15 & 1.487284 & 1.3244 & 0 & 0 & 1.333333 & 2.666667 & 4 \\
\hline
upper\_face\_depth & 3108 & 3 & 86 & 10.0705 & 0.983102 & 6.7 & 9.4 & 10.1 & 10.7 & 14.9 \\
\hline
mid\_face\_depth & 3106 & 5 & 85 & 10.7872 & 0.980476 & 7.8 & 10.1 & 10.8 & 11.4 & 15.1 \\
\hline
low\_face\_depth & 3110 & 1 & 94 & 12.91204 & 1.149739 & 9.5 & 12.1 & 12.9 & 13.6 & 19 \\
\hline
max\_depth\_angle & 3104 & 7 & 355 & 84.44965 & 6.580233 & 8.9 & 80.2 & 84.4 & 88.7 & 107.6 \\
\hline
mand\_depth\_angle & 3108 & 3 & 337 & 77.01638 & 6.260158 & 8.9 & 72.8 & 76.95 & 81.3 & 97.6 \\
\hline
max\_mand\_rel\_angle & 3106 & 5 & 191 & 7.445419 & 2.978375 & -3.3 & 5.5 & 7.3 & 9.4 & 18.3 \\
\hline
facial\_axis\_angle & 3108 & 3 & 417 & 30.97631 & 7.939419 & 6.1 & 25.5 & 30.7 & 36.4 & 60.8 \\
\hline
photo\_upper\_face\_ht & 3109 & 2 & 60 & 7.211444 & 0.626629 & 5.3 & 6.8 & 7.2 & 7.6 & 9.9 \\
\hline
nose\_ht & 3105 & 6 & 52 & 4.995208 & 0.501182 & 3.4 & 4.7 & 5 & 5.3 & 11.4 \\
\hline
photo\_low\_face\_ht & 3107 & 4 & 67 & 6.168555 & 0.695872 & 3.9 & 5.7 & 6.1 & 6.6 & 9.3 \\
\hline
mand\_length & 3108 & 3 & 89 & 7.77684 & 1.028879 & 4.4 & 7.1 & 7.7 & 8.4 & 13 \\
\hline
mand\_length\_dia & 3108 & 3 & 85 & 9.548668 & 1.001957 & 6.6 & 8.9 & 9.5 & 10.1 & 13.8 \\
\hline
post\_mand\_ht & 3110 & 1 & 74 & 6.737971 & 0.947212 & 3.7 & 6.1 & 6.7 & 7.4 & 10.5 \\
\hline
mand\_plane\_angle & 3109 & 2 & 422 & 11.29391 & 7.808057 & -15.9 & 6 & 11.3 & 16.3 & 46.5 \\
\hline
photo\_mand\_angle & 3073 & 38 & 512 & 114.158 & 10.24584 & 10.1 & 108.6 & 114.3 & 120.1 & 159.5 \\
\hline
max\_mand\_box\_area & 3106 & 5 & 479 & 54.66754 & 9.55144 & 10.1 & 47.9 & 54.2 & 60.875 & 105.6 \\
\hline
cranial\_base\_tri\_area & 3109 & 2 & 463 & 66.1554 & 9.033765 & 4.2 & 60.2 & 65.5 & 71.6 & 113.9 \\
\hline
max\_tri\_area & 3107 & 4 & 478 & 70.8823 & 9.490196 & 4.3 & 64.4 & 70.4 & 76.7 & 117.1 \\
\hline
mand\_tri\_area & 3111 & 0 & 419 & 39.48905 & 8.109848 & 7.2 & 33.8 & 38.8 & 43.9 & 82 \\
\hline
mid\_cran\_fossa\_vol & 3105 & 6 & 889 & 109.475 & 21.77432 & 7.3 & 94.6 & 107.2 & 122.39 & 217.2 \\
\hline
head\_posit\_angle & 3109 & 2 & 386 & 12.65574 & 7.282256 & -11.2 & 8.2 & 13.2 & 17.5 & 35.5 \\
\hline
max\_volume & 3104 & 7 & 1344 & 159.8664 & 69.70423 & 5.9 & 130.375 & 152 & 177.6 & 1565.3 \\
\hline
mand\_vol & 3111 & 0 & 1513 & 142.8338 & 49.46381 & 6.9 & 107.2 & 138.3 & 172.3 & 365.8 \\
\hline
max\_mand\_vol & 3104 & 7 & 1892 & 302.4981 & 94.83212 & 6.9 & 244.3 & 290.2 & 345.15 & 1753.6 \\
\hline
ant\_mand\_ht & 3111 & 0 & 58 & 3.918666 & 0.545831 & 2.1 & 3.6 & 3.9 & 4.3 & 7 \\
\hline
man\_nas\_ang & 3106 & 5 & 290 & 40.04895 & 4.977194 & 8.5 & 36.8 & 39.9 & 43.4 & 60.9 \\
\hline
man\_sub\_ang & 3104 & 7 & 364 & 57.18392 & 6.654066 & 7 & 52.7 & 57.1 & 61.4 & 84.4 \\
\hline
hea\_inc\_ang & 3108 & 3 & 377 & 19.68565 & 7.200343 & -5.3 & 14.8 & 19.8 & 24.7 & 42 \\
\hline
cer\_ang & 2836 & 275 & 751 & 154.3691 & 18.35592 & 8.3 & 141.985 & 154.345 & 166.2 & 262.3 \\
\hline
face\_width & 3111 & 0 & 88 & 13.11694 & 0.954734 & 4.8 & 12.5 & 13.1 & 13.7 & 17.6 \\
\hline
mand\_width & 3111 & 0 & 97 & 11.00771 & 1.208481 & 5.4 & 10.2 & 10.9 & 11.8 & 17.2 \\
\hline
intercant\_width & 3099 & 12 & 44 & 3.341488 & 0.402077 & 2.3 & 3.1 & 3.3 & 3.6 & 5.1 \\
\hline
biocular\_width & 3099 & 12 & 67 & 8.711707 & 0.728143 & 6.7 & 8.2 & 8.6 & 9.1 & 12 \\
\hline
nose\_width & 3110 & 1 & 44 & 3.952251 & 0.425781 & 2.7 & 3.7 & 3.9 & 4.2 & 6.2 \\
\hline
lat\_fac\_ht & 3107 & 4 & 80 & 10.31225 & 0.984748 & 7.5 & 9.6 & 10.3 & 11 & 14.3 \\
\hline
man\_wid\_len\_ang & 3111 & 0 & 405 & 74.33594 & 7.909022 & 10.6 & 68.8 & 73.9 & 79.5 & 103.9 \\
\hline
fac\_wid\_fac\_dep\_ang & 3107 & 4 & 282 & 62.32664 & 4.798661 & 12.3 & 59.2 & 62.2 & 65.4 & 79.2 \\
\hline
fac\_wid\_low\_fac\_dep\_ang & 3111 & 0 & 258 & 54.6636 & 4.162312 & 12.4 & 51.9 & 54.6 & 57.2 & 71.8 \\
\hline
eye\_width & 3099 & 12 & 62 & 2.68511 & 0.352662 & 1.65 & 2.45 & 2.65 & 2.85 & 4.15 \\
\hline
face\_eye\_width & 3099 & 12 & 960 & 4.952678 & 0.616253 & 3.2 & 4.529776 & 4.938776 & 5.348332 & 7.210526 \\
\hline
face\_height\_width & 3105 & 6 & 1170 & 0.854631 & 0.101542 & 0.584416 & 0.8 & 0.848 & 0.907143 & 4.270833 \\
\hline
mand\_width\_length & 3108 & 3 & 1333 & 0.711906 & 0.104744 & 0.398438 & 0.642857 & 0.709677 & 0.776596 & 2.407407 \\
\hline
face\_upper\_lower & 3105 & 6 & 611 & 0.816738 & 0.098053 & 0.530864 & 0.75 & 0.8125 & 0.877193 & 1.282051 \\
\hline
face\_height\_depth & 3104 & 7 & 1125 & 1.039679 & 0.100866 & 0.766667 & 0.971963 & 1.031747 & 1.098928 & 2.594937 \\
\hline
cranial\_mand\_area & 3109 & 2 & 2945 & 1.714797 & 0.272733 & 0.583333 & 1.527972 & 1.690073 & 1.87766 & 4.04 \\
\hline
face\_nose\_width & 3110 & 1 & 686 & 3.346333 & 0.33184 & 2.22807 & 3.121951 & 3.333333 & 3.552632 & 5.032258 \\
\hline
open\_wid\_tim\_nph & 3085 & 26 & 2909 & 5.66042 & 0.854845 & 3.516 & 5.042679 & 5.599256 & 6.227113 & 10.02625 \\
\hline
ver\_mou\_ope & 3080 & 31 & 2908 & 4.671383 & 0.916921 & 2.141 & 4.012679 & 4.616064 & 5.261602 & 8.531904 \\
\hline
open\_area\_tim\_nph & 3080 & 31 & 3040 & 19.71676 & 6.062227 & 4.523 & 15.3255 & 18.88386 & 23.3019 & 47.19128 \\
\hline
tongue\_width\_tim\_nph & 2754 & 357 & 2606 & 4.744686 & 0.596651 & 2.179 & 4.36924 & 4.756935 & 5.141194 & 6.713906 \\
\hline
open\_wid\_f\_te\_m & 3000 & 111 & 2838 & 5.820935 & 0.715308 & 3.688 & 5.322976 & 5.79759 & 6.286839 & 9.03 \\
\hline
tongue\_length\_f & 2955 & 156 & 2820 & 5.382951 & 1.107808 & 1.28297 & 4.6385 & 5.378722 & 6.129566 & 9.477 \\
\hline
tongue\_width\_f\_te\_m & 3026 & 85 & 2831 & 5.035716 & 0.600539 & 3.145094 & 4.626886 & 5.02608 & 5.412897 & 7.64877 \\
\hline
ton\_are & 2954 & 157 & 2922 & 22.84309 & 6.453854 & 5.264 & 18.3635 & 22.44592 & 26.68929 & 63.407 \\
\hline
tongue\_length\_p\_te\_m & 3042 & 69 & 2865 & 4.272597 & 0.805157 & 1.36538 & 3.763103 & 4.281134 & 4.799724 & 8.895 \\
\hline
tongue\_area\_p\_te\_m & 3041 & 70 & 2956 & 5.698665 & 1.979176 & 0.082545 & 4.31 & 5.530136 & 6.842665 & 29.382 \\
\hline
tongue\_thickness\_p\_te\_m & 3036 & 75 & 2717 & 1.533286 & 0.311344 & 0.48 & 1.324 & 1.511668 & 1.723153 & 2.923578 \\
\hline
tongue\_curvature\_p\_te\_m & 3042 & 69 & 2915 & 5.012658 & 1.208547 & 1.258 & 4.116 & 4.983812 & 5.860505 & 8.987571 \\
\hline
mal\_score\_ftem\_c1 & 2836 & 275 & 2 & 0.048307 & 0.214453 & 0 & 0 & 0 & 0 & 1 \\
\hline
mal\_score\_ftem\_c2 & 2836 & 275 & 2 & 0.068759 & 0.253088 & 0 & 0 & 0 & 0 & 1 \\
\hline
mal\_score\_ftem\_c3 & 2836 & 275 & 2 & 0.150917 & 0.358031 & 0 & 0 & 0 & 0 & 1 \\
\hline
mal\_score\_ftem\_c4 & 2836 & 275 & 2 & 0.732017 & 0.442987 & 0 & 0 & 1 & 1 & 1 \\
\hline
mal\_score\_timnph\_c1 & 2943 & 168 & 2 & 0.06524 & 0.24699 & 0 & 0 & 0 & 0 & 1 \\
\hline
mal\_score\_timnph\_c2 & 2943 & 168 & 2 & 0.067618 & 0.251132 & 0 & 0 & 0 & 0 & 1 \\
\hline
mal\_score\_timnph\_c3 & 2943 & 168 & 2 & 0.086646 & 0.281364 & 0 & 0 & 0 & 0 & 1 \\
\hline
mal\_score\_timnph\_c4 & 2943 & 168 & 2 & 0.780496 & 0.413981 & 0 & 1 & 1 & 1 & 1 \\
\hline
\caption*{\textbf{Table S3: Summary Statistics for SAGIC Obstructive Sleep Apnea Data.} Includes the entire original dataset prior to replication data partitioning, and internal STREAMLINE k-fold CV partitioning.}
\end{longtable}
\addcontentsline{toc}{subsubsection}{Table S3: Summary Statistics for SAGIC Obstructive Sleep Apnea Data.}

\normalsize
\endgroup

\newpage
\begingroup
\setlength{\LTleft}{-20cm plus -1fill}
\setlength{\LTright}{\LTleft}
\scriptsize
\begin{longtable}{| p{3cm} | p{1cm} | p{6.8cm} | p{5.3cm} |}
\hline
\cellcolor{gray!20}\textbf{Column Name} & \cellcolor{gray!20}\textbf{Feature Group} & \cellcolor{gray!20}\textbf{Description} & \cellcolor{gray!20}\textbf{Feature Type/Encoding} \\
\hline
subnum & NA & unique SAGIC instance ID including site identifier & NA \\
\hline
class (AHI$\geq$15) & NA & OSA binary outcome encoded by AHI$\geq$15 or AHI$\geq$5 & Binary/Categorical (0=No OSA, 1=OSA) \\
\hline
class (AHI$\geq$5) & NA & OSA binary outcome encoded by AHI$\geq$5 & Binary/Categorical (0=No OSA, 1=OSA) \\
\hline
age & DEM & What is your age? & Quantitative \\
\hline
male & DEM & Are you male? & Binary/Categorical (0=Female,1=Male) \\
\hline
bmi & DEM & Body Mass Index kg/m**2 & Quantitative \\
\hline
caucasian & DEM & Caucasian Race/Ethnicity & Binary/Categorical(0=Not, 1=Caucasian) \\
\hline
aframer & DEM & African American Race/Ethnicity & Binary/Categorical(0=Not, 1=African Am.) \\
\hline
asian & DEM & Asian Race/Ethnicity & Binary/Categorical(0=Not, 1=Asian) \\
\hline
csamer & DEM & Central/South American Race/Ethnicity & Binary/Categorical(0=Not, 1=Cen/South Am.) \\
\hline
other & DEM & Other Race/Ethnicity & Binary/Categorical(0=Not, 1=Other) \\
\hline
medhx\_hbp & DX & Have you been diagnosed with high blood pressure? & Binary/Categorical(0=No, 1=Yes) \\
\hline
medhx\_cad & DX & Ever Diagnosed with CAD (Angina, MI, Heart Attack)? & Binary/Categorical(0=No, 1=Yes) \\
\hline
medhx\_heart\_fail & DX & Have you ever been diagnosed with heart failure? & Binary/Categorical(0=No, 1=Yes) \\
\hline
medhx\_stroke & DX & Have you ever had a stroke? & Binary/Categorical(0=No, 1=Yes) \\
\hline
medhx\_ldl & DX & Have you ever been told that you have high cholesterol and/or triglycerides? & Binary/Categorical(0=No, 1=Yes) \\
\hline
medhx\_dm & DX & Have you been diagnosed with diabetes? & Binary/Categorical(0=No, 1=Yes) \\
\hline
medhx\_afib & DX & Have you ever been diagnosed with atrial fibrillation? & Binary/Categorical(0=No, 1=Yes) \\
\hline
loud\_snore & SYM & Frequency of loud snoring symptom from the Multivariable Apnea Prediction (MAP) Index & Ordinal/Quantitative: 0 = None/Don’t Know, 1 = Rarely (less than once a week), 2 = Sometimes (1-2 times a week), 3 = Frequently (3-4 times a week), 4 = Always (5-7 times a week) \\
\hline
gasping & SYM & Frequency of snorting or gasping symptom from the Multivariable Apnea Prediction (MAP) Index & Ordinal/Quantitative: 0 = None/Don’t Know, 1 = Rarely (less than once a week), 2 = Sometimes (1-2 times a week), 3 = Frequently (3-4 times a week), 4 = Always (5-7 times a week) \\
\hline
choke & SYM & Frequency of breathing stops or witnessed apneas from the Multivariable Apnea Prediction (MAP) Index & Ordinal/Quantitative: 0 = None/Don’t Know, 1 = Rarely (less than once a week), 2 = Sometimes (1-2 times a week), 3 = Frequently (3-4 times a week), 4 = Always (5-7 times a week) \\
\hline
map\_index & SYM & Multivariable Apnea Prediction (MAP) Index Score & Quantitative (0-4) \\
\hline
upper\_face\_depth & CF & Upper Face Depth & Quantitative \\
\hline
mid\_face\_depth & CF & Middle Face Depth & Quantitative \\
\hline
low\_face\_depth & CF & Lower Face Depth & Quantitative \\
\hline
max\_depth\_angle & CF & Maxillary Depth Angle & Quantitative \\
\hline
mand\_depth\_angle & CF & Mandibular Depth Angle & Quantitative \\
\hline
max\_mand\_rel\_angle & CF & Maxillary-Mandibular Relationship Angle & Quantitative \\
\hline
facial\_axis\_angle & CF & Facial Axis Angle & Quantitative \\
\hline
photo\_upper\_face\_ht & CF & Upper Face Height & Quantitative \\
\hline
nose\_ht & CF & Nose Height & Quantitative \\
\hline
photo\_low\_face\_ht & CF & Lower Face Height & Quantitative \\
\hline
mand\_length & CF & Mandibular Length & Quantitative \\
\hline
mand\_length\_dia & CF & Mandibular Length Diagonal & Quantitative \\
\hline
post\_mand\_ht & CF & Posterior Mandibular Height & Quantitative \\
\hline
mand\_plane\_angle & CF & Mandibular Plane Angle & Quantitative \\
\hline
photo\_mand\_angle & CF & Mandibular Angle & Quantitative \\
\hline
max\_mand\_box\_area & CF & Maxillary-Mandibular Box Area & Quantitative \\
\hline
cranial\_base\_tri\_area & CF & Cranial Base Triangular Area & Quantitative \\
\hline
max\_tri\_area & CF & Maxillary Triangular Area & Quantitative \\
\hline
mand\_tri\_area & CF & Mandibular Triangular Area & Quantitative \\
\hline
mid\_cran\_fossa\_vol & CF & Middle Cranial Fossa Volume & Quantitative \\
\hline
head\_posit\_angle & CF & Natural Head Position Angle & Quantitative \\
\hline
max\_volume & CF & Maxillary Volume & Quantitative \\
\hline
mand\_vol & CF & Mandibular Volume & Quantitative \\
\hline
max\_mand\_vol & CF & Maxillary-Mandibular Volume & Quantitative \\
\hline
ant\_mand\_ht & CF & Anterior Mandibular Height & Quantitative \\
\hline
man\_nas\_ang & CF & Mandibular-Nasion Angle & Quantitative \\
\hline
man\_sub\_ang & CF & Mandibular-Subnasion Angle & Quantitative \\
\hline
hea\_inc\_ang & CF & Head Base Inclination Angle & Quantitative \\
\hline
cer\_ang & CF & Cervicomental Angle & Quantitative \\
\hline
face\_width & CF & Face Width & Quantitative \\
\hline
mand\_width & CF & Mandibular Width & Quantitative \\
\hline
intercant\_width & CF & Intercanthal Width & Quantitative \\
\hline
biocular\_width & CF & Biocular Width & Quantitative \\
\hline
nose\_width & CF & Nose Width & Quantitative \\
\hline
lat\_fac\_ht & CF & Lateral Face Height & Quantitative \\
\hline
man\_wid\_len\_ang & CF & Mandibular Width to Length Angle & Quantitative \\
\hline
fac\_wid\_fac\_dep\_ang & CF & Face Width to Middle Face Depth Angle & Quantitative \\
\hline
fac\_wid\_low\_fac\_dep\_ang & CF & Face Width to Lower Face Depth Angle & Quantitative \\
\hline
eye\_width & CF & Eye Width ([biocular\_width - intercant\_width]/2) & Quantitative \\
\hline
face\_eye\_width & CF & Face Width :: Eye Width & Quantitative \\
\hline
face\_height\_width & CF & Face Height::Width & Quantitative \\
\hline
mand\_width\_length & CF & Mandibular Width::Length & Quantitative \\
\hline
face\_upper\_lower & CF & Upper Face Height::Lower Face Height & Quantitative \\
\hline
face\_height\_depth & CF & Face Height::Depth & Quantitative \\
\hline
cranial\_mand\_area & CF & Cranial Mandibular Area & Quantitative \\
\hline
face\_nose\_width & CF & Face Width :: Nose Width & Quantitative \\
\hline
open\_wid\_tim\_nph & IO & Mouth Opening Width (Tongue in Mouth, No Phonation) & Quantitative \\
\hline
ver\_mou\_ope & IO & Vertical Mouth Opening (Tongue in Mouth, No Phonation) & Quantitative \\
\hline
open\_area\_tim\_nph & IO & Mouth Opening Area (Tongue in Mouth, No Phonation) & Quantitative \\
\hline
tongue\_width\_tim\_nph & IO & Tongue Width (Tongue in Mouth, No Phonation) & Quantitative \\
\hline
open\_wid\_f\_te\_m & IO & Mouse Opening Width (Frontal, Tongue Extended) & Quantitative \\
\hline
tongue\_length\_f & IO & Tongue Length (Frontal, Tongue Extended) & Quantitative \\
\hline
tongue\_width\_f\_te\_m & IO & Tongue Width (Frontal, Tonuge Extended) & Quantitative \\
\hline
ton\_are & IO & Tongue Area (Frontal, Tongue Extended) & Quantitative \\
\hline
tongue\_length\_p\_te\_m & IO & Tongue Length (Profile, Tongue Extended) & Quantitative \\
\hline
tongue\_area\_p\_te\_m & IO & Tongue Area (Profile, Tongue Extended) & Quantitative \\
\hline
tongue\_thickness\_p\_te\_m & IO & Tongue Thickness (Profile, Tongue Extended) & Quantitative \\
\hline
tongue\_curvature\_p\_te\_m & IO & Tongue Curvature (Profile, Tongue Extended) & Quantitative \\
\hline
mal\_score\_ftem\_c1 & IO & Mallampati Score Class I (FTEM) & Binary/Categorical (0=No, 1=Yes) \\
\hline
mal\_score\_ftem\_c2 & IO & Mallampati Score Class II (FTEM) & Binary/Categorical (0=No, 1=Yes) \\
\hline
mal\_score\_ftem\_c3 & IO & Mallampati Score Class III (FTEM) & Binary/Categorical (0=No, 1=Yes) \\
\hline
mal\_score\_ftem\_c4 & IO & Mallampati Score Class IV (FTEM) & Binary/Categorical (0=No, 1=Yes) \\
\hline
mal\_score\_timnph\_c1 & IO & Mallampati Score Class I (TIMNPH) & Binary/Categorical (0=No, 1=Yes) \\
\hline
mal\_score\_timnph\_c2 & IO & Mallampati Score Class II (TIMNPH) & Binary/Categorical (0=No, 1=Yes) \\
\hline
mal\_score\_timnph\_c3 & IO & Mallampati Score Class III (TIMNPH) & Binary/Categorical (0=No, 1=Yes) \\
\hline
mal\_score\_timnph\_c4 & IO & Mallampati Score Class IV (TIMNPH) & Binary/Categorical (0=No, 1=Yes) \\
\hline
\caption*{\textbf{Table S4: Variable Key - SAGIC Obstructive Sleep Apnea Data}}
\end{longtable}
\normalsize
\endgroup
\addcontentsline{toc}{subsubsection}{Table S4: Variable Key - SAGIC Obstructive Sleep Apnea Data}

\newpage
\begin{table}
\centering
\scriptsize
\begin{tabular}{| l | l | l | l |}
\hline
\rowcolor{lightgray}
\textbf{Dataset ID} & \textbf{Feature Set(s) Included} & \textbf{AHI cutoff} & \textbf{Full Dataset Name} \\
\hline
A & DEM & AHI$\geq$15 events/hour & A\_GE15\_DEM \\
\hline
B & DX & AHI$\geq$15 events/hour & B\_GE15\_DX \\
\hline
C & SYM & AHI$\geq$15 events/hour & C\_GE15\_SYM \\
\hline
D & CF & AHI$\geq$15 events/hour & D\_GE15\_CF \\
\hline
E & IO & AHI$\geq$15 events/hour & E\_GE15\_IO \\
\hline
F & DEM+DX & AHI$\geq$15 events/hour & F\_GE15\_DEMDX \\
\hline
G & DEM+DX+SYM & AHI$\geq$15 events/hour & G\_GE15\_DEMDXSYM \\
\hline
H & DEM+DX+SYM+CF & AHI$\geq$15 events/hour & H\_GE15\_DEMDXSYMCF \\
\hline
I & DEM+DX+SYM+IO & AHI$\geq$15 events/hour & I\_GE15\_DEMDXSYMIO \\
\hline
J & DEM+DX+SYM+CF+IO & AHI$\geq$15 events/hour & J\_GE15\_DEMDXSYMCFIO \\
\hline
\rowcolor{gray!20}
K & DEM & AHI$\geq$5 events/hour & K\_GE5\_DEM \\
\hline
\rowcolor{gray!20}
L & DX & AHI$\geq$5 events/hour & L\_GE5\_DX \\
\hline
\rowcolor{gray!20}
M & SYM & AHI$\geq$5 events/hour & M\_GE5\_SYM \\
\hline
\rowcolor{gray!20}
N & CF & AHI$\geq$5 events/hour & N\_GE5\_CF \\
\hline
\rowcolor{gray!20}
O & IO & AHI$\geq$5 events/hour & O\_GE5\_IO \\
\hline
\rowcolor{gray!20}
P & DEM+DX & AHI$\geq$5 events/hour & P\_GE5\_DEMDX \\
\hline
\rowcolor{gray!20}
Q & DEM+DX+SYM & AHI$\geq$5 events/hour & Q\_GE5\_DEMDXSYM \\
\hline
\rowcolor{gray!20}
R & DEM+DX+SYM+CF & AHI$\geq$5 events/hour & R\_GE5\_DEMDXSYMCF \\
\hline
\rowcolor{gray!20}
S & DEM+DX+SYM+IO & AHI$\geq$5 events/hour & S\_GE5\_DEMDXSYMIO \\
\hline
\rowcolor{gray!20}
T & DEM+DX+SYM+CF+IO & AHI$\geq$5 events/hour & T\_GE5\_DEMDXSYMCFIO \\
\hline

\end{tabular}
\normalsize
\caption*{\textbf{Table S5: Constructed SAGIC Datasets for Analysis}}
\end{table}
\addcontentsline{toc}{subsubsection}{Table S5: Constructed Datasets for Analysis}

\normalsize

\addcontentsline{toc}{subsubsection}{S.2.3.1: Development and Replication Datasets}
\subsubsection*{S.2.3.1: Development and Replication Datasets}
Included in the SAGIC dataset instance identifiers is the site/location where the data was collected. 
 
Below we provide the unique study/data collection site IDs:
\begin{itemize}  \itemsep0em 
\item BER = Germany (Berlin)
\item BRA = Brazil
\item ICE = Iceland
\item OSU = USA (Ohio State)
\item PEN = USA (UPenn)
\item WES = Australia (Perth)
\item SYD = Australia (Sydney)
\item TAI = Taiwan
\item BEI = China (Beijing)
\end{itemize}

Once the 20 datasets listed in Table S4 were constructed we applied a Jupyter Notebook ‘OSA\_Replication\_Data\_-Splitter.ipynb’ (available upon request), which split each of these datasets into respective development and replication sets. All data partitioning in this study (both here and during internal STREAMLINE CV partitioning) randomly assigned instances to partitions, i.e. instances from a given site are not kept together as a hold out evaluation set, in order to avoid possible underlying site-specific collection differences/biases. For datasets with the same binary OSA class definition (i.e. AHI$\geq$15 or AHI$\geq$5), the exact same set of instances comprise the development and replication sets to ensure within-group comparability. However, between these two dataset groups, different instances are included in development and replication sets because different instances are labeled as having either a 0 (no OSA) or 1 (OSA) outcome with a subsequently different degree of ‘class balance’. 

Specifically, AHI$\geq$15 datasets all have \~45.3\% OSA (i.e. class 1), and AHI$\geq$5 datasets all have \~ 68.9\% OSA (i.e. class 1). Thus while AHI$\geq$15 datasets are relatively ‘balanced’, AHI$\geq$5 datasets have a more significant degree of imbalance which is taken into account during model evaluation. After this initial partitioning, AHI$\geq$15 development datasets have 2178 instances and replication datasets have 933 instances, while AHI$\geq$5 development datasets have 2179 instances and replication datasets have 932 instances. 

\addcontentsline{toc}{subsubsection}{S.2.3.2: OSA Data Analyses with STREAMLINE}
\subsubsection*{S.2.3.2: OSA Data Analyses with STREAMLINE}

Refer to ‘sagic\_exp\_STREAMLINE\_Report.pdf’ (Page 1) for all STREAMLINE run parameter settings used in this analysis. We ran this complete analysis across all 20 development datasets using STREAMLINE phases 1-7, and then ran separate phase 8 replication analyses for each of the 20 respective replication datasets. 

 Subsequently, we re-ran phase 7 ‘dataset comparison’ (on it’s own, without retraining models) on different subsets of the 20 datasets to generate simpler and more targeted comparisons between (1) AHI$\geq$15 vs. AHI$\geq$5 datasets, (2) all datasets with just one feature group included, and (3) datasets with just one feature group included for AHI$\geq$15 vs. AHI$\geq$5 datasets, separately. 

\addcontentsline{toc}{subsection}{S.2.4: Study-Wide STREAMLINE Run Parameters}
\subsection*{S.2.4: Study-Wide STREAMLINE Run Parameters}

While STREAMLINE run parameters for each ‘experiment’ are documented on the first page of their respective PDF reports (in supplemental materials), here we document the relevant STREAMLINE run parameters that were consistent across all experiments and analyses. 

\begin{itemize}  \itemsep0em 
\item CV Partitions: 10
\item Partition Method: Stratified
\item Statistical Significance Cutoff: 0.05
\item Random Seed: 42
\item Use Mutual Information: True
\item Use MultiSURF: True
\item Use TURF: False
\item Max Features to Keep: 2000
\item Top Features to Display: 40
\item Export Feature Importance Plot: True
\item Engineering Missingness Cutoff: 0.5
\item Cleaning Missingness Cutoff: 0.5
\item Correlation Removal Threshold: 1.0
\item List of Exploratory Analysis Ran: ['Describe', 'Univariate Analysis', 'Feature Correlation']
\item List of Exploratory Plots Saved: ['Describe', 'Univariate Analysis', 'Feature Correlation']
\item Use Data Scaling: True
\item Use Data Imputation: True
\item Use Multivariate Imputation: True
\item Overwrite CV Datasets: True
\item Artificial Neural Network: True
\item Category Gradient Boosting: True
\item Decision Tree: True
\item Elastic Net: True
\item ExSTraCS: True
\item Extreme Gradient Boosting: True
\item Genetic Programming: True
\item Gradient Boosting: True
\item K-Nearest Neighbors: True
\item Light Gradient Boosting: True
\item Logistic Regression: True
\item Naive Bayes: True
\item Random Forest: True
\item Support Vector Machine: True
\item XCS: False
\item eLCS: False
\item Primary Metric: balanced\_accuracy
\item Uniform Feature Importance Estimation (Models): True
\item Hyperparameter Sweep Number of Trials: 200
\item Hyperparameter Timeout: 900
\item Stats and Figure Settings:
\item Export ROC Plot: True
\item Export PRC Plot: True
\item Export Metric Boxplots: True
\item Export Feature Importance Boxplots: True
\item Metric Weighting Composite FI Plots: balanced\_accuracy
\item Top Model Features To Display: 40
\end{itemize}

\addcontentsline{toc}{section}{S.3: Results}
\section*{S.3 Results}
Note, that all results figures below and in main text were automatically generated by STREAMLINE (some were minorly edited in Adobe Illustrator to move legends or create multi-panel figures. STREAMLINE is also set up with supporting code in the form of ‘Useful Notebooks’ allowing users to re-generate plots to their own formatting specifications. Included in these supplementary materials are PDF reports for each STREAMLINE ‘experiment’. However, detailed results files, models, and other automatically saved STREAMLINE output files can also be made available upon request (\href{mailto:ryan.urbanowiz@cshs.org}{ryan.urbanowiz@cshs.org}).

\addcontentsline{toc}{subsection}{S.3.1: HCC Benchmark Results}
\subsection*{S.3.1: HCC Benchmark Results}

\addcontentsline{toc}{subsection}{S.3.1.1: Simulated ‘Custom’ HCC Validates Automated Data Processing}
\subsubsection*{S.3.1.1: Simulated ‘Custom’ HCC Validates Automated Data Processing }

First, examination of the hcc\_data\_custom data processing results validated the efficacy of the STREAMLINE phase 1 data processing (i.e. cleaning and feature engineering. Specifically, all simulated features and instances designed to be cleaned or engineered were done so correctly. Referring to ‘hcc\_exp\_STREAMLINE\_Report.pdf’ (Page 3) note that hcc\_data does not have any changes to feature or instance count as the result of cleaning or engineering. However, on (Page 4) we observe the following correct data processing of the simulated hcc\_data\_custom dataset:

\begin{itemize}  \itemsep0em 
\item \textbf{C1:} 2 instances missing an outcome label are removed (note that class counts don’t change because those class labels were missing to begin with)
\item \textbf{E1:} 3 categorical features are added to the dataset representing newly engineered missingness features for the 3 features in this dataset with a missingness > 0.5.
\item \textbf{C2:} 6 features are removed (4 categorical and 2 quantitative) including the original 3 features with high missingness and 3 features with invariant values.
\item \textbf{C3:} 2 instances are removed with missingness > 0.5
\item \textbf{E2:} One-hot-encoding replaces 4 non-binary categorical features with binary features for each feature’s possible categories. Specifically 14 features are added and 4 are removed yielding a net gain of 10 binary categorical features. 
\item \textbf{C4:} 2 pairs of features in this data are perfectly correlated. One of each pair is removed for a net reduction of two quantitative features
\end{itemize}

Now referring to ‘hcc\_exp\_STREAMLINE\_Replication\_Report.pdf’ (Page 2) which summarizes the replication evaluation of hcc\_data\_custom using hcc\_data\_custom\_rep, we correctly observed that C1 removes 3 instances missing the class label, and R1 processes the rest of this data in the same way as for hcc\_data\_custom, yielding the same number of features as well as confirming Phase 8’s ability to handle as-of-yet unobserved categorical values occurring in the replication data (i.e. replacing them with NA for binary categorical features, and adding 0 values for all one-hot-encoded features for that original feature. 

\addcontentsline{toc}{subsubsection}{S.3.1.2: Original HCC Benchmark Results}
\subsubsection*{S.3.1.2: Original HCC Benchmark Results}

Refer to ‘hcc\_exp\_STREAMLINE\_Report.pdf’ (Pages 2,3,5,6,\&7) for EDA and modeling results for the hcc\_data dataset. We review major results below.

\textbf{Figure S1} includes ROC and PRC plots comparing mean algorithm performance. XGB yielded the best ROC-AUC of 0.800 (0.127). In this dataset there is a reasonable degree of class imbalance (63 deceased, and 102 survived in the original dataset) so it is best here to focus instead balanced accuracy and on PRC-AUC results. EN yielded the best balanced accuracy of 0.728 (0.067) and SVM yielded the best PRC-AUC of 0.718 (0.107). \textbf{Figure S2} gives the PRC plot for SVM comparing each of 10-fold CV models trained, and illustrating reasonably high variance in performance across partitions. This is reasonably expected given the small sample size of this dataset and highlights the importance of evaluating models across CV partitions in order to evaluate the impact of partitioning-based sample bias. \textbf{Figure S3} gives permutation-based model feature importance estimates for each algorithm (averaged across all 10-CV models for each algorithm). 

\begin{figure} [H]
    \centering
    \includegraphics[width=\textwidth]{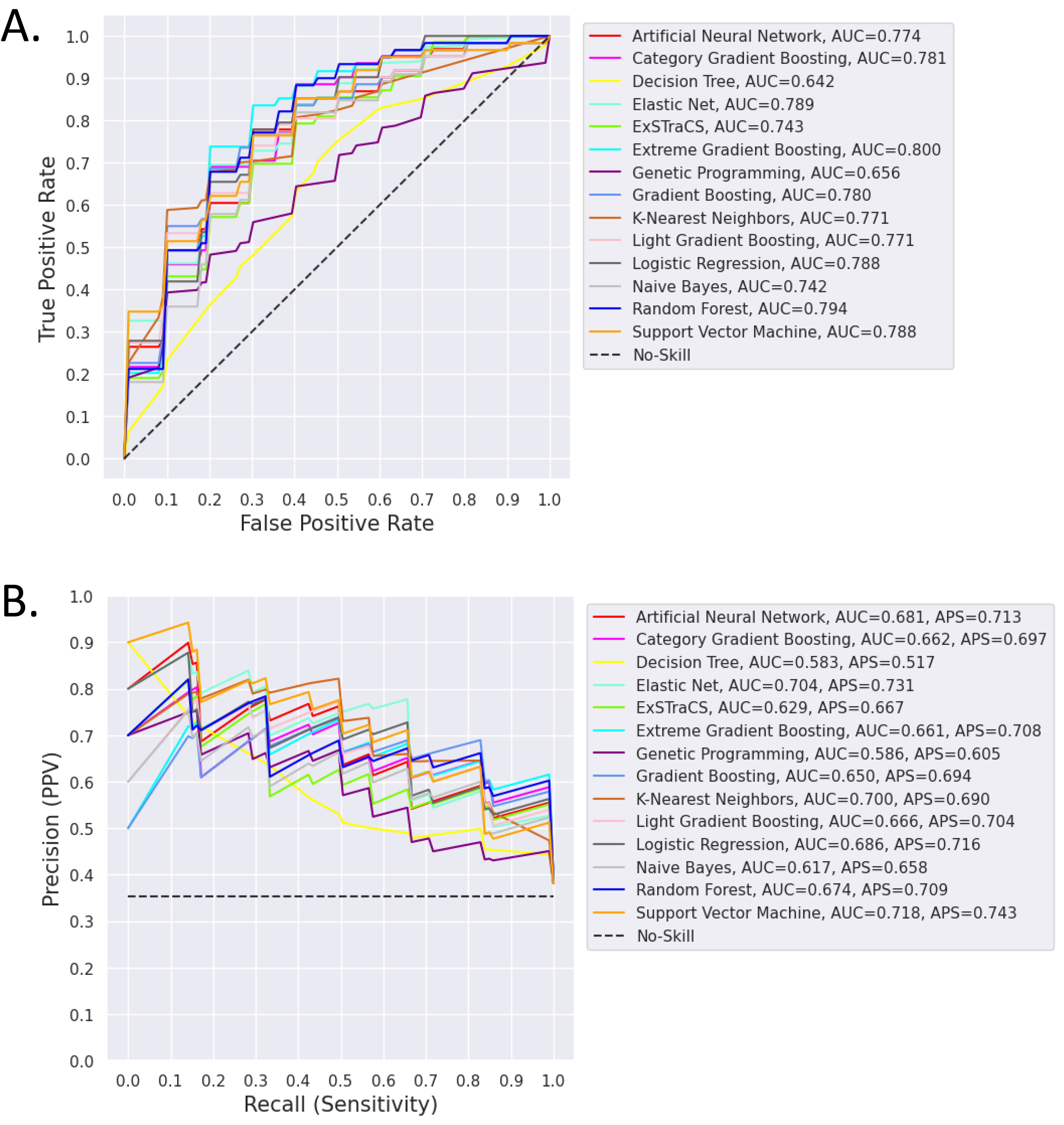}
    \caption*{\textbf{Figure S1: Original HCC Data - Comparing Mean Algorithm Performance.} (A) ROC plot comparing algorithm performance. (B) PRC plot comparing algorithm performance. For both, each line represents mean algorithm performance across 10 CV models. Area under the curve (AUC) is provided for both, and average precision (APS) is included for PRC.}
\end{figure}
\addcontentsline{toc}{subsubsection}{Figure S1: Original HCC Data - Comparing Mean Algorithm Performance.}

\begin{figure} [H]
    \centering
    \includegraphics[width=\textwidth]{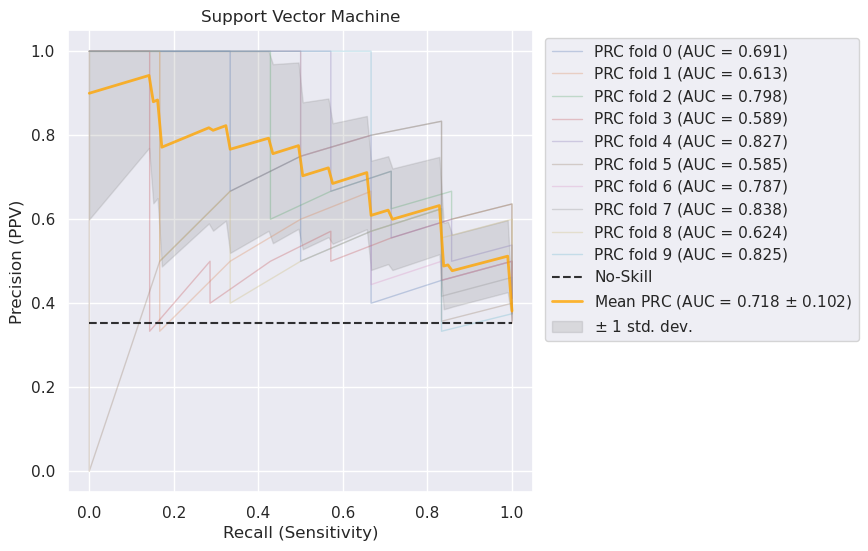}
    \caption*{\textbf{Figure S2: Original HCC Data - PRC Plot of SVM performance across individual CV Models. }}
\end{figure}
\addcontentsline{toc}{subsubsection}{Figure S2: Original HCC Data - PRC Plot of SVM performance across individual CV Models.}

\begin{figure} [H]
    \centering
    \includegraphics[width=\textwidth]{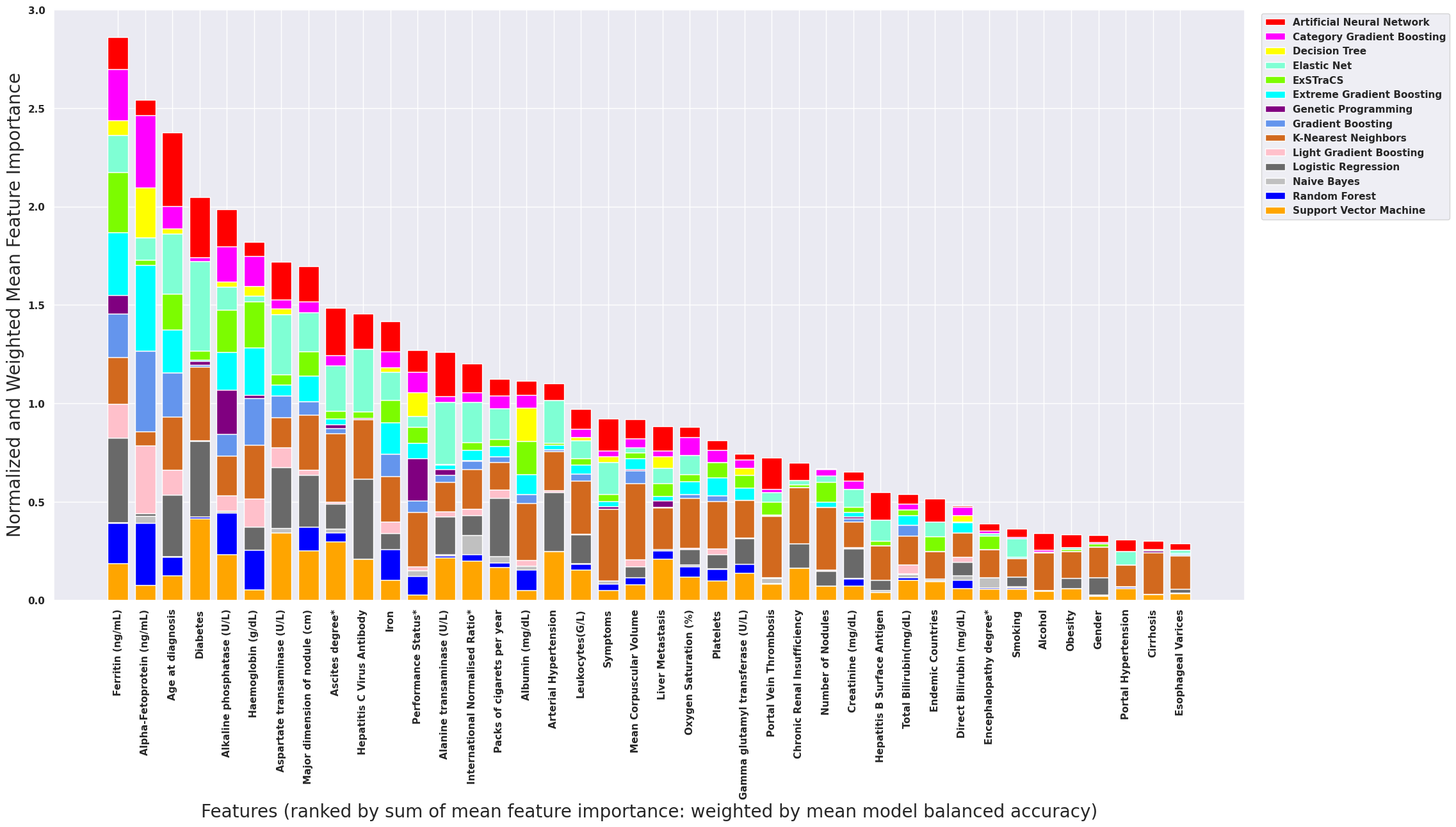}
    \caption*{\textbf{Figure S3: Original HCC Data – Mean Model Feature Importance Estimates Across all Algorithms.} This composite feature importance plot illustrates normalized and (balanced accuracy)-weighted mean FI scores summed across all algorithms. Only the top 40 features are displayed.}
\end{figure}
\addcontentsline{toc}{subsubsection}{Figure S3: Original HCC Data – Mean Model Feature Importance Estimates Across all Algorithms.}

\textbf{Table S6} compares the performance of our previous STREAMLINE HCC benchmarking (v0.2.5) [1] with the current (v0.3.4). Mean model performances were similar, if not slightly higher, however the performance metric standard deviation was also significantly larger, suggesting a larger degree of random partitioning sample bias than in in the earlier benchmarking run. Replication data was not available for the HCC benchmarking, and the dataset was too small to reasonably hold out further data for replication, however these results suggest that this new release of STREAMLINE is performing as expected, yielding similar top performance metrics, and similar top model feature importance estimates. 

\begin{table} [H]
\centering
\scalebox{0.85}{
\begin{tabular}{| l | l | l | l | l |}
\hline
\rowcolor{light-gray}
 & \multicolumn{2}{|c|}{\textbf{v0.2.5}} & \multicolumn{2}{|c|}{\textbf{v0.3.4}} \\
\hline
 \rowcolor{light-gray}
\textbf{Metric} & \textbf{Best Algorithm} & \textbf{Benchmark Testing Evaluation} & \textbf{Best Algorithm} & \textbf{Benchmark Testing Evaluation} \\
\hline
\textbf{ROC AUC} & SVM & 0.777 (0.049) & XGB & 0.800 (0.127) \\
\hline
\textbf{PRC AUC} & CGB & 0.635 (0.076) & SVM & 0.718 (0.107) \\
\hline
\textbf{Balanced Accuracy} & RF & 0.724 (0.013) & EN & 0.728 (0.067) \\
\hline
\textbf{F1 Score} & RF & 0.662 (0.008) & EN & 0.667 (0.094) \\
\hline
\end{tabular}}
\caption*{\textbf{Table S6: Best Mean Algorithm Performance Between STEAMLINE versions on HCC.} Each value represents the mean metric value (with standard deviation) across 10-fold CV trained models for that algorithm. Here the ‘best’ model is selected for each individual metric for comparison.}
\end{table}
\addcontentsline{toc}{subsubsection}{Table S6: Best Mean Algorithm Performance Between STEAMLINE versions on HCC.}

Notably the results highlight the fact that there is no clear ‘winning’ algorithm that performs best across all evaluation metrics or CV partitions, and on this particular dataset most algorithms perform similarly well with respect to PRC-AUC, with the exception of DT and GP. 

\addcontentsline{toc}{subsection}{S.3.2: GAMETES Benchmark Results}
\subsection*{S.3.2: GAMETES Benchmark Results}
Refer to ‘gametes\_exp\_STREAMLINE\_Report.pdf’ for EDA and modeling results for all 6 of the simulated GAMETES datasets. We review major results below. These datasets are balanced, so we can confidently focus on ROC-AUC as the key performance metric. \textbf{Table S7} compares the performance of our previous STREAMLINE GAMETES benchmarking (v0.2.5) [1] with the current (v0.3.4). As expected, mean model ROC-AUCs were very similar between the two analyses.

\begin{table}
\centering
\scalebox{0.85}{
\begin{tabular}{| l | l | l | l | l |}
\hline
 \rowcolor{light-gray}
  & \multicolumn{2}{|c|}{\textbf{v0.2.5}} & \multicolumn{2}{|c|}{\textbf{v0.3.4}} \\
\hline
 \rowcolor{light-gray}
\textbf{GAMETES Dataset} & \textbf{Best Algorithm} & \textbf{Benchmark Testing Evaluation} & \textbf{Best Algorithm} & \textbf{Benchmark Testing Evaluation} \\
\hline
\textbf{A - univariate} & RF & 0.842 (0.022) & XGB & 0.847 (0.035) \\
\hline
\textbf{B – uni-4add} & XGB & 0.983 (0.009) & XGB & 0.983 (0.007) \\
\hline
\textbf{C – uni-4het} & CGB & 0.672 (0.040) & CGB & 0.671 (0.042) \\
\hline
\textbf{D - 2way-epistasis} & ExSTraCS & 0.854 (0.028) & ExSTraCS & 0.852 (0.025) \\
\hline
\textbf{E - 2way epi+2het} & ExSTraCS & 0.740 (0.040) & ExSTraCS & 0.746 (0.026) \\
\hline
\textbf{F – 3way epistasis} & ExSTraCS & 0.564 (0.114) & SVM & 0.511 (0.042) \\
\hline
\end{tabular}}
\caption*{\textbf{Table S7: Best Mean Algorithm ROC-AUC Between STEAMLINE versions on GAMETES Datasets.} Each value represents the mean metric value (with standard deviation) across 10-fold CV trained models for that algorithm. Here the ‘best’ model is selected for each individual metric for comparison.}
\end{table}
\addcontentsline{toc}{subsubsection}{Table S7: Best Mean Algorithm ROC-AUC Between STEAMLINE versions on GAMETES Datasets.}

\textbf{Figure S4} similarly compares mean model performance across all algorithms and datasets using STREAMLINE v0.2.5 vs v0.3.4. Similar to what was observed for v0.2.5 [1], dataset (A), i.e. univariate, all algorithms performed similarly well, with NB (instead of KNN) performing the least well. For dataset (B), i.e. additive univariate, all algorithms performed similarly well, with SVM (instead of NB) performing the least well. For dataset (C), i.e. heterogeneous univariate, CGB, XGB, EN, LGB, RF, GB, and ExSTraCS performed best, while NB, KNN, and GP performed the least well. For dataset (D), i.e. pure 2-way epistasis, we observed the first dramatic performance differences with ExSTraCS, CGB, LGB, XGB, and GB performing similarly at the top, while NB, LR, and DT entirely failed to detect the 2-feature interaction. For dataset (E), i.e. heterogeneous pure 2-way epistasis, ExSTraCS and CGB stood out as top performers, while again NB, LR, and DT entirely failed. Lastly, for dataset (F), i.e. pure 3-way epistasis, all algorithms struggled with
this noisy complex association. Notably in this run even ExSTraCS failed to detect the correct noisy complex association. As would be expected, no algorithm stands out as being ideal across all data scenarios.

\begin{figure}
    \centering
    \includegraphics[width=0.8\textwidth]{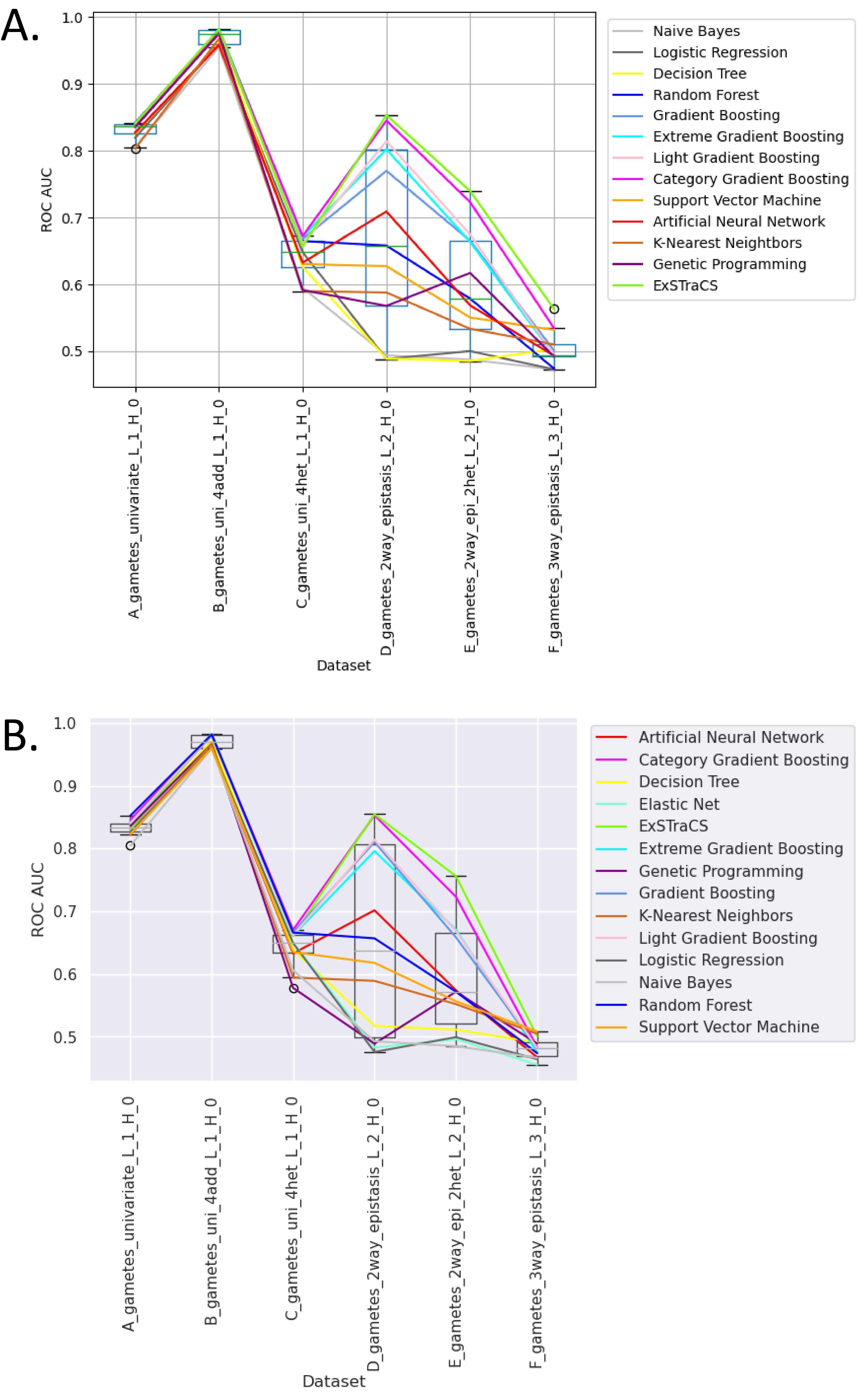}
    \caption*{\textbf{Figure S4: GAMETES Dataset Performance Comparison.} (A). STREAMLINE v0.2.5 analysis (B) STREAMLINE v0.3.4 analysis. Each algorithm line represents mean ROC-AUC across 10-fold CV models. Each plot’s formatting is different given that it was automatically generated with different versions of the AutoML, and the newest version has added the EN algorithm. }
\end{figure}
\addcontentsline{toc}{subsubsection}{Figure S4: GAMETES Dataset Performance Comparison.}

Also, as before, model feature importance estimates correctly prioritized the ground-truth important features in these simulations. For example, \textbf{Figure S5} illustrates the correct identification of features M0P0 and M0P1 (involved in a pure, epistatic interaction) out of a total 100 features in the original dataset. Additionally, \textbf{Figure S6} illustrates the correct identification of features M0P0, M0P1, M1P0, and M1P1 (which represent two independently simulated 2-way epistatic interactions, i.e. M0P0xM0P1, M1P0xP1P1, that are each predictive in half of the simulated instances of the original dataset). We can observe here how different algorithms succeed or fail in their ability to handle both interactions and heterogeneous associations. 

\begin{figure}
    \centering
    \includegraphics[width=\textwidth]{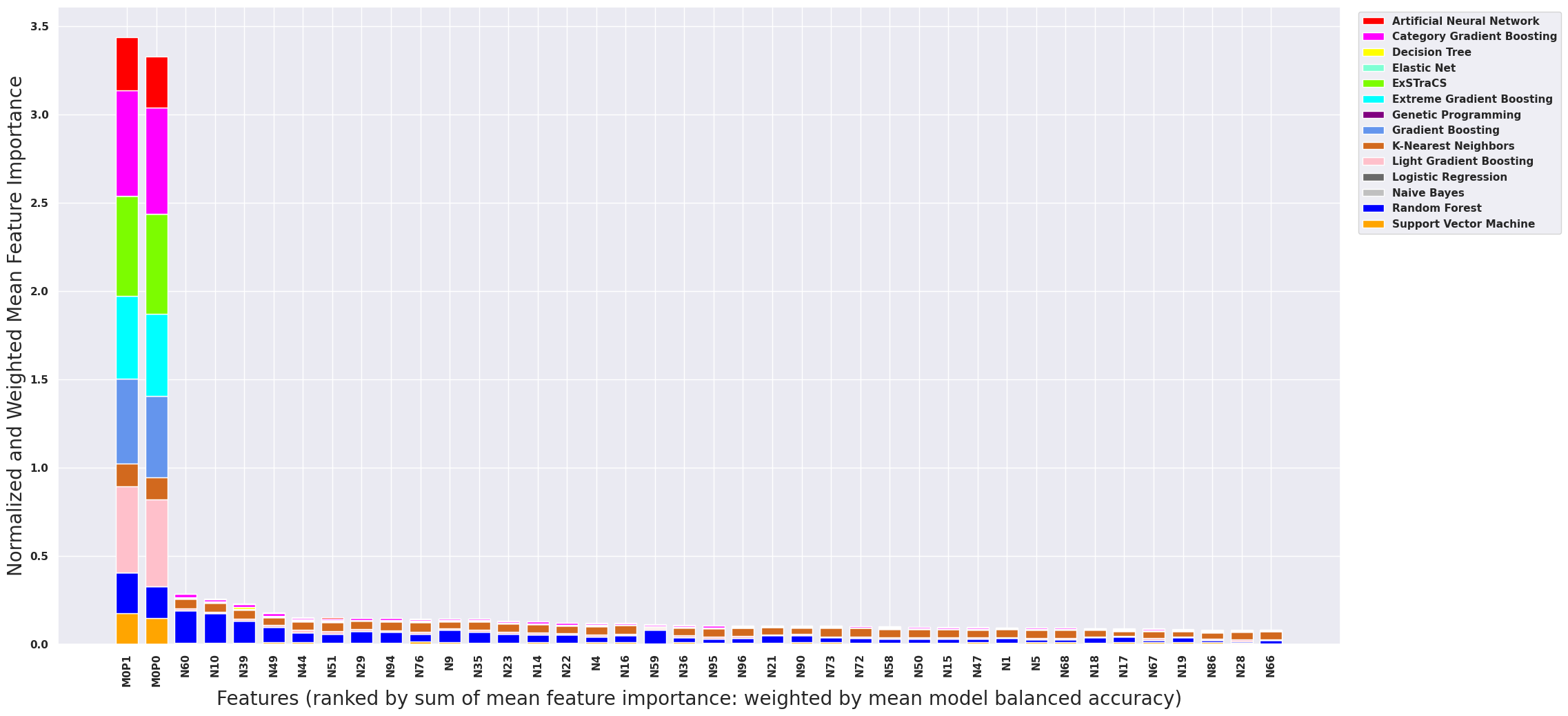}
    \caption*{\textbf{Figure S5: GAMETES Dataset D (2-Way Epistasis) Model Feature Importance.} This composite feature importance plot illustrates normalized and (balanced accuracy)-weighted mean FI scores summed across all algorithms. Only the top 40 features are displayed.}
\end{figure}
\addcontentsline{toc}{subsubsection}{Figure S5: GAMETES Dataset D (2-Way Epistasis) Model Feature Importance.}

\begin{figure}
    \centering
    \includegraphics[width=\textwidth]{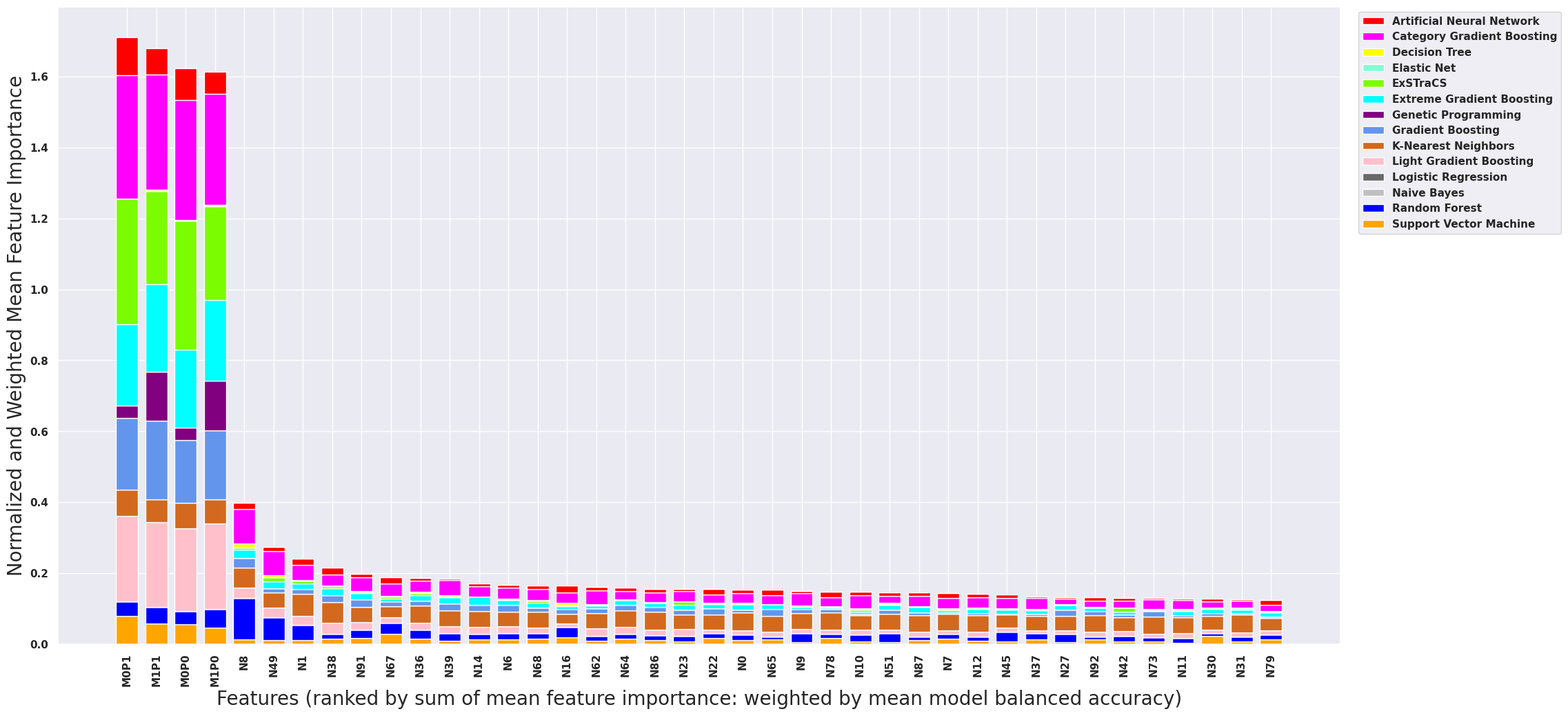}
    \caption*{\textbf{Figure S6: GAMETES Dataset E (2-Way Epi+2 Het) Model Feature Importance.} This composite feature importance plot illustrates normalized and (balanced accuracy)-weighted mean FI scores summed across all algorithms. Only the top 40 features are displayed. In this dataset only M0P0, M0P1, M1P0, and M1P1 are predictive.}
\end{figure}
\addcontentsline{toc}{subsubsection}{Figure S6: GAMETES Dataset E (2-Way Epi+2 Het) Model Feature Importance.}

In summary, these results suggest that this new release of STREAMLINE is performing as expected, yielding similar top performance metrics, and similar top model feature importance estimates. 

\addcontentsline{toc}{subsection}{S.3.3: Multiplexer Benchmark Results}
\subsection*{S.3.3: Multiplexer Benchmark Results}
Refer to ‘multiplexer\_exp\_STREAMLINE\_Report.pdf’ for EDA and modeling results for all 6 of the benchmark multiplexer datasets. We review major results below. These datasets are also balanced, so we can confidently focus on ROC-AUC as the key performance metric. \textbf{Table S8} compares the performance of our previous STREAMLINE multiplexer benchmarking (v0.2.5) [1] with the current (v0.3.4). As expected, mean model ROC-AUCs were very similar between the two analyses. We attribute minor performance differences to any pipeline variations based on the use of ‘random’, i.e. stochasticity, including the partitioning of training/testing CV sets, which will not be exactly the same in running v0.2.5 and v0.3.4. 

\begin{table} [H]
\centering
\scalebox{0.85}{
\begin{tabular}{| l | l | l | l | l |}
\hline
 \rowcolor{light-gray}
  & \multicolumn{2}{|c|}{\textbf{v0.2.5}} & \multicolumn{2}{|c|}{\textbf{v0.3.4}} \\
\hline
 \rowcolor{light-gray}
\textbf{Multiplexer Dataset} & \textbf{Best Algorithm} & \textbf{Benchmark Testing Evaluation} & \textbf{Best Algorithm} & \textbf{Benchmark Testing Evaluation} \\
\hline
\textbf{A – 6-bit} & 10-way tie & 1.000 (0.000) & 11-way tie & 1.000 (0.000) \\
\hline
\textbf{B – 11-bit} & 8-way tie & 1.000 (0.000) & RF & 0.982 (0.013) \\
\hline
\textbf{C – 20-bit} & 5-way tie & 1.000 (0.000) & ExSTraCS & 0.968 (0.026) \\
\hline
\textbf{D – 37-bit} & ExSTraCS & 1.000 (0.000) & ExSTraCS & 0.938 (0.021) \\
\hline
\textbf{E – 70-bit} & ExSTraCS & 0.992 (0.016) & ExSTraCS & 0.908 (0.051) \\
\hline
\textbf{F – 135-bit} & LGB & 0.559 (0.016) & SVM & 0.556 (0.009) \\
\hline
\end{tabular}}
\caption*{\textbf{Table S8: Best Mean Algorithm ROC-AUC Between STEAMLINE versions on Multiplexer Datasets.} Each value represents the mean metric value (with standard deviation) across 10-fold CV trained models for that algorithm. Here the ‘best’ model is selected for each individual metric for comparison.}
\end{table}
\addcontentsline{toc}{subsubsection}{Table S8: Best Mean Algorithm ROC-AUC Between STEAMLINE versions on Multiplexer Datasets.}

\textbf{Figure S7} similarly compares mean model performance across all algorithms and datasets using STREAMLINE v0.2.5 vs v0.3.4. Similar to what was observed in v0.2.5 \cite{urbanowicz_streamline_2023}, for the 6-bit dataset, most algorithms solved the problem similarly well with the exception of EN, LR, and NB which are not expected to handle complex interactions in data. These algorithms performed similarly poorly on all other (more complicated) multiplexer datasets. For the 11-bit dataset, GP joined the list of poor performing algorithms, and unlike in v0.2.5, no algorithm solved this dataset with 100\% testing accuracy (likely due to variations in CV partitioning), with RF performing slightly best. For the 20-bit dataset, CGB, ExSTraCS, XGB, GB, LGB, and RF performed best, with DT and K-NN significantly dropping in performance. For the 37-bit dataset, ExSTraCS and LGB performed best with XGB and GB seeing some more significant drop in performance. For the 70-bit dataset, ExSTraCS performed far better than any other algorithm (see \textbf{Figure S8}). 

\begin{figure}
    \centering
    \includegraphics[width=0.8\textwidth]{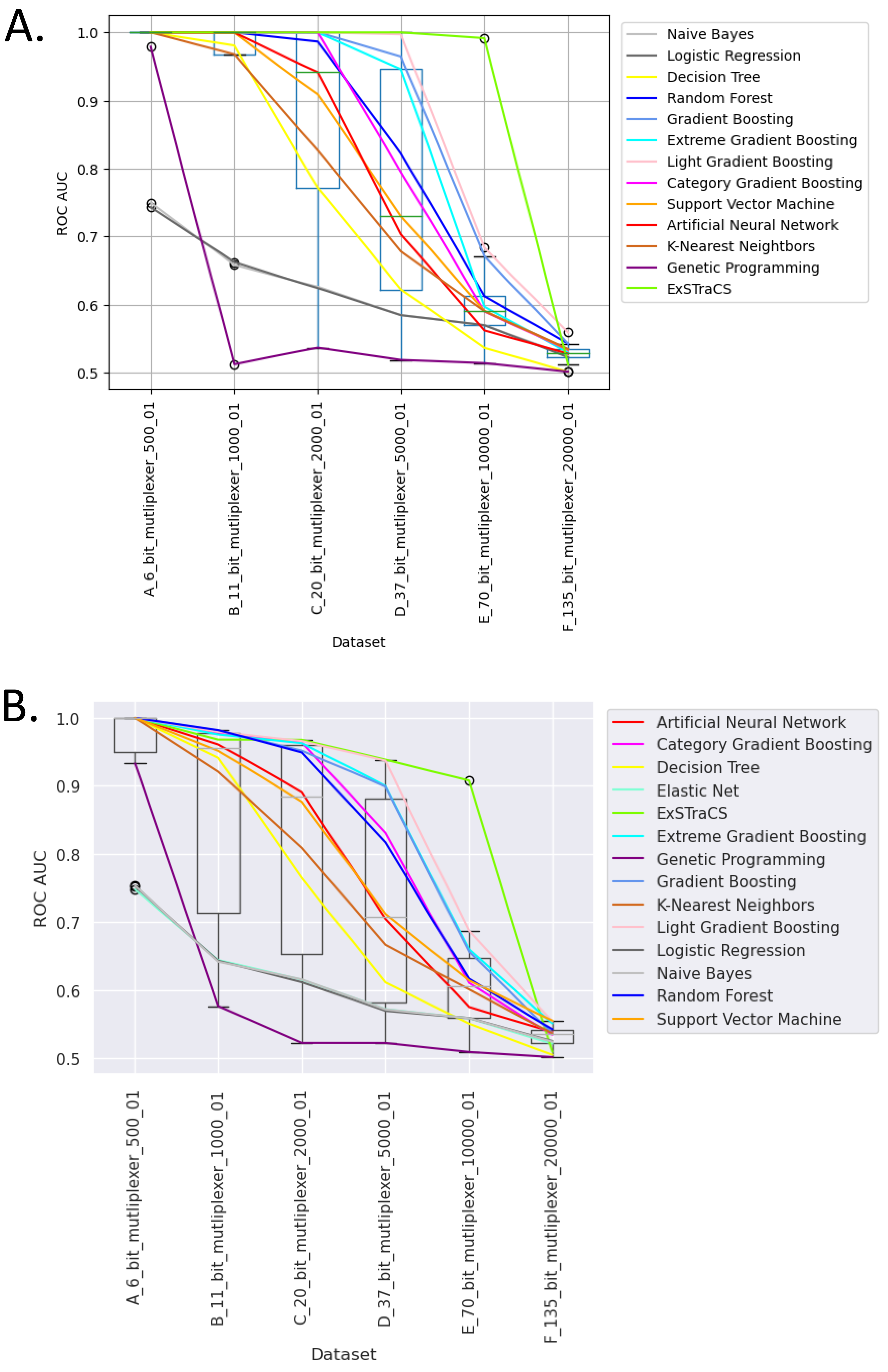}
    \caption*{\textbf{Figure S7: Multiplexer Dataset Performance Comparison.} (A). STREAMLINE 0.2.5 analysis (B) STREAMLINE 0.3.4 analysis. Each algorithm line represents mean ROC-AUC across 10-fold CV models. Each plot’s formatting is different given that it was automatically generated with different versions of the AutoML, and the newest version has added the EN algorithm.}
\end{figure}
\addcontentsline{toc}{subsubsection}{Figure S7: Multiplexer Dataset Performance Comparison.}

\begin{figure}
    \centering
    \includegraphics[width=\textwidth]{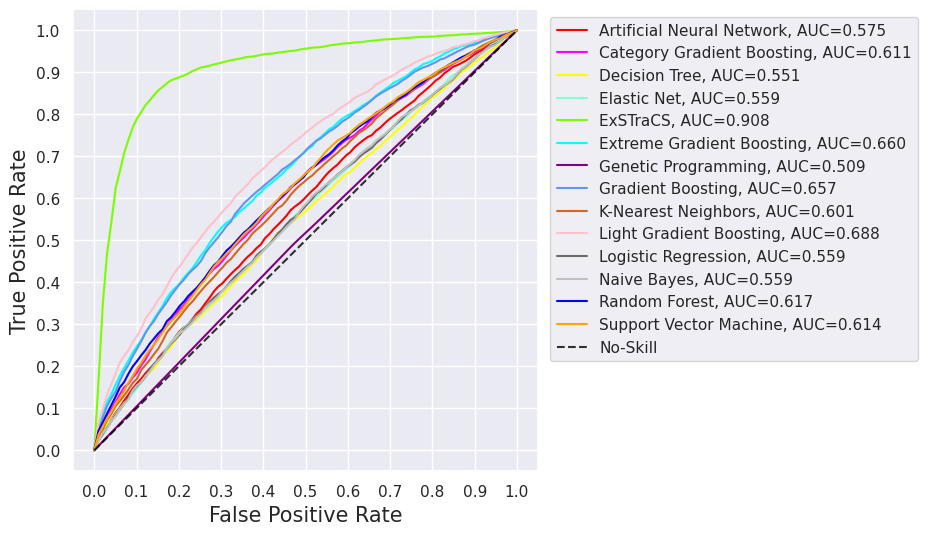}
    \caption*{\textbf{Figure S8: 70-bit Multiplexer ROC Plot Comparing Mean Algorithm Performance.} A composite feature importance plot illustrating the mean, normalized and balanced-accuracy weighted permutation feature importance estimates for each algorithm. Each line represents mean algorithm performance across 10 CV models. Area under the curve (AUC) is provided.}
\end{figure}
\addcontentsline{toc}{subsubsection}{Figure S8: 70-bit Multiplexer ROC Plot Comparing Mean Algorithm Performance.}

\textbf{Figure S9} illustrates how ExSTraCS correctly prioritizes ‘address’ bit features in model testing. And lastly, for the 135-bit dataset, no algorithm performed well, but SVM performed slightly better than others. While ExSTraCS had previously been demonstrated to be able to perform well even on the 135-bit multiplexer problem, in this analysis only 18K instances were available for model training, when previous work required 40K instances for ExSTraCS to closely solve the 135-bit MUX where a staggering 4.36e40 unique binary instances make up the problem space \cite{urbanowicz_exstracs_2015}. In this particular benchmark analysis, ExSTraCS stands out clearly as the best-performing algorithm. Overall this analysis highlights the capability of RBML algorithms such as ExSTraCS to perform competitively and sometimes better than other well-established ML modeling approaches. 

\begin{figure}
    \centering
    \includegraphics[width=\textwidth]{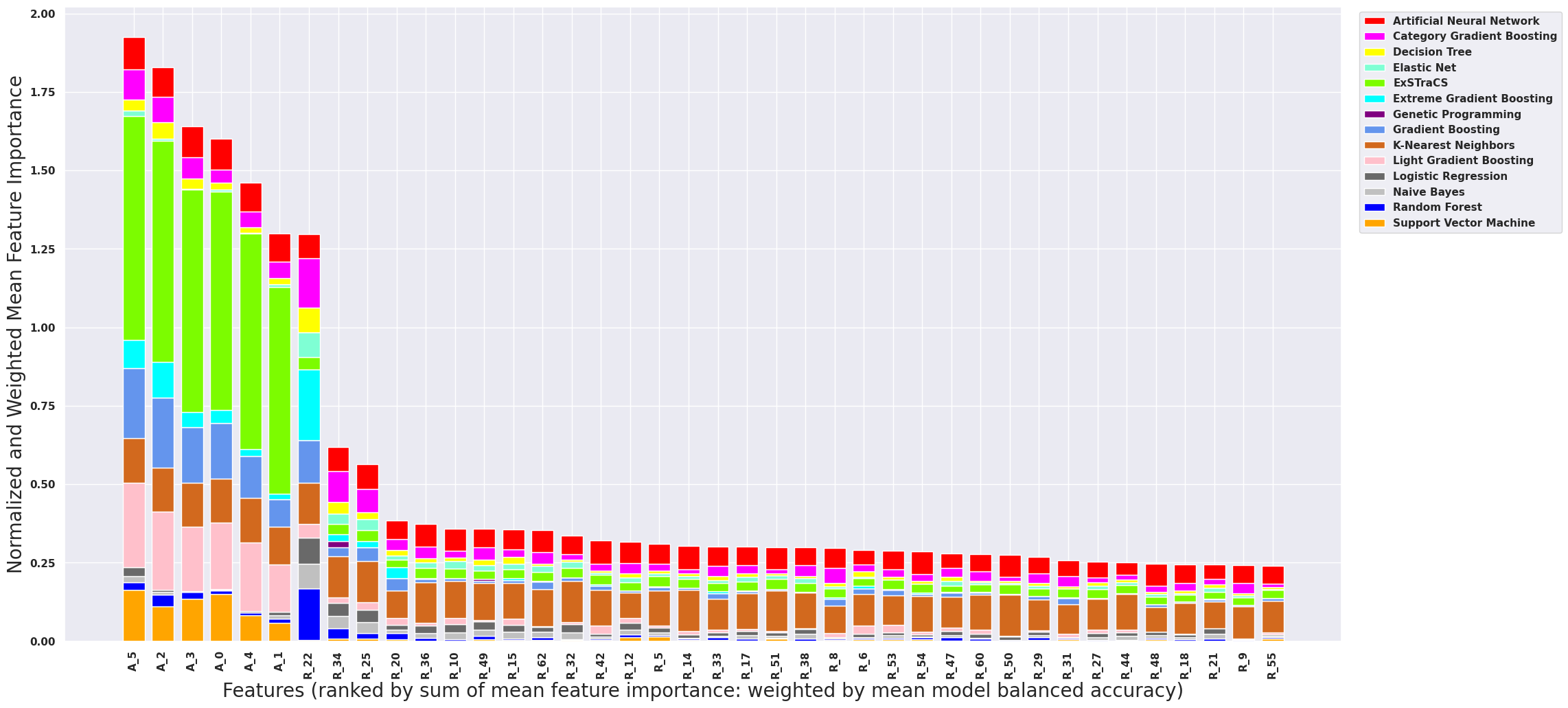}
    \caption*{\textbf{Figure S9: 70-bit Multiplexer – Mean Model Feature Importance Estimates Across all Algorithms.} This composite feature importance plot illustrates normalized and (balanced accuracy)-weighted mean FI scores summed across all algorithms. Only the top 40 features are displayed. In this dataset all features are predictive, but A\_0, A\_1, A\_2, A\_3, A\_4, and A\_5 are expected to be most predictive. }
\end{figure}
\addcontentsline{toc}{subsubsection}{Figure S9: 70-bit Multiplexer – Mean Model Feature Importance Estimates Across all Algorithms.}

\addcontentsline{toc}{subsection}{S.3.4: XOR Benchmark Results}
\subsection*{S.3.4: XOR Benchmark Results}

Refer to ‘xor\_exp\_STREAMLINE\_Report.pdf’ for EDA and modeling results for all 4 of the benchmark xor datasets. We review major results below. These datasets are also balanced, so we can confidently focus on ROC-AUC as the key performance metric. \textbf{Table S9} and \textbf{Figure S10} highlight the performance of STREAMLINE v0.3.4 on these datasets that had not been previously evaluated using STREAMLINE v0.2.5. For the 2-way dataset, 8 algorithms solve it perfectly, and DT, KNN, and SVM do reasonably well, with EN, LR, and NB failing. For the 3-way dataset, 5 algorithms (including CGB, ExSTraCS, XGB, GB, and LGB) solve it perfectly, and RF performing reasonably well, with all other algorithms failing. For the 4 and 5-way datasets, no algorithm performed well. This result was not initially expected, as previous work applying ExSTraCS suggested it should be well suited to detecting even higher-order interactions. 

\begin{table} [H]
\centering
\begin{tabular}{| l | l | l |}
\hline
 \rowcolor{light-gray}
  & \multicolumn{2}{|l|}{\textbf{v0.3.4}} \\
\hline
 \rowcolor{light-gray}
\textbf{Multiplexer Dataset} & \textbf{Best Algorithm} & \textbf{Benchmark Testing Evaluation} \\
\hline
\textbf{A – 2-way} & 8-way tie & 1.000 (0.000) \\
\hline
\textbf{B – 3-way} & 5-way tie & 1.000 (0.000) \\
\hline
\textbf{C – 4-way} & DT & 0.519 \\
\hline
\textbf{D – 5-way} & LR & 0.542 \\
\hline
\end{tabular}
\caption*{\textbf{Table S9: Best Mean Algorithm ROC-AUC in STEAMLINE v0.3.4 on XOR Datasets.} Each value represents the mean metric value (with standard deviation) across 10-fold CV trained models for that algorithm. Here the ‘best’ model is selected for each individual metric for comparison.}
\end{table}
\addcontentsline{toc}{subsubsection}{Table S9: Best Mean Algorithm ROC-AUC in STEAMLINE v0.3.4 on XOR Datasets.}

\begin{figure} [H]
    \centering
    \includegraphics[width=\textwidth]{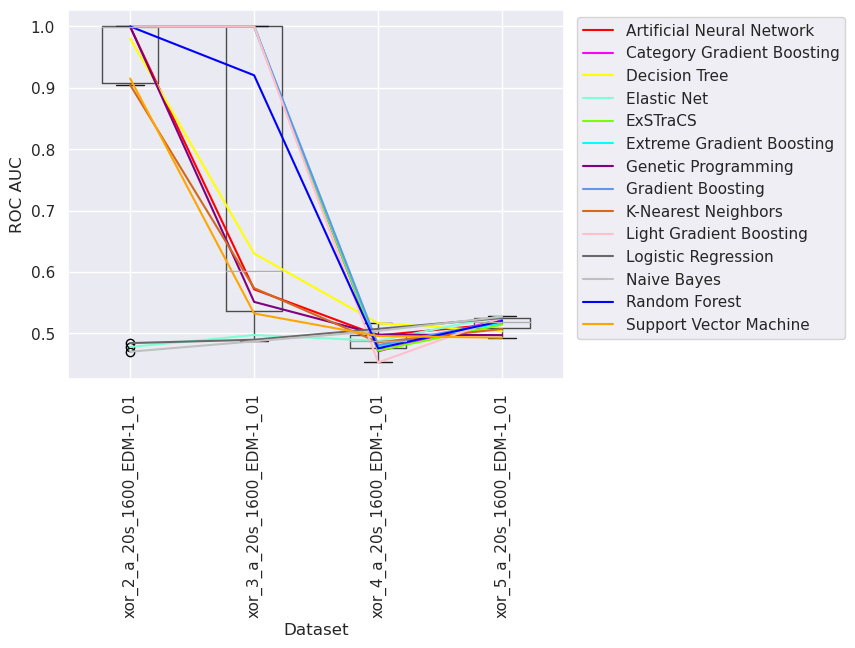}
    \caption*{\textbf{Figure S10: XOR Dataset Performance Comparison.} STREAMLINE v0.3.4 analysis. Each algorithm line represents mean ROC-AUC across 10-fold CV models. }
\end{figure}
\addcontentsline{toc}{subsubsection}{Figure S10: XOR Dataset Performance Comparison.}

However, this failure can be explained (not as an issue with all modeling algorithms) but due to the default strategy implemented by STREAMLINE to remove features during feature selection. Specifically, in the current implementation of STREAMLINE, phase 4 automatically removes any features from the respective training and testing datasets that yielded a score of 0 or lower for both the mutual information and MultiSURF algorithms (when filter-feat is set to True, i.e. the current default setting). This is based on the current working assumption that only a positive score for these algorithms is indicative of a potentially important feature. As mentioned above, we previously observed that MultiSURF would yield highly negative scores for predictive features involved in a high order interaction (i.e. 4-way and above) \cite{urbanowicz_benchmarking_2018}.  We thus noticed that predictive features were being removed prior to ML modeling in the 3-way and 4-way dataset analyses (see \textbf{Figure S11}). 

\begin{figure} [H]
    \centering
    \includegraphics[width=\textwidth]{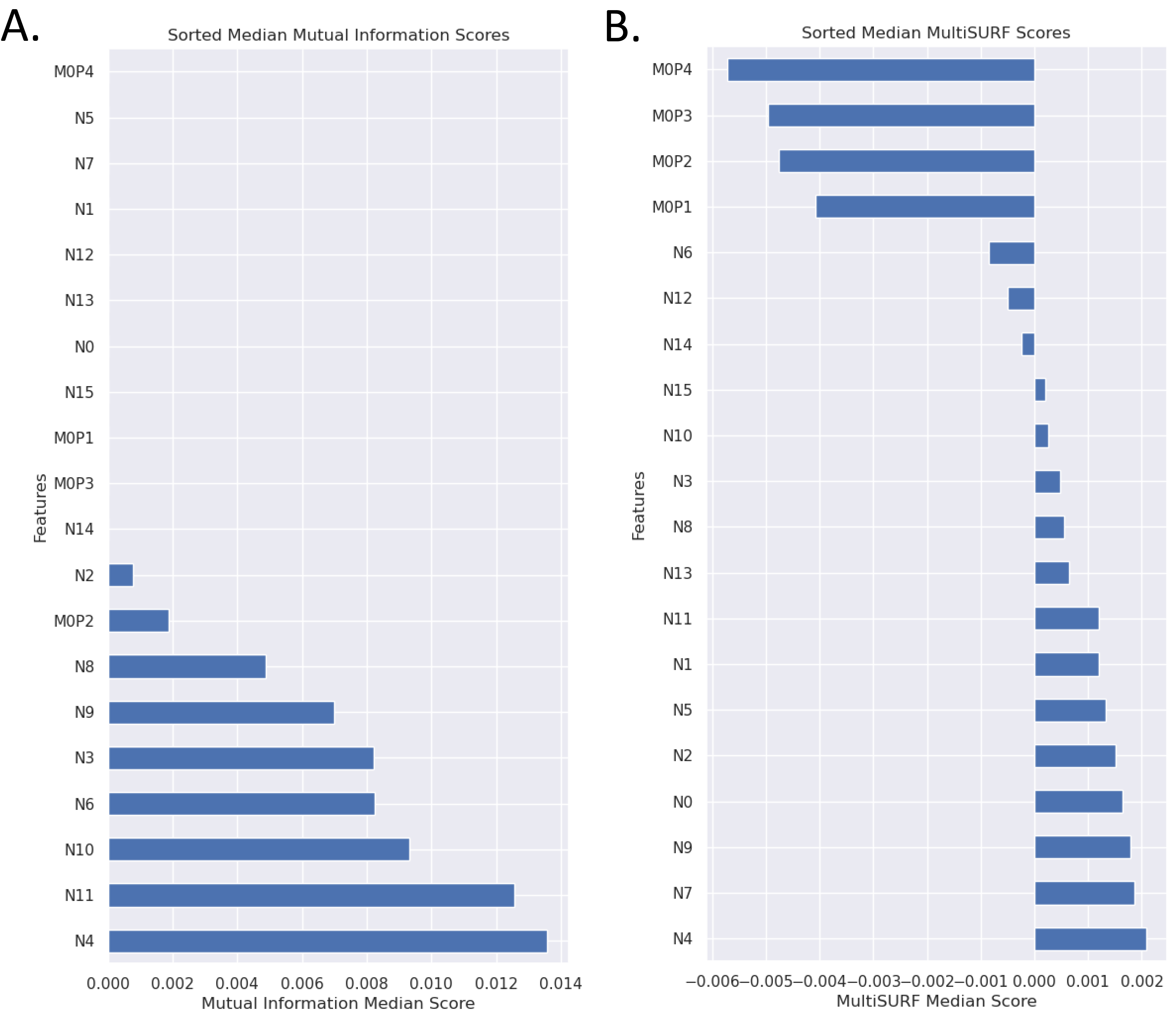}
    \caption*{\textbf{Figure S11: 4-way XOR Pre-Modeling Median Feature Importance Scores.} (A) Mutual Information, (B) MultiSURF. In this dataset, simulated predictive features include M0P1, M0P2, M0P3, and M0P4.}
\end{figure}
\addcontentsline{toc}{subsubsection}{Figure S11: 4-way XOR Pre-Modeling Median Feature Importance Scores.}

Thus, we re-ran this experiment setting STREAMLINE’s filter-feat parameter to False, to keep all features in the dataset regardless of initial feature importance scoring by mutual information or MultiSURF. \textbf{Table S10} and \textbf{Figure S12} presents the results of this follow-up analysis. In this secondary analysis, performance results for algorithms were very similar for 2-way and 3-way, but as expected for the 4-way XOR, ExSTraCS now stood out as performing close to 100\%, with CGB and GB doing fairly well, at around 82\%. Further behind was XGB and LGB achieving closer to 60\% and all algorithms failing. \textbf{Figure S13} illustrates model feature importance estimates for this 4-way XOR dataset, showing how the models that performed best focused on the correct interacting predictive features. On the 5-way XOR dataset, no algorithms performed well, but out of these, ExSTraCS performed best. This is most likely a result of the ‘curse of dimensionality’ and having a relatively small sample size available to detect such a high order interaction \cite{niel_survey_2015}. It is likely that with a larger number of instances, ExSTraCS (and possibly other ML algorithms evaluated) would be able to perform better on this benchmark problem. 

\begin{table} [H]
\centering
\begin{tabular}{| l | l | l |}
\hline
 \rowcolor{light-gray}
  & \multicolumn{2}{|l|}{\textbf{v0.3.4 – no feature selection}} \\
\hline
 \rowcolor{light-gray}
\textbf{Multiplexer Dataset} & \textbf{Best Algorithm} & \textbf{Benchmark Testing Evaluation} \\
\hline
\textbf{A – 2-way} & 7-way tie & 1.000 (0.000) \\
\hline
\textbf{B – 3-way} & 5-way tie & 1.000 (0.000) \\
\hline
\textbf{C – 4-way} & ExSTraCS & 0.98 () \\
\hline
\textbf{D – 5-way} & ExSTraCS & 0.522 () \\
\hline
\end{tabular}
\caption*{\textbf{Table S10: Best Mean Algorithm ROC-AUC in STEAMLINE v0.3.4 on XOR Datasets (no feature selection).} Each value represents the mean metric value (with standard deviation) across 10-fold CV trained models for that algorithm. Here the ‘best’ model is selected for each individual metric for comparison. }
\end{table}
\addcontentsline{toc}{subsubsection}{Table S10: Best Mean Algorithm ROC-AUC in STEAMLINE v0.3.4 on XOR Datasets (no feature selection).}

\begin{figure} [H]
    \centering
    \includegraphics[width=\textwidth]{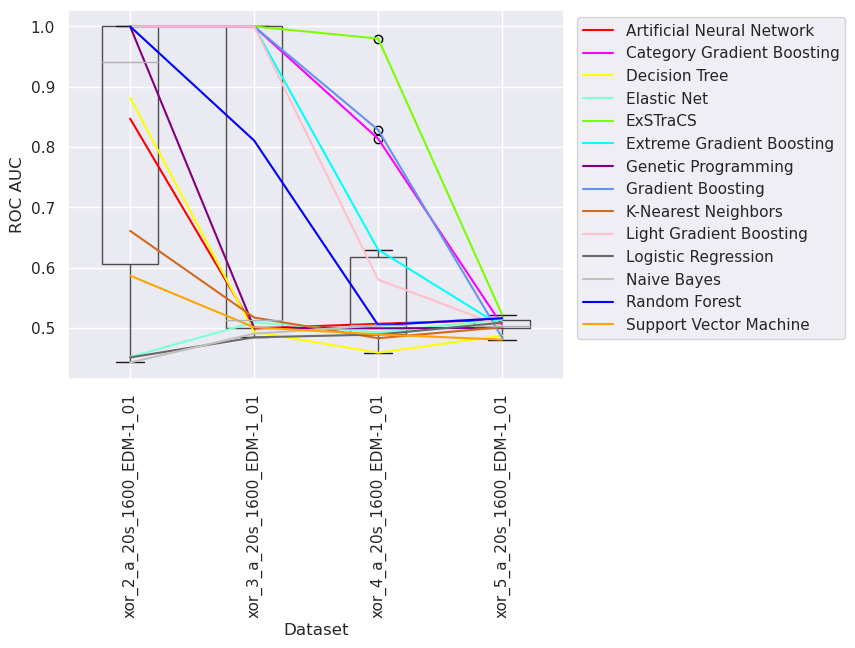}
    \caption*{\textbf{Figure S12: XOR Dataset Performance Comparison (no feature selection).} STREAMLINE v0.3.4 analysis. Each algorithm line represents mean ROC-AUC across 10-fold CV models. }
\end{figure}
\addcontentsline{toc}{subsubsection}{Figure S12: XOR Dataset Performance Comparison (no feature selection).}

\begin{figure} [H]
    \centering
    \includegraphics[width=\textwidth]{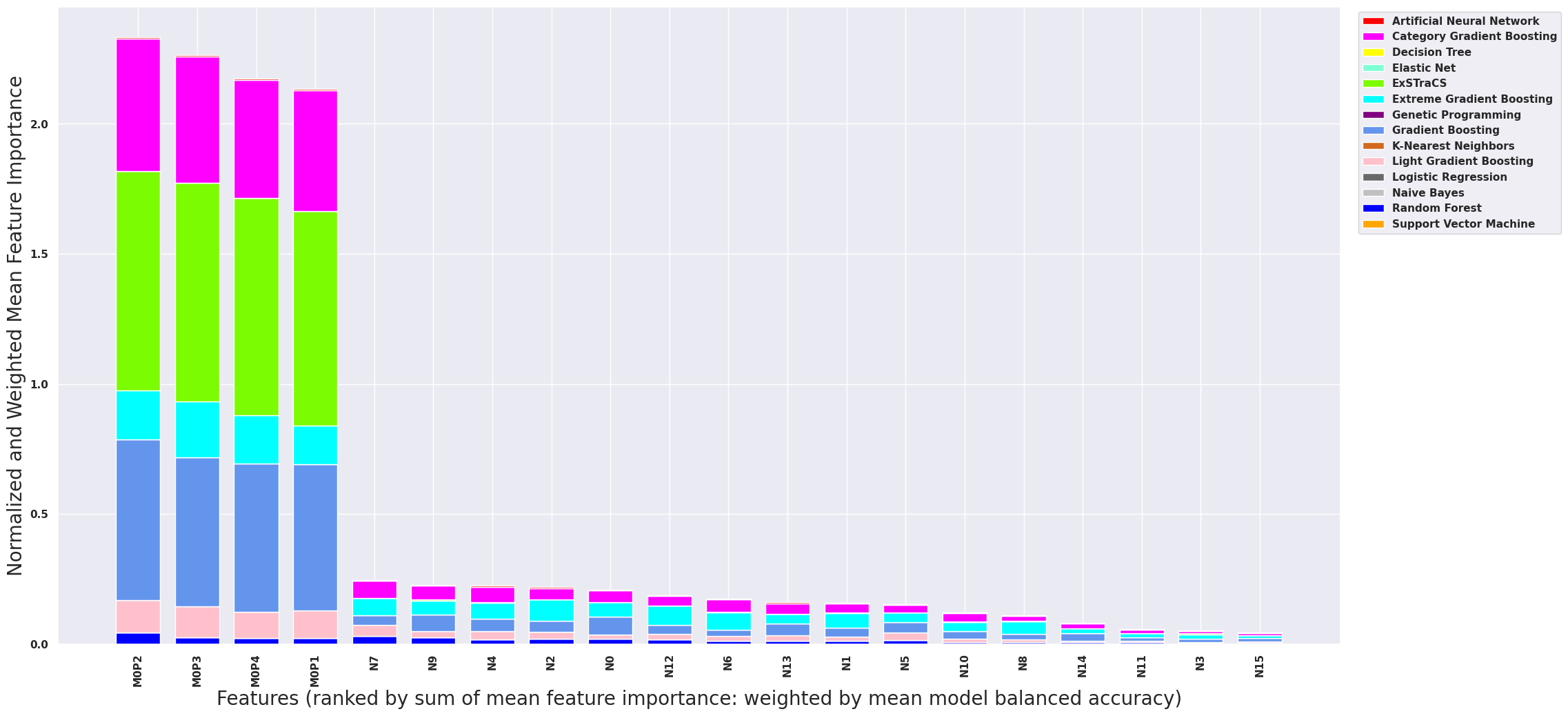}
    \caption*{\textbf{Figure S13: XOR 4-way Dataset Mean Model Feature Importance Estimates Across all Algorithms.} This composite feature importance plot illustrates normalized and (balanced accuracy)-weighted mean FI scores summed across all algorithms. In this dataset, simulated predictive features include M0P1, M0P2, M0P3, and M0P4.}
\end{figure}
\addcontentsline{toc}{subsubsection}{Figure S13: XOR 4-way Dataset Mean Model Feature Importance Estimates Across all Algorithms.}

These results highlight the need to further improve the feature selection strategy utilized in Phase 4 of STREAMLINE in future releases, in particular to address the (likely) rare, but possible, presence of features involved in high order interactions with no univariate effects. Future work will specifically examine the reliability of employing absolute value feature ranking with MultiSURF and it’s impact on the detection of simple univariate effects, as well as increasingly complex interactions, and then adopt the most effective strategy within STREAMLINE. 

\addcontentsline{toc}{subsection}{S.3.5: Obstructive Sleep Apnea Results}
\subsection*{S.3.5: Obstructive Sleep Apnea Results}

Refer to ‘sagic\_exp\_STREAMLINE\_Report.pdf’ for EDA and modeling results for all 20 variations of the SAGIC datasets. EDA and model performance results for respective replication datasets (A-T) are available upon request to the corresponding author (as their inclusion exceeded arxiv file size limits). Select plots and results are presented below to (1) demonstrate the diverse set of automated figures generated by STREAMLINE over subsequent phases, and (2) highlight major findings.

\addcontentsline{toc}{subsubsection}{S.3.5.1: Initial Exploratory Data Analysis}
\subsubsection*{S.3.5.1: Initial Exploratory Data Analysis}

\textbf{Figure S14} first illustrates the difference in class balance in SAGIC development datasets (before data processing) where OSA is defined as either AHI$\geq$15 or AHI$\geq$5 events/hour. While AHI$\geq$15 datasets are close to balanced, i.e. Class 0 = 1098 and Class 1 = 1080 (and can be primarily evaluated with ROC-AUC as the primary metric), AHI$\geq$5 have reasonably imbalanced classes, i.e. Class 0 = 606 and Class 1 = 1573 (and should be primarily evaluated with PRC-AUC as the primary metric). Note, these respective class balances have been maintained in the corresponding replication datasets for each (stratified partitioning used). 

\begin{figure} [H]
    \centering
    \includegraphics[width=\textwidth]{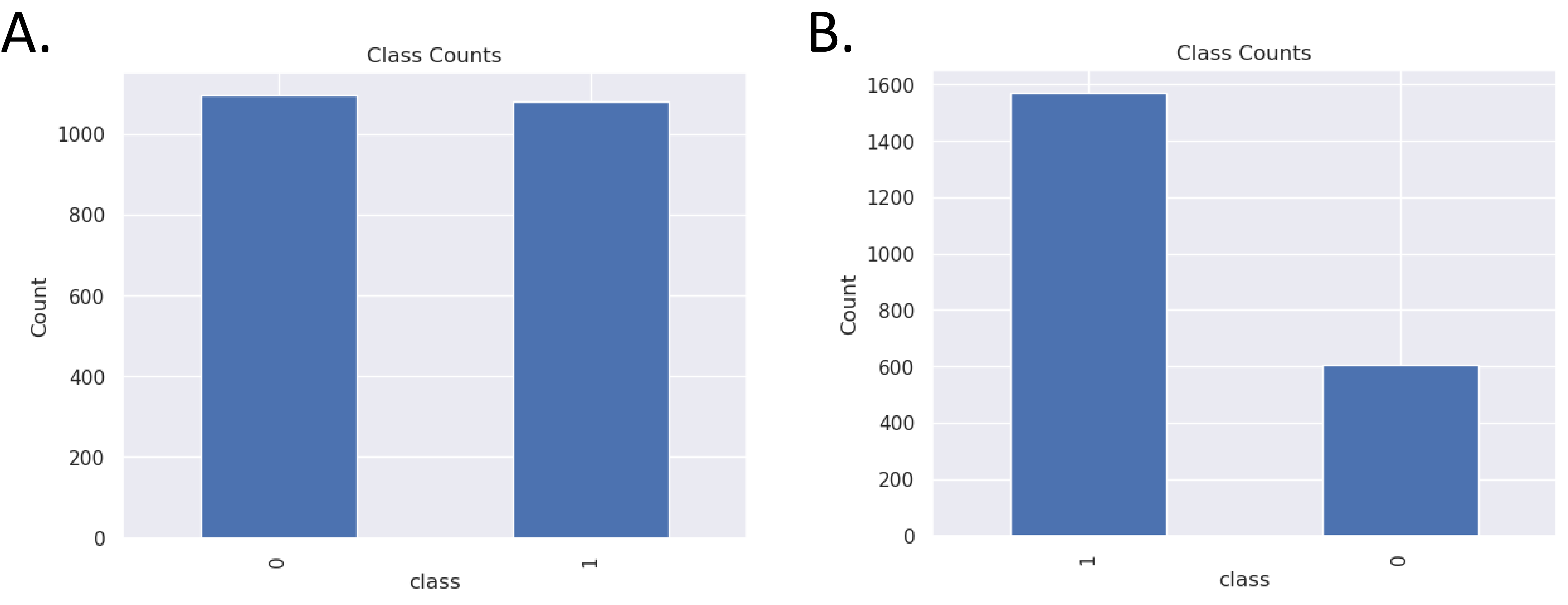}
    \caption*{\textbf{Figure S14: Bar Plot of Class Balance in OSA Development Datasets (Before Processing).} (A) Class counts for all 10 development datasets where OSA is defined as having AHI$\geq$15 events/hour. These are closely balanced datasets. (B) Class counts for all 10 development datasets where OSA is defined as having AHI$\geq$5 events/hour. These datasets have a reasonably large degree of class imbalance with more OSA than non-OSA. }
\end{figure}
\addcontentsline{toc}{subsubsection}{Figure S14: Bar Plot of Class Balance in OSA Development Datasets (Before Processing).}

\textbf{Figure S15} give an example of missing value counts in one of the 20 OSA datasets analyzed (i.e. J) prior to processing. This plot is the same for all 20 datasets (prior to processing). The majority of the 85 total features had no missing values, however some other features had a fairly high degree of missingness, with ‘medhx\_afib’ having the largest missing value count of 548.

\begin{figure} [H]
    \centering
    \includegraphics[width=0.6\textwidth]{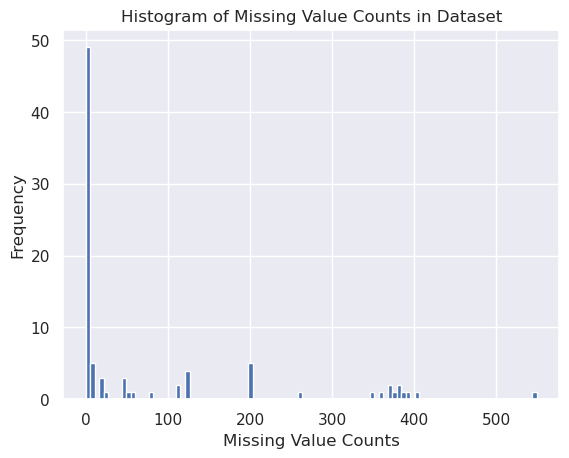}
    \caption*{\textbf{Figure S15: Histogram of Missing Value Counts Across Features in J\_GE15\_DEMDXSYMCFIO Development Dataset.} This dataset includes all features (across all feature groups) and a total of 2178 instances prior to data processing.}
\end{figure}
\addcontentsline{toc}{subsubsection}{Figure S15: Histogram of Missing Value Counts Across Features in J\_GE15\_DEMDXSYMCFIO Development Dataset.}

\textbf{Figure S16} gives the Pearson correlation heatmap for one of the 20 OSA datasets analyzed (i.e. J) across all 85 features in the OSA data prior to processing. These correlations are the same for all AHI$\geq$15 datasets, and very similar for all AHI$\geq$5 datasets. There were no perfect correlations in any of the datasets, so no features were automatically removed by STREAMLINE using the default correlation cutoff of 1.0. However, we note that there are a number of reasonably high positive and negative correlations among features in this dataset with the strongest correlation observed between \textit{mand\_tri\_area} and \textit{mand\_length\_dia} (0.959), in all datasets.

\begin{figure} [H]
    \centering
    \includegraphics[width=\textwidth]{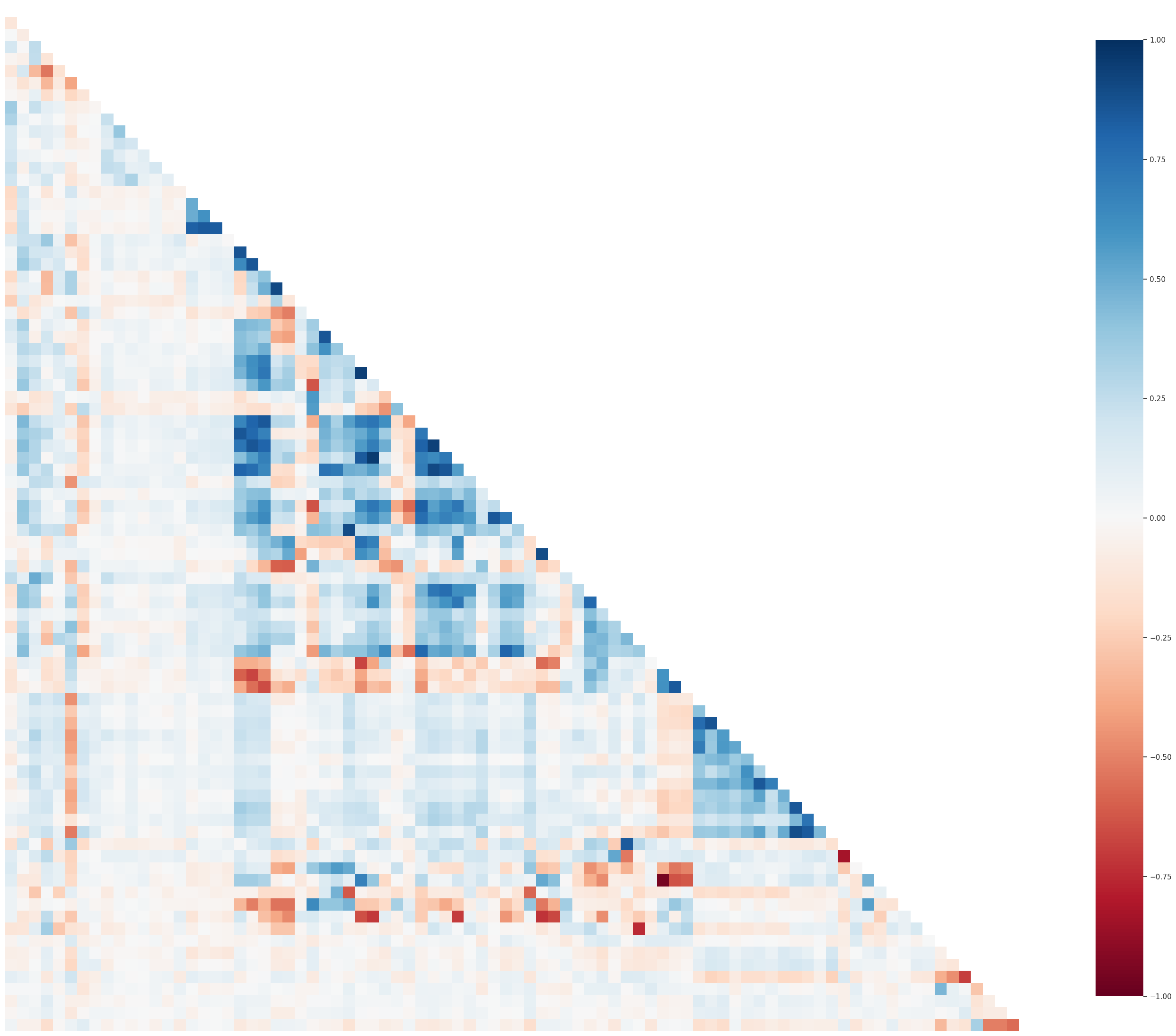}
    \caption*{\textbf{Figure S16: Pearson Feature Correlations in J\_GE15\_DEMDXSYMCFIO Development Dataset.} This dataset includes all features (across all feature groups) and a total of 2178 instances prior to processing. }
\end{figure}
\addcontentsline{toc}{subsubsection}{Figure S16: Pearson Feature Correlations in J\_GE15\_DEMDXSYMCFIO Development Dataset.}

\addcontentsline{toc}{subsubsection}{S.3.5.2: Automated Data Processing}
\subsubsection*{S.3.5.2: Automated Data Processing}

\textbf{Table S11} and \textbf{Table S12} summarize the development datasets characteristics before and after automated STREAMILNE data processing (i.e. cleaning and feature engineering). Since in these datasets we had already one-hot-encoded categorical features, and no features had a missingness > 50\%, the only processing to take place on any of these datasets was removing instances with high missingness (i.e. > 50\%). STREAMLINE automatically removes instances with more than 50\% missing values (by default), intended to avoid potential data quality issues when imputing missing values with multiple imputation when many other feature values for an instance are also missing. Note that even after processing, AHI$\geq$15 datasets remained relatively balanced, and AHI$\geq$5 datasets remained largely imbalanced (i.e. class ratio).

\begin{table} [H]
\centering
\scriptsize
\begin{tabular}{| p{0.8cm} | p{1cm} | p{1cm} | p{1cm} | p{1.2cm} | p{1cm} | p{1cm} | p{1cm} | p{1cm} | p{1cm} | p{0.7cm} |}
\hline
 \rowcolor{lightgray}
Dataset & Pre/Post
Proc. & Instances & Total Features & Categorical Features & Quant. Features & Missing Values & Missing Percent & Class 0 & Class 1 & Class Ratio \\
\hline
A & Pre & 2178 & 8 & 6 & 2 & 20 & 0..0011 & 1098 & 1080 & 1.02 \\
\hline
 \rowcolor{light-gray}
A & Post & 2174 & 8 & 6 & 2 & 0 & 0 & 1097 & 1077 & 1.02 \\
\hline
B & Pre & 2178 & 7 & 7 & 0 & 2877 & 0.1887 & 1098 & 1080 & 1.02 \\
\hline
 \rowcolor{light-gray}
B & Post & 1794 & 7 & 7 & 0 & 229 & 0.0182 & 840 & 954 & 0.88 \\
\hline
C & Pre & 2178 & 4 & 0 & 4 & 1442 & 0.1655 & 1098 & 1080 & 1.02 \\
\hline
 \rowcolor{light-gray}
C & Post & 1810 & 4 & 0 & 4 & 10 & 0.0014 & 855 & 955 & 0.90 \\
\hline
D & Pre & 2178 & 46 & 0 & 46 & 360 & 0.0036 & 1098 & 1080 & 1.02 \\
\hline
 \rowcolor{light-gray}
D & Post & 2178 & 46 & 0 & 46 & 360 & 0.0036 & 1098 & 1080 & 1.02 \\
\hline
E & Pre & 2178 & 20 & 8 & 12 & 2183 & 0.0501 & 1098 & 1080 & 1.02 \\
\hline
 \rowcolor{light-gray}
E & Post & 2139 & 20 & 8 & 12 & 1719 & 0.0402 & 1090 & 1049 & 1.04 \\
\hline
F & Pre & 2178 & 15 & 13 & 2 & 2897 & 0.0887 & 1098 & 1080 & 1.02 \\
\hline
 \rowcolor{light-gray}
F & Post & 1810 & 15 & 13 & 2 & 306 & 0.0113 & 852 & 958 & 0.89 \\
\hline
G & Pre & 2178 & 19 & 13 & 6 & 4339 & 0.1048 & 1098 & 1080 & 1.02 \\
\hline
 \rowcolor{light-gray}
G & Post & 1834 & 19 & 13 & 6 & 546 & 0.0157 & 867 & 967 & 0.90 \\
\hline
H & Pre & 2178 & 65 & 13 & 52 & 4699 & 0.0332 & 1098 & 1080 & 1.02 \\
\hline
 \rowcolor{light-gray}
H & Post & 2178 & 65 & 13 & 52 & 4699 & 0.0332 & 1098 & 1080 & 1.02 \\
\hline
I & Pre & 2178 & 39 & 21 & 18 & 6522 & 0.0768 & 1098 & 1080 & 1.02 \\
\hline
 \rowcolor{light-gray}
I & Post & 2166 & 39 & 21 & 18 & 6281 & 0.0743 & 1092 & 1074 & 1.02 \\
\hline
J & Pre & 2178 & 85 & 21 & 64 & 6882 & 0.0372 & 1098 & 1080 & 1.02 \\
\hline
 \rowcolor{light-gray}
J & Post & 2178 & 85 & 21 & 64 & 6882 & 0.0372 & 1098 & 1080 & 1.02 \\
\hline
\end{tabular}
\normalsize
\caption*{\textbf{Table S11: AHI$\geq$15 Development Dataset Characteristics Pre vs. Post STREAMLINE Processing. }}
\end{table}
\addcontentsline{toc}{subsubsection}{Table S11: AHI$\geq$15 Development Dataset Characteristics Pre vs. Post STREAMLINE Processing.}

\normalsize

\begin{table} [H]
\centering
\scriptsize
\begin{tabular}{| p{0.8cm} | p{1cm} | p{1cm} | p{1cm} | p{1.2cm} | p{1cm} | p{1cm} | p{1cm} | p{1cm} | p{1cm} | p{0.7cm} |}
\hline
 \rowcolor{lightgray}
Dataset & Pre/Post Proc. & Instances & Total Features & Categorical Features & Quant.
Features & Missing Values & Missing Percent & Class 0 & Class 1 & Class Ratio \\
\hline
K & Pre & 2179 & 8 & 6 & 2 & 25 & 0.0014 & 606 & 1573 & 0.39 \\
\hline
 \rowcolor{light-gray}
K & Post & 2174 & 8 & 6 & 2 & 0 & 0 & 605 & 1569 & 0.39 \\
\hline
L & Pre & 2179 & 7 & 7 & 0 & 2866 & 0.1879 & 606 & 1573 & 0.39 \\
\hline
 \rowcolor{light-gray}
L & Post & 1798 & 7 & 7 & 0 & 233 & 0.0185 & 447 & 1351 & 0.33 \\
\hline
M & Pre & 2179 & 4 & 0 & 4 & 1408 & 0.1615 & 606 & 1573 & 0.39 \\
\hline
 \rowcolor{light-gray}
M & Post & 1820 & 4 & 0 & 4 & 10 & 0.0014 & 449 & 1371 & 0.33 \\
\hline
N & Pre & 2179 & 46 & 0 & 46 & 336 & 0.0034 & 606 & 1573 & 0.39 \\
\hline
 \rowcolor{light-gray}
N & Post & 2179 & 46 & 0 & 46 & 336 & 0.0034 & 606 & 1573 & 0.39 \\
\hline
O & Pre & 2179 & 20 & 8 & 12 & 2170 & 0.0498 & 606 & 1573 & 0.39 \\
\hline
 \rowcolor{light-gray}
O & Post & 1961 & 20 & 8 & 12 & 1673 & 0.0391 & 603 & 1534 & 0.41 \\
\hline
P & Pre & 2179 & 15 & 13 & 2 & 2891 & 0.0885 & 606 & 1573 & 0.39 \\
\hline
 \rowcolor{light-gray}
P & Post & 1812 & 15 & 13 & 2 & 302 & 0.0111 & 450 & 1362 & 0.39 \\
\hline
Q & Pre & 2179 & 19 & 13 & 6 & 4299 & 0.1028 & 606 & 1573 & 0.39 \\
\hline
 \rowcolor{light-gray}
Q & Post & 1840 & 19 & 13 & 6 & 566 & 0.0162 & 457 & 1383 & 0.33 \\
\hline
R & Pre & 2179 & 65 & 13 & 52 & 4635 & 0.0327 & 606 & 1573 & 0.39 \\
\hline
 \rowcolor{light-gray}
R & Post & 2179 & 65 & 13 & 52 & 4635 & 0.0327 & 606 & 1573 & 0.39 \\
\hline
S & Pre & 2179 & 39 & 21 & 18 & 6469 & 0.0761 & 606 & 1573 & 0.39 \\
\hline
 \rowcolor{light-gray}
S & Post & 2168 & 39 & 21 & 18 & 6244 & 0.0738 & 600 & 1668 & 0.36 \\
\hline
J & Pre & 2179 & 85 & 21 & 64 & 6805 & 0.0367 & 606 & 1573 & 0.39 \\
\hline
 \rowcolor{light-gray}
J & Post & 2179 & 85 & 21 & 64 & 6805 & 0.0367 & 606 & 1573 & 0.39 \\
\hline
\end{tabular}
\normalsize
\caption*{\textbf{Table S12: AHI$\geq$5 Development Dataset Characteristics Pre vs. Post STREAMLINE Processing. }}
\end{table}
\addcontentsline{toc}{subsubsection}{Table S12: AHI$\geq$5 Development Dataset Characteristics Pre vs. Post STREAMLINE Processing. }
\normalsize

\addcontentsline{toc}{subsubsection}{S.3.5.3: EDA Univariate Analyses on Processed Development Data}
\subsubsection*{S.3.5.3: EDA Univariate Analyses on Processed Development Data}
Here, we focus on the univariate analyses conducted in the two development datasets (after processing) with all feature subgroups (i.e. J and T). \textbf{Table S13} and \textbf{Table S14} give the top features univariately associated with outcome for AHI$\geq$15 and AHI$\geq$5 datasets, respectively. In these datasets, top features were all quantitative (i.e. evaluated with the Mann-Whitney U Test), however all categorical features were automatically evaluated with Chi-Square Test.

\begin{table} [H]
\centering
\begin{tabular}{| l | l | l | l |}
\hline
\rowcolor{light-gray}
\textbf{Feature} & \textbf{p-value} & \textbf{Test-Statistic} & \textbf{Test-Name} \\
\hline
face\_width & 1.046e-51 & 371046.5 & Mann-Whitney U Test \\
\hline
mand\_width & 1.414e-50 & 373520.5 & Mann-Whitney U Test \\
\hline
bmi & 1.082e-38 & 402018.5 & Mann-Whitney U Test \\
\hline
max\_mand\_vol & 9.671e-36 & 407717.0 & Mann-Whitney U Test \\
\hline
max\_tri\_area & 2.541e-34 & 412405.0 & Mann-Whitney U Test \\
\hline
cer\_ang & 8.807e-31 & 342422.0 & Mann-Whitney U Test \\
\hline
cranial\_base\_tri\_area & 1.530e-28 & 429862.5 & Mann-Whitney U Test \\
\hline
mand\_vol & 3.194e-28 & 431267.5 & Mann-Whitney U Test \\
\hline
mand\_tri\_area & 6.906e-27 & 435382.0 & Mann-Whitney U Test \\
\hline
max\_mand\_box\_area & 1.250e-26 & 434456.5 & Mann-Whitney U Test \\
\hline
\end{tabular}
\caption*{\textbf{Table S13: Top Univariate Analyses for J\_GE15\_DEMDXSYMCFIO.} Top 10 features, ranked by p-value.}
\end{table}
\addcontentsline{toc}{subsubsection}{Table S13: Top Univariate Analyses for J\_GE15\_DEMDXSYMCFIO.}

\begin{table} [H]
\centering
\begin{tabular}{| l | l | l | l |}
\hline
\rowcolor{light-gray}
\textbf{Feature} & \textbf{p-value} & \textbf{Test-Statistic} & \textbf{Test-Name} \\
\hline
bmi & 1.060e-42 & 296375.5 & Mann-Whitney U Test \\
\hline
mand\_width & 1.339e-40 & 301119 & Mann-Whitney U Test \\
\hline
face\_width & 5.267e-38 & 307115 & Mann-Whitney U Test \\
\hline
cer\_ang & 6.350e-33 & 258706.5 & Mann-Whitney U Test \\
\hline
max\_mand\_vol & 1.464e-30 & 323896.5 & Mann-Whitney U Test \\
\hline
mand\_vol & 1.623e-27 & 333591 & Mann-Whitney U Test \\
\hline
max\_tri\_area & 1.865e-26 & 336101 & Mann-Whitney U Test \\
\hline
loud\_snore & 1.721e-25 & 214783.5 & Mann-Whitney U Test \\
\hline
mand\_tri\_area & 4.203e-24 & 343357 & Mann-Whitney U Test \\
\hline
lat\_fac\_ht & 4.635e-24 & 342626 & Mann-Whitney U Test \\
\hline

\end{tabular}
\caption*{\textbf{ Table S14: Top Univariate Analyses for T\_GE5\_DEMDXSYMCFIO.} Top 10 features, ranked by p-value.}
\end{table}
\addcontentsline{toc}{subsubsection}{Table S14: Top Univariate Analyses for T\_GE5\_DEMDXSYMCFIO.}

Focusing on the most significant univariate association for each table, \textbf{Figure S17} gives a boxplot summary between classes for \textit{face\_width}, and \textbf{Figure S18} gives a boxplot summary between classes for \textit{bmi}. 

\begin{figure} [H]
    \centering
    \includegraphics[width=.7\textwidth]{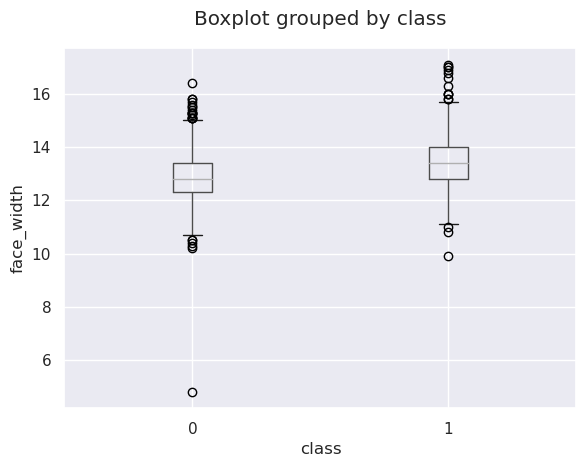}
    \caption*{\textbf{Figure S17: Most Significant Univariate Association for J\_GE15\_DEMDXSYMCFIO Development Dataset (face\_width).} Boxplot comparing \textit{face\_width} values between class 0 and class 1 as defined by AHI$\geq$15.}
\end{figure}
\addcontentsline{toc}{subsubsection}{Figure S17: Most Significant Univariate Association for J\_GE15\_DEMDXSYMCFIO Development Dataset (face\_width).}

\begin{figure} [H]
    \centering
    \includegraphics[width=.7\textwidth]{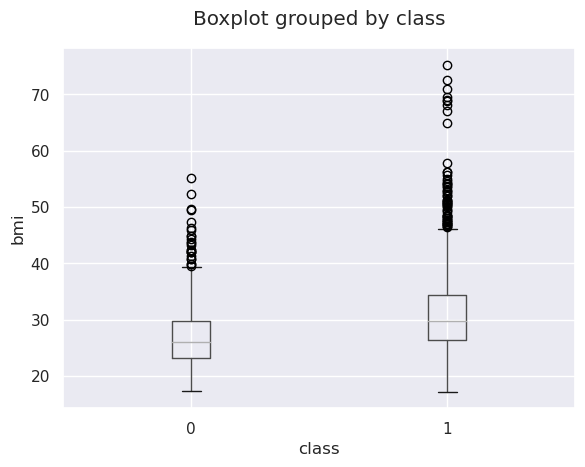}
    \caption*{\textbf{Figure S18: Most Significant Univariate Association for T\_GE5\_DEMDXSYMCFIO Development Dataset (bmi).} Boxplot comparing bmi values between class 0 and class 1 as defined by AHI$\geq$5.}
\end{figure}
\addcontentsline{toc}{subsubsection}{Figure S18: Most Significant Univariate Association for T\_GE5\_DEMDXSYMCFIO Development Dataset (bmi).}

\addcontentsline{toc}{subsubsection}{S.3.5.4: Pre-Modeling Feature Importance Estimation}
\subsubsection*{S.3.5.4: Pre-Modeling Feature Importance Estimation}

 \textbf{Figure S19} and \textbf{Figures S20} give pre-modeling FI estimates for the two development datasets with all feature subgroups (i.e. J and T) after processing, CV partitioning, imputation, and scaling. These scores represent the median FI calculated by either Mutual Information (sensitive to univariate effects) or MultiSURF (sensitive to both univariate and interaction effects) over 10-fold CV training datasets. For dataset J in \textbf{Figure S19}, the three top scoring features are the same for both FI estimation algorithms, i.e. \textit{mand\_width}, \textit{face\_width}, and \textit{bmi}. For dataset T in \textbf{Figure S20}, the top scoring feature for mutual information was \textit{bmi}, but for MultiSURF it was \textit{cer\_ang}. 

 \begin{figure} [H]
    \centering
    \includegraphics[width=\textwidth]{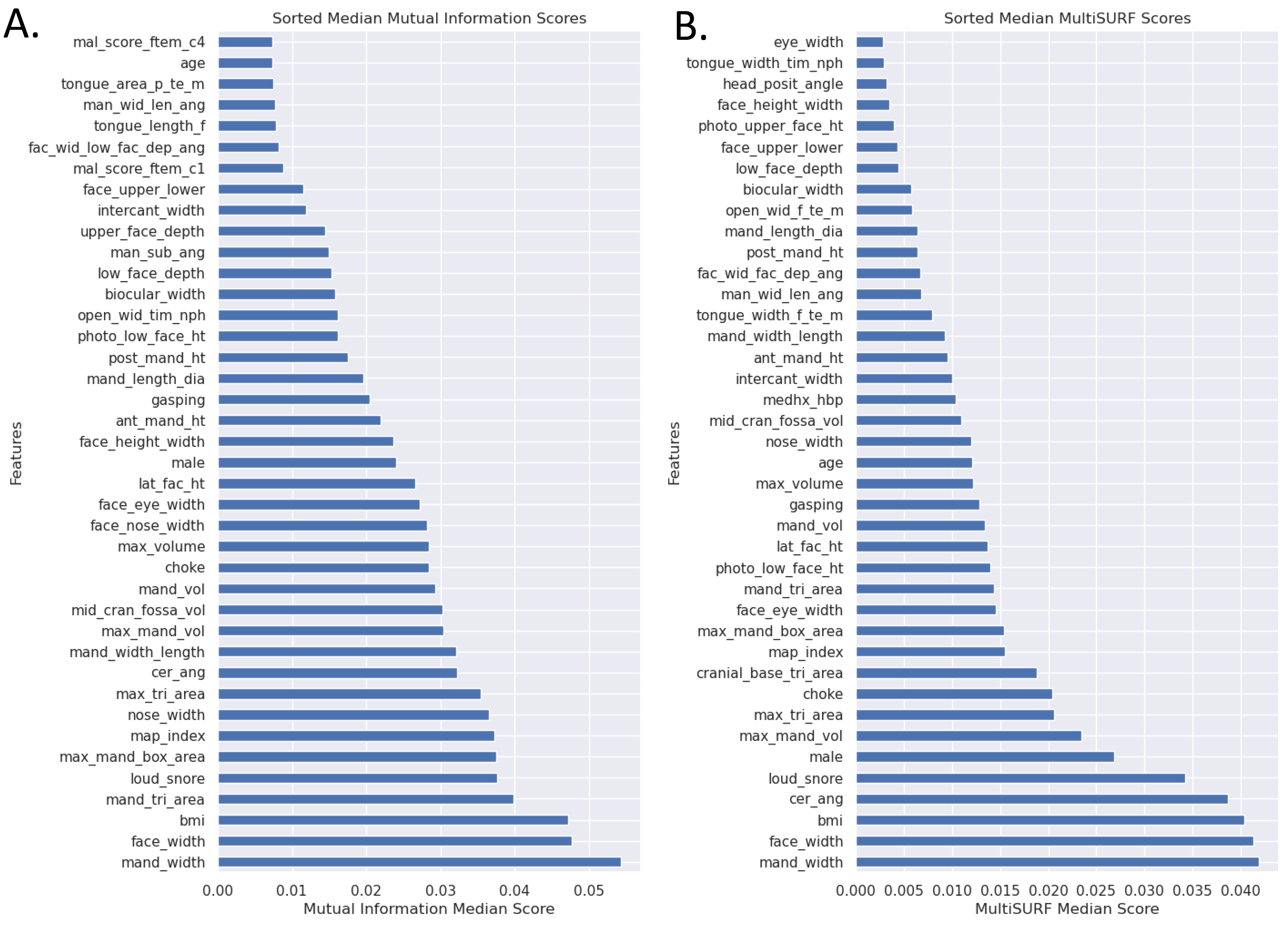}
    \caption*{\textbf{Figure S19: J\_GE15\_DEMDXSYMCFIO Pre-Modeling Median FI Scores.}Median FI scores across 10-fold CV training datasets (A) Mutual Information, (B) MultiSURF. }
\end{figure}
\addcontentsline{toc}{subsubsection}{Figure S19: J\_GE15\_DEMDXSYMCFIO Pre-Modeling Median FI Scores.}

\begin{figure} [H]
    \centering
    \includegraphics[width=\textwidth]{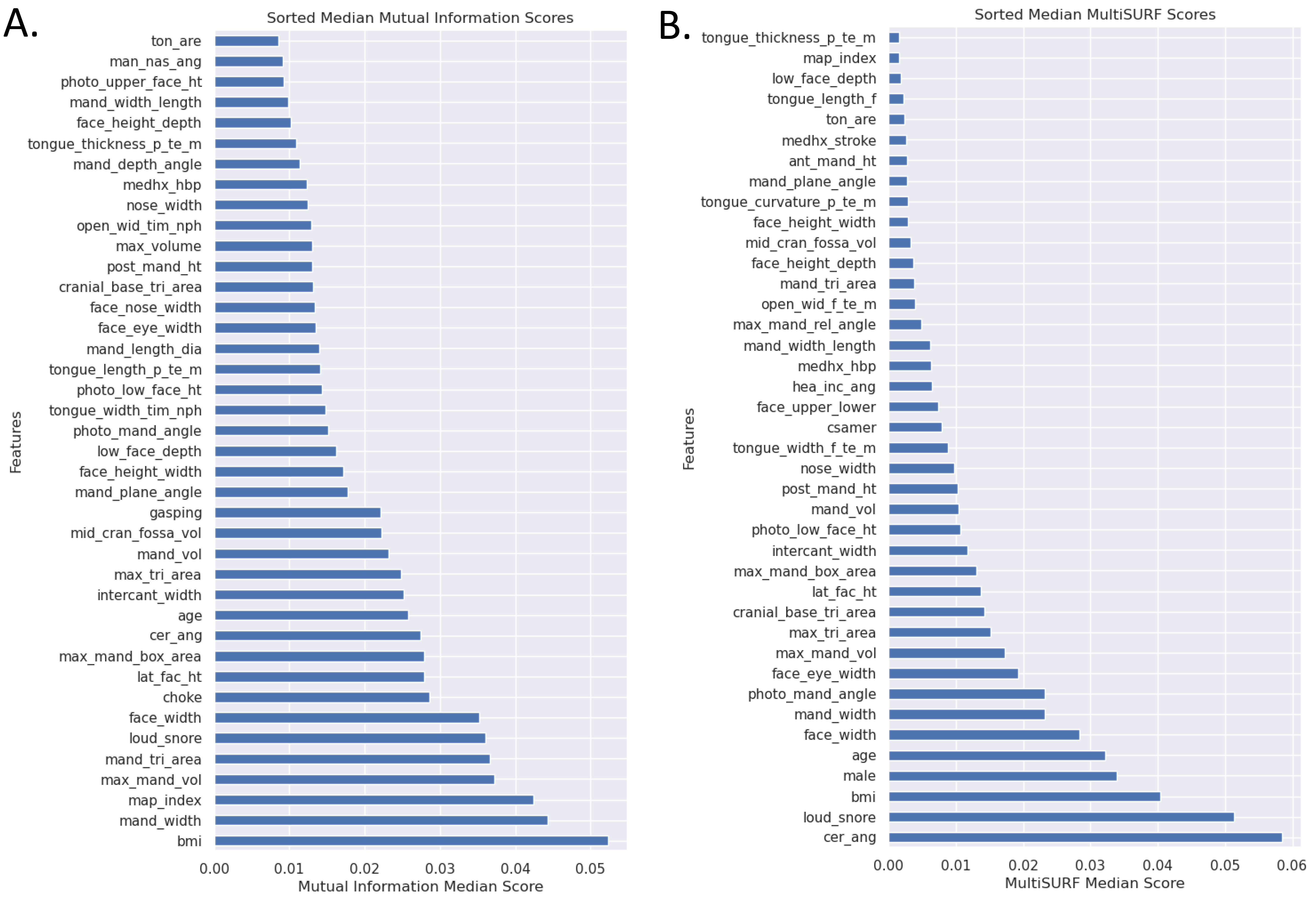}
    \caption*{\textbf{Figure S20: T\_GE5\_DEMDXSYMCFIO Pre-Modeling Median FI Scores.} Median FI scores across 10-fold CV training datasets (A) Mutual Information, (B) MultiSURF. }
\end{figure}
\addcontentsline{toc}{subsubsection}{Figure S20: T\_GE5\_DEMDXSYMCFIO Pre-Modeling Median FI Scores.}

\addcontentsline{toc}{subsubsection}{S.3.5.5: Development Data Testing Evaluations of Modeling}
\subsubsection*{S.3.5.5: Development Data Testing Evaluations of Modeling}

\textbf{Figure 2} (main text) gives a global summary of model testing performance (with 14 ML algorithms) across all 20 OSA datasets differing in their encoding of OSA outcome (as either AHI$\geq$15 or AHI$\geq$5), and/or their inclusion of different available features sets. \textbf{Table S15} gives pairwise performance metric comparisons (using Wilcoxon Signed-Rank Test) between datasets having the same underlying features, e.g. A vs. K, indicate that balanced accuracy and ROC-AUC were consistently (but not always significantly) higher for models trained on AHI$\geq$5 data. Notably, when focusing on PRC-AUC, GP models perform far better than any other algorithm for datasets B, C, E, L, M, and O (see \textbf{Figure S21}). However, closer inspection of other evaluation metrics for these GP models reveal that they are predicting class 1 for all testing instances (i.e. they have the maximum number of possible false positives, and thus are not useful models). Therefore, we replaced these GP models in our assessment with the next best scoring algorithm (as indicated by GP->’AlgorithmName’ in \textbf{Table S15}. Even after this substitution, AHI$\geq$5 datasets still yielded consistently and significantly higher PRC-AUC.

\begin{table} [H]
\begin{adjustbox}{center}
\scalebox{.82}{
\begin{tabular}{| p{1.2cm} | l | l | l | l | l | p{1.7cm} | p{1.7cm} | p{1.8cm} | p{1.7cm} |}
\hline
\rowcolor{light-gray}
&&&&&& Data 1 & Data 1 & Data 2 & Data 2 \\
\rowcolor{light-gray}
Metric & Data 1 & Data 2 & Feature Set(s) & Statistic & p-value & Best &Median &Best&Median \\
\rowcolor{light-gray}
&&&&&& Algorithm & (AHI$\geq$15) & Algorithm & (AHI$\geq$5) \\ \hline \hline
\multirow{10}{*}{{\rotatebox[origin=c]{90}{\textbf{Balanced Accuracy}}}} & A & K & DEM & 15 & 0.232422 & CGB & 0.674879 & XGB & \cellcolor{green!25}0.702773 \\ \cline{2-10}
 & B & L & DX & 22 & 0.625 & GP & 0.55878 & XGB & \cellcolor{green!25}0.57963 \\ \cline{2-10}
 & C & M & SYM & 8 & \cellcolor{yellow!50}0.048828 & LGP & 0.623776 & XGB & \cellcolor{green!25}0.654582 \\ \cline{2-10}
 & D & N & CF & 22 & 0.625 & LR & 0.678416 & XGB & \cellcolor{green!25}0.693432 \\ \cline{2-10}
 & E & O & IO & 15 & 0.232422 & LR & 0.60168 & EN & \cellcolor{green!25}0.612221 \\ \cline{2-10}
 & F & P & DEM+DX & 24 & 0.769531 & XGB & 0.661979 & SVM & \cellcolor{green!25}0.688487 \\ \cline{2-10}
 & G & Q & DEM+DX+SYM & 19 & 0.431641 & ExSTraCS & 0.690846 & ExSTraCS & \cellcolor{green!25}0.695652 \\ \cline{2-10}
 & H & R & DEM+DX+SYM+CF & 5 & \cellcolor{yellow!50}0.019531 & ExSTraCS & 0.694347 & ExSTraCS & \cellcolor{green!25}0.718696 \\ \cline{2-10}
 & I & S & DEM+DX+SYM+IO & 16 & 0.275391 & ExSTraCS & 0.685458 & ExSTraCS & \cellcolor{green!25}0.709368 \\ \cline{2-10}
 & J & T & \cellcolor{blue!25}DEM+DX+SYM+CF+IO & 8 & \cellcolor{yellow!50}0.048828 & CGB & 0.692929 & \cellcolor{blue!25}EN & \cellcolor{green!25}0.718936 \\ 
\hline \hline
\multirow{10}{*}{{\rotatebox[origin=c]{90}{\textbf{ROC-AUC}}}} & A & K & DEM & 14 & 0.193359 & ExSTraCS & 0.738696 & RF & \cellcolor{green!25}0.768363 \\ \cline{2-10}
 & B & L & DX & 19 & 0.431641 & LGB & 0.561623 & LGB & \cellcolor{green!25}0.58465 \\ \cline{2-10}
 & C & M & SYM & 16 & 0.275391 & GB & 0.660045 & EN & \cellcolor{green!25}0.679587 \\ \cline{2-10}
 & D & N & CF & 18 & 0.375 & RF & 0.734304 & XGB & \cellcolor{green!25}0.767255 \\ \cline{2-10}
 & E & O & IO & 25 & 0.845703 & SVM & 0.640454 & RF & \cellcolor{green!25}0.653842 \\ \cline{2-10}
 & F & P & DEM+DX & 8 & \cellcolor{yellow!50} 
 0.048828 & LGB & 0.720548 & ANN & \cellcolor{green!25}0.761928 \\ \cline{2-10}
 & G & Q & DEM+DX+SYM & 16 & 0.275391 & LGB & 0.742128 & EN & \cellcolor{green!25}0.761972 \\ \cline{2-10}
 & H & R & \cellcolor{blue!25}DEM+DX+SYM+CF & 3 & \cellcolor{yellow!50}0.009766 & ANN & 0.766822 & \cellcolor{blue!25}XGB & \cellcolor{green!25}0.79297 \\ \cline{2-10}
 & I & S & DEM+DX+SYM+IO & 16 & 0.275391 & LGB & 0.7554 & CGB & \cellcolor{green!25}0.778981 \\ \cline{2-10}
 & J & T & DEM+DX+SYM+CF+IO & 7 & \cellcolor{yellow!50} 0.037109 & ANN & 0.760819 & GB & \cellcolor{green!25}0.792941 \\
\hline \hline
\multirow{10}{*}{{\rotatebox[origin=c]{90}{\textbf{PRC-AUC}}}} & A & K & DEM & 0 & \cellcolor{yellow!50}0.001953 & CGB & 0.726915 & CGB & \cellcolor{green!25}0.873214 \\ \cline{2-10}
 & B & L & DX & 3 & \cellcolor{yellow!50}0.001152 & GP & 0.678069 & GP -> DT & \cellcolor{green!25}0.819413 \\ \cline{2-10}
 & C & M & SYM & 3 & \cellcolor{yellow!50}0.000157 & GP & 0.708505 & GP -> LGB & \cellcolor{green!25}0.862589 \\ \cline{2-10}
 & D & N & CF & 0 & \cellcolor{yellow!50}0.001953 & XGB & 0.719624 & RF & \cellcolor{green!25}0.87526 \\ \cline{2-10}
 & E & O & IO & 4 & \cellcolor{yellow!50}0.000157 & EN & 0.625999 & GP -> RF & \cellcolor{green!25}0.815426 \\ \cline{2-10}
 & F & P & DEM+DX & 0 & \cellcolor{yellow!50}0.001953 & RF & 0.735926 & CGB & \cellcolor{green!25}0.889826 \\ \cline{2-10}
 & G & Q & DEM+DX+SYM & 0 & \cellcolor{yellow!50}0.001953 & LGB & 0.749936 & EN & \cellcolor{green!25}0.89404 \\ \cline{2-10}
 & H & R & DEM+DX+SYM+CF & 0 & \cellcolor{yellow!50}0.001953 & CGB & 0.752545 & XGB & \cellcolor{green!25}0.894904 \\ \cline{2-10}
 & I & S & DEM+DX+SYM+IO & 0 & \cellcolor{yellow!50}0.001953 & XGB & 0.732568 & SVM & \cellcolor{green!25}0.892398 \\ \cline{2-10}
 & J & T & \cellcolor{blue!25}DEM+DX+SYM+CF+IO & 0 & \cellcolor{yellow!50}0.001953 & RF & 0.756443 & \cellcolor{blue!25}EN & \cellcolor{green!25}0.896618 \\
\hline
\end{tabular}}
\end{adjustbox}
\caption*{\textbf{Table S15: Pairwise Performance Comparisons of AHI$\geq$15 vs. AHI$\geq$5 Datasets with Same Feature Set(s).} Pairwise tests using the Wilcoxon Signed Rank Test. The testing metrics of the 10 CV models trained by the algorithm with the best median metric is the basis for this comparison. Metrics in this table include balanced accuracy, ROC-AUC, and PRC-AUC. Cells in green denote the larger median performance of the pair. The cells in blue point out the dataset/algorithm combination where the highest median metric value was observed. Rows with p-value highlighted in yellow fall under the 0.05 significance cutoff. ‘GP->Algorithm’ indicates where GP was replaced by algorithm with next highest median performance (new algorithm metrics given).}
\end{table}
\addcontentsline{toc}{subsubsection}{Table S15: Pairwise Performance Comparisons of AHI$\geq$15 vs. AHI$\geq$5 Datasets with Same Feature Set(s).}

\begin{figure} [H]
    \centering
    \includegraphics[width=\textwidth]{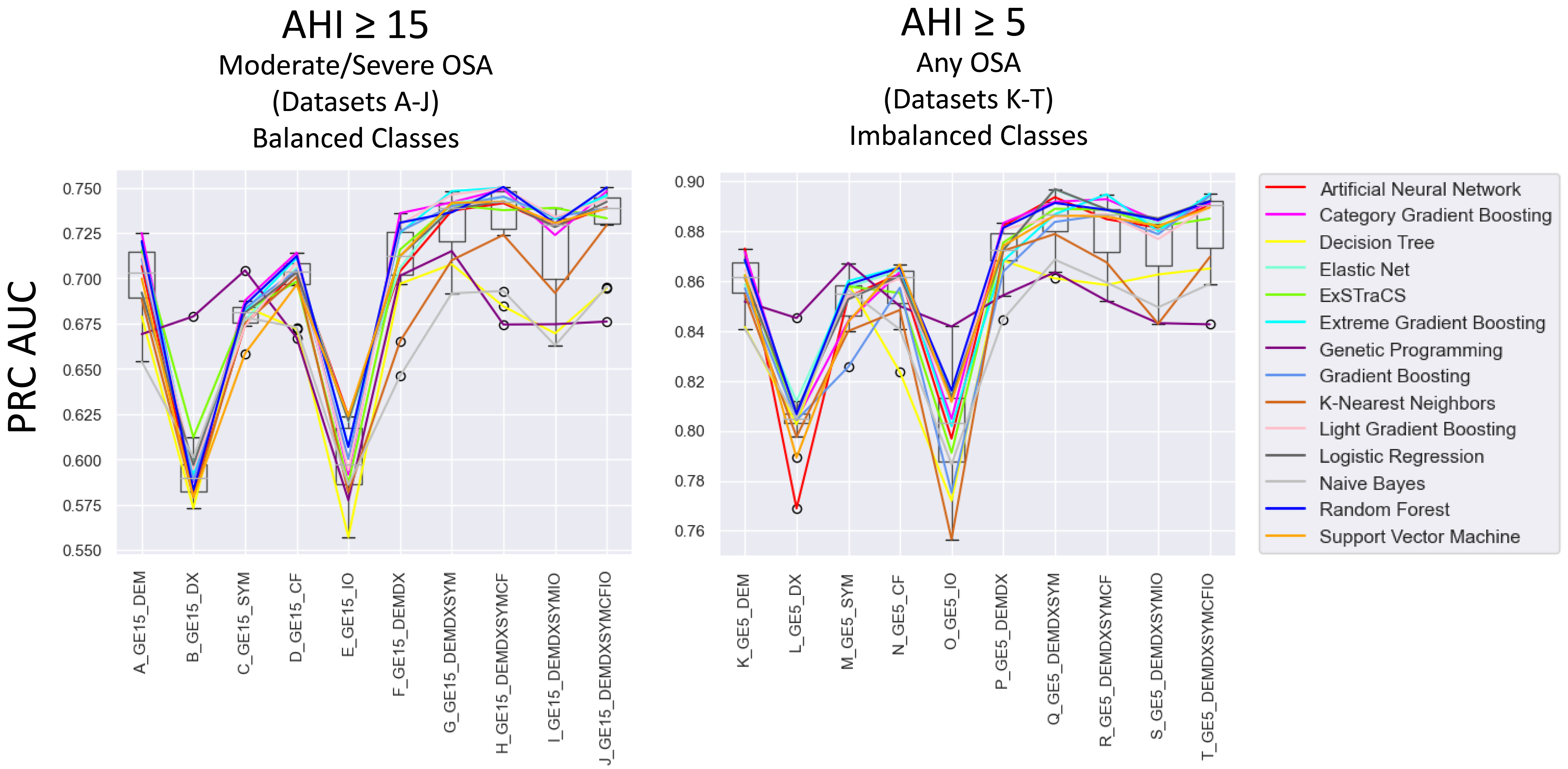}
    \caption*{\textbf{Figure S21: PRC-AUC Model Testing Evaluation Summary Across All OSA Datasets.} These 2 subplots report PRC-AUC in datasets where OSA outcome is defined by either AHI $\geq$ 15 or AHI $\geq$ 5. In each subplot the y-axis is not fixed (to improve algorithm performance visibility) and the x-axis is the target dataset (with dataset names specifying included feature sets, i.e. DEM, DX, SYM, CF, and IO). Each of the 20 individual boxplots give the spread of target metric performance (averaged over 10-fold CV models) across the 14 ML algorithms. Mean CV testing performance of a given algorithm is identified by the colored lines in the key. }
\end{figure}
\addcontentsline{toc}{subsubsection}{Figure S21: PRC-AUC Model Testing Evaluation Summary Across All OSA Datasets.}

\textbf{Table S16} and \textbf{Table S17} give pairwise performance metric comparisons between datasets having only a single feature set (i.e. A-E and K-O) identified DEM and CF features as being significantly more informative of OSA outcome (AHI$\geq$15 and AHI$\geq$5) than DX, SYM, or IO based on balanced accuracy and ROC-AUC. For PRC-AUC in AHI$\geq$15, DEM was significantly more informative than DX or IO and non-significantly better than SYM or CF. CF was significantly more informative than IO, and non-significantly better than SYM or DX. For PRC-AUC in AHI$\geq$5, DEM and CF were both significantly more informative than DX or IO, and non-significantly better than SYM. SYM was consistently (but not always significantly) higher than DX or IO for AHI$\geq$15 and AHI$\geq$5 for all three metrics. DEM and CF yielded similar performance for the three metrics and both outcome encodings. IO performed significantly better than DX for AHI$\geq$15 and AHI$\geq$5 using balanced accuracy and ROC-AUC, but yielded a significantly lower PRC-AUC for AHI$\geq$15 and a similar PRC-AUC for AHI$\geq$5. To generalize, DEM and CF were similarly most informative on their own, followed by SYM, IO, and lastly DX.

\begin{table} [H]
\begin{adjustbox}{center}
\scalebox{0.87}{
\begin{tabular}{| p{1.2cm} | l | l | l | l | p{1.7cm} | p{1.7cm} | p{1.8cm} | p{1.7cm} |}
\hline
\rowcolor{light-gray}
&&&&& Data 1 & Data 1 & Data 2 & Data 2 \\
\rowcolor{light-gray}
Metric & Data 1 & Data 2  & Statistic & p-value & Best &Median &Best&Median \\
\rowcolor{light-gray}
&&&&& Algorithm &  & Algorithm &  \\ \hline
\hline
\multirow{10}{*}{{\rotatebox[origin=c]{90}{\textbf{Balanced Accuracy}}}} & A (DEM) & B (DX) & 0 & \cellcolor{yellow!50}0.001953 & CGB & \cellcolor{green!25}0.674879 & GP & 0.55878 \\ \cline{2-9}
 & A (DEM) & C (SYM) & 0 & \cellcolor{yellow!50}0.001953 & CGB & \cellcolor{green!25}0.674879 & LGB & 0.623776 \\ \cline{2-9}
 & A (DEM) & \cellcolor{blue!25}D (CF) & 27 & 1 & CGB & 0.674879 & \cellcolor{blue!25}LR & \cellcolor{green!25}0.678416 \\ \cline{2-9}
 & A (DEM) & E (IO) & 0 & \cellcolor{yellow!50}0.001953 & CGB & \cellcolor{green!25}0.674879 & LR & 0.60168 \\ \cline{2-9}
 & B (DX) & C (SYM) & 0 & \cellcolor{yellow!50}0.001953 & GP & 0.55878 & LGB & \cellcolor{green!25}0.623776 \\ \cline{2-9}
 & B (DX) & \cellcolor{blue!25}D (CF) & 0 & \cellcolor{yellow!50}0.001953 & GP & 0.55878 & \cellcolor{blue!25}LR & \cellcolor{green!25}0.678416 \\ \cline{2-9}
 & B (DX) & E (IO) & 0 & \cellcolor{yellow!50}0.001953 & GP & 0.55878 & LR & \cellcolor{green!25}0.60168 \\ \cline{2-9}
 & C (SYM) & \cellcolor{blue!25}D (CF) & 1 & \cellcolor{yellow!50}0.003906 & LGB & 0.623776 & \cellcolor{blue!25}LR & \cellcolor{green!25}0.678416 \\ \cline{2-9}
 & C (SYM) & E (IO) & 13 & 0.160156 & LGB & \cellcolor{green!25}0.623776 & LR & 0.60168 \\ \cline{2-9}
 & \cellcolor{blue!25}D (CF) & E (IO) & 1 & \cellcolor{yellow!50}0.003906 & \cellcolor{blue!25}LR & \cellcolor{green!25}0.678416 & LR & 0.60168 \\
\hline \hline
\multirow{10}{*}{{\rotatebox[origin=c]{90}{\textbf{ROC-AUC}}}} & \cellcolor{blue!25}A (DEM) & B (DX) & 0 & \cellcolor{yellow!50}0.001953 & \cellcolor{blue!25}ExSTraCS & \cellcolor{green!25}0.738696 & LGB & 0.561623 \\ \cline{2-9}
 &\cellcolor{blue!25}A (DEM) & C (SYM) & 0 &\cellcolor{yellow!50} 0.001953 & \cellcolor{blue!25}ExSTraCS & \cellcolor{green!25}0.738696 & GB & 0.660045 \\ \cline{2-9}
 & \cellcolor{blue!25}A (DEM) & D (CF) & 21 & 0.556641 & \cellcolor{blue!25}ExSTraCS & \cellcolor{green!25}0.738696 & RF & 0.734304 \\ \cline{2-9}
 & \cellcolor{blue!25}A (DEM) & E (IO) & 0 & \cellcolor{yellow!50}0.001953 & \cellcolor{blue!25}ExSTraCS & \cellcolor{green!25}0.738696 & SVM & 0.640454 \\ \cline{2-9}
 & B (DX) & C (SYM) & 1 & \cellcolor{yellow!50}0.003906 & LGB & 0.561623 & GB & \cellcolor{green!25}0.660045 \\ \cline{2-9}
 & B (DX) & D (CF) & 0 & \cellcolor{yellow!50}0.001953 & LGB & 0.561623 & RF & \cellcolor{green!25}0.734304 \\ \cline{2-9}
 & B (DX) & E (IO) & 0 & \cellcolor{yellow!50}0.001953 & LGB & 0.561623 & SVM & \cellcolor{green!25}0.640454 \\ \cline{2-9}
 & C (SYM) & D (CF) & 0 & \cellcolor{yellow!50}0.001953 & GB & 0.660045 & RF & \cellcolor{green!25}0.734304 \\ \cline{2-9}
 & C (SYM) & E (IO) & 25 & 0.845703 & GB & \cellcolor{green!25}0.660045 & SVM & 0.640454 \\ \cline{2-9}
 & D (CF) & E (IO) & 0 & \cellcolor{yellow!50}0.001953 & RF & \cellcolor{green!25}0.734304 & SVM & 0.640454 \\
\hline \hline
\multirow{10}{*}{{\rotatebox[origin=c]{90}{\textbf{PRC-AUC}}}} & \cellcolor{blue!25}A (DEM) & B (DX) & 5 & \cellcolor{yellow!50}0.019531 & \cellcolor{blue!25}CGB & 0.\cellcolor{green!25}726915 & GP & 0.678069 \\ \cline{2-9}
 & \cellcolor{blue!25}A (DEM) & C (SYM) & 11 & 0.105469 & \cellcolor{blue!25}CGB & \cellcolor{green!25}0.726915 & GP & 0.708505 \\ \cline{2-9}
 & \cellcolor{blue!25}A (DEM) & D (CF) & 15 & 0.232422 & \cellcolor{blue!25}CGB & \cellcolor{green!25}0.726915 & XGB & 0.719624 \\ \cline{2-9}
 & \cellcolor{blue!25}A (DEM) & E (IO) & 0 & \cellcolor{yellow!50}0.001953 & \cellcolor{blue!25}CGB & \cellcolor{green!25}0.726915 & EN & 0.625999 \\ \cline{2-9}
 & B (DX) & C (SYM) & 10 & 0.083984 & GP & 0.678069 & GP & \cellcolor{green!25}0.708505 \\ \cline{2-9}
 & B (DX) & D (CF) & 14 & 0.193359 & GP & 0.678069 & XGB & \cellcolor{green!25}0.719624 \\ \cline{2-9}
 & B (DX) & E (IO) & 0 & \cellcolor{yellow!50}0.001953 & GP & \cellcolor{green!25}0.678069 & EN & 0.625999 \\ \cline{2-9}
 & C (SYM) & D (CF) & 25 & 0.845703 & GP & 0.708505 & XGB & \cellcolor{green!25}0.719624 \\ \cline{2-9}
 & C (SYM) & E (IO) & 0 & \cellcolor{yellow!50}0.001953 & GP & \cellcolor{green!25}0.708505 & EN & 0.625999 \\ \cline{2-9}
 & D (CF) & E (IO) & 2 & \cellcolor{yellow!50}0.005859 & XGB & \cellcolor{green!25}0.719624 & EN & 0.625999 \\
\hline
\end{tabular}}
\end{adjustbox}
\caption*{\textbf{Table S16: Pairwise Performance Comparisons of AHI$\geq$15 Datasets with One Feature Set.} Pairwise tests using the Wilcoxon Signed Rank Test. The testing metrics of the 10 CV models trained by the algorithm with the best median metric is the basis for this comparison. Metrics in this table include balanced accuracy, ROC-AUC, and PRC-AUC. Cells in green denote the larger median performance of the pair. Rows with p-value highlighted in yellow fall under the 0.05 significance cutoff. The cells in blue point out the dataset/algorithm combination where the highest median metric value was observed. ‘GP->Algorithm’ indicates where GP was replaced by algorithm with next highest median performance (new algorithm metrics given).}
\end{table}
\addcontentsline{toc}{subsubsection}{Table S16: Pairwise Performance Comparisons of AHI$\geq$15 Datasets with One Feature Set.}

\begin{table} [H]
\begin{adjustbox}{center}
\scalebox{0.87}{
\begin{tabular}{| p{1.2cm} | l | l | l | l | p{1.7cm} | p{1.7cm} | p{1.8cm} | p{1.7cm} |}
\hline
\rowcolor{light-gray}
&&&&& Data 1 & Data 1 & Data 2 & Data 2 \\
\rowcolor{light-gray}
Metric & Data 1 & Data 2  & Statistic & p-value & Best &Median &Best&Median \\
\rowcolor{light-gray}
&&&&& Algorithm &  & Algorithm &  \\ \hline \hline
\multirow{10}{*}{{\rotatebox[origin=c]{90}{\textbf{Balanced Accuracy}}}} & \cellcolor{blue!25}K (DEM) & L (DX) & 0 & \cellcolor{yellow!50}0.001953 & \cellcolor{blue!25}XGB & \cellcolor{green!25}0.702773 & XGB & 0.57963 \\ \cline{2-9}
 & \cellcolor{blue!25}K (DEM) & M (SYM) & 6 & \cellcolor{yellow!50}0.027344 & \cellcolor{blue!25}XGB & \cellcolor{green!25}0.702773 & XGB & 0.654582 \\ \cline{2-9}
 & \cellcolor{blue!25}K (DEM) & N (CF) & 18 & 0.375 & \cellcolor{blue!25}XGB & \cellcolor{green!25}0.702773 & XGB & 0.693432 \\ \cline{2-9}
 & \cellcolor{blue!25}K (DEM) & O (IO) & 2 & \cellcolor{yellow!50}0.005859 & \cellcolor{blue!25}XGB & \cellcolor{green!25}0.702773 & EN & 0.612221 \\ \cline{2-9}
 & L (DX) & M (SYM) & 0 & \cellcolor{yellow!50}0.001953 & XGB & 0.57963 & XGB & \cellcolor{green!25}0.654582 \\ \cline{2-9}
 & L (DX) & N (CF) & 0 & \cellcolor{yellow!50}0.001953 & XGB & 0.57963 & XGB & \cellcolor{green!25}0.693432 \\ \cline{2-9}
 & L (DX) & O (IO) & 3 & \cellcolor{yellow!50}0.009766 & XGB & 0.57963 & EN & \cellcolor{green!25}0.612221 \\ \cline{2-9}
 & M (SYM) & N (CF) & 13 & 0.160156 & XGB & 0.654582 & XGB & \cellcolor{green!25}0.693432 \\ \cline{2-9}
 & M (SYM) & O (IO) & 6 & \cellcolor{yellow!50}0.027344 & XGB & \cellcolor{green!25}0.654582 & EN & 0.612221 \\ \cline{2-9}
 & N (CF) & O (IO) & 5 & \cellcolor{yellow!50}0.019531 & XGB & \cellcolor{green!25}0.693432 & EN & 0.612221 \\ 
\hline \hline
\multirow{10}{*}{{\rotatebox[origin=c]{90}{\textbf{ROC-AUC}}}} & \cellcolor{blue!25}K (DEM) & L (DX) & 0 & \cellcolor{yellow!50}0.001953 & \cellcolor{blue!25}RF & \cellcolor{green!25}0.768363 & LGB & 0.58465 \\ \cline{2-9}
 & \cellcolor{blue!25}K (DEM) & M (SYM) & 1 & \cellcolor{yellow!50}0.003906 & \cellcolor{blue!25}RF & \cellcolor{green!25}0.768363 & EN & 0.679587 \\ \cline{2-9}
 & \cellcolor{blue!25}K (DEM) & N (CF) & 21 & 0.556641 & \cellcolor{blue!25}RF & \cellcolor{green!25}0.768363 & XGB & 0.767255 \\ \cline{2-9}
 & \cellcolor{blue!25}K (DEM) & O (IO) & 0 & \cellcolor{yellow!50}0.001953 & \cellcolor{blue!25}RF & \cellcolor{green!25}0.768363 & RF & 0.653842 \\ \cline{2-9}
 & L (DX) & M (SYM) & 0 & \cellcolor{yellow!50}0.001953 & LGB & 0.58465 & EN & \cellcolor{green!25}0.679587 \\ \cline{2-9}
 & L (DX) & N (CF) & 0 & \cellcolor{yellow!50}0.001953 & LGB & 0.58465 & XGB & \cellcolor{green!25}0.767255 \\ \cline{2-9}
 & L (DX) & O (IO) & 3 & \cellcolor{yellow!50}0.009766 & LGB & 0.58465 & RF & \cellcolor{green!25}0.653842 \\ \cline{2-9}
 & M (SYM) & N (CF) & 3 & \cellcolor{yellow!50}0.009766 & EN & 0.679587 & XGB & \cellcolor{green!25}0.767255 \\ \cline{2-9}
 & M (SYM) & O (IO) & 12 & 0.130859 & EN & \cellcolor{green!25}0.679587 & RF & 0.653842 \\ \cline{2-9}
 & N (CF) & O (IO) & 0 & \cellcolor{yellow!50}0.001953 & XGB & \cellcolor{green!25}0.767255 & RF & 0.653842 \\
\hline \hline
\multirow{10}{*}{{\rotatebox[origin=c]{90}{\textbf{PRC-AUC}}}} & K (DEM) & L (DX) & 3 & \cellcolor{yellow!50}0.000669 & CGB & \cellcolor{green!25}0.873214 & GP->DT & 0.819413 \\ \cline{2-9}
 & K (DEM) & M (SYM) & 1 & 0.150926 & CGB & \cellcolor{green!25}0.873214 & GP->LGB & 0.862589 \\ \cline{2-9}
 & K (DEM) & \cellcolor{blue!25}N (CF) & 23 & 0.695312 & CGB & 0.873214 & \cellcolor{blue!25}RF & \cellcolor{green!25}0.87526 \\ \cline{2-9}
 & K (DEM) & O (IO) & 3 & \cellcolor{yellow!50}0.0008807 & CGB & \cellcolor{green!25}0.873214 & GP->RF & 0.815426 \\ \cline{2-9}
 & L (DX) & M (SYM) & 3 & \cellcolor{yellow!50}0.005158 & GP->DT & 0.819413 & GP->LGB & \cellcolor{green!25}0.862589 \\ \cline{2-9}
 & L (DX) & \cellcolor{blue!25}N (CF) & 3 & \cellcolor{yellow!50}0.001498 & GP->DT & 0.819413 & \cellcolor{blue!25}RF & \cellcolor{green!25}0.87526 \\ \cline{2-9}
 & L (DX) & O (IO) & 0 & 0.939742 & GP->DT & \cellcolor{green!25}0.819413 & GP->RF & 0.815426 \\ \cline{2-9}
 & M (SYM) & \cellcolor{blue!25}N (CF) & 1 & 0.405678 & GP->LGB & 0.862589 & \cellcolor{blue!25}RF & \cellcolor{green!25}0.87526 \\ \cline{2-9}
 & M (SYM) & O (IO) & 2 & \cellcolor{yellow!50}0.012611 & GP->LGB & \cellcolor{green!25}0.862589 & GP->RF & 0.815426 \\ \cline{2-9}
 & \cellcolor{blue!25}N (CF) & O (IO) & 3 & \cellcolor{yellow!50}0.001498 & \cellcolor{blue!25}RF & \cellcolor{green!25}0.87526 & GP->RF & 0.815426 \\
\hline
\end{tabular}}
\end{adjustbox}
\caption*{\textbf{Table S17: Pairwise Performance Comparisons of AHI$\geq$5 Datasets with One Feature Set.} Pairwise tests using the Wilcoxon Signed Rank Test. The testing metrics of the 10 CV models trained by the algorithm with the best median metric is the basis for this comparison. Metrics in this table include balanced accuracy, ROC-AUC, and PRC-AUC. Cells in green denote the larger median performance of the pair. Rows with p-value highlighted in yellow fall under the 0.05 significance cutoff. The cells in blue point out the dataset/algorithm combination where the highest median metric value was observed. ‘GP -> Algorithm’ indicates where GP was replaced by algorithm with next highest median performance (new algorithm metrics given). }
\end{table}
\addcontentsline{toc}{subsubsection}{Table S17: Pairwise Performance Comparisons of AHI$\geq$5 Datasets with One Feature Set.}

Next, we examined pairwise performance metric comparisons between datasets progressively adding feature sets (i.e. DEM, +DX, +SYM, +CF, +IO). \textbf{Figure 3} (main text) focuses on these specific comparisons (narrowing the results from \textbf{Figure 2} – main text), using ROC-AUC for AHI$\geq$15 and PRC-AUC for AHI$\geq$5. For AHI$\geq$15 datasets, the only step-wise feature set addition that led to a significant increase in ROC-AUC was CF added to DEM+DX+SYM (dataset G vs. H; p=0.0137). Significant ROC-AUC increases were also observed between (1) A vs. either H or J, (2) F vs. either H or J, and (3) G vs. J (see \textbf{Table S18}). The highest overall median ROC-AUC was obtained on dataset H (DEM+DX+SYM+CF) using the artificial neural network (ANN) algorithm. \textbf{Figure 3B} (main text) reveals a similar pattern of ROC-AUC performance across AHI$\geq$15 datasets when exclusively focusing on the 10 ANN models trained for each CV partition.

\begin{table} [H]
\begin{adjustbox}{center}
\scalebox{0.79}{
\begin{tabular}{| p{1.2cm} | l | l | l | l | p{1.7cm} | p{1.7cm} | p{1.8cm} | p{1.7cm} |}
\hline
\rowcolor{light-gray}
&&&&& Data 1 & Data 1 & Data 2 & Data 2 \\
\rowcolor{light-gray}
Metric & Data 1 & Data 2  & Statistic & p-value & Best &Median &Best&Median \\
\rowcolor{light-gray}
&&&&& Algorithm &  & Algorithm &  \\ \hline \hline
\multirow{10}{*}{{\rotatebox[origin=c]{90}{Balanced Accuracy}}}  & A (DEM) & F (DEMDX) & 19 & 0.431641 & CGB & \cellcolor{green!25}0.674879 & XGB & 0.661979 \\ \cline{2-9}
 & F (DEMDX) & G (DEMDXSYM) & 17 & 0.322266 & XGB & 0.661979 & ExSTraCS & \cellcolor{green!25}0.690846 \\ \cline{2-9}
 & G (DEMDXSYM) & \cellcolor{blue!25}H (DEMDXSYMCF) & 6 & \cellcolor{yellow!50}0.027344 & ExSTraCS & 0.690846 & \cellcolor{blue!25}ExSTraCS & \cellcolor{green!25}0.694347 \\ \cline{2-9}
 & \cellcolor{blue!25}H (DEMDXSYMCF) & J (DEMDXSYMCFIO) & 23 & 0.695312 & \cellcolor{blue!25}ExSTraCS & \cellcolor{green!25}0.694347 & CGB & 0.692929 \\ \cline{2-9}
 & A (DEM) & G (DEMDXSYM) & 17 & 0.322266 & CGB & 0.674879 & ExSTraCS & \cellcolor{green!25}0.690846 \\ \cline{2-9}
 & A (DEM) & \cellcolor{blue!25}H (DEMDXSYMCF) & 7 & \cellcolor{yellow!50}0.037109 & CGB & 0.674879 & \cellcolor{blue!25}ExSTraCS & \cellcolor{green!25}0.694347 \\ \cline{2-9}
 & A (DEM) & J (DEMDXSYMCFIO) & 2 & \cellcolor{yellow!50}0.005859 & CGB & 0.674879 & CGB & \cellcolor{green!25}0.692929 \\ \cline{2-9}
 & F (DEMDX) & \cellcolor{blue!25}H (DEMDXSYMCF) & 10 & 0.083984 & XGB & 0.661979 & \cellcolor{blue!25}ExSTraCS & \cellcolor{green!25}0.694347 \\ \cline{2-9}
 & F (DEMDX) & J (DEMDXSYMCFIO) & 9 & 0.064453 & XGB & 0.661979 & CGB &\cellcolor{green!25} 0.692929 \\ \cline{2-9}
 & G (DEMDXSYM) & J (DEMDXSYMCFIO) & 8 & \cellcolor{yellow!50}0.048828 & ExSTraCS & 0.690846 & CGB & \cellcolor{green!25}0.692929 \\ \cline{2-9}
 & \cellcolor{blue!25}H (DEMDXSYMCF) & I (DEMDXSYMIO) & 20 & 0.492188 & \cellcolor{blue!25}ExSTraCS & \cellcolor{green!25}0.694347 & ExSTraCS & 0.685458 \\ \cline{2-9}
 & I (DEMDXSYMIO) & J (DEMDXSYMCFIO) & 14 & 0.193359 & ExSTraCS & 0.685458 & CGB & \cellcolor{green!25}0.692929 \\
\hline \hline
\multirow{10}{*}{{\rotatebox[origin=c]{90}{\textbf{ROC-AUC}}}} & A (DEM) & F (DEMDX) & 14 & 0.193359 & ExSTraCS & \cellcolor{green!25}0.738696 & LGB & 0.720548 \\ \cline{2-9}
& F (DEMDX) & G (DEMDXSYM) & 11 & 0.105469 & LGB & 0.720548 & LGB & \cellcolor{green!25}0.742128 \\ \cline{2-9}
 & G (DEMDXSYM) & \cellcolor{blue!25}H (DEMDXSYMCF) & 4 & \cellcolor{yellow!50}0.013672 & LGB & 0.742128 & \cellcolor{blue!25}ANN & \cellcolor{green!25}0.766822 \\ \cline{2-9}
& \cellcolor{blue!25}H (DEMDXSYMCF) & J (DEMDXSYMCFIO) & 15 & 0.232422 & \cellcolor{blue!25}ANN & \cellcolor{green!25}0.766822 & ANN & 0.760819 \\ \cline{2-9}
 & A (DEM) & G (DEMDXSYM) & 24 & 0.769531 & ExSTraCS & 0.738696 & LGB & \cellcolor{green!25}0.742128 \\ \cline{2-9}
 & A (DEM) & \cellcolor{blue!25}H (DEMDXSYMCF) & 0 & \cellcolor{yellow!50}0.001953 & ExSTraCS & 0.738696 & \cellcolor{blue!25}ANN & \cellcolor{green!25}0.766822 \\ \cline{2-9}
 & A (DEM) & J (DEMDXSYMCFIO) & 0 & \cellcolor{yellow!50}0.001953 & ExSTraCS & 0.738696 & ANN & \cellcolor{green!25}0.760819 \\ \cline{2-9}
 & F (DEMDX) & \cellcolor{blue!25}H (DEMDXSYMCF) & 4 & \cellcolor{yellow!50}0.013672 & LGB & 0.720548 & \cellcolor{blue!25}ANN & \cellcolor{green!25}0.766822 \\ \cline{2-9}
 & F (DEMDX) & J (DEMDXSYMCFIO) & 3 & \cellcolor{yellow!50}0.009766 & LGB & 0.720548 & ANN & \cellcolor{green!25}0.760819 \\ \cline{2-9}
 & G (DEMDXSYM) & J (DEMDXSYMCFIO) & 8 & \cellcolor{yellow!50}0.048828 & LGB & 0.742128 & ANN & \cellcolor{green!25}0.760819 \\ \cline{2-9}
 & \cellcolor{blue!25}H (DEMDXSYMCF) & I (DEMDXSYMIO) & 15 & 0.232422 & \cellcolor{blue!25}ANN & \cellcolor{green!25}0.766822 & ANN & 0.7554 \\ \cline{2-9}
 & I (DEMDXSYMIO) & J (DEMDXSYMCFIO) & 19 & 0.431641 & LGB & 0.7554 & ANN & \cellcolor{green!25}0.760819 \\
\hline \hline
\multirow{10}{*}{{\rotatebox[origin=c]{90}{PRC-AUC}}} & A (DEM) & F (DEMDX) & 23 & 0.695312 & CGB & 0.726915 & RF & \cellcolor{green!25}0.735926 \\ \cline{2-9}
 & F (DEMDX) & G (DEMDXSYM) & 19 & 0.431641 & RF & 0.735926 & LGB & \cellcolor{green!25}0.749936 \\ \cline{2-9}
 & G (DEMDXSYM) & H (DEMDXSYMCF) & 25 & 0.845703 & LGB & 0.749936 & CGB & \cellcolor{green!25}0.752545 \\ \cline{2-9}
 & H (DEMDXSYMCF) & \cellcolor{blue!25}J (DEMDXSYMCFIO) & 25 & 0.845703 & CGB & 0.752545 & \cellcolor{blue!25}RF & \cellcolor{green!25}0.756443 \\ \cline{2-9}
 & A (DEM) & G (DEMDXSYM) & 7 & \cellcolor{yellow!50}0.037109 & CGB & 0.726915 & LGB & \cellcolor{green!25}0.749936 \\ \cline{2-9}
 & A (DEM) & H (DEMDXSYMCF) & 9 & 0.064453 & CGB & 0.726915 & CGB & \cellcolor{green!25}0.752545 \\ \cline{2-9}
 & A (DEM) & \cellcolor{blue!25}J (DEMDXSYMCFIO) & 6 & \cellcolor{yellow!50}0.027344 & CGB & 0.726915 & \cellcolor{blue!25}RF & \cellcolor{green!25}0.756443 \\ \cline{2-9}
 & F (DEMDX) & H (DEMDXSYMCF) & 17 & 0.322266 & RF & 0.735926 & CGB & \cellcolor{green!25}0.752545 \\ \cline{2-9}
 & F (DEMDX) & \cellcolor{blue!25}J (DEMDXSYMCFIO) & 15 & 0.232422 & RF & 0.735926 & \cellcolor{blue!25}RF & \cellcolor{green!25}0.756443 \\ \cline{2-9}
 & G (DEMDXSYM) & \cellcolor{blue!25}J (DEMDXSYMCFIO) & 25 & 0.845703 & LGB & 0.749936 & \cellcolor{blue!25}RF & \cellcolor{green!25}0.756443 \\ \cline{2-9}
 & H (DEMDXSYMCF) & I (DEMDXSYMIO) & 6 & \cellcolor{yellow!50}0.027344 & CGB & \cellcolor{green!25}0.752545 & XGB & 0.732568 \\ \cline{2-9}
 & I (DEMDXSYMIO) & \cellcolor{blue!25}J (DEMDXSYMCFIO) & 9 & 0.064453 & XGB & 0.732568 & \cellcolor{blue!25}RF & \cellcolor{green!25}0.756443 \\
\hline
\end{tabular}}
\end{adjustbox}
\caption*{\textbf{Table S18: Pairwise Performance Comparisons of AHI$\geq$15 Datasets Progressively Adding Feature Sets.} Pairwise tests using the Wilcoxon Signed Rank Test. The testing metrics of the 10 CV models trained by the algorithm with the best median metric is the basis for this comparison. Metrics in this table include balanced accuracy, ROC-AUC, and PRC-AUC. Cells in green denote the larger median performance of the pair. Rows with p-value highlighted in yellow fall under the 0.05 significance cutoff. The cells in blue point out the dataset/algorithm combination where the highest median metric value was observed.}
\end{table}
\addcontentsline{toc}{subsubsection}{Table S18: Pairwise Performance Comparisons of AHI$\geq$15 Datasets Progressively Adding Feature Sets.}

 Shifting to AHI$\geq$5 datasets, no step-wise feature set additions led to a significant increase in PRC-AUC. However, we observe consistently increasing median best-algorithm performance with the addition of each subsequent feature set, with CF and IO yielding minimal increases. The only significant PRC-AUC increases were observed between datasets K and R (i.e. DEM vs. DEM+DX+SYM+CF) and K and T (i.e. DEM vs. DEM+DX+SYM+CF+IO) (see \textbf{Table S19}). The highest overall median PRC-AUC across both AHI$\geq$5 and AHI$\geq$15 datasets was obtained on dataset T (DEM+DX+SYM+CF+IO) using elastic net. Figure 3D reveals a similar pattern of PRC-AUC performance across AHI$\geq$5 datasets when exclusively focusing on the 10 elastic net models trained for each CV partition. In comparing all algorithm PRC-AUCs on dataset T, elastic net performed significantly better than decision tree, genetic programming, K-nearest neighbors, and Naïve Bayes algorithms, and similarly well to the rest (see \textbf{Table S20}). 

\begin{table} [H]
\begin{adjustbox}{center}
\scalebox{0.8}{
\begin{tabular}{| p{1.2cm} | l | l | l | l | p{1.7cm} | p{1.7cm} | p{1.8cm} | p{1.7cm} |}
\hline
\rowcolor{light-gray}
&&&&& Data 1 & Data 1 & Data 2 & Data 2 \\
\rowcolor{light-gray}
Metric & Data 1 & Data 2  & Statistic & p-value & Best &Median &Best&Median \\
\rowcolor{light-gray}
&&&&& Algorithm &  & Algorithm &  \\ \hline \hline
\multirow{10}{*}{{\rotatebox[origin=c]{90}{Balanced Accuracy}}} & K (DEM) & P (DEMDX) & 18 & 0.375 & XGB & \cellcolor{green!25}0.702773 & SVM & 0.688487 \\ \cline{2-9}
 & P (DEMDX) & Q (DEMDXSYM) & 16 & 0.275391 & SVM & 0.688487 & ExSTraCS & \cellcolor{green!25}0.695652 \\ \cline{2-9}
 & Q (DEMDXSYM) & R (DEMDXSYMCF) & 14 & 0.193359 & ExSTraCS & 0.695652 & ExSTraCS & \cellcolor{green!25}0.718696 \\ \cline{2-9}
 & R (DEMDXSYMCF) & \cellcolor{blue!25}T (DEMDXSYMCFIO) & 21 & 0.556641 & ExSTraCS & 0.718696 & \cellcolor{blue!25}EN & \cellcolor{green!25}0.718936 \\ \cline{2-9}
 & K (DEM) & Q (DEMDXSYM) & 22 & 0.625 & XGB & \cellcolor{green!25}0.702773 & ExSTraCS & 0.695652 \\ \cline{2-9}
 & K (DEM) & R (DEMDXSYMCF) & 11 & 0.105469 & XGB & 0.702773 & ExSTraCS & \cellcolor{green!25}0.718696 \\ \cline{2-9}
 & K (DEM) & \cellcolor{blue!25}T (DEMDXSYMCFIO) & 14 & 0.193359 & XGB & 0.702773 & \cellcolor{blue!25}EN & \cellcolor{green!25}0.718936 \\ \cline{2-9}
 & P (DEMDX) & R (DEMDXSYMCF) & 11 & 0.105469 & SVM & 0.688487 & ExSTraCS & \cellcolor{green!25}0.718696 \\ \cline{2-9}
 & P (DEMDX) & \cellcolor{blue!25}T (DEMDXSYMCFIO) & 11 & 0.105469 & SVM & 0.688487 & \cellcolor{blue!25}EN & \cellcolor{green!25}0.718936 \\ \cline{2-9}
 & Q (DEMDXSYM) & \cellcolor{blue!25}T (DEMDXSYMCFIO) & 17 & 0.322266 & ExSTraCS & 0.695652 & \cellcolor{blue!25}EN & \cellcolor{green!25}0.718936 \\ \cline{2-9}
 & R (DEMDXSYMCF) & S (DEMDXSYMIO) & 22 & 0.625 & ExSTraCS & \cellcolor{green!25}0.718696 & ExSTraCS & 0.709368 \\ \cline{2-9}
 & S (DEMDXSYMIO) & \cellcolor{blue!25}T (DEMDXSYMCFIO) & 24 & 0.769531 & ExSTraCS & 0.709368 & \cellcolor{blue!25}EN & \cellcolor{green!25}0.718936 \\
\hline \hline
\multirow{10}{*}{{\rotatebox[origin=c]{90}{ROC-AUC}}} & K (DEM) & P (DEMDX) & 25 & 0.845703 & RF & \cellcolor{green!25}0.768363 & ANN & 0.761928 \\ \cline{2-9}
 & P (DEMDX) & Q (DEMDXSYM) & 21 & 0.556641 & ANN & 0.761928 & EN & \cellcolor{green!25}0.761972 \\ \cline{2-9}
 & Q (DEMDXSYM) & \cellcolor{blue!25}R (DEMDXSYMCF) & 14 & 0.193359 & EN & 0.761972 & \cellcolor{blue!25}XGB & \cellcolor{green!25}0.79297 \\ \cline{2-9}
 & \cellcolor{blue!25}R (DEMDXSYMCF) & T (DEMDXSYMCFIO) & 20 & 0.492188 & \cellcolor{blue!25}XGB & \cellcolor{green!25}0.79297 & GB & 0.792941 \\ \cline{2-9}
 & K (DEM) & Q (DEMDXSYM) & 22 & 0.625 & RF & \cellcolor{green!25}0.768363 & EN & 0.761972 \\ \cline{2-9}
 & K (DEM) & \cellcolor{blue!25}R (DEMDXSYMCF) & 13 & 0.160156 & RF & 0.768363 & \cellcolor{blue!25}XGB & \cellcolor{green!25}0.79297 \\ \cline{2-9}
 & K (DEM) & T (DEMDXSYMCFIO) & 15 & 0.232422 & RF & 0.768363 & GB & \cellcolor{green!25}0.792941 \\ \cline{2-9}
 & P (DEMDX) & \cellcolor{blue!25}R (DEMDXSYMCF) & 7 & \cellcolor{yellow!50}0.037109 & ANN & 0.761928 & \cellcolor{blue!25}XGB & \cellcolor{green!25}0.79297 \\ \cline{2-9}
 & P (DEMDX) & T (DEMDXSYMCFIO) & 10 & 0.083984 & ANN & 0.761928 & GB & \cellcolor{green!25}0.792941 \\ \cline{2-9}
 & Q (DEMDXSYM) & T (DEMDXSYMCFIO) & 14 & 0.193359 & EN & 0.761972 & GB & \cellcolor{green!25}0.792941 \\ \cline{2-9}
 & \cellcolor{blue!25}R (DEMDXSYMCF) & S (DEMDXSYMIO) & 20 & 0.492188 & \cellcolor{blue!25}XGB & \cellcolor{green!25}0.79297 & CGB & 0.778981 \\ \cline{2-9}
 & S (DEMDXSYMIO) & T (DEMDXSYMCFIO) & 22 & 0.625 & CGB & 0.778981 & GB & \cellcolor{green!25}0.792941 \\
\hline \hline
\multirow{10}{*}{{\rotatebox[origin=c]{90}{\textbf{PRC-AUC}}}} & K (DEM) & P (DEMDX) & 17 & 0.322266 & CGB & 0.873214 & CGB & \cellcolor{green!25}0.889826 \\ \cline{2-9}
 & P (DEMDX) & Q (DEMDXSYM) & 21 & 0.556641 & CGB & 0.889826 & EN & \cellcolor{green!25}0.89404 \\ \cline{2-9}
 & Q (DEMDXSYM) & R (DEMDXSYMCF) & 24 & 0.769531 & EN & 0.89404 & XGB & \cellcolor{green!25}0.894904 \\ \cline{2-9}
 & R (DEMDXSYMCF) & \cellcolor{blue!25}T (DEMDXSYMCFIO) & 25 & 0.845703 & XGB & 0.894904 & \cellcolor{blue!25}EN & \cellcolor{green!25}0.896618 \\ \cline{2-9}
 & K (DEM) & Q (DEMDXSYM) & 9 & 0.064453 & CGB & 0.873214 & EN &\cellcolor{green!25} 0.89404 \\ \cline{2-9}
 & K (DEM) & R (DEMDXSYMCF) & 7 & \cellcolor{yellow!50}0.037109 & CGB & 0.873214 & XGB & \cellcolor{green!25}0.894904 \\ \cline{2-9}
 & K (DEM) & \cellcolor{blue!25}T (DEMDXSYMCFIO) & 8 & \cellcolor{yellow!50}0.048828 & CGB & 0.873214 & \cellcolor{blue!25}EN &\cellcolor{green!25} 0.896618 \\ \cline{2-9}
 & P (DEMDX) & R (DEMDXSYMCF) & 15 & 0.232422 & CGB & 0.889826 & XGB & \cellcolor{green!25}0.894904 \\ \cline{2-9}
 & P (DEMDX) & \cellcolor{blue!25}T (DEMDXSYMCFIO) & 19 & 0.431641 & CGB & 0.889826 & \cellcolor{blue!25}EN & \cellcolor{green!25}0.896618 \\ \cline{2-9}
 & Q (DEMDXSYM) & \cellcolor{blue!25}T (DEMDXSYMCFIO) & 22 & 0.625 & EN & 0.89404 & \cellcolor{blue!25}EN & \cellcolor{green!25}0.896618 \\ \cline{2-9}
 & R (DEMDXSYMCF) & S (DEMDXSYMIO) & 21 & 0.556641 & XGB & \cellcolor{green!25}0.894904 & SVM & 0.892398 \\ \cline{2-9}
 & S (DEMDXSYMIO) & \cellcolor{blue!25}T (DEMDXSYMCFIO) & 23 & 0.695312 & SVM & 0.892398 & \cellcolor{blue!25}EN & \cellcolor{green!25}0.896618 \\
\hline
\end{tabular}}
\end{adjustbox}
\caption*{\textbf{Table S19: Pairwise Performance Comparisons of AHI$\geq$5 Datasets Progressively Adding Feature Sets. }
Pairwise tests using the Wilcoxon Signed Rank Test. The testing metrics of the 10 CV models trained by the algorithm with the best median metric is the basis for this comparison. Metrics in this table include balanced accuracy, ROC-AUC, and PRC-AUC. Cells in green denote the larger median performance of the pair. Rows with p-value highlighted in yellow fall under the 0.05 significance cutoff. The cells in blue point out the dataset/algorithm combination where the highest median metric value was observed.}
\end{table}
\addcontentsline{toc}{subsubsection}{Table S19: Pairwise Performance Comparisons of AHI$\geq$5 Datasets Progressively Adding Feature Sets.}

\begin{table} [H]
\begin{adjustbox}{center}
\scalebox{0.8}{
\begin{tabular}{| l | l | l | l | l | l |}
\hline
\rowcolor{light-gray}
\textbf{Algorithm 1} & \textbf{Algorithm 1}
\textbf{Median PRC-AUC} & \textbf{Algorithm 2} & \textbf{Algorithm 2}
\textbf{Median PRC-AUC} & \textbf{Statistic} & \textbf{p-value} \\
\hline
EN & 0.896618 & ANN & 0.893487 & 18 & 0.375 \\
\hline
EN & 0.896618 & CGB & 0.889757 & 16 & 0.275391 \\
\hline
EN & 0.896618 & DT & 0.871028 & 0 & \cellcolor{yellow!50}0.001953 \\
\hline
EN & 0.896618 & LR & 0.896397 & 23 & 0.695313 \\
\hline
EN & 0.896618 & ExSTraCS & 0.889757 & 11 & 0.105469 \\
\hline
EN & 0.896618 & XGB & 0.871028 & 25 & 0.845703 \\
\hline
EN & 0.896618 & GP & 0.889757 & 1 & \cellcolor{yellow!50}0.003906 \\
\hline
EN & 0.896618 & GB & 0.871028 & 25 & 0.845703 \\
\hline
EN & 0.896618 & KNN & 0.889757 & 2 & \cellcolor{yellow!50}0.005859 \\
\hline
EN & 0.896618 & LGB & 0.871028 & 17 & 0.322266 \\
\hline
EN & 0.896618 & NB & 0.861903 & 0 & \cellcolor{yellow!50}0.001953 \\
\hline
EN & 0.896618 & RF & 0.887067 & 16 & 0.275391 \\
\hline
EN & 0.896618 & SVM & 0.895224 & 15 & 0.232422 \\
\hline
\end{tabular}}
\end{adjustbox}
\caption*{\textbf{Table S20: T\_GE5\_DEMDXSYMCFIO Algorithm Performance Comparison.} Wilcoxon Rank Sum comparisons are made in contrast with elastic net (EN) as the top median performer. Comparisons highlighted in yellow had a p-value lower than the 0.05 cutoff.}
\end{table}
\addcontentsline{toc}{subsubsection}{Table S20: T\_GE5\_DEMDXSYMCFIO Algorithm Performance Comparison.}

Looking more closely at dataset T (DEM+DX+SYM+CF+IO) which yielded the top study-wide median PRC-AUC, \textbf{Figure S22} examines ROC-AUC across all algorithms, and \textbf{Figure S23} examines PRC-AUC across all algorithms. Median performance of all algorithms across 16 classification metrics are given in \textbf{Table S21}. Mean performance of all algorithms across 16 classification metrics are given in \textbf{Table S22}. \textbf{Figure S24} more closely examines EN testing performance across individual 10-fold CV trained models with ROC and PRC plots. 

\begin{figure} [H]
    \centering
    \includegraphics[width=\textwidth]{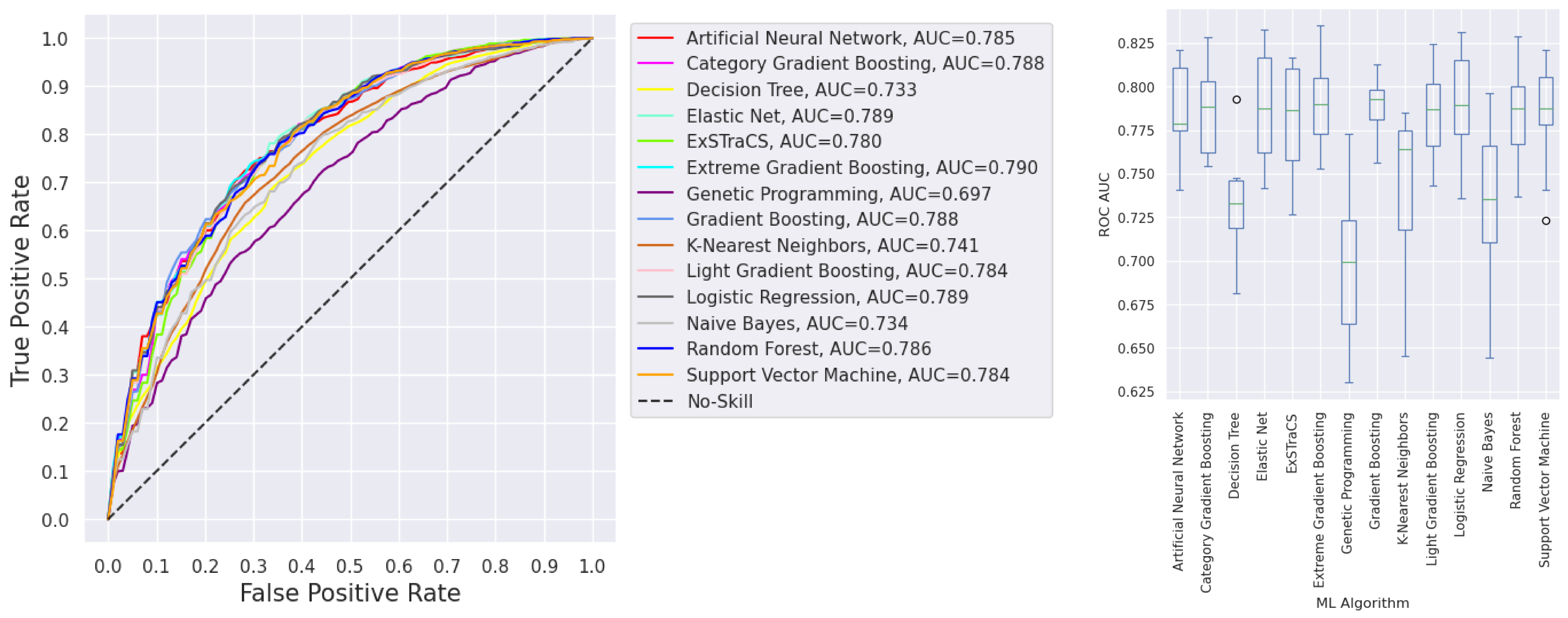}
    \caption*{\textbf{Figure S22: T\_GE5\_DEMDXSYMCFIO ROC Performance Across Algorithms.} (Left) ROC plot comparing 14 algorithm testing performance based on the average of 10-fold CV models trained for each algorithm. (Right) Boxplot of ROC-AUC testing performance for each algorithm across 10-fold CV models.}
\end{figure}
\addcontentsline{toc}{subsubsection}{Figure S22: T\_GE5\_DEMDXSYMCFIO ROC Performance Across Algorithms.}

\begin{figure} [H]
    \centering
    \includegraphics[width=\textwidth]{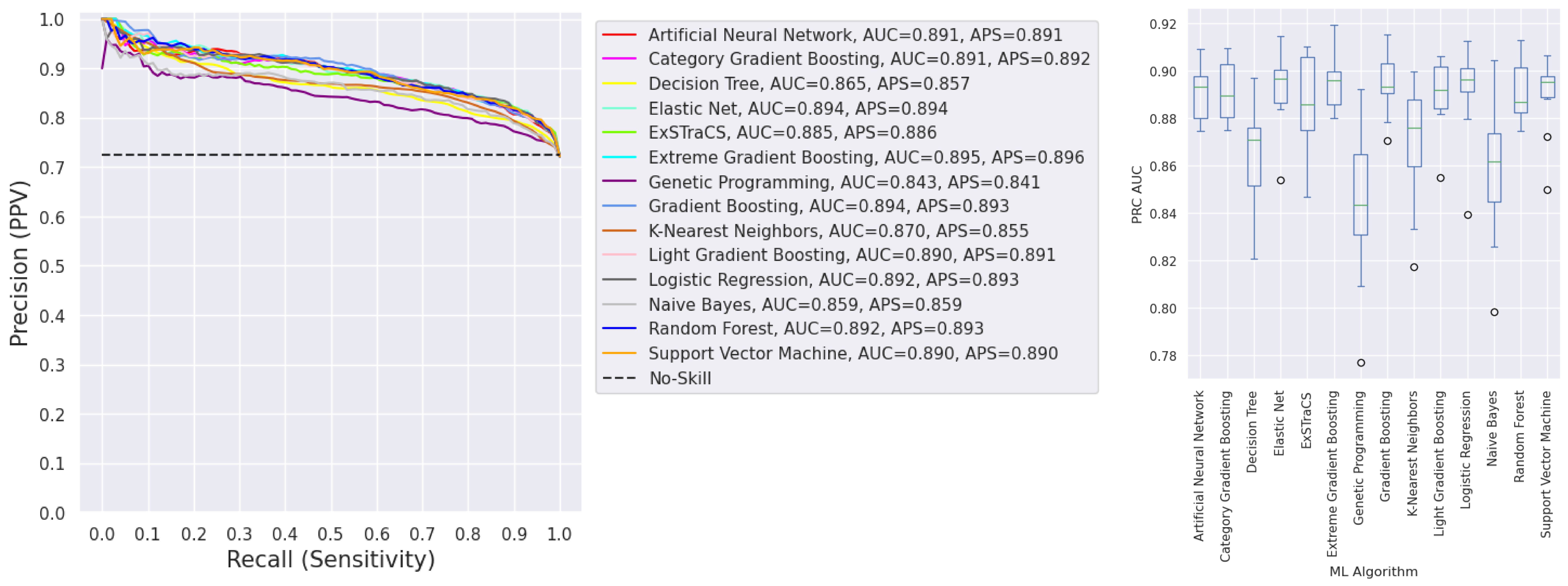}
    \caption*{\textbf{Figure S23: T\_GE5\_DEMDXSYMCFIO PRC Plot Across Algorithms.} (Left) PRC plot comparing 14 algorithm testing performance based on the average of 10-fold CV models trained for each algorithm. (Right) Boxplot of PRC-AUC testing performance for each algorithm across 10-fold CV models.}
\end{figure}
\addcontentsline{toc}{subsubsection}{Figure S23: T\_GE5\_DEMDXSYMCFIO PRC Plot Across Algorithms.}

\begin{table} [H]
\begin{adjustbox}{center}
\scalebox{0.65}{
\begin{tabular}{| p{1.8cm} | p{1.5cm} | p{1.4cm} | p{1.3cm} | p{1.5cm} | p{1.5cm} | p{1.3cm} | p{0.7cm} | p{0.7cm} | p{0.7cm} | p{0.7cm} | p{1.1cm} | p{1.1cm} | p{1.1cm} | p{1cm} | p{1cm} | p{1cm} |}
\hline
\rowcolor{light-gray}
Algorithm & Balanced Accuracy & Accuracy & F1 Score & Sensitivity (Recall) & Specificity & Precision (PPV) & TP & TN & FP & FN & NPV & LR+ & LR- & ROC AUC & PRC AUC & PRC APS \\
\hline
Artificial Neural Network & 0.6663 & 0.7683 & 0.8490 & 0.9016 & 0.4380 & 0.8052 & 142 & 26.5 & 34 & 15.5 & 0.6239 & 1.5881 & 0.2336 & 0.7789 & 0.8935 & 0.8943 \\
\hline
Category Gradient Boosting & 0.6537 & 0.7936 & 0.8714 & 0.9651 & 0.3525 & 0.7906 & 152 & 21.5 & 39.5 & 5.5 & 0.7894 & 1.4676 & 0.1025 & 0.7887 & 0.8898 & 0.8904 \\
\hline
Decision Tree & 0.6791 & 0.6874 & 0.7614 & 0.6911 & 0.6448 & 0.8360 & 108.5 & 39 & 21.5 & 48.5 & 0.4581 & 1.9576 & 0.4558 & 0.7328 & 0.8710 & 0.8620 \\
\hline
\rowcolor{yellow!50}
Elastic Net & 0.7189 & 0.7248 & 0.7953 & 0.7333 & 0.6967 & 0.8635 & 115.5 & 42.5 & 18.5 & 42 & 0.5022 & 2.4641 & 0.3825 & 0.7874 & 0.8966 & 0.8972 \\
\hline
ExSTraCS & 0.7073 & 0.7477 & 0.8183 & 0.8032 & 0.6311 & 0.8480 & 126.5 & 38.5 & 22.5 & 31 & 0.5391 & 2.1677 & 0.3322 & 0.7867 & 0.8860 & 0.8870 \\
\hline
Extreme Gradient Boosting & 0.7166 & 0.7500 & 0.8193 & 0.7898 & 0.6363 & 0.8507 & 124 & 38.5 & 22 & 33 & 0.5405 & 2.1930 & 0.3279 & 0.7898 & 0.8959 & 0.8964 \\
\hline
Genetic Programming & 0.6055 & 0.7454 & 0.8411 & 0.9302 & 0.2833 & 0.7711 & 146.5 & 17 & 43 & 11 & 0.5900 & 1.2934 & 0.2670 & 0.6993 & 0.8436 & 0.8427 \\
\hline
Gradient Boosting & 0.6751 & 0.7816 & 0.8584 & 0.9175 & 0.4214 & 0.8082 & 144.5 & 25.5 & 35 & 13 & 0.6667 & 1.6020 & 0.1927 & 0.7929 & 0.8934 & 0.8940 \\
\hline
K-Nearest Neighbors & 0.6692 & 0.7638 & 0.8461 & 0.8667 & 0.5082 & 0.8123 & 136.5 & 31 & 30 & 21 & 0.6061 & 1.6821 & 0.2496 & 0.7640 & 0.8759 & 0.8681 \\
\hline
Light Gradient Boosting & 0.7066 & 0.7333 & 0.8066 & 0.7746 & 0.6281 & 0.8452 & 122 & 38 & 22.5 & 35.5 & 0.5167 & 2.0966 & 0.3605 & 0.7869 & 0.8920 & 0.8925 \\
\hline
Logistic Regression & 0.7173 & 0.7248 & 0.7940 & 0.7293 & 0.7049 & 0.8645 & 114.5 & 43 & 18 & 42.5 & 0.5024 & 2.4896 & 0.3806 & 0.7893 & 0.8964 & 0.8969 \\
\hline
Naive Bayes & 0.6725 & 0.6628 & 0.7346 & 0.6497 & 0.6941 & 0.8474 & 102 & 42 & 18.5 & 55 & 0.4322 & 2.1574 & 0.5078 & 0.7356 & 0.8619 & 0.8617 \\
\hline
Random Forest & 0.7043 & 0.7615 & 0.8349 & 0.8344 & 0.5820 & 0.8370 & 131 & 35.5 & 25.5 & 26 & 0.5700 & 1.9953 & 0.2899 & 0.7875 & 0.8871 & 0.8879 \\
\hline
Support Vector Machine & 0.7083 & 0.7271 & 0.7980 & 0.7587 & 0.6611 & 0.8529 & 119.5 & 40 & 20.5 & 38 & 0.5086 & 2.2339 & 0.3756 & 0.7876 & 0.8952 & 0.8958 \\
\hline
\end{tabular}}
\end{adjustbox}
\caption*{\textbf{Table S21: T\_GE5\_DEMDXSYMCFIO Median Performance Metrics Across all Algorithms. }}
\end{table}
\addcontentsline{toc}{subsubsection}{Table S21: T\_GE5\_DEMDXSYMCFIO Median Performance Metrics Across all Algorithms.}

\begin{table} [H]
\begin{adjustbox}{center}
\scalebox{0.65}{
\begin{tabular}{| p{1.8cm} | p{1.5cm} | p{1.4cm} | p{1.3cm} | p{1.5cm} | p{1.5cm} | p{1.3cm} | p{0.7cm} | p{0.7cm} | p{0.7cm} | p{0.7cm} | p{1.1cm} | p{1.1cm} | p{1.1cm} | p{1cm} | p{1cm} | p{1cm} |}
\hline
\rowcolor{light-gray}
Algorithm & Balanced Accuracy & Accuracy & F1 Score & Sensitivity (Recall) & Specificity & Precision (PPV) & TP & TN & FP & FN & NPV & LR+ & LR- & ROC AUC & PRC AUC & PRC APS \\
\hline
Artificial Neural Network & 0.6726 & 0.7719 & 0.8502 & 0.8964 & 0.4488 & 0.8089 & 141 & 27.2 & 33.4 & 16.3 & 0.6267 & 1.6495 & 0.2339 & 0.7845 & 0.8905 & 0.8913 \\
\hline
Category Gradient Boosting & 0.6498 & 0.7903 & 0.8694 & 0.9663 & 0.3334 & 0.7902 & 152 & 20.2 & 40.4 & 5.3 & 0.7934 & 1.4596 & 0.1055 & 0.7875 & 0.8914 & 0.8921 \\
\hline
Decision Tree & 0.6664 & 0.6847 & 0.7631 & 0.7075 & 0.6253 & 0.8318 & 111.3 & 37.9 & 22.7 & 46 & 0.4547 & 1.9414 & 0.4669 & 0.7330 & 0.8651 & 0.8571 \\
\hline
\rowcolor{yellow!50}
Elastic Net & 0.7158 & 0.7237 & 0.7928 & 0.7336 & 0.6980 & 0.8632 & 115.4 & 42.3 & 18.3 & 41.9 & 0.5040 & 2.5043 & 0.3845 & 0.7889 & 0.8937 & 0.8944 \\
\hline
ExSTraCS & 0.7164 & 0.7531 & 0.8233 & 0.7991 & 0.6337 & 0.8504 & 125.7 & 38.4 & 22.2 & 31.6 & 0.5511 & 2.2466 & 0.3188 & 0.7805 & 0.8852 & 0.8859 \\
\hline
Extreme Gradient Boosting & 0.7128 & 0.7471 & 0.8179 & 0.7902 & 0.6354 & 0.8487 & 124.3 & 38.5 & 22.1 & 33 & 0.5444 & 2.1816 & 0.3313 & 0.7902 & 0.8951 & 0.8956 \\
\hline
Genetic Programming & 0.5977 & 0.7421 & 0.8376 & 0.9231 & 0.2724 & 0.7677 & 145.2 & 16.5 & 44.1 & 12.1 & 0.5880 & 1.2790 & 0.2796 & 0.6968 & 0.8427 & 0.8414 \\
\hline
Gradient Boosting & 0.6717 & 0.7802 & 0.8574 & 0.9161 & 0.4273 & 0.8062 & 144.1 & 25.9 & 34.7 & 13.2 & 0.6655 & 1.6130 & 0.1990 & 0.7879 & 0.8943 & 0.8927 \\
\hline
K-Nearest Neighbors & 0.6719 & 0.7517 & 0.8317 & 0.8519 & 0.4919 & 0.8135 & 134 & 29.8 & 30.8 & 23.3 & 0.5674 & 1.7053 & 0.3038 & 0.7410 & 0.8698 & 0.8546 \\
\hline
Light Gradient Boosting & 0.7046 & 0.7361 & 0.8090 & 0.7756 & 0.6337 & 0.8463 & 122 & 38.4 & 22.2 & 35.3 & 0.5230 & 2.1445 & 0.3545 & 0.7842 & 0.8902 & 0.8910 \\
\hline
Logistic Regression & 0.7148 & 0.7201 & 0.7891 & 0.7266 & 0.7029 & 0.8642 & 114.3 & 42.6 & 18 & 43 & 0.4989 & 2.5370 & 0.3913 & 0.7890 & 0.8922 & 0.8930 \\
\hline
Naive Bayes & 0.6748 & 0.6696 & 0.7432 & 0.6630 & 0.6866 & 0.8463 & 104.3 & 41.6 & 19 & 53 & 0.4399 & 2.2029 & 0.4954 & 0.7338 & 0.8589 & 0.8591 \\
\hline
Random Forest & 0.7039 & 0.7563 & 0.8292 & 0.8220 & 0.5858 & 0.8371 & 129.3 & 35.5 & 25.1 & 28 & 0.5635 & 2.0025 & 0.3064 & 0.7863 & 0.8921 & 0.8928 \\
\hline
Support Vector Machine & 0.7051 & 0.7279 & 0.8005 & 0.7566 & 0.6536 & 0.8505 & 119 & 39.6 & 21 & 38.3 & 0.5083 & 2.2481 & 0.3762 & 0.7843 & 0.8896 & 0.8904 \\
\hline
\end{tabular}}
\end{adjustbox}
\caption*{\textbf{Table S22: T\_GE5\_DEMDXSYMCFIO Mean Performance Metrics Across all Algorithms. }}
\end{table}
\addcontentsline{toc}{subsubsection}{Table S22: T\_GE5\_DEMDXSYMCFIO Mean Performance Metrics Across all Algorithms.}
\normalsize

\begin{figure} [H]
    \includegraphics[width=\textwidth]{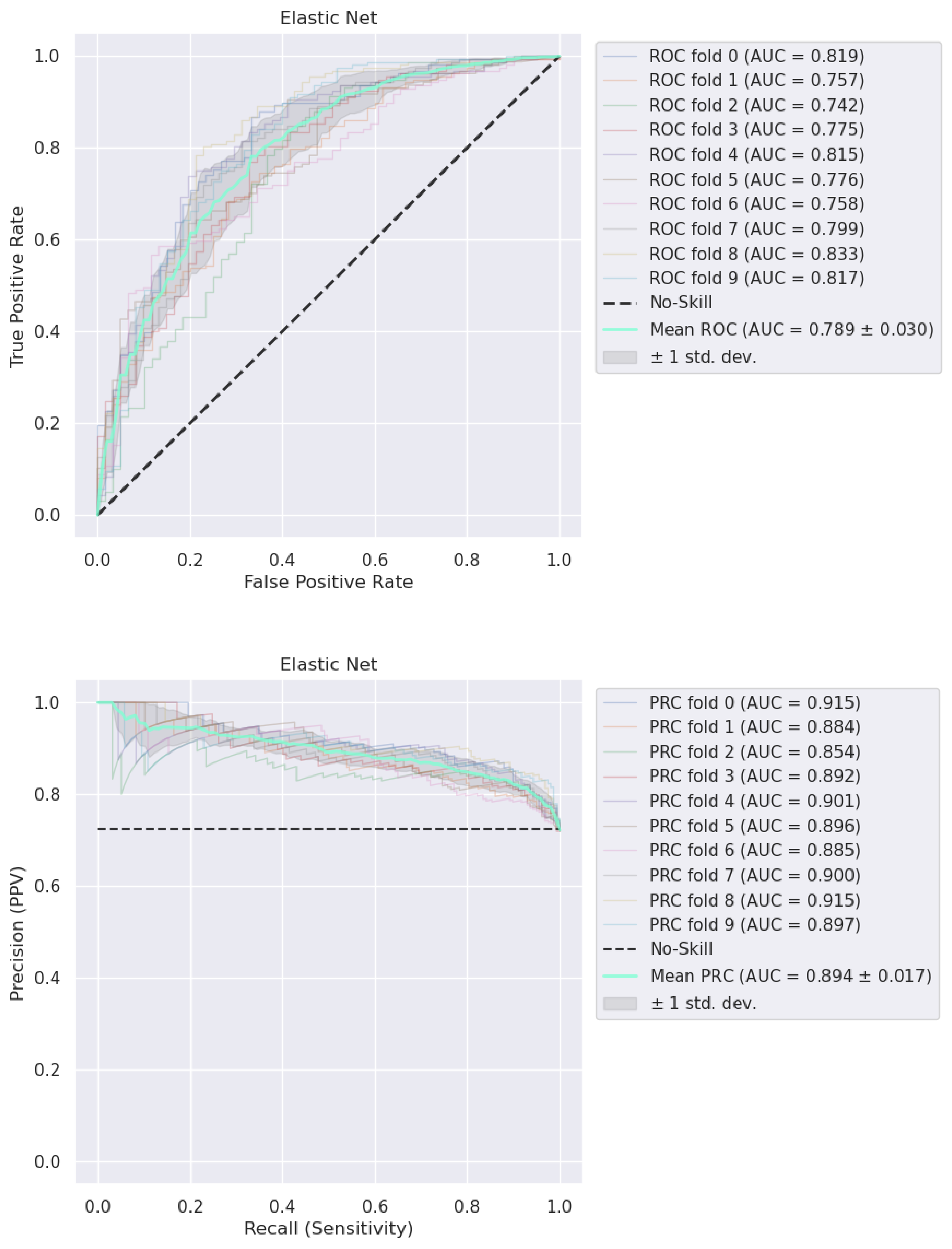}
    \caption*{\textbf{Figure S24: T\_GE5\_DEMDXSYMCFIO ROC and PRC Plots EN Performance Across CV Test Sets.} (Top) ROC plot showing individual logistic regression model performance trained on 10 separate CV partitions. (Bottom) PRC plot showing individual logistic regression model performance trained on 10 separate CV partitions. }
\end{figure}
\addcontentsline{toc}{subsubsection}{Figure S24: T\_GE5\_DEMDXSYMCFIO ROC and PRC Plots EN Performance Across CV Test Sets.}

\addcontentsline{toc}{subsubsection}{S.3.5.6: Replication Data Evaluations of Modeling}
\subsubsection*{S.3.5.6: Replication Data Evaluations of Modeling}

This last assessment was completed outside of STREAMLINE utilizing the output files it generated. Note that the number of instances in the replication datasets are much larger than that of the testing partitions generated from the respective development datasets. Up until this point, all OSA model evaluations have been conducted on respective hold-out testing datasets (different for each CV partition). Next, STREAMLINE was applied to re-evaluate all ML models (i.e 10 individual CV models trained by each of 14 ML algorithms) on the same replication dataset.

 For each of the 20 OSA datasets we compare the algorithm’s models that previously yielded the best median testing performance metric (either ROC-AUC for AHI$\geq$15, or PRC-AUC for AHI$\geq$5) to the performance of that same algorithm’s models on the replication data (see \textbf{Table S23}). For AHI$\geq$15 datasets, the best replication performance was again observed for dataset H with ANN models, having a significantly increased ROC-AUC of 0.807 (p=0.0002) in contrast with the testing evaluation. For all other AHI$\geq$15 datasets, we observed a mix of significant and non-significant increases and decreases in ROC-AUC replication performance. For AHI$\geq$5 datasets, the best replication performance was again observed for dataset T with elastic net models, having a significantly decreased PRC-AUC of 0.830 (p=0.0002), in contrast with the testing evaluation. For all other AHI$\geq$5 datasets we similarly observed a significant decrease in PRC-AUC replication performance in contrast with the testing evaluation.

\begin{table} [H]
\begin{adjustbox}{center}
\scalebox{0.9}{
\begin{tabular}{| l | l | l | l | l | l | l | l |}
\cline{3-6}
  \multicolumn{2}{ l }{}   & \multicolumn{2}{| l |}{\cellcolor{gray!25}Testing Data} & \multicolumn{2}{| l |}{\cellcolor{gray!25}Replication Data} & \multicolumn{2}{ l }{ } \\
\hline 
\rowcolor{gray!25}
Metric & Data & Best Algorithm & Median & Same Algorithm & Median  & Statistic & p-value \\
\hline \hline
\multirow{10}{*}{{\rotatebox[origin=c]{90}{\textbf{ROC-AUC(AHI$\geq$15)}}}} & A & ExSTraCS & 0.738696 & ExSTraCS & \cellcolor{green!25}0.765741 & 3 & \cellcolor{yellow!50}0.003197 \\ \cline{2-8}
 & B & LGB & 0.561623 & LGB & \cellcolor{green!25}0.589809 & 2 & \cellcolor{yellow!50}0.049366 \\ \cline{2-8}
 & C & GB & \cellcolor{green!25}0.660045 & GB & 0.638504 & 1 & 0.364346 \\ \cline{2-8}
 & D & RF & \cellcolor{green!25}0.734304 & RF & 0.731760 & 0 & 0.820595 \\ \cline{2-8}
 & E & SVM & 0.640454 & SVM & \cellcolor{green!25}0.651070 & 1 & 0.173617 \\ \cline{2-8}
 & F & LGB & 0.720548 & LGB & \cellcolor{green!25}0.760259 & 3 & \cellcolor{yellow!50}0.001152 \\ \cline{2-8}
 & G & LGB & 0.742128 & LGB & \cellcolor{green!25}0.781781 & 3 & \cellcolor{yellow!50}0.008151 \\ \cline{2-8}
 & \cellcolor{blue!25}H & ANN & 0.766822 & \cellcolor{blue!25}ANN & \cellcolor{green!25}0.807005 & 4 & \cellcolor{yellow!50}0.000212 \\ \cline{2-8}
 & I & LGB & 0.7554 & LGB & \cellcolor{green!25}0.785684 & 4 & \cellcolor{yellow!50}0.000381 \\ \cline{2-8}
 & J & ANN & 0.760819 & ANN & \cellcolor{green!25}0.797987 & 3 & \cellcolor{yellow!50}0.000507 \\ 
\hline \hline
\multirow{10}{*}{{\rotatebox[origin=c]{90}{\textbf{PRC-AUC(AHI$\geq$5)}}}}
 & K & CGB & \cellcolor{green!25}0.873214 & CGB & 0.796177 & 4 & \cellcolor{yellow!50}0.000157 \\ \cline{2-8}
 & L & GP->DT & \cellcolor{green!25}0.819413 & GP->DT & 0.718561 & 3 & \cellcolor{yellow!50}0.002497 \\ \cline{2-8}
 & M & GP->LGB & \cellcolor{green!25}0.862589 & GP->LGB & 0.652634 & 4 & \cellcolor{yellow!50}0.000157 \\ \cline{2-8}
 & N & RF & \cellcolor{green!25}0.87526 & RF & 0.799786 & 4 & \cellcolor{yellow!50}0.000157 \\ \cline{2-8}
 & O & GP->RF & \cellcolor{green!25}0.815426 & GP->RF & 0.705872 & 4 & \cellcolor{yellow!50}0.000157 \\ \cline{2-8}
 & P & CGB & \cellcolor{green!25}0.889826 & CGB & 0.775304 & 4 & \cellcolor{yellow!50}0.000157 \\ \cline{2-8}
 & Q & EN & \cellcolor{green!25}0.89404 & EN & 0.817442 & 4 & \cellcolor{yellow!50}0.000157 \\ \cline{2-8}
 & R & XGB & \cellcolor{green!25}0.894904 & XGB & 0.824617 & 4 & \cellcolor{yellow!50}0.000157 \\ \cline{2-8}
 & S & SVM & \cellcolor{green!25}0.892398 & SVM & 0.819281 & 4 & \cellcolor{yellow!50}0.000212 \\ \cline{2-8}
 & \cellcolor{blue!25}T & \cellcolor{blue!25}EN & \cellcolor{green!25}0.896618 & EN & 0.830697 & 4 & \cellcolor{yellow!50}0.000157 \\
\hline
\end{tabular}}
\end{adjustbox}
\caption*{\textbf{Table S23: Testing vs Replication Data Evaluation Comparisons Using the Same Models.} Pairwise tests using the Wilcoxon Signed Rank Test. For each of the 20 OSA datasets we compare the algorithm that previously yielded the best median testing performance metric (either ROC-AUC for AHI$\geq$15, or PRC-AUC for AHI$\geq$5) to the performance of that same algorithm’s models using the replication data held out for the respective dataset (A-T). Cells in green denote the larger median performance of the pair. Rows with p-value highlighted in yellow fall under the 0.05 significance cutoff. The cells in blue point out the dataset/algorithm combination where the highest median metric value was observed.}
\end{table}
\addcontentsline{toc}{subsubsection}{Table S23: Testing vs Replication Data Evaluation Comparisons Using the Same Models.}

Additionally, for each of the 20 OSA datasets we compare the algorithm’s models that previously yielded the best median testing performance metric (either ROC-AUC for AHI$\geq$15, or PRC-AUC for AHI$\geq$5) to the algorithm’s models that yielded the best median performance on the replication data (see \textbf{Table S24}). For AHI$\geq$15 datasets, the best replication ROC-AUC was again observed for dataset H with ANN models and all other AHI$\geq$15 datasets yielded a consistently (but not always significantly) higher median ROC-AUC on replication data in contrast with testing evaluations. For AHI$\geq$5 datasets, the best median replication performance was again observed for dataset T with elastic net models, and all other AHI$\geq$5 datasets yielded significantly lower median PRC-AUC on replication data in contrast with testing evaluations.

\textbf{Table S24} also identifies the individual best performing individual model on the replication data across all 20 OSA datasets. For AHI$\geq$15, a top ROC-AUC of 0.813 was obtained by CV model ‘1’ using the ANN algorithm on dataset H (DEM+DX+SYM+CF). For AHI$\geq$5, a top PRC-AUC of 0.847 was obtained by CV model ‘8’ using the category gradient boosting algorithm on dataset R (DEM+DX+SYM+CF).

\begin{table} [H]
\begin{adjustbox}{center}
\scalebox{0.8}{

\begin{tabular}{| l| l | l | l| l | l | l | l | p{3cm} | p{3cm} |}
\cline{3-6} \cline{9-10}
  \multicolumn{2}{l}{}   & \multicolumn{2}{|l|}{\cellcolor{gray!25}Testing Data} & \multicolumn{2}{|l|}{\cellcolor{gray!25}Replication Data} & \multicolumn{2}{l}{ } & \multicolumn{2}{|l|}{\cellcolor{gray!25}Replication Data} \\
\hline
\rowcolor{gray!25}
Metric & Data & Best Algorithm & Median & Best Algorithm & Median  & Statistic & p-value & Best Model: Algorithm + (CV Partition) & Target Metric Of Single Model \\
\hline \hline
\multirow{10}{*}{ROC-AUC} & A & ExSTraCS & 0.738696 & ExSTraCS & \cellcolor{green!25}0.765741 & 3 & \cellcolor{yellow!50}0.003197 & ANN (1) & 0.771473 \\ \cline{2-10}
 & B & LGB & 0.561623 & KNN & \cellcolor{green!25}0.602896 & 2 & \cellcolor{yellow!50}0.023342 & ExSTraCS (1) & 0.619898 \\ \cline{2-10}
 & C & GB & 0.660045 & NB & \cellcolor{green!25}0.668883 & 2 & 0.112411 & ANN (0) & 0.674619 \\ \cline{2-10}
 & D & RF & 0.734304 & ANN & \cellcolor{green!25}0.745548 & 2 & 0.096304 & LGB (9) & 0.750042 \\ \cline{2-10}
 & E & SVM & 0.640454 & EN & \cellcolor{green!25}0.654427 & 1 & 0.290846 & EN (9) & 0.666300 \\ \cline{2-10}
 & F & LGB & 0.720548 & ANN & \cellcolor{green!25}0.759872 & 3 & \cellcolor{yellow!50}0.000670 & ANN (0) & 0.769845 \\ \cline{2-10}
 & G & LGB & 0.742128 & ANN & \cellcolor{green!25}0.791045 & 3 & \cellcolor{yellow!50} 0.008274 & ANN (2) & 0.795970 \\ \cline{2-10}
 & \cellcolor{blue!25}H & ANN & 0.766822 & \cellcolor{blue!25}ANN & \cellcolor{green!25}0.807005 & 4 & \cellcolor{yellow!50}0.000212 & \cellcolor{blue!25}ANN (1) & \cellcolor{blue!25}0.813081 \\ \cline{2-10}
 & I & LGB & 0.7554 & EN & \cellcolor{green!25}0.788949 & 4 & \cellcolor{yellow!50}0.000212 & ExSTraCS (4) & 0.798181 \\ \cline{2-10}
 & J & ANN & 0.760819 & LR & \cellcolor{green!25}0.801513 & 4 & \cellcolor{yellow!50}0.000212 & LR (2) & 0.804307 \\
\hline \hline
\multirow{10}{*}{PRC-AUC} & K & CGB & \cellcolor{green!25}0.873214 & SVM & 0.804350 & 4 & \cellcolor{yellow!50}0.000157 & ExSTraCS (8) & 0.809892 \\ \cline{2-10}
 & L & GP->DT & \cellcolor{green!25}0.819413 & GP-> RF & 0.722144 & 3 & \cellcolor{yellow!50}0.002497 & GP->DT (5) & 0.769239 \\ \cline{2-10}
 & M & GP->LGB & \cellcolor{green!25}0.862589 & GP - >LR & 0.735277 & 4 & \cellcolor{yellow!50}0.000157 & GP->XGB (5) & 0.741470 \\ \cline{2-10}
 & N & RF & \cellcolor{green!25}0.87526 & SVM & 0.800717 & 4 & \cellcolor{yellow!50}0.000157 & SVM (2) & 0.809353 \\ \cline{2-10}
 & O & GP->RF & \cellcolor{green!25}0.815426 & GP-> SVM & 0.719038 & 4 & \cellcolor{yellow!50}0.000157 & GP->ANN (1) & 0.726045 \\ \cline{2-10}
 & P & CGB & \cellcolor{green!25}0.889826 & LR & 0.798189 & 4 & \cellcolor{yellow!50}0.000157 & ExSTraCS (4) & 0.805650 \\ \cline{2-10}
 & Q & EN & \cellcolor{green!25}0.89404 & EN & 0.817443 & 4 & \cellcolor{yellow!50}0.000157 & EN (5) & 0.824741 \\ \cline{2-10}
 & R & XGB & \cellcolor{green!25}0.894904 & SVM & 0.830600 & 4 & \cellcolor{yellow!50}0.000157 & \cellcolor{blue!25}CGB (8) & \cellcolor{blue!25}0.846841 \\ \cline{2-10}
 & S & SVM & \cellcolor{green!25}0.892398 & EN & 0.824477 & 3 & \cellcolor{yellow!50}0.000881 & ExSTraCS (0) & 0.834440 \\ \cline{2-10}
 & \cellcolor{blue!25}T & \cellcolor{blue!25}EN & \cellcolor{green!25}0.896618 & EN & 0.830697 & 4 & \cellcolor{yellow!50}0.000157 & ANN (9) & 0.837726 \\
\hline
\end{tabular}}
\end{adjustbox}
\caption*{\textbf{Table S24: Testing vs Replication Data Evaluation Comparisons Using Top Performing Algorithms.} Pairwise tests using the Wilcoxon Signed Rank Test. For each of the 20 OSA datasets we compare the algorithm that previously yielded the best median testing performance metric (either ROC-AUC for AHI$\geq$15, or PRC-AUC for AHI$\geq$5) to the algorithm that yields the best median performance metric models using the replication data held out for the respective dataset (A-T). Cells in green denote the larger median performance of the pair. Rows with p-value highlighted in yellow fall under the 0.05 significance cutoff. The cells in blue point out the dataset/algorithm combination where the highest median metric value was observed. The far-right side of the table identifies the single best performing model (specified by the name of the algorithm and the CV partition the model was trained on; numbered 0-9).}
\end{table}
\addcontentsline{toc}{subsubsection}{Table S24: Testing vs Replication Data Evaluation Comparisons Using Top Performing Algorithms.}

\paragraph{S.3.5.6.1: Hyperparameter Settings for Top Performing Models on Replication Data (Optuna Optimized)} 
\hfill
Top Performing AHI$\geq$15 model (dataset H): \textbf{ANN (1)}
\begin{itemize}  \itemsep0em 
\item Activation = logistic
\item Learning Rate = constant
\item Momentum = 0.810133
\item Solver = Adam
\item Batch Size = Auto
\item Alpha = 0.42784
\item Max iter = 200
\item Uses a default single hidden layer with a width of 100 nodes
\end{itemize}

Top Performing AHI$\geq$5 model (dataset R): \textbf{CGB (8)}
\begin{itemize}  \itemsep0em 
\item Learning Rate = 0.148981
\item Iterations = 231
\item Depth = 2
\item l2\_leaf\_reg = 7
\item loss\_function = Logloss
\end{itemize}

\addcontentsline{toc}{subsubsection}{S.4 Discussion}
\section*{S.4 Discussion}

\addcontentsline{toc}{subsubsection}{S.4.1 Benchmarking Challenges}
\subsection*{S.4.1 Benchmarking Challenges}
First, attempting to demonstrate the superiority of a single modeling algorithm through benchmarking is challenging in itself, and often in conflict with the “no free lunch theorem \cite{wolpert_no_1997}, discussed in the main text. ML benchmark datasets often focus on a more specific collection of data types or problems \cite{thiyagalingam_scientific_2022,olson_pmlb_2017,takamoto_pdebench_2022}, but ‘how’ to benchmark effectively remains an open topic of debate \cite{liao_are_2021}. These challenges are only exaggerated in seeking to compare the performance of different AutoML tools which effectively becomes a meta-optimization problem. Ultimately, we expect convincing AutoML comparison to necessarily focus on multiple objectives including (1) predictive performance across multiple metrics using a wide variety of simulated and real-world datasets with different underlying association complexities, data dimensions, signal/noise ratio, feature type mixes, class balance/imbalance, feature value frequencies and distributions, and ratio of informative to non-informative features, (2) computational complexity/run time, (3) model/pipeline interpretability and transparency, (4) what pipeline elements are automated and (5) ease of use. 

Second, running a single AutoML tool can be computationally expensive, thus running a rigorous comparison across several over a ‘representative’ array of benchmark datasets may be extremely time consuming even with significant computational resources. STREAMLINE limits the overall expected run-time in comparison to some AutoML tools by adopting a fixed, pre-configured pipeline design rather than exploring many possible configurations.

Third, some AutoML tools adopt a different ‘theory’ of pipeline assembly and target output, making the design of fair comparisons problematic. For example, most other AutoML tools focus on optimizing a single best model requiring a single data partitioning of training/evaluation data. Differently, STREAMLINE’s design focuses on being able to compare the performance of models trained by different ML algorithms with statistical significance comparisons. This requires an additional level of data partitioning in order to assess a globally best-performing model, i.e. external partitioning of data into development and replication sets, and further partitioning of the development data into k-fold CV training and testing sets. Starting with a dataset of the same size, this means that some other AutoML tools would have a larger set of data to ‘learn’ from in model/pipeline optimization, at the expense of not being able to rigorously assess (1) performance differences between ML algorithms or (2) sample bias from partitioning. 

Fourth, all AutoML tools likely have some degree of stochasticity, i.e. if you ran the same tool on the same dataset with the same parameter settings (excluding random seed) you would likely get somewhat different results. Therefore, AutoML benchmarking comparisons should ideally also run each tool on each dataset multiple times and ensure the same development vs. evaluation partitions were used on each. While some attempts have been made to compare the performance of a subset of AutoML tools \cite{nikitin_automated_2022,gijsbers_gama_2021,H2OAutoML20,zoller_benchmark_2021,balaji_benchmarking_2018,ferreira_comparison_2021,gijsbers_amlb_2022}, no study has yet addressed all of the above issues. 

In lieu of reliable cross-AutoML benchmarking, we recommend users select an AutoML tool that can handle the demands of the given task (e.g. data type, intended model use), and which is transparent enough to understand exactly how the data is being processed, how the model is trained vs. evaluated, and whether best practices are being properly adopted.  

\addcontentsline{toc}{subsubsection}{S.4.2 Feature Importance Consistency}
\subsection*{S.4.2 Feature Importance Consistency}

\begin{figure} [H]
    \centering
    \includegraphics[width=0.8\textwidth]{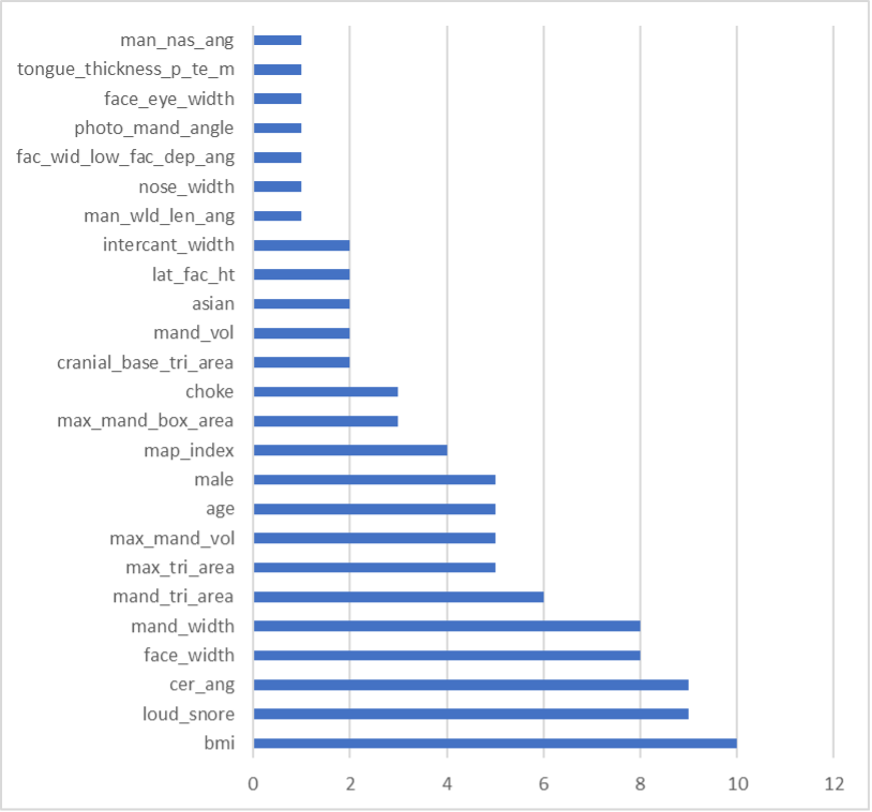}
    \caption*{\textbf{Figure S25: Consistency of the Top 10 Features (AHI$\geq$15 and AHI$\geq$5).} The consistency with which any of the 85 total features examined in this study are ranked in the top 10 based on (1) univariate analyses, (2) mutual information, (3) MultiSURF, (4) composite model feature importance plot ranking over all algorithms, and (5) mean model feature importance ranking for the algorithm yielding the top median performance. }
\end{figure}
\addcontentsline{toc}{subsubsection}{Figure S25: Consistency of the Top 10 Features (AHI$\geq$15 and AHI$\geq$5).}

\begin{figure} [H]
    \centering
    \includegraphics[width=0.5\textwidth]{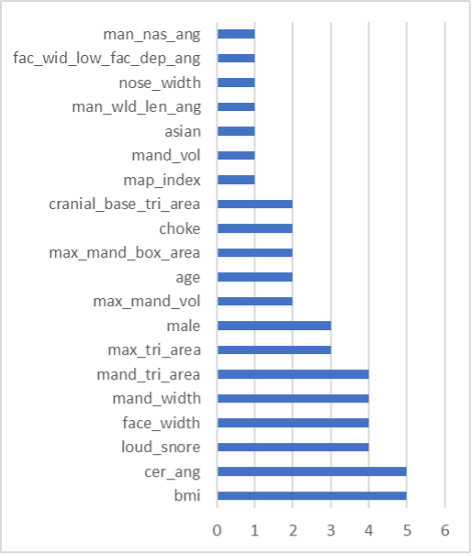}
    \caption*{\textbf{Figure S26: Consistency of the Top 10 Features (AHI$\geq$15).} The consistency with which any of the 85 total features examined in this study are ranked in the top 10 based on (1) univariate analyses, (2) mutual information, (3) MultiSURF, (4) composite model feature importance plot ranking over all algorithms, and (5) mean model feature importance ranking for the algorithm yielding the top median performance. }
\end{figure}
\addcontentsline{toc}{subsubsection}{Figure S26: Consistency of the Top 10 Features (AHI$\geq$15).}

\begin{figure} [H]
    \centering
    \includegraphics[width=0.5\textwidth]{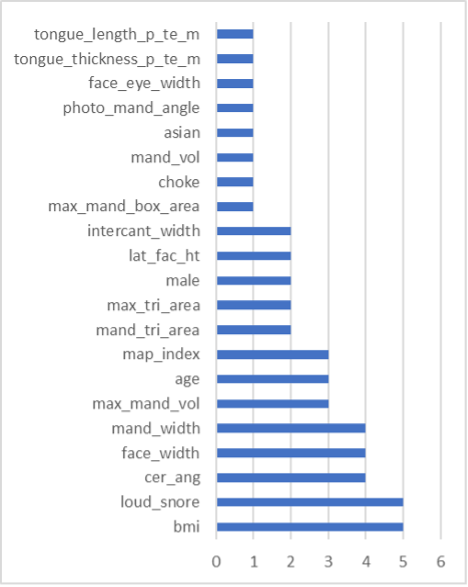}
    \caption*{\textbf{Figure S27: Consistency of the Top 10 Features (AHI$\geq$5).} The consistency with which any of the 85 total features examined in this study are ranked in the top 10 based on (1) univariate analyses, (2) mutual information, (3) MultiSURF, (4) composite model feature importance plot ranking over all algorithms, and (5) mean model feature importance ranking for the algorithm yielding the top median performance.}
\end{figure}
\addcontentsline{toc}{subsubsection}{Figure S27: Consistency of the Top 10 Features (AHI$\geq$5).}

\addcontentsline{toc}{subsubsection}{S.4.3 STREAMLINE Future Development}
\subsection*{S.4.3 STREAMLINE Future Development}

Beyond the future developments discussed in the main text, the following items are under active development to improve future STREAMLINE releases.

 (1) Expand the repertoire of available feature importance estimation algorithms as well as ML modeling algorithms (e.g. deep learning and other new promising algorithms). (2) Facilitate the application of dataset comparisons (phase 7) between any subset of datasets run in a STREAMLINE analysis. (3) Automatically exclude algorithms that achieve top PRC-AUC from dataset significance comparisons if they have a true negative rate of zero. (4) Expand the replication data evaluation (phase 8) to include reassessment of model feature importance estimates, and automated statistical comparisons between replication analyses. (5) Add a multiple imputation approach suited to categorical features (currently only mode imputation is available for categorical features). (6) Add a new phase to automatically pick the overall best-performing algorithm and provide a detailed performance summary. (7) Add a new phase focused on deployment of a best performing model that facilitates making predictions on unlabeled data including explanation. (8) Provide support to easily run STREAMLINE on cloud computing platforms. (9) Add Shapley value estimators of model feature importance and corresponding visualizations in addition to existing permutation-based feature importance.

\bibliographystyle{unsrtnat}  

\include{bibliography}